\RequirePackage[svgnames,table]{xcolor}
\documentclass[11pt,letterpaper]{mystyle}

\usepackage[all]{hypcap}
\usepackage[numbers,sort&compress]{natbib}
\usepackage{hyperref}
\hypersetup{colorlinks=true, citecolor={DodgerBlue}, linkcolor={DodgerBlue}, urlcolor={DodgerBlue}}
\usepackage{multicol}
\usepackage{setspace}
\usepackage{caption}
\usepackage{ragged2e}
\newcommand{\x}{\mathbf{x}}

\newcommand{\y}{\mathbf{y}}

\newcommand{\loss}{\mathcal{L}}

\newcommand{\bx}{\mathbf{x}}

\definecolor{blanchedalmond}{rgb}{1.0, 0.92, 0.8}
\definecolor{carmine}{rgb}{0.59, 0.0, 0.09}
\definecolor{lightblue}{rgb}{0.22,0.45,0.70}%

\renewcommand{\mathbf}{\boldsymbol}

\newcommand{\bb}{\mathbb}
\newcommand{\msf}{\mathsf}

\newcommand{\reals}{\bb R}

\newcommand{\R}{\reals}

\makeatletter
\def\Ddots{\mathinner{\mkern1mu\raise\p@
\vbox{\kern7\p@\hbox{.}}\mkern2mu
\raise4\p@\hbox{.}\mkern2mu\raise7\p@\hbox{.}\mkern1mu}}
\makeatother

\newcommand{\clip}{\msf{clipped}}

\definecolor{amaranth}{rgb}{0.9, 0.17, 0.31}
\definecolor{antiquebrass}{rgb}{0.8, 0.58, 0.46}
\definecolor{antiquefuchsia}{rgb}{0.57, 0.36, 0.51}
\definecolor{chromeyellow}{rgb}{0.31, 0.47, 0.26}

\usepackage{amssymb,mathrsfs,amsmath}
\usepackage[T1]{fontenc}
\usepackage{url}
\usepackage{booktabs}
\usepackage{amsfonts}
\usepackage{nicefrac}
\usepackage{microtype}
\usepackage{array}
\usepackage{graphicx}
\usepackage{wrapfig}
\usepackage{pifont}
\usepackage{subcaption}
\usepackage{enumitem}
\usepackage[skip=12pt plus 2pt minus 1pt]{parskip}
\usepackage{multirow}
\usepackage{makecell}
\usepackage{tabularx}

\usepackage{newunicodechar}
\newunicodechar{γ}{$\gamma$}
\usepackage{longtable}
\usepackage{float}
\usepackage{xltabular}
\usepackage{pdflscape}
\usepackage{threeparttable}
\usepackage{rotating}
\usepackage{siunitx}
\usepackage{hyperref}

\usepackage{adjustbox}
\usepackage{tikz}
\usetikzlibrary{arrows.meta,positioning,fit,backgrounds,calc,fadings}

\definecolor{fgInk}  {HTML}{000000}   
\definecolor{fgRule} {HTML}{8FA3B8}   
\definecolor{fgMute} {HTML}{68788C}   
\definecolor{fgB0}   {HTML}{DBE6F3}   
\definecolor{fgB1}   {HTML}{C6DAEE}
\definecolor{fgB2}   {HTML}{B3CDE6}
\definecolor{fgB3}   {HTML}{7FAAD4}
\definecolor{fgB4}   {HTML}{3A79B8}   
\definecolor{fgAcc}  {HTML}{C24552}   
\definecolor{fgAccBg}{HTML}{F5E1E3}   
\definecolor{fgWash} {HTML}{E8EBF1}   

\definecolor{catA}   {HTML}{0072B2}   
\definecolor{catB}   {HTML}{D55E00}   
\definecolor{catC}   {HTML}{009E73}   
\definecolor{catD}   {HTML}{A63A7B}   
\tikzset{
  fg/.style={
    font=\normalsize,
    text=fgInk,
    line width=0.6pt,
    >={Latex[length=1.7mm,width=1.3mm]},
  },
  fgbox/.style   = {draw=fgRule, line width=0.6pt, rounded corners=1.5pt,
                    fill=white, align=center, inner sep=3.5pt,
                    minimum height=8.5mm},
  fghid/.style   = {fgbox, dash pattern=on 1.7pt off 1.3pt},
  fgcirc/.style  = {circle, draw=fgRule, line width=0.6pt,
                    minimum size=10mm, inner sep=0pt, font=\normalsize},
  fgedge/.style  = {draw=fgInk, line width=0.6pt, ->},
  fgfeed/.style  = {draw=fgInk!55, line width=0.6pt, ->},
  fgdash/.style  = {draw=fgInk, line width=0.6pt, ->,
                    dash pattern=on 2.4pt off 1.8pt},
  fgacce/.style  = {draw=fgAcc,  line width=0.6pt, ->,
                    dash pattern=on 2.4pt off 1.8pt},
  fgaccs/.style  = {draw=fgAcc,  line width=0.6pt},
  fglab/.style   = {font=\small, text=fgInk,  inner sep=1.5pt},
  fgmut/.style   = {font=\small, text=fgMute, inner sep=1.5pt},
  fgred/.style   = {font=\small, text=fgAcc,  inner sep=1.5pt},
  fgttl/.style   = {font=\normalsize\bfseries, text=fgInk, inner sep=1.5pt},
}

\definecolor{accent}{HTML}{16695E}
\definecolor{accentbg}{HTML}{EEF4F2}
\definecolor{rulegrey}{HTML}{D9E1DE}
\definecolor{accent}{HTML}{16695E}
\definecolor{accentbg}{HTML}{EDF4F2}
\definecolor{toolblue}{HTML}{1F5C7A}
\definecolor{toolbg}{HTML}{EDF2F6}
\definecolor{failred}{HTML}{A8442C}
\definecolor{failbg}{HTML}{FAEEEA}
\definecolor{subgold}{HTML}{7A6320}
\definecolor{subbg}{HTML}{F7F3E6}
\definecolor{greyink}{HTML}{5B6A66}
\definecolor{obsgrey}{HTML}{6E7B77}
\tcbset{stepbase/.style={enhanced,breakable,sharp corners,boxrule=0pt,leftrule=2.6pt,
  left=8pt,right=8pt,top=5pt,bottom=5pt,boxsep=0pt,before skip=6pt,after skip=6pt,
  fontupper=\footnotesize,fonttitle=\scriptsize\bfseries\sffamily,titlerule=0pt,
  attach title to upper={\par\vspace{2pt}}}}
\newtcolorbox{stepR}[1]{stepbase,colframe=greyink,colback=greyink!4!white,
  coltitle=greyink,colbacktitle=greyink!4!white,title={#1}}
\newtcolorbox{stepT}[1]{stepbase,colframe=toolblue,colback=toolbg,
  coltitle=toolblue,colbacktitle=toolbg,title={#1}}
\newtcolorbox{stepX}[1]{stepbase,colframe=failred,colback=failbg,
  coltitle=failred,colbacktitle=failbg,title={#1}}
\newtcolorbox{stepS}[1]{stepbase,colframe=subgold,colback=subbg,
  coltitle=subgold,colbacktitle=subbg,title={#1}}
\newtcolorbox{taskbox}[1]{stepbase,colframe=accent,colback=accentbg,
  coltitle=accent,colbacktitle=accentbg,title={#1}}
\newcommand{\stepsub}[1]{%
  \par\addvspace{3pt}\noindent{\color{obsgrey}\scriptsize\sffamily #1}\par
  \addvspace{1pt}}
\newenvironment{payload}{\par\addvspace{2pt}\begingroup
  \scriptsize\ttfamily\raggedright\setlength{\leftskip}{8pt}\setlength{\parindent}{0pt}}%
 {\par\endgroup\addvspace{3pt}}
\newtcolorbox{cmdbox}{enhanced,breakable,sharp corners,boxrule=0.4pt,
  colframe=black!12,colback=black!3,left=5pt,right=5pt,top=3pt,bottom=3pt,boxsep=0pt,
  before skip=2pt,after skip=3pt,fontupper=\scriptsize\ttfamily\raggedright}
\newcommand{\bandbox}[1]{\tikz[baseline=(c.base)]{\node[fill=accentbg,text=accent,
  rounded corners=2pt,inner xsep=4pt,inner ysep=1.5pt,
  font=\bfseries\scriptsize\ttfamily](c){#1};}}
\newcommand{\critbadge}{\tikz[baseline=(c.base)]{\node[fill=failbg,text=failred,
  rounded corners=2pt,inner xsep=3pt,inner ysep=1pt,
  font=\bfseries\tiny\sffamily](c){CRITICAL};}}
\newtcolorbox{stepboxenv}[1]{
  enhanced, breakable, sharp corners=downhill, boxrule=0pt,
  leftrule=2.2pt, colframe=accent, colback=accentbg,
  left=7pt, right=7pt, top=4pt, bottom=4pt, boxsep=0pt,
  before skip=3pt, after skip=3pt,
  title={\footnotesize\bfseries\color{accent}#1},
  fonttitle=\footnotesize, coltitle=accent,
  attach title to upper={\quad}, colbacktitle=accentbg, titlerule=0pt}
\newcommand{\capbadge}[1]{\tikz[baseline=(c.base)]{\node[fill=accentbg,text=accent,rounded corners=2pt,inner xsep=4pt,inner ysep=2pt,font=\bfseries\footnotesize\ttfamily](c){#1};}}

\newcolumntype{Y}[1]{>{\RaggedRight\arraybackslash\hsize=#1\hsize}X}
\newcommand{\sub}[1]{{\scriptsize\itshape\color{black!62}#1}}
\newcommand{\hdr}[1]{\textbf{#1}}

\usetikzlibrary{positioning,arrows.meta,fit,backgrounds}
\definecolor{apodexrow}{HTML}{FBF1E4}   
\definecolor{ccrow}{HTML}{E9F1F9}        
\definecolor{baserow}{HTML}{F3F3F3}      
\definecolor{headband}{HTML}{ECEEF1}     
\newcommand{\best}[1]{\textbf{#1}}       
\newcommand{\na}{--}                     

\newcommand{\gain}[1]{\textcolor{green!50!black}{$\blacktriangle$\,#1}}
\renewcommand{\loss}[1]{\textcolor{red!70!black}{$\blacktriangledown$\,#1}}

\newif\ifshowcomments
\showcommentstrue      

\NewDocumentCommand{\heng}
{ mO{} }{\ifshowcomments\textcolor{red}{\textsuperscript{\textit{Heng}}\textsf{\textbf{\small[#1]}}}\fi}
\NewDocumentCommand{\zhenhailong}
{ mO{} }{\ifshowcomments\textcolor{blue}{\textsuperscript{\textit{Zhenhailong}}\textsf{\textbf{\small[#1]}}}\fi}

\NewDocumentCommand{\cflog}
{ mO{} }{\ifshowcomments\textcolor{purple}{\textsf{\textbf{\small[#1]}}}\fi}

\title{
Apodex Discovery:
Reality Benchmarks and Environments
for Evaluating and Building
\mbox{Discoverative Artificial Intelligence}
}
\author{Brian Wang, Bin Feng, Xiaoman Pan, Chenyang An, Felix Liu, Tangqi Fang, Gongbo Sun, Lingfeng Shen, Ning Wang, Handuo Zhang, Feng Chen, Fuchao Yang, Xiang Wang, Jiacheng Lin, Siting Li, Zixuan Liu, Chi Han, Zhenhailong Wang, Kunlun Zhu, Lawrence Zhao, Yueqi Guo, Kailong Wen, Feng Xing, Yiling Guo, Lidong Bing, David Tan, Bo An, Heng Ji, Sheng Wang$^{\dagger}$
\\
{\small Apodex}
}

\runningtitle{Apodex Discovery Technical Report}
\graphicspath{{figures/}{./}}
\begin{document}

\begin{abstract}
Apollo did not reach the Moon merely because its engineers could solve difficult equations. It succeeded by turning a distant ambition into a mission architecture of explicit objectives, coordinated systems, simulation, telemetry, verification, and repeated correction. AI now faces a similar transition: frontier models can solve increasingly difficult tasks once the problem, tools, and success criteria have been specified, yet the most consequential real-world challenges rarely arrive in an executable or verifiable form. We introduce \textbf{Apodex Discovery}, a framework for building and evaluating \emph{discoverative AI}, referring to AI directed at making genuine discoveries, realized through its operational unit, the \emph{heavy-duty solver}, together with \textbf{TRACES}, a reality benchmark we built.
Such heavy-duty solvers are complete systems comprising a foundation model, harness, tools, and control policies that pursue extended, stateful, and verifiable investigations. 

Apodex Discovery makes three principal contributions toward \emph{discoverative AI}: turning open-ended real-world problems into tractable, verifiable discovery. First, we develop a systematic problem-scouting process that surveyed 561 industries across 16 sectors, assembled 423 high-value real-world problems, and selected 20 for the initial release, spanning retrospective tasks with hidden outcomes and \emph{discovery challenges} evaluated prospectively as new evidence emerges. Second, we formalize \emph{executable environments} through a common environment--task--episode abstraction that provides solvers with the data, tools, constraints, and feedback needed to act, while recording trajectories and verifying key intermediate artifacts and final submissions; these efforts yield \textbf{TRACES}, the reality benchmark we released together with this paper. Third, we introduce \textbf{HDS6}, a process-verification framework assessing Tools, Repair, Alternatives, Coherence, Evidence, and Scope independently of final-task success. This enables meaningful evaluation even when definitive ground truth is delayed, incomplete, or unavailable.

In adeno-associated virus (AAV) capsid design, Apodex surpassed the
task-level published state of the art by 7\% across all four tasks: viability
prediction, tropism prediction, structure prediction, and generative design. In drug repurposing and reformulation, adding a task-specific biomedical environment improved the mean normalized prediction score of GPT-5.5 and GPT-5.6-sol by 2.5 and 7.6 points, respectively, over the same closed-book backbone. 
Controlled ablations further demonstrate that the fixed TRACES episode interface enables performance differences to be attributed to specific components of the solver system. By leveraging real-world problems, executable environments, and process evaluation, Apodex Discovery defines a new paradigm that lays a foundation for \emph{discoverative AI}: moving beyond solving predefined benchmarks toward conducting the consequential investigations by which discoveries are made. The TRACES benchmark is available at \href{https://discovery.apodex.com}{discovery.apodex.com}.
\par\vspace{3mm}
\end{abstract}

\maketitle

\begingroup
\makeatletter
\renewcommand{\thefootnote}{}
\renewcommand{\@makefntext}[1]{%
  \noindent
  \makebox[0pt][r]{\textsuperscript{$\dagger$}\hspace{0.4em}}%
  #1%
}
\footnotetext{%
Technical Lead. Corresponding author:
\href{mailto:sheng@apodex.com}{sheng@apodex.com}.

We gratefully acknowledge Mr. Tianqiao Chen, the project lead of this work,
who proposed the concept of the \emph{heavy-duty solver}, defined the six
capabilities of the HDS6 process metric, and shaped the overall architecture of
Apodex Discovery. We sincerely thank him for the vision, guidance, and support that made
this work possible.
}
\makeatother
\endgroup

\newpage
\tableofcontents
\newpage
\section{Introduction}
\label{sec:intro}
The Apollo program did not succeed merely because its engineers could solve difficult equations. The ambition to ``reach the Moon'' first had to be transformed into a mission architecture: a precise objective, explicit constraints, coordinated subsystems, simulation environments, telemetry, failure criteria, and repeated cycles of testing and correction. Only within this architecture could individual acts of problem solving accumulate into a consequential real-world achievement.

Artificial intelligence may now be approaching a similar transition. On advanced examinations, mathematical challenges, and software-engineering benchmarks, increasingly capable models can solve difficult problems once the objective, environment, tools, and success criteria have already been defined. Yet humanity's most consequential ambitions, from developing curative therapies for currently untreatable diseases, to achieving commercially viable fusion energy, to discovering new materials with transformative properties, rarely arrive with such an architecture. The next critical challenge for AI may therefore lie not only in building more capable models, but also in constructing the infrastructure that translates their capabilities into solutions to open and consequential problems.

{The aim that ultimately motivates this infrastructure is \emph{discovery}: enabling AI systems to reach conclusions that are not yet known, rather than to reproduce answers that already exist. Discovery is the goal, and the question this paper addresses is \emph{how} it is carried out---how an open-ended aspiration becomes a sequence of grounded, verifiable investigative acts that a system can execute and that others can check. We refer to this goal-directed paradigm as \emph{discoverative AI}. Discovery in practice proceeds as a loop, forming a hypothesis, acting on the world, verifying the result, and revising, and automating this discovery loop is the shared ambition of a rapidly expanding body of work, from autonomous scientific agents~\cite{boiko2023coscientist,gottweis2025aicoscientist,lu2024aiscientist} to dedicated efforts to automate the experimental loop of research itself, such as Discovery Loop~\cite{discoveryloop2026}. What these efforts require, and what remains largely absent, is the infrastructure that makes each turn of the loop executable, verifiable, and improvable.}

We identify four components of this missing infrastructure: \emph{problem formulation, reality-based environments, verification mechanisms, and repair loops}. Problem formulation translates an open-ended human ambition into explicit objectives, constraints, assumptions, and criteria for success. A reality-based environment equips the solver with the data, tools, computational resources, experimental interfaces, and human interactions required to act on the problem. Verification determines whether intermediate claims and final outcomes are supported, correct, and relevant to the original objective. Repair loops return the results of computation, experiments, external evaluation, or human judgment to the solver, enabling it to identify failures, revise its hypotheses, and improve its solution. Together, these components transform an under-specified ambition into a tractable, iterative, and evaluable problem that a \emph{heavy-duty solver} can pursue: 
a solver which is essentially the \emph{operational unit} of discoverative AI.

\begin{figure}[t]
\centering
\begin{adjustbox}{max width=\textwidth}
\input{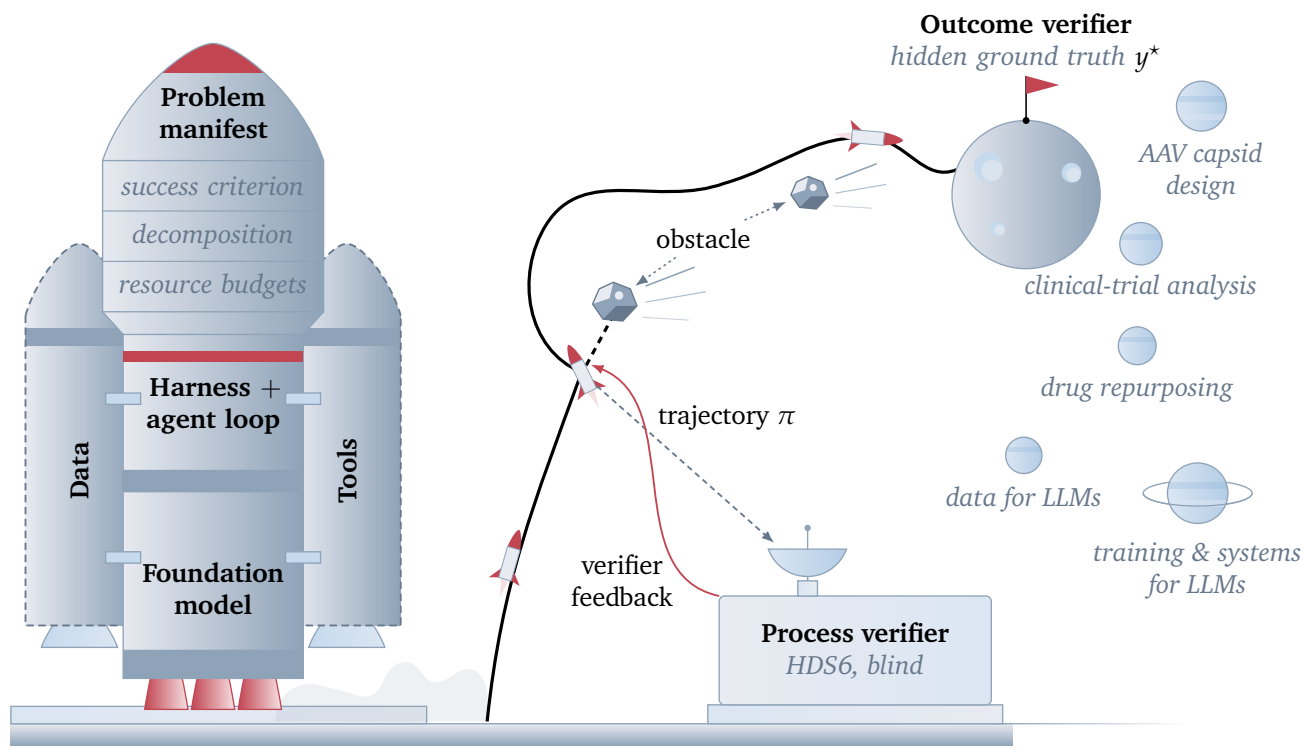}
\end{adjustbox}
\caption{
\textbf{Apodex Discovery mission architecture for a heavy-duty solver.}
The fairing contains the \emph{problem manifest}; the solid core represents the \emph{solver system under evaluation}; and the dashed side boosters form the task-specific \emph{reality-based environment}, which can be exchanged across problems. These components meet at a common \emph{episode interface}. During execution, failed checks trigger verifier feedback and repair, while the full trajectory $\pi$ is recorded for blind process verification. The final submission $y$ is evaluated separately by the outcome verifier against hidden ground truth $y^*$. The surrounding planets denote other problem families in the Apodex Discovery registry.
}
\label{fig:overview}
\end{figure}

Moreover, current evaluation paradigms are not designed to measure this form of problem solving. Most benchmarks fall into two broad families. Static question-answering benchmarks, including MMLU~\cite{hendrycks2021mmlu}, GPQA~\cite{rein2023gpqa}, MATH~\cite{hendrycks2021math}, FrontierMath~\cite{glazer2024frontiermath}, and Humanity's Last Exam~\cite{phan2025hle}, evaluate whether a model maps a fixed prompt to a correct answer. They probe important capabilities, but provide no persistent environment, external action, evolving state, or opportunity to respond to intermediate failure. Interactive benchmarks such as SWE-bench~\cite{jimenez2024swebench}, WebArena~\cite{zhou2023webarena}, OSWorld~\cite{xie2024osworld}, and $\tau$-bench~\cite{yao2024taubench} move closer to realistic problem solving by placing models in executable environments. Nevertheless, they typically remain bounded to narrowly specified tasks and reduce an episode to a final success signal. They rarely evaluate whether a solver decomposed the problem correctly, grounded its claims in evidence, considered viable alternatives, or repaired its approach after receiving verification feedback. What is missing is an evaluation framework whose unit of analysis is not an isolated answer, but an entire episode of tool-using, evidence-seeking, verifiable, and self-correcting problem solving.

Guided by the Apodex Discovery framework and methodology, we release \textbf{TRACES}, our reality benchmark of executable environments, episodes, and tasks with hidden verifiers, together with this paper at \href{https://discovery.apodex.com}{discovery.apodex.com}.

\subsection{What is a Heavy-Duty Solver?}
\label{sec:intro-hds}

We use the term \emph{heavy-duty solver} for a complete system, a foundation model together with the harness, tools, memory, and control policy that drive it, charged with carrying an open-ended, real-world ambition all the way to a verifiable result (\textbf{Fig. \ref{fig:overview}}). What sets such a solver apart from a model answering a prompt is not raw reasoning power but the ability to sustain a long, stateful, and self-correcting investigation against an external world that, at many times, yields some quite surprising results. That ability can be exercised only once an ambition has been equipped with the four pieces of infrastructure named above---a problem manifest, a reality-based environment, verification, and repair. We take each in turn; the sections that follow develop the machinery previewed here in full,
while a more detailed description of these important components is provided later in this paper.

\subsubsection{Problem Manifest}

A heavy-duty problem begins as an ambition rather than a concrete question or task, and the first piece of infrastructure turns it into something a solver can be graded on: a \emph{problem manifest}, the static, declarative contract that fixes everything invariant about the problem and from which concrete runnable instances are drawn. A manifest specifies:
\begin{itemize}
\item a \textbf{question and success criterion}---what would count as a solution. This need not be an answer known in advance, but it must be a criterion by which a proposed solution can be objectively recognized, verified, or falsified when it appears.
\item a \textbf{task decomposition}---the scoped tasks the problem is broken into, each with its own input, expected output, and success criteria, so that load-bearing intermediate results---those on which the final answer depends---are checked where they arise rather than inferred from a terminal artifact. Our adeno-associated virus (AAV) capsid-design problem, for instance, decomposes into four tasks: predicting whether a variant is viable, predicting its tissue tropism, reconstructing its three-dimensional structure, and generating novel on-target sequences under a limited experimental-feedback budget.
\item a \textbf{target outcome and hidden verifier}---the quantity that defines success, together with the hidden procedure and scoring rule that grade a submission against ground truth, or against a measurable proxy when the definitive outcome is delayed or unobservable.
\item the \textbf{data and tools}---the inputs and the action interface the solver is granted, and nothing beyond it, held separately from the hidden ground truth on which the submission is graded.
\item the \textbf{resource budgets}---the time, compute, tool-call, safety, and data-access limits under which the solver must operate.
\end{itemize}
Concrete \emph{episodes}, which are individual runnable instances, are instantiated from the manifest, each carrying its own instance data and hidden ground truth while sharing the tools, verifier, and scoring rule the manifest declares. A solver system is therefore measured across many instances of a task rather than on a single case.

\subsubsection{Reality-Based Environment}

The manifest says what must be solved; the environment is where the solving happens. A reality-based environment is the stateful, interactive substrate that realizes a manifest and through which a solver acts on the problem: tools (coding, search, specialized software), data (public or proprietary), and other interactions allowed by the environment manifest. Its defining property is that it returns fresh observations in response to the solver's actions: tool outputs, execution errors, retrieved documents, simulation results, experimental measurements, or verifier feedback. An environment is therefore not a prompt. A prompt supplies static context, whereas an environment produces new information as the solver acts.

Two commitments make the environment a faithful stand-in for reality. First, \emph{isolation}: the hidden ground truth on which a submission is graded is structurally prevented from reaching the solver, and execution is confined so that answers cannot be retrieved from outside nor results exfiltrated. Second, the unit of evaluation is the complete \emph{solver system under test}---the foundation model together with its harness and agent loop---rather than the model in isolation, because heavy-duty work is carried out by the whole system. Solver and environment meet only through a fixed \emph{episode interface}, which grants the inputs, tools, budget, and submission format on one side and applies the hidden verifier, hard gates, and outcome metrics on the other. Because every system and every baseline passes through the identical interface, a difference in outcome can be attributed to a specific component---the model, the harness, the agent loop, the tools, or the handling of feedback---rather than to the environment, and the same environments serve to compare harnesses at a fixed model as readily as models at a fixed harness.

\subsubsection{Verification Mechanisms}

An environment is only as trustworthy as the checks applied to what happens inside it, and verification takes two complementary forms. \emph{Outcome verification} asks what the solver produced, scoring the final submission---and, where a problem is decomposed, each load-bearing intermediate result---against success criteria that are known by the solver, but detailed scoring rubrics/formulas and data held inaccessible by the solver. For many of the problems that motivate this work the definitive check is unavailable, delayed, or only a shadow of the objective that ultimately matters, so the verifier falls back on a measurable proxy, such as a held-out experimental fitness in place of eventual \emph{in-vivo} therapeutic benefit. Because a deliverable is often a trained model, a curated corpus, or a data pipeline rather than a direct answer, each outcome verifier also carries explicit anti-gaming gates---leakage and contamination checks that disqualify an episode outright rather than merely lowering its score---and every outcome is normalized to a common scale from $0$ to $1$, between a trivial baseline and the best attainable result.

\emph{Process verification}, on the other hand, asks how the solver produced its result, reading the recorded trajectory rather than the submitted artifact.
Process verification complements outcome verification by detecting procedural deficiencies, maintaining integrity, and may sometimes be the only feasible form of evaluation for truly open-ended questions.
We instantiate it through \textbf{HDS6}, a metric that scores the trajectory along the six capabilities whose initials spell \textbf{TRACES}, defined in full below, which together separate a system able to sustain a long, self-correcting investigation from one that merely assembles a plausible account. The process verifier operates blind: it cannot consult the verified outcome, the solver's private chain-of-thought is stripped from the record before judging, and the solver's identity is withheld, so that a score reflects only the externally visible behavior of the trajectory, while a single integrity gate consults the outcome solely to detect fabrication. The two forms are complementary. Outcome scores measure success but are sparse and, for the highest-value problems, may not resolve for months or years; process scores supply the finer-grained and immediately available evidence of whether the investigation was conducted reliably, and are what allow progress to be assessed even when definitive ground truth is delayed, incomplete, or not yet available.

In this paper, we introduce Apodex Discovery, a framework for building and evaluating \emph{discoverative AI} through heavy-duty solvers, together with its reality benchmark \textbf{TRACES}. Apodex Discovery transforms an open-ended problem into a budgeted, stateful, and partially observable environment in which the evaluated object is the complete solver system under test—including the foundation model, harness, and agent loop—operating through the tools and feedback exposed by the environment, rather than the foundation model in isolation. The framework is \emph{interactive}: the solver acts on an environment whose state changes in response. It is \emph{verifiable}: episode-level and task-level outcome verifiers evaluate the final submission and load-bearing intermediate outputs, respectively, against hidden ground truth or measurable proxies. It is \emph{repairable}: structured diagnostic feedback can be returned to the solver without exposing hidden ground truth, enabling failures to be localized, corrected, and re-checked. Finally, it supports process verification: HDS6 evaluates the recorded trajectory independently of the verified outcome, allowing Apodex Discovery to distinguish a reliable investigation from a submission that succeeds without adequate evidential support.

To evaluate the process capabilities required for sustained heavy-duty work, we introduce \textbf{HDS6}, a six-dimensional process-verification metric whose dimensions spell \textbf{TRACES}: Tools, Repair,
Alternatives, Coherence, Evidence, and Scope.
\begin{itemize}
\item \textbf{Tools}: selecting, invoking, and correctly interpreting external tools;
\item \textbf{Repair}: responding to verification feedback or observed failure, correcting the underlying error, and confirming the fix;
\item \textbf{Alternatives}: making competing hypotheses explicit and adjudicating among them as evidence accumulates;
\item \textbf{Coherence}: maintaining state, constraints, and logical consistency over an extended horizon;
\item \textbf{Evidence}: grounding claims in observations, tool outputs, data, experiments, or references; and
\item \textbf{Scope}: identifying the conditions under which a conclusion holds and the boundaries beyond which it should not be applied.
\end{itemize}
We refer to the resulting six-capability evaluation as \emph{HDS6} (Heavy-Duty Solver Six Capabilities). Outcome verification asks whether the solver ultimately succeeded; HDS6 process verification asks whether the solution was produced through a reliable, evidence-grounded, and self-correcting investigation. This distinction is particularly important for high-value real-world problems, where definitive outcome ground truth may be delayed, incomplete, or initially unavailable. Together, outcome verification and HDS6 provide complementary views of whether an AI system can not only produce an answer, but also conduct the sustained investigation required to earn one.

\subsubsection{Repair Loops}

The final component turns evaluation from a verdict into a loop. A repair loop returns the results of verification---computation, experiment, external evaluation, or human judgment---to the solver so that failures can be localized, corrected, and re-checked rather than merely recorded. Within an episode this already occurs: verifier feedback passed back across the episode interface is what lets a solver detect a failed check and revise the work that produced it, and the Repair capability of HDS6 measures whether it does so by targeting the actual error and confirming the fix.

Verification also supports a repair loop \emph{across} attempts, because process verification yields a structured diagnosis rather than only a number. From the subrubric assessment, each capability that scored below its top band contributes a proposed band transition---the specific behavior a higher band would require---and from the load-bearing analysis, each faulty step contributes feedback localized to that step. These findings are synthesized into a compact \emph{repair note} and returned to the solver for a fresh attempt, and re-scoring the repaired trajectory quantifies the effect of the guidance. The note withholds the verified outcome, the hidden ground truth, and the outcome score, so that it guides the process without disclosing the answer; repair is in this sense a fixed procedure that closes a measure--diagnose--improve loop without ever relaxing the isolation between solver and ground truth. It is this loop that makes a heavy-duty problem not merely measurable but progressively solvable.

We begin by systematically scouting real-world problems that are both consequential and suitable for rigorous AI evaluation. A two-month effort by a ten-person team of STEM Ph.D. researchers surveyed 561 industries across sixteen sectors, assembled a registry of 423 high-value problems, and selected twenty for the initial TRACES benchmark. On AAV capsid design, Apodex exceeded the task-level published state of the art across all four stages of the pipeline: viability, tropism, structure prediction, and generative design; in drug repurposing, adding the biomedical environment improved the mean normalized prediction score of GPT-5.5 and GPT-5.6-sol by 2.5 and 7.6 points, respectively, over the same closed-book backbone. Beyond measuring aggregate performance, the fixed TRACES episode interface and controlled ablations make it possible to attribute success or failure to specific components of the solver system, including the foundation model, harness, initial skill guidance, and terminal verifier-guided repair, rather than treating every failure as an undifferentiated limitation of model capability.

\section{Scouting High-Value Real-World Problems}
Discoverative AI is only as consequential as the problems it is directed at. Identifying suitable real-world problems for AI evaluation requires a systematic process of domain selection, problem discovery, and environment construction. Although unresolved problems are abundant across science, medicine, engineering, and industry, only a subset possesses the reasoning depth, technical structure, verification pathway, and real-world value required to evaluate heavy-duty AI systems.

\begin{center}
\begin{tcolorbox}[colback=black!4,colframe=black!15,boxrule=0.4pt,arc=3mm,left=6mm,right=6mm,top=3mm,bottom=3mm,width=\linewidth,halign=center,fontupper=\itshape]
We only care about the unknowns: they are hard, uncertain and may take years to solve.\\
But they are highly valuable to mankind. And they are the only problems we care about.
\end{tcolorbox}
\end{center}

To identify such problems, a ten-person research team conducted a two-month survey spanning scientific research, engineering, healthcare, finance, and industrial applications (\textbf{Fig.~\ref{fig:selection} and Appendix~\ref{app:domains}}). All members of the team hold Ph.D. degrees in STEM disciplines and collectively contribute expertise in scientific research, AI development, engineering, computational methods, and domain-specific problem formulation. The survey combined systematic literature review, analysis of industry reports and technical materials, participation in domain-specific meetings and workshops, discussions with researchers and practitioners, and examination of problems arising from partner programs and internal deployment efforts.

This process produced an initial collection of \textbf{423 high-value real-world problems}, each of which was reviewed for reasoning depth, technical feasibility, verifiability, availability of the necessary data and tools, expected verification latency, and potential real-world impact. The resulting benchmark is therefore not a collection of questions assembled through lightweight brainstorming, but the outcome of a structured process involving domain analysis, problem collection, expert review, verifier design, and environment construction.

\begin{figure}[t]
\centering
\begin{adjustbox}{max width=\linewidth}
\begin{tikzpicture}[fg,
  ttl/.style  ={font=\normalsize\bfseries, text=fgInk, anchor=west,
                inner sep=0pt},
  num/.style  ={font=\large\bfseries, text=fgInk, inner sep=0pt},
  sub/.style  ={font=\small, text=fgMute, inner sep=0pt},
  body/.style ={font=\small, text=fgInk, anchor=north west,
                align=left, text width=76mm, inner sep=0pt},
  hd/.style   ={font=\small\bfseries, text=fgInk, anchor=north west,
                inner sep=0pt},
  kw/.style   ={font=\small, text=fgMute, anchor=north west,
                align=left, text width=76mm, inner sep=0pt},
  dom/.style  ={font=\small, text=fgInk, anchor=west, inner sep=0pt},
  arw/.style  ={font=\small\itshape, text=fgMute, inner sep=1.5pt},
  flow/.style ={draw=fgInk, line width=0.7pt, ->},
  crit/.style ={draw=fgRule, line width=0.5pt, rounded corners=2pt,
                fill=fgWash!40!white},
  lead/.style ={draw=fgMute, line width=0.5pt,
                dash pattern=on 0.5pt off 1.2pt},
]
\hbadness=10000 \hyphenpenalty=10000 \exhyphenpenalty=10000
\tikzset{every pic/.style={draw=fgB3, line width=0.65pt, scale=1.25,
                           line cap=round, line join=round, fill=none}}

\tikzset{
  icdna/.pic={
  \draw (0.1125,0.3262) .. controls (0.2025,0.2775) and (0.2025,0.2287) .. (0.1125,0.1800) .. controls (0.2025,0.1313) and (0.2025,0.0825) .. (0.1125,0.0338);
  \draw (0.2475,0.3262) .. controls (0.1575,0.2775) and (0.1575,0.2287) .. (0.2475,0.1800) .. controls (0.1575,0.1313) and (0.1575,0.0825) .. (0.2475,0.0338);
  \draw (0.1440,0.2903) -- (0.2160,0.2903);
  \draw (0.1440,0.1800) -- (0.2160,0.1800);
  \draw (0.1440,0.0697) -- (0.2160,0.0697);
  },
  icnet/.pic={
    \draw (0.0765,0.2700) circle (0.0382);
    \draw (0.2835,0.2700) circle (0.0382);
    \draw (0.1800,0.0765) circle (0.0382);
    \draw (0.1800,0.1935) circle (0.0338);
  \draw (0.1103,0.2520) -- (0.1530,0.2070);
  \draw (0.2497,0.2520) -- (0.2070,0.2070);
  \draw (0.1800,0.1597) -- (0.1800,0.1147);
  },
  iccode/.pic={
  \draw (0.1260,0.2790) -- (0.0405,0.1800) -- (0.1260,0.0810);
  \draw (0.2340,0.2790) -- (0.3195,0.1800) -- (0.2340,0.0810);
  },
  icpulse/.pic={
  \draw (0.0338,0.1800) -- (0.1035,0.1800) -- (0.1395,0.2655) -- (0.1935,0.0945) -- (0.2295,0.1800) -- (0.3262,0.1800);
  },
  iccloud/.pic={
  \draw (0.0945,0.1215) -- (0.2655,0.1215) .. controls (0.2871,0.1325) and (0.2982,0.1569) .. (0.2922,0.1803) .. controls (0.2863,0.2038) and (0.2649,0.2200) .. (0.2407,0.2194) .. controls (0.2337,0.2502) and (0.2074,0.2728) .. (0.1759,0.2751) .. controls (0.1443,0.2774) and (0.1150,0.2589) .. (0.1035,0.2295) .. controls (0.0737,0.2320) and (0.0475,0.2098) .. (0.0450,0.1800) .. controls (0.0425,0.1502) and (0.0647,0.1240) .. (0.0945,0.1215) -- cycle;
  \draw (0.1215,0.0720) -- (0.1215,0.0360);
  \draw (0.1800,0.0675) -- (0.1800,0.0270);
  \draw (0.2385,0.0720) -- (0.2385,0.0360);
  },
  icbridge/.pic={
  \draw (0.0270,0.0900) -- (0.3330,0.0900);
  \draw (0.0495,0.0900) .. controls (0.1125,0.2565) and (0.2475,0.2565) .. (0.3105,0.0900);
  \draw (0.1125,0.0900) -- (0.1125,0.1732);
  \draw (0.1800,0.0900) -- (0.1800,0.2093);
  \draw (0.2475,0.0900) -- (0.2475,0.1732);
  },
  icchip/.pic={
    \draw[rounded corners=0.0225cm] (0.0967,0.0968) rectangle (0.2632,0.2632);
  \draw (0.1440,0.2632) -- (0.1440,0.3172);
  \draw (0.2160,0.2632) -- (0.2160,0.3172);
  \draw (0.1440,0.0968) -- (0.1440,0.0428);
  \draw (0.2160,0.0968) -- (0.2160,0.0428);
  \draw (0.0967,0.2160) -- (0.0427,0.2160);
  \draw (0.0967,0.1440) -- (0.0427,0.1440);
  \draw (0.2632,0.2160) -- (0.3172,0.2160);
  \draw (0.2632,0.1440) -- (0.3172,0.1440);
  },
  icchart/.pic={
  \draw (0.0450,0.3150) -- (0.0450,0.0495) -- (0.3150,0.0495);
  \draw (0.0945,0.1125) -- (0.1575,0.1980) -- (0.2160,0.1530) -- (0.2970,0.2745);
  },
  ichex/.pic={
    \draw (0.1800,0.1800) circle (0.0338);
  \draw (0.1800,0.3195) -- (0.2970,0.2497) -- (0.2970,0.1103) -- (0.1800,0.0405) -- (0.0630,0.1103) -- (0.0630,0.2497) -- cycle;
  },
  icgear/.pic={
    \draw (0.1800,0.1800) circle (0.0765);
  \draw (0.1800,0.2565) -- (0.1800,0.3060);
  \draw (0.1800,0.1035) -- (0.1800,0.0540);
  \draw (0.1035,0.1800) -- (0.0540,0.1800);
  \draw (0.2565,0.1800) -- (0.3060,0.1800);
  \draw (0.1260,0.2340) -- (0.0945,0.2655);
  \draw (0.2340,0.1260) -- (0.2655,0.0945);
  \draw (0.2340,0.2340) -- (0.2655,0.2655);
  \draw (0.1260,0.1260) -- (0.0945,0.0945);
  },
}

\node[ttl] at (0.15,9.10) {(a) Selection pipeline};

\foreach \bx/\n/\lab/\fl/\dr in {%
  0.15/561/{industries surveyed}/{fgWash!55!white}/fgRule,
  4.97/10/{domains selected}/fgB0/fgB3,
  9.78/423/{candidate questions}/{fgWash!55!white}/fgRule,
  14.60/20/{registry published}/fgB1/fgB4}{
  \draw[draw=\dr, line width=0.5pt, rounded corners=2pt, fill=\fl]
       (\bx,7.18) rectangle (\bx+2.55,8.80);
  \node[num] at (\bx+1.275,8.44) {\n};
  \node[sub, align=center, text width=20mm] at (\bx+1.275,7.74) {\lab};
}
\foreach \ax/\lab in {2.90/scored, 7.72/expanded, 12.53/screened}{
  \draw[flow] (\ax,8.05) -- (\ax+1.87,8.05);
  \node[arw] at (\ax+0.935,8.38) {\lab};
}

\foreach \x/\tx in {0.15/3.84, 8.75/13.47}{
  \filldraw[draw=fgRule, line width=0.5pt, fill=white, rounded corners=4pt]
    (\x,6.85) -- (\tx-0.26,6.85) [sharp corners]
      -- (\tx,7.24) -- (\tx+0.26,6.85) [rounded corners=4pt]
    -- (\x+8.40,6.85) -- (\x+8.40,2.01) -- (\x,2.01) -- cycle;
}
\node[hd] at (0.45,6.61) {scored on three dimensions};
\node[hd] at (9.05,6.61) {screened against three requirements};

\foreach \x/\y/\term/\rest/\kw in {%
  0.35/6.09/{Technical fit}/{--- suits long-horizon tasks}/%
       {reasoning depth $\cdot$ verifiability $\cdot$ deployability},
  0.35/4.69/{Value}/{--- a landmark result or a large market}/%
       {the maximum of the two, never the average},
  0.35/3.29/{Commercial access}/{--- can a solution scale}/%
       {enter $\cdot$ embed $\cdot$ defend $\cdot$ beyond one customer},
  8.95/6.09/{A verifier}/{--- a rubric and a runnable instance}/%
       {``an expert eyeballs it'' was rejected},
  8.95/4.69/{Impact and a buyer}/{--- both, not either}/%
       {a paying problem with real difficulty},
  8.95/3.29/{A closed loop}/{--- from attempt to outcome}/%
       {each solve can improve the next}}{
  \draw[crit] (\x,\y-1.10) rectangle (\x+8.00,\y);
  \node[body] at (\x+0.20,\y-0.17) {\textbf{\term} \rest};
  \node[kw]   at (\x+0.20,\y-0.60) {\kw};
}

\draw[draw=fgRule, line width=0.5pt] (0.15,1.71) -- (17.15,1.71);
\node[ttl] at (0.15,1.23) {(b) The ten selected domains};

\pic at (0.13,0.45) {icdna};   \node[dom] at (0.72,0.68) {Bioinformatics \& Genomic Medicine};
\pic at (9.15,0.45) {icbridge};   \node[dom] at (9.74,0.68) {Structural \& Civil Engineering Services};
\pic at (0.13,-0.13) {icnet};   \node[dom] at (0.72,0.10) {Foundation Models for Scientific Reasoning};
\pic at (9.15,-0.13) {icchip};   \node[dom] at (9.74,0.10) {Chip Design \& Verification (EDA)};
\pic at (0.13,-0.71) {iccode};   \node[dom] at (0.72,-0.48) {Scientific HPC Code Generation \& Optimization};
\pic at (9.15,-0.71) {icchart};   \node[dom] at (9.74,-0.48) {Market Data \& Financial Indices};
\pic at (0.13,-1.29) {icpulse};   \node[dom] at (0.72,-1.06) {Healthcare IT \& Digital Health};
\pic at (9.15,-1.29) {ichex};   \node[dom] at (9.74,-1.06) {Materials Discovery, Design \& Qualification};
\pic at (0.13,-1.87) {iccloud};   \node[dom] at (0.72,-1.64) {Weather-Driven Commodity \& Energy Forecasting};
\pic at (9.15,-1.87) {icgear};   \node[dom] at (9.74,-1.64) {Industrial Automation \& Control};

\end{tikzpicture}

\end{adjustbox}

\caption{The selection pipeline. From 561 industries surveyed to 10 domains,
           expanded to 423 candidate questions and screened to the published
           registry of 20.}
\label{fig:selection}
\end{figure}

\subsection{Domain Selection}
Apodex Discovery is designed to evaluate heavy-duty problems whose solutions cannot be obtained through simple memorization or direct retrieval. A question with a published answer may still require substantial reasoning, but performance on such a question can also reflect memorization, benchmark contamination, or sophisticated information retrieval. In contrast, when a problem has no established answer, the solver must be evaluated through the work it performs, including the evidence it collects, the tools it invokes, the hypotheses it considers, the intermediate artifacts it produces, and the final outcome it proposes.

We began the domain-selection process by constructing a master taxonomy of \textbf{561 mutually exclusive industries} spanning \textbf{sixteen sectors}. This bottom-up representation of the global industrial landscape was then cross-checked against the scientific and technological challenge areas identified by the U.S. Government Genesis Mission, providing a complementary top-down view of strategically important problems.

Each industry was assessed along three dimensions:

\begin{enumerate}
\item \textbf{Technical fit.} We evaluated whether the most difficult problems in the industry had an appropriate structure for a long-horizon reasoning system, considering reasoning depth, formalizability, executability, verifier fidelity, verification latency, data availability, and last-mile deployability.

\item \textbf{Value.} We evaluated whether solving the industry's hardest problems could produce either a landmark scientific or technological result or a large commercial opportunity. Rather than averaging these two forms of value, we used their maximum, such that exceptional scientific significance or exceptional commercial value was independently sufficient for a domain to qualify for further consideration.

\item \textbf{Commercial access.} We evaluated whether a solution could realistically enter the domain, become embedded in an existing workflow, develop a defensible advantage, and scale beyond a single demonstration, partner, or customer.

\end{enumerate}

The three dimensions were assessed independently rather than collapsed into a fully compensatory score, because, for example, strengths in market size, data availability, or commercial access should not obscure weaknesses in verifier quality or reasoning depth. Two further principles shaped the scoring. First, we judged each domain by what an advanced discoverative AI system could ultimately achieve on its hardest problems, not by how widely AI is used there today, since current adoption reveals little about that potential. Second, the depth of heavy-duty reasoning and tool use capped the overall score: a problem ranked highly only if solving it genuinely demanded such deep, multi-step work, not merely because it offered abundant data, convenient evaluation, or an established market.

Because commercial access depends heavily on institutional structure, procurement processes, data ownership, regulatory constraints, and deployment context, it was more difficult to estimate reliably than technical fit or scientific value. Quantitative scores were therefore used to organize and narrow the search rather than to determine the final selection automatically. Human reviewers then examined the highest-ranked domains and selected ten domains across eight sectors, with no more than two domains drawn from any single sector.

The resulting domains are:

\begin{itemize}
\item Bioinformatics, Computational Biology, and Genomic Medicine;
\item AI Foundation Models for Scientific Reasoning;
\item AI for Scientific HPC Code Generation and Optimization;
\item Healthcare IT and Digital Health;
\item Weather-Driven Commodity and Energy Market Forecasting;
\item Structural and Civil Engineering Services;
\item AI-Assisted Chip Design and Verification;
\item Market Data and Financial Indices;
\item AI-Driven Materials Discovery, Design, and Qualification; and
\item Industrial Automation and Control.
\end{itemize}

\subsection{Question Collection and Screening}
The ten selected domains were expanded into concrete candidate problems through literature review, industry research, domain-specific meetings, expert discussions, partner programs, and internal deployment experience. The team also considered questions proposed directly by scientists, engineers, clinicians, and other domain practitioners, particularly when those questions arose from active research programs or operational workflows rather than from benchmark design alone.

For each candidate, the team documented not only the problem statement but also the data, tools, computational interfaces, verification procedures, and external dependencies required to make the problem operational. Candidates were therefore evaluated as prospective environments rather than as isolated natural-language questions, and an appealing or consequential problem description was not sufficient for inclusion unless a credible path existed from the solver's actions to an inspectable and verifiable outcome.

At minimum, each candidate problem was required to specify:

\begin{itemize}
\item a concrete objective and a clearly defined output artifact;
\item a decomposition into substantive intermediate tasks;
\item a measurable verification rubric;
\item at least one runnable input--output instance;
\item an identifiable source of data, evidence, or experimental feedback;
\item a path from the solver's attempt to an externally verifiable outcome; and
\item both a consequential impact and an identifiable user, customer, or beneficiary.
\end{itemize}

Questions whose success conditions ultimately reduced to unstructured expert judgment were rejected. Expert review may remain an important component of evaluation, particularly for frontier scientific problems, but it cannot serve as the sole basis for determining success; instead, a benchmark problem must expose sufficient observable structure for its claims, intermediate artifacts, experimental assumptions, and final outcomes to be independently inspected, challenged, and, where possible, falsified.

We also applied a strict value gate under which each problem had to demonstrate both meaningful impact and a plausible beneficiary or buyer. A problem that no identifiable party would invest resources to solve is more likely to constitute an academic exercise than a reality-facing challenge, whereas a problem with an obvious buyer but limited technical depth is more appropriately treated as a product feature than as a heavy-duty benchmark problem.

This process resulted in a registry of \textbf{423 high-value real-world problems}. From this registry, we selected \textbf{20 problems} for the initial TRACES release, spanning frontier-model development, biomedical discovery, clinical translation, scientific engineering, and deployment-oriented intelligence. 

The selected problems fall into two evaluation settings. In the retrospective setting, the benchmark creator has access to a known outcome that is deliberately withheld from the solver, and the task is constructed so that the outcome cannot be recovered through ordinary lookup but must instead be reconstructed through evidence gathering, modeling, experimentation, or tool use. In the prospective setting, represented by discovery challenges, no authoritative answer exists when the evaluation begins, and the quality of the solver's work is assessed as external evidence and real-world outcomes subsequently become available.

Although these settings differ in whether the final outcome is already known, they share the same fundamental property: the solver cannot succeed by recalling or retrieving an answer. A correct solution need not be known in advance, but it must be \emph{objectively recognizable when it appears}, because the ability to recognize, verify, or falsify a proposed solution is what makes an open-ended real-world problem suitable for evaluation.

\subsection{Reality-Facing Problems in the Initial Benchmark}
The initial release of TRACES focuses on five problem families selected for their combination of substantial real-world value, long-horizon reasoning, heterogeneous tool use, meaningful intermediate verification, and external outcomes capable of ultimately confirming or falsifying the solver's conclusions. Table~\ref{tab:env-tasks-episodes} lists the concrete tasks within each family, with brief descriptions and the number of episodes currently available.

\begin{enumerate}
\item \textbf{Adeno-associated virus capsid design.}
This problem evaluates a four-stage pipeline for the assessment and
design of adeno-associated virus capsids for gene delivery.
The tasks comprise predicting whether a capsid variant remains viable,
predicting its tissue or cellular tropism, reconstructing its
three-dimensional structure, and generating novel, manufacturable, and
on-target capsid sequences under a limited experimental-feedback
budget.

A solver must integrate sequence--function data, structural and
biological constraints, predictive models, folding and analysis tools,
and experimentally derived feedback. Each task is evaluated against
leak-resistant held-out evidence, including extrapolation across
mutational load, species, structural-deposition date, and unseen
candidate sequences.

\item \textbf{Drug Repurposing and Reformulation.}

This task evaluates the identification of new therapeutic indications and optimized formulations for existing, clinically characterized drugs, rather than the discovery of new drugs. Specifically, it focuses on repurposing drugs approved for one disease to treat different diseases. A solver must synthesize mechanistic evidence, disease biology, clinical-trial records, pharmacology, safety information, regulatory evidence, repurposing procedure and real-world outcomes to prioritize drug--disease hypotheses. The reformulation aspects include population, biomarker, drug combination, and formulation/delivery.

Performance is assessed against held-out clinical, regulatory, or real-world outcomes rather than against a published answer available to the solver. The objective is not to generate a plausible therapeutic narrative, but to produce a prioritized and evidence-grounded recommendation that survives prospective or retrospective verification.

\item \textbf{Clinical-trial analysis and translation.}

This problem covers statistical analysis, clinical reporting, safety- and efficacy-signal discovery, trial-advancement forecasting, and preclinical-to-human translation. Representative tasks include generating analysis code and tables under a prespecified statistical analysis plan, identifying emerging safety or efficacy signals, predicting whether a program will advance to the next clinical stage, and diagnosing why promising preclinical findings fail to translate to humans.

\item \textbf{Data for large language models.}

This problem addresses the acquisition, construction, filtering, and evaluation of the data underlying language-model development, with tasks including pretraining-data procurement, quality estimation, deduplication, contamination detection, benchmark construction, and the discovery of data mixtures that improve downstream capabilities under fixed compute and budget constraints.

The objective is not simply to collect more data, but to construct data pipelines whose downstream effects can be measured through controlled experiments and whose decisions can be traced to explicit evidence.

\item \textbf{Training and systems for large language models.}

This problem evaluates the discovery, diagnosis, and repair of the recipes and systems required to train and deploy language models, with tasks including synthetic-data fine-tuning, reinforcement-learning recipe discovery, optimizer and schedule selection, inference-determinism repair, distributed-training debugging, and solution-verifier construction.

Solvers must operate across codebases, logs, experiments, hardware constraints, and evaluation results, such that success requires not only proposing a training recipe or code modification but also executing it, diagnosing failures, interpreting measurements, and demonstrating that the resulting system satisfies the stated constraints.
\end{enumerate}

Together, these problem families illustrate the intended scope of Apodex Discovery. They are not short-form questions expressed in domain-specific language, but extended investigations that require a solver to maintain state over a long horizon, interact with tools and data, produce inspectable intermediate artifacts, respond to verification feedback, and ultimately create work whose value can be judged outside the benchmark.

The initial executable environments represent only a small fraction of the 423-problem registry, and future releases will progressively expand into additional scientific, engineering, financial, healthcare, and industrial domains. The growing collection is intended both as an evaluation framework and as an invitation to the broader community to test models and solver systems not only on questions whose answers are already known, but also on difficult, consequential, and verifiable problems that remain to be solved.

\subsection{Discovery Challenges}
We define a \emph{discovery challenge} as a reality-facing problem for which no authoritative reference answer is available when evaluation begins. Unlike a retrospective benchmark instance, in which a known outcome is withheld from the solver, a discovery challenge is evaluated prospectively as experimental measurements, external evidence, expert adjudication, or real-world outcomes become available. The solver is therefore assessed not by comparison with a pre-existing answer, but by whether it produces an auditable body of work whose assumptions, intermediate conclusions, proposed intervention, and final claims can be independently confirmed or falsified over time.

Our first discovery challenge focuses on \textbf{causal drug repurposing by functional-screen compound matching}. Given a protective-versus-detrimental gene signature from an unbiased genome-wide survival screen, in which each gene's effect on survival was established causally rather than correlatively, together with a public compound resource, the solver must produce a confidence-scored ranking of candidate compounds and, for each highly ranked compound of unestablished mechanism, a defended target and mechanism-of-action hypothesis supported by an auditable evidence chain. It must also judge whether the ranking transfers to an independent screen in a second disease; and where the evidence cannot support a committed call, that conclusion is itself scored, not forced.

Every claim is committed before any answer is revealed: each mechanistic hypothesis must trace to a citable source verified to exist and to support it, and a claim without such grounding is scored as unsupported rather than as a hedge. A control benchmark and a withheld slice of the ranking are sealed until after a submission freeze, and the scoring metrics and baselines are fixed in advance. Because the leading candidates' mechanisms and the eventual survival outcome are not yet established, the exercise rewards a defensible causal argument over pattern-matching to the literature.

The first problem instance is a Parkinson's-disease model, where a genome-wide survival screen in disease-model neurons yields protective and detrimental gene hits and a preliminary compound ranking that already sorts sensibly, with known neuroprotectants near the top and known cytotoxins near the bottom. It addresses a structural weakness in repurposing, where reversing a disease signature that mixes pathogenic, incidental, and compensatory changes can undo a protective adaptation as readily as it corrects a harmful one. The solver must integrate the screen signature with pharmacological, chemoinformatic, and patent literature and with independent human and model-organism evidence into a calibrated ranking and a defensible mechanistic account.

Evaluation is scored by a verifier on a shared auditable execution backbone once the sealed material is revealed. Each prediction is scored against held-out ground truth rather than a correlative proxy, matching the ranking against control-set enrichment and measured survival effect, a masked-mechanism set against known targets, and the reasoning trace against a validation protocol. It is benchmarked against a conventional signature-reversal floor, a random lower bound, and the partner's preliminary ranking, run inside and outside the environment; and on a longer horizon, the top candidates are advanced to a single-cell survival assay whose outcome does not yet exist.

This challenge exemplifies the intended role of discovery challenges in TRACES: the benchmark does not assume that the correct scientific answer is already available, but it does require the solver to produce a causal argument, a confidence-scored repurposing candidate, and a falsifiable mechanistic hypothesis whose validity can be progressively tested as sealed evidence is revealed and new experiments are run.

\section{Process Verification and Reality-Based Environment}
\label{sec:methods}

In this section, we describe the verification framework and executable environment infrastructure underlying Apodex Discovery, the mechanisms through which a discoverative AI's investigation becomes verifiable rather than merely plausible. We first introduce outcome and process verification, and then formalize the environment–task–episode abstraction that supports reproducible and repairable solver evaluation.
In our view, the main ingredient of our benchmark is its \textbf{verification} process: not only verifying the final output, but a complete framework that showcases verification for \emph{partial results} on \emph{decomposable tasks},
and a novel conceptual framework that captures agents' process integrity from six important aspects desirable for any heavy-duty solvers (HDS).
After introducing our verification framework, we introduce \textbf{executable environments}, a powerful, general-purpose repository that blends \emph{environments}, \emph{tasks}, \emph{episodes} and \emph{verifiers} into a single runnable, reproducible and repairable whole.

\subsection{Verification}
\label{sec:methods-verification}

Verification is the mechanism by which solver behavior is evaluated. We distinguish two complementary forms: \emph{outcome verification} and \emph{process verification}. Outcome verification evaluates what the solver produced. Process verification evaluates how the solver produced it. Both are necessary. Outcome scores measure task success, but they often provide sparse diagnostic information.
Additionally, in many situations outcome verification may be difficult or time-consuming to obtain due to proprietary data issues or wet-lab verification requirements.
On the other hand, 
process scores supply finer-grained evidence about tool use, evidence handling, hypothesis management, repair behavior, and other capabilities that determine whether a trajectory is reliable and trustworthy.

\begin{table}[tp]
\centering\small
\caption{Real-world questions available in our initial benchmark.}
\label{tab:env-tasks-episodes}
\setlength{\tabcolsep}{6pt}
\begin{tabularx}{\linewidth}{@{}l>{\raggedright\arraybackslash}X r@{}}
\toprule
Task & Description & Episodes \\
\midrule

\multicolumn{3}{@{}l}{\textbf{AAV capsid design}} \\

Viability prediction
& Predict whether an AAV capsid variant remains viable, meaning that it can assemble and package its genome, from its amino-acid sequence.
& 4 \\

Tropism prediction
& Predict the tissue- or cell-specific enrichment of AAV capsid variants, including generalization across organs, species, and human in-vitro assays.
& 3 \\

Structure prediction
& Predict the three-dimensional structure of an AAV capsid from sequence at the level of the surface loops, the full 60-mer assembly, and capsid--ligand complexes.
& 3 \\

Generative design
& Design a diverse set of novel, manufacturable, and on-target AAV capsid sequences under a limited experimental-feedback budget.
& 8 \\

\addlinespace
\multicolumn{3}{@{}l}{\textbf{Drug repurposing and reformulation}} \\

Drug--disease assessment
& Given each drug--disease pair and the evidence available at prediction time, estimate its promisingness and confidence against held-out clinical and regulatory outcomes, together with a detailed repurposing and reformulation procedure.
& 100 \\

\addlinespace
\multicolumn{3}{@{}l}{\textbf{Clinical trials}} \\

Statistical analysis and reporting
& Produce analysis tables, listings, and figures for a trial dataset under a pre-specified statistical analysis plan.
& 3 \\

\addlinespace
\multicolumn{3}{@{}l}{\textbf{Data for large language models}} \\

Pretraining-data procurement
& Assemble a high-value pretraining corpus from candidate sources under a fixed budget while avoiding contamination.
& 6 \\

Pretraining-data quality filtering
& Train a scorer that predicts multiple document-quality dimensions used to filter a corpus.
& 2 \\

Deduplication and decontamination
& Build a pipeline that removes duplicate and evaluation-contaminating documents from a corpus.
& 3 \\

Benchmark construction
& Generate difficult, self-contained evaluation items that strong models cannot easily solve.
& 16 \\

\addlinespace
\multicolumn{3}{@{}l}{\textbf{Training and systems for large language models}} \\

Pretraining speedrun
& Edit a training script to reach a target validation loss in the least wall-clock time.
& 3 \\

Reinforcement-learning recipe discovery
& Discover a training recipe that maximizes held-out gain from a limited-budget proxy.
& 4 \\

Inference determinism repair
& Locate and remove sources of nondeterminism in an inference stack so that outputs become reproducible.
& 3 \\

SWE-agent training
& Fine-tune a model to resolve real software-engineering issues.
& 1 \\

Solution selection and ranking
& Select or rank the correct solution among candidates without access to visible tests.
& 22 \\

Solution-verifier authoring
& Author a program that judges whether a model's solution to a problem is correct.
& 4 \\

Operator alignment
& Determine whether ported numerical kernels match their reference implementations.
& 33 \\

\bottomrule
\end{tabularx}
\end{table}

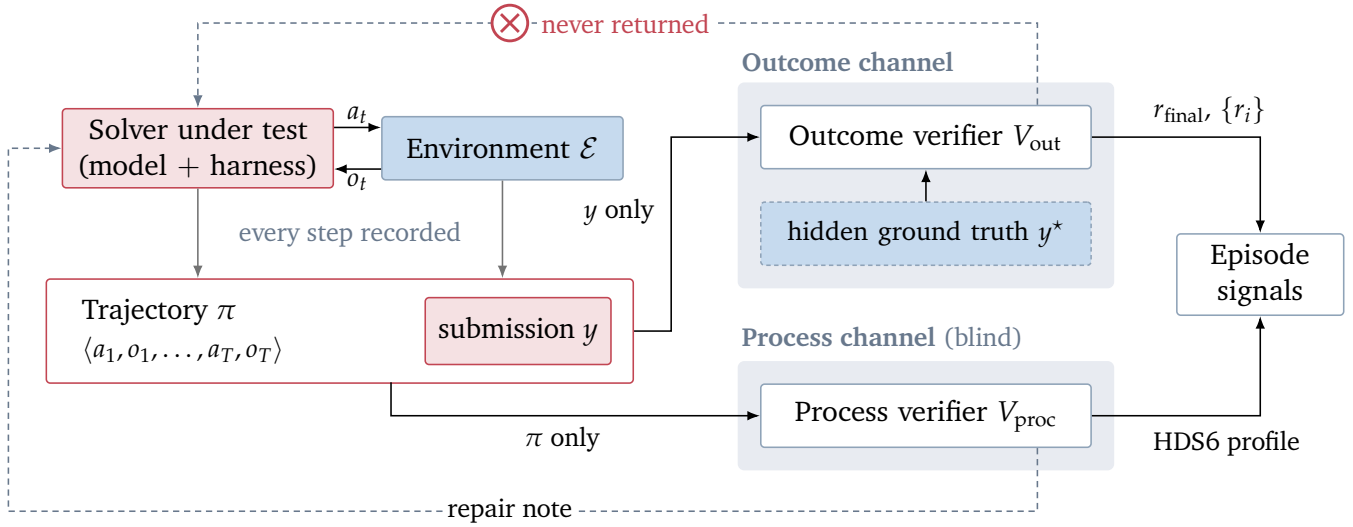
\begin{figure}[t]
\centering
\begin{adjustbox}{max width=\linewidth}
\begin{tikzpicture}[fg]

  \node[fgbox, text width=34mm, fill=fgAccBg, draw=fgAcc] (sut)
       at (1.4,0) {Solver under test\\(model $+$ harness)};
  \node[fgbox, text width=30mm, fill=fgB1] (env)
       at (5.5,0) {Environment $\mathcal{E}$};

  \draw[fgedge] ([yshift= 2.8mm]sut.east) --
        node[fglab, above]{$a_t$} ([yshift= 2.8mm]env.west);
  \draw[fgedge] ([yshift=-2.8mm]env.west) --
        node[fglab, below]{$o_t$} ([yshift=-2.8mm]sut.east);

  \node[fgbox, anchor=north west, minimum width=79mm, minimum height=14mm,
        fill=white, draw=fgAcc] (traj) at (-0.65,-1.75) {};
  \node[anchor=west, align=left] at ([xshift=3.5mm]traj.west)
       {Trajectory $\pi$\\[1pt]
        {\small$\langle a_1,o_1,\dots,a_T,o_T\rangle$}};
  \node[fgbox, anchor=east, minimum width=25mm, minimum height=9mm,
        fill=fgAccBg, draw=fgAcc] (ysub)
       at ([xshift=-3mm]traj.east) {submission $y$};

  \draw[fgfeed] (sut.south) -- (sut.south|-traj.north);
  \draw[fgfeed] (env.south) -- (env.south|-traj.north);
  \node[fgmut] at (3.45,-1.15) {every step recorded};

  \node[fgbox, text width=42mm, fill=white] (vout) at (11.2,0.15)
       {Outcome verifier $V_{\mathrm{out}}$};
  \node[fghid, text width=42mm, fill=fgB1] (ystar) at (11.2,-1.15)
       {\small hidden ground truth $y^\star$};
  \node[fgbox, text width=42mm, fill=white] (vproc) at (11.2,-3.6)
       {Process verifier $V_{\mathrm{proc}}$};

  \begin{scope}[on background layer]
    \node[fit=(vout)(ystar), inner sep=3mm, rounded corners=2.5pt,
          draw=none, fill=fgWash] (laneO) {};
    \node[fit=(vproc), inner sep=3mm, rounded corners=2.5pt,
          draw=none, fill=fgWash] (laneP) {};
  \end{scope}
  \node[fgttl, font=\small\bfseries, text=fgMute, anchor=south west]
       at ([yshift=0.8mm]laneO.north west) {Outcome channel};
  \node[fgttl, font=\small\bfseries, text=fgMute, anchor=south west]
       at ([yshift=0.8mm]laneP.north west) {Process channel \textnormal{(blind)}};

  \draw[fgedge] (ystar.north) -- (vout.south);

  \draw[fgedge] (traj.east) -- ++(0.5,0) |- (vout.west);
  \node[fglab, anchor=east] at (7.6,-0.85) {$y$ only};

  \draw[fgedge] (4.0,-3.15) |- (vproc.west);
  \node[fglab, anchor=north] at (6.3,-3.68) {$\pi$ only};

  \node[fgbox, text width=20mm, fill=white] (sig) at (15.7,-1.7)
       {Episode signals};
  \draw[fgedge] (vout.east)  -- (15.7,0.15) -- (sig.north);
  \draw[fgedge] (vproc.east) -- (15.7,-3.6) -- (sig.south);
  \node[fglab, anchor=west] at (14.2,0.52)  {$r_{\mathrm{final}},\,\{r_i\}$};
  \node[fglab, anchor=west] at (14.2,-3.97) {HDS6 profile};

  \draw[fgdash, draw=fgMute] ([xshift=15mm]vproc.south) -- ++(0,-0.85) -| (-1.15,0) -- (sut.west);
  \node[fglab, fill=white, inner sep=1.6pt] at (5.6,-4.875) {repair note};

  \draw[fgdash, draw=fgMute] ([xshift=15mm]vout.north) -- ++(0,1.1)
        coordinate (wh) -| (sut.north);
  \coordinate (whb) at (wh -| 5.6,0);
  \node[circle, draw=fgAcc, line width=1.1pt, fill=white,
        inner sep=0pt, minimum size=4.8mm] (stop) at (whb) {};
  \draw[fgAcc, line width=1.1pt]
        ([shift={(-1.2mm,-1.2mm)}]stop.center) -- ([shift={( 1.2mm, 1.2mm)}]stop.center)
        ([shift={(-1.2mm, 1.2mm)}]stop.center) -- ([shift={( 1.2mm,-1.2mm)}]stop.center);
  \node[fgred, anchor=west, fill=white, inner sep=1.6pt]
       at ([xshift=1.0mm]stop.east) {never returned};

\end{tikzpicture}
\end{adjustbox}
\caption{Verification channels. The solver and the environment interact through actions $a_t$ and observations $o_t$, and their interleaved sequence is the trajectory $\pi=\langle a_1,o_1,\dots,a_T,o_T\rangle$. The outcome verifier reads only the submitted artifact $y$ together with hidden ground truth $y^\star$, and its result is withheld from the solver (crossed edge across the top); the blind process verifier reads the trajectory alone, without the outcome or the solver's identity, and its diagnosis does return to the solver as a repair note that still discloses neither the outcome nor $y^\star$. Both feed the episode signals: the outcome scores $\{r_i\},\,r_{\mathrm{final}}$ and the HDS6 process profile.}
\label{fig:verification-channels}
\end{figure}

Figure~\ref{fig:verification-channels} summarizes the information flow across these two channels: the solver and the environment interact through actions and observations whose interleaved sequence is the trajectory, the hidden outcome verifier reads the submission against ground truth it never exposes, and the blind process verifier reads the trajectory alone.

\subsubsection{Outcome Verification}
\label{sec:methods-outcome-verification}

Outcome verification asks what the solver ultimately produced, scoring the final submission against success criteria known to the solver using scoring procedures and ground truth that are kept hidden from the solver. The form of the check follows the task: unit tests or exact-answer keys for code and mathematics, held-out labels or benchmark accuracy for prediction problems, a simulation endpoint for a proposed policy or design, and expert adjudication where no automatic criterion suffices.

\paragraph{Task decomposition and proxies.}
An episode is seldom a single answer. Formally, an episode-level verifier scores the final submission, $V_{\mathrm{out}}^{\mathrm{final}}(y, y^\star) \rightarrow r_{\mathrm{final}}$, and when a complex problem is decomposed into tasks $\tau_1,\ldots,\tau_n$ each task carries its own verifier, $V_{\mathrm{out}}^{(i)}(y_i, y_i^\star) \rightarrow r_i$, so that important intermediate results or milestones are checked without waiting for the completion of the final artifact. In a drug-repurposing episode, for instance, the retrieved mechanistic evidence, the constructed mechanism chain, and the final promisingness estimate are each verified in turn; in a capsid-design episode for the design of adeno-associated virus (AAV), the validity of a proposed sequence under the encoding and mutation rules is confirmed before its predicted fitness is scored. Such intermediate verification gives the diagnostic resolution to localize where in a long investigation a solver succeeded or failed, which a single terminal score cannot provide. 

For many of the problems that motivate this work, even the final check is unavailable, delayed, or only a \emph{shadow} of the objective that ultimately matters, and the verifier must fall back on a measurable proxy. For example, in AAV capsid engineering the aim is a vector that manufactures efficiently, evades pre-existing immunity, and transduces a target human tissue \emph{in vivo}. Since these definitive outcomes cannot be directly observed in a dry lab, a natural verification target is a held-out experimental proxy such as packaging fitness measured by high-throughput selection~\cite{bryant2021aav}. 

\paragraph{Anti-gaming as part of the verifier.}
Because many deliverables are trained models, curated corpora, or data pipelines rather than direct answers, the cheapest way to earn a high score without genuinely solving the task is to smuggle the hidden evaluation data into the submission---for example, by training on the very examples used to grade it. Each outcome verifier for environments or tasks where such gaming is likely therefore includes an explicit leakage check. A data-curation environment measures how much of the hidden evaluation text reappears in the solver's training material and voids the episode once that overlap is too large; a deduplication environment requires the submitted pipeline to remove every copy of the evaluation data; and a data-procurement environment penalizes spam or duplicated filler padded into the assembled corpus. Each check is a hard gate, so a violation disqualifies the episode outright rather than lowering its score.

\paragraph{Normalization against baselines.}
Every outcome score is reported on a common scale from $0$ to $1$. Two anchors, fixed when the instance is built rather than chosen by hand, make this possible: a floor that a trivial policy would reach---a random guess or a default recipe---which maps to $0$, and a ceiling representing the best result known to be attainable---an oracle or the strongest recorded solution---which maps to $1$; the raw measurement is then linearly rescaled between the two. A score is thus read as a position between ``no better than a trivial baseline'' and ``at the known frontier,'' rather than as a raw quantity whose magnitude is hard to compare across tasks. When several quality dimensions are combined, as in a document-scoring environment, each is rescaled against its own floor and ceiling and measured by the statistic suited to its label distribution---correlation for broad-support dimensions, mean absolute error for narrow ones, and detection scores for rare positives---so that no dimension dominates through an accident of scale.

\subsubsection{Process Verification}
\label{sec:methods-process-verification}

Process verification reads the recorded trajectory rather than the submitted artifact, asking not what the solver produced but how it produced it. The trajectory is treated as an authoritative log: each step carries its action, the resource it charged, and the observation or error it returned, and rejected or malformed attempts are retained alongside successful ones.

We operationalize process verification through a single metric, HDS6, which we take to be the bedrock of heavy-duty-solver (HDS) capability, since its dimensions name the competencies that separate a system able to sustain a long, self-correcting investigation from one that merely assembles a plausible account. These six capabilities are the process competencies on which any discoverative AI must be judged. HDS6 scores a trajectory along six dimensions, denoted by the mnemonic TRACES:
\begin{itemize}
    \item \textbf{Tools}: whether the solver selects, invokes, and interprets tools correctly. Heavy-duty work is carried out through tools, so a wrong tool, a malformed call, or a result never absorbed into or improperly interpreted by the solver's state leaves the reasoning running open-loop, disconnected from the real state of the world.
    \item \textbf{Repair}: whether the solver can detect and self-correct incorrect steps, or respond to verification feedback or failures in an effective manner, rather than repeating the same mistakes; such repair or self-correction capabilities are central to a discoverative AI's ability to sustain ultra long-chain reasoning and working efforts;
    \item \textbf{Alternatives}: whether the solver makes competing hypotheses explicit, ranks them, and updates the ranking as evidence arrives. Premature commitment to a single explanation is the dominant failure mode in open-ended discovery, and disciplined management of alternatives is how a solver avoids confident but wrong conclusions.
    \item \textbf{Coherence}: whether the solver maintains long-horizon state, constraints, and consistency across the trajectory. Over many steps a solver that loses track of its own commitments produces a final answer that silently contradicts its intermediate work, which should be avoided by a disocoverative AI system that can properly maintain long-term coherence over hundreds of steps or turns;
    \item \textbf{Evidence}: whether every claim is faithfully grounded in observations, tool outputs, references, or data. A single unsupported or fabricated claim invalidates the conclusions that depend on it, so grounding is what distinguishes a defensible result from a plausible or even hallucinated guess.
    \item \textbf{Scope}: whether the solver states the boundaries, uncertainty, failure modes, and applicability limits of its result. A conclusion offered without the conditions under which it holds is unsafe to act on, however correct it may be within those conditions.
\end{itemize}
Taken together, these dimensions assess whether a conclusion was \emph{earned} by the process that produced it.

We score the trajectory against these six capabilities in two complementary ways, described below, and in both the process verifier operates blind: it cannot access the verified outcome, the solver's hidden or proprietary chain-of-thought is stripped from the record before judging, and the solver's identity is withheld, so that a score reflects only the externally visible behavior of the trajectory. There is also a single ``integrity gate'' that detects the most serious process deficiency such as phantom tool calls or fabricated tool call results, which once detected would invalidate the entire trajectory.

\paragraph{HDS6 subrubric-based scoring.}
The first approach scores the trajectory against a fixed checklist of these six capabilities. Each capability is separated into several subcapabilities, twenty-five in total across the six, and each subcapability carries a \emph{subrubric}: a scoring item graded on a short ordinal band ($0/1/2$) whose anchors are authored per task, so that the capability being measured is shared across environments while the behavior that earns each band is specific to the task at hand. Table~\ref{tab:subrubric-example} illustrates one such subrubric, claim grounding within the Evidence capability, authored for two unrelated environments: the capability, the three-band scale, and the aggregation are identical, yet the trajectory behavior that earns each band differs entirely. A separate integrity gate detects the most serious process deficiency such as phantom tool calls or fabricated tool call results, which if detected would invalidate the entire trajectory. The grading is carried out by an agentic procedure rather than a single prompt: one model proposes a band with quotes cited to specific steps, a second is required to argue the strongest case for a different band, and a third arbitrates any disagreement that remains, and every quote is re-grounded against the log so that a judge cannot support a band with text the trajectory does not contain.

\begin{table}[t]
\centering\small
\caption{One subrubric, two environments. Both columns instantiate the same subrubric, \emph{claim grounding} (a subcapability of Evidence), scored at bands $0/1/2$. The capability being scored is identical; the trajectory evidence that earns each band is authored to the task.}
\label{tab:subrubric-example}
\begin{tabularx}{\linewidth}{@{}c>{\raggedright\arraybackslash}X>{\raggedright\arraybackslash}X@{}}
\toprule
Band & Protein engineering (model viability) & Inference nondeterminism (root-cause repair) \\
\midrule
0 & Claims model quality with no held-out evidence and no comparison to a naive baseline. & A key conclusion (root cause, or which configurations diverge or are fixed) has no supporting probe evidence, or contradicts a probe result already shown in the trace. \\[2pt]
1 & Reports a validation check but does not position it against the stated additive and distance baselines. & Has evidence but the chain is incomplete, or draws a determinacy conclusion over unswept axes (other backends, cache dtypes) from a few probed corners. \\[2pt]
2 & Grounds claims in validation performance and explicitly positions the model against the naive baselines. & Every load-bearing claim traces to a specific probe output across mode and configuration, covering confirming re-runs and residual divergences, and declares unswept axes rather than assuming coverage. \\
\bottomrule
\end{tabularx}
\end{table}

\paragraph{Load-bearing step-based scoring.}
The second approach does not apply a fixed checklist but first identifies which steps of a particular trajectory actually matter. It reconstructs the run as a dependency graph and propagates influence backward from the committed answer, weighting each step by how much the final result rests on it; the steps that carry most of this weight are the \emph{load-bearing} ones. Each load-bearing step is then re-solved independently, by a panel of models that never sees the solver's own reasoning, and the step is scored by how far it departs from that independent solution---a continuous gap rather than a discrete band. These per-step gaps, weighted by load-bearing influence and by the panel's confidence, aggregate into a process-quality score and, more usefully, into targeted feedback that localizes where a trajectory went wrong. 
Note that here ``load-bearing steps'' should not be confused with task decomposition and hidden outcome verifiers for intermediate results introduced in the previous section: load-bearing steps are organically detected and constructed by the process verifier for a single trajectory, while task decomposition and the design of verifiers for intermediate outcomes is locked in when environment is designed and is out of the control from both the system under test, or the process verifier.

\paragraph{Structured repair feedback.}
Both scoring approaches yield not only a number but a structured diagnostic that can be returned to the solver, which is what closes the measure--diagnose--improve loop essential for a discoverative AI system. From the subrubric checklist, each subcapability that scored below its top band contributes a proposed \emph{band transition}: the band the trajectory received, paired with the specific behavior a higher band would require. From the load-bearing analysis, each faulty step contributes feedback localized to that step, whose precision tracks the re-solving panel's confidence, with a confident departure yielding a full correction stating where the step went wrong, why, what it should have been, and what to check next.

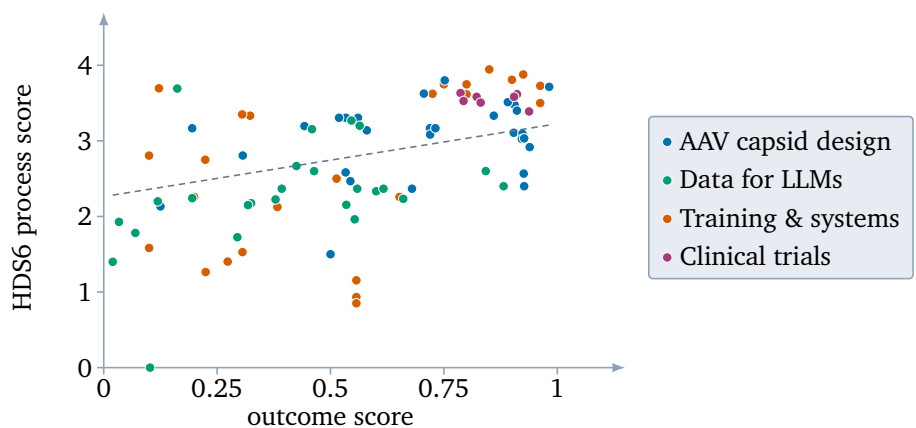
\begin{figure}[t]
\centering
\begin{adjustbox}{max width=\linewidth}
\begin{tikzpicture}[fg]
  \draw[draw=fgRule,->] (0,0)--(6.9,0);
  \draw[draw=fgRule,->] (0,0)--(0,4.7);
  \foreach \x/\lab in {0/0,1.5/0.25,3/0.5,4.5/0.75,6/1}{
    \draw[draw=fgRule] (\x,0)--(\x,-0.09) node[fglab, below]{\lab};}
  \foreach \y in {0,1,2,3,4}{
    \draw[draw=fgRule] (0,\y)--(-0.09,\y) node[fglab, left]{\y};}
  \node[fglab, below=5mm] at (3,0) {outcome score};
  \node[fglab, rotate=90, anchor=center] at (-1.05,2.25) {HDS6 process score};
  \foreach \x/\y in {3.913/2.258, 1.196/2.258, 0.600/2.805, 0.600/1.583, 3.342/1.155, 3.342/0.934, 3.342/0.851, 5.774/3.500, 1.348/1.264, 0.731/3.695, 3.080/2.500, 1.835/1.528, 2.297/2.125, 1.640/1.403, 1.343/2.750, 1.936/3.333, 1.831/3.350, 4.500/3.748, 5.550/3.878, 4.800/3.615, 5.775/3.728, 5.400/3.808, 4.800/3.748, 4.350/3.623, 5.100/3.945}
    {\filldraw[fill=catB, draw=white, line width=0.3pt](\x,\y)circle(1.8pt);}
  \foreach \x/\y in {5.434/3.571, 3.000/1.500, 5.160/3.333, 5.893/3.714, 5.424/3.107, 5.435/3.476, 5.346/3.512, 5.635/2.917, 5.539/3.095, 5.531/3.024, 5.556/2.567, 0.750/2.133, 4.235/3.625, 5.562/2.400, 3.263/2.467, 4.511/3.800, 5.466/3.400, 1.170/3.167, 5.562/3.033, 4.078/2.367, 3.364/3.305, 3.204/3.305, 4.315/3.167, 1.839/2.806, 3.481/3.139, 3.113/3.305, 4.318/3.083, 3.200/2.583, 4.388/3.167, 2.656/3.195}
    {\filldraw[fill=catA, draw=white, line width=0.3pt](\x,\y)circle(1.8pt);}
  \foreach \x/\y in {0.613/0.000, 2.355/2.367, 3.604/2.333, 3.354/2.367, 3.962/2.233, 5.290/2.400, 0.715/2.200, 5.054/2.600, 2.783/2.600, 3.701/2.367, 0.119/1.400, 1.950/2.175, 1.171/2.242, 0.202/1.928, 1.909/2.150, 2.273/2.225, 1.768/1.725, 0.418/1.782, 0.973/3.692, 3.277/3.269, 3.209/2.154, 2.549/2.667, 3.320/1.962, 3.386/3.200, 2.755/3.154}
    {\filldraw[fill=catC, draw=white, line width=0.3pt](\x,\y)circle(1.8pt);}
  \foreach \x/\y in {4.933/3.583, 5.627/3.389, 5.467/3.617, 4.720/3.633, 5.428/3.583, 4.760/3.528, 4.987/3.506}
    {\filldraw[fill=catD, draw=white, line width=0.3pt](\x,\y)circle(1.8pt);}
  \draw[draw=fgMute, dash pattern=on 2.4pt off 1.8pt] (0.119,2.282)--(5.893,3.209);
  \node[draw=fgRule, line width=0.6pt, rounded corners=1.5pt, fill=fgWash,
        inner sep=5pt, anchor=west, align=left,
        font=\small] at (7.2,2.25) {%
    \tikz[baseline=-0.6ex]\filldraw[fill=catA,draw=white,line width=0.3pt](0,0)circle(1.8pt);~AAV capsid design\\[2.5pt]
    \tikz[baseline=-0.6ex]\filldraw[fill=catC,draw=white,line width=0.3pt](0,0)circle(1.8pt);~Data for LLMs\\[2.5pt]
    \tikz[baseline=-0.6ex]\filldraw[fill=catB,draw=white,line width=0.3pt](0,0)circle(1.8pt);~Training \& systems\\[2.5pt]
    \tikz[baseline=-0.6ex]\filldraw[fill=catD,draw=white,line width=0.3pt](0,0)circle(1.8pt);~Clinical trials};
\end{tikzpicture}
\end{adjustbox}
\caption{Each point is a judged trajectory with outcome and HDS6 scores. For clarity, fully-failed (outcome $0$) and fully-solved (outcome $1$) trajectories are omitted, and trajectories are subsampled to at most ten per task and coloured by problem family. The dashed line is the least-squares fit over all $283$ intermediate-outcome trajectories (HDS6 $\approx 2.26 + 0.96\times$ outcome).
}
\label{fig:calibration-scatter}
\end{figure}

\paragraph{Calibration and alignment with outcome scores}
Albeit being primarily process centric, we also check the calibration and alignment of the HDS6 metric with outcome scores (from hidden outcome verifiers) to complement our understanding of the TRACES scores. Over the $409$ judged trajectories that carry a numeric outcome we correlate each trajectory's outcome score with its HDS6 process score, pooled within each environment (Table~\ref{tab:calibration}) and shown trajectory by trajectory in Figure~\ref{fig:calibration-scatter}. All four problem families show positive correlations---Spearman $\rho$ ranges from $+0.24$ on the AAV capsid-design pipeline to $+0.65$ on training-and-systems tasks, with comparable Pearson $r$, and the pooled coefficient is $+0.51$ ($r=+0.56$)---so the blind process judge and the hidden verifier largely concur on which trajectories are good even though neither can see the other. Because each task carries its own outcome verifier, these pooled figures mix per-task scales and are read as agreement summaries rather than calibrated magnitudes; taken one task at a time the agreement is often sharper, with a median $\rho$ near $0.5$ that reaches $0.8$--$0.9$ where the task admits a clear success signal.


\begin{table}[t]
\centering\small
\setlength{\tabcolsep}{9pt}
\caption{Outcome--HDS6 rank correlation (Spearman $\rho$) and linear correlation (Pearson $r$) over the $409$ judged trajectories that carry a numeric outcome, pooled within each environment of Table~\ref{tab:env-tasks-episodes}; 
Each environment pools several tasks whose outcome verifiers use different scales, so the coefficients are read as agreement summaries rather than calibrated magnitudes.}
\label{tab:calibration}
\begin{tabular}{@{}l r r r@{}}
\toprule
Environment & $n$ & Spearman $\rho$ & Pearson $r$ \\
\midrule
AAV capsid design                              & $220$ & $+0.24$ & $+0.26$ \\
Clinical trials                                & $13$  & $+0.47$ & $+0.62$ \\
Data for large language models                 & $62$  & $+0.46$ & $+0.51$ \\
Training and systems for large language models & $114$ & $+0.65$ & $+0.74$ \\
\midrule
All environments (pooled)                      & $409$ & $+0.51$ & $+0.56$ \\
\bottomrule
\end{tabular}
\end{table}

\subsection{Environments, Tasks, and Episodes}
\label{sec:methods-environments}

This section describes the machinery of an \emph{executable environment}, our foundation to a heavy-duty system for discoverative AI. There are three nested constructs---\emph{environments}, \emph{tasks}, and \emph{episodes}---which, together with the verification agent teams that grade the work done inside them, make up a complete runnable, verifiable, reproducible environment. From a solver system's perspective, an environment exposes an action interface, the available tools, and the resource budgets, and it returns fresh observations as the solver acts rather than serving a fixed prompt. A task is a problem posed within an environment, representing the entire question or an important step at which final or intermediate results can be verified where they arise; an episode is a single runnable instance of that task, provisioned with its own data, budgets, and hidden ground truth, and it records the complete trajectory the solver produces. Overlaying all three are the verification agent teams of Section~\ref{sec:methods-verification}---a hidden outcome verifier over the submitted artifact and the blind process judges over the trajectory---which score each episode. An environment, its tasks and episodes, and these verifiers together form one runnable, reproducible, and repairable realization of an otherwise open-ended problem.
The remainder of this section defines each construct in turn.

\begin{figure}[t]
\centering
%
%
\newcommand{\bandicon}[1]{\tikz[baseline=0pt]{%
  \foreach \i in {1,2}{%
    \ifnum\i>#1\relax\def\cc{fgB0}\else\def\cc{fgB3}\fi
    \draw[fill=\cc, draw=fgRule, line width=0.4pt]
          (\i*1.5mm-1.5mm,0) rectangle ++(1.1mm,2.2mm);}}}
\newcommand{\objicon}{\tikz[baseline=-0.55ex]{%
  \draw[draw=fgAcc, line width=0.7pt, fill=white] (0,0) circle (1.45mm);
  \draw[draw=fgAcc, line width=0.7pt, line cap=round] (0,0.62mm) -- (0,-0.15mm);
  \fill[fgAcc] (0,-0.72mm) circle (0.24mm);}}
\newcommand{\okicon}{\tikz[baseline=-0.55ex]{%
  \draw[draw=fgMute, line width=0.7pt, fill=white] (0,0) circle (1.45mm);
  \draw[draw=fgMute, line width=0.8pt, line cap=round, line join=round]
        (-0.68mm,0.05mm) -- (-0.15mm,-0.6mm) -- (0.8mm,0.72mm);}}
\ifdefined\hdscolh\else\newlength{\hdscolh}\fi
\setlength{\hdscolh}{78mm}
\begin{minipage}[b][\hdscolh][c]{0.29\linewidth}
\centering
\begin{tikzpicture}[fg, remember picture, font=\small, scale=1.0]
  \def\R{1.95}
  \foreach \g in {1,2,3,4}{
    \draw[fgRule!30, line width=0.5pt]
      (90:{\R*\g/4}) -- (30:{\R*\g/4}) -- (-30:{\R*\g/4}) --
      (-90:{\R*\g/4}) -- (-150:{\R*\g/4}) -- (150:{\R*\g/4}) -- cycle;
  }
  \foreach \a in {90,30,-30,-90,-150}{
    \draw[fgRule!55, line width=0.5pt] (0,0) -- (\a:\R); }
  \draw[fgRule!55, line width=0.5pt, densely dashed] (0,0) -- (150:\R);

  \node[above]            at (90:{\R+0.10})   {C};
  \node[right] (Elab)     at (30:{\R+0.10})   {E};
  \node[right]            at (-30:{\R+0.10})  {A};
  \node[below]            at (-90:{\R+0.10})  {S};
  \node[left]             at (-150:{\R+0.10}) {T};
  \node[left, text=fgMute] at (150:{\R+0.10})  {R\,{\small(n/a)}};

  \draw[line width=0.9pt, draw=fgB4, fill=fgB4, fill opacity=0.14]
    (90:{\R*4.0/4}) -- (30:{\R*4.0/4}) -- (-30:{\R*4.0/4}) --
    (-90:{\R*4.0/4}) -- (-150:{\R*3.6/4}) -- (0,0) -- cycle;
  \draw[line width=0.9pt, densely dashed, draw=fgAcc, fill=fgAcc, fill opacity=0.10]
    (90:{\R*3.0/4}) -- (30:{\R*2.0/4}) -- (-30:{\R*4.0/4}) --
    (-90:{\R*3.5/4}) -- (-150:{\R*4.0/4}) -- (0,0) -- cycle;
\end{tikzpicture}\\[4pt]
{\small
 \tikz[baseline=-0.6ex]\draw[fgB4, line width=0.9pt] (0,0)--(1.2em,0);~traj.\ A (overall $3.9$)\\[2pt]
 \tikz[baseline=-0.6ex]\draw[fgAcc, line width=0.9pt, densely dashed] (0,0)--(1.2em,0);~traj.\ B (overall $3.2$)}
\end{minipage}\hspace{0.05\linewidth}%
\begin{minipage}[b]{0.20\linewidth}
\centering
\begin{tikzpicture}[fg, remember picture, font=\small,
  srow/.style={draw=fgRule, line width=0.6pt, rounded corners=1.5pt, fill=white,
               align=left, inner sep=3pt, text width=27mm, minimum height=11.5mm,
               execute at begin node={\hyphenpenalty=10000\exhyphenpenalty=10000\relax}}]
  \node[fgttl, font=\small\bfseries] (ehdr) at (0,0) {Evidence $=2.0$};
  \node[srow, below=3.2mm of ehdr] (r1)
       {claim grounding\\[2pt]\hbox to 27mm{\bandicon{1}~band 1\hfill w\,2}};
  \node[srow, below=3.2mm of r1, draw=fgAcc] (r2)
       {source integrity\\[2pt]\hbox to 27mm{\bandicon{0}~band \textbf{0}\hfill w\,2}};
  \node[srow, below=3.2mm of r2] (r3)
       {key-fact correctness\\[2pt]\hbox to 27mm{\bandicon{2}~band 2\hfill w\,2}};
  \node[srow, below=3.2mm of r3] (r4)
       {fact / inference / unknown\\[2pt]\hbox to 27mm{\bandicon{1}~band 1\hfill w\,1}};
  \node[fgmut, below=3.2mm of r4, align=center, font=\footnotesize] (esum)
       {$2\times\frac{2\cdot1+2\cdot0+2\cdot2+1\cdot1}{7}=2.0$};
\end{tikzpicture}\\[4pt]
{\small traj.\ B, four subrubrics}
\end{minipage}\hspace{0.04\linewidth}%
\begin{minipage}[b]{0.42\linewidth}
\centering
\begin{tikzpicture}[fg, remember picture, font=\small,
  turn/.style={draw=fgRule, line width=0.6pt, rounded corners=1.5pt,
               align=left, inner sep=4pt, text width=70mm,
               execute at begin node={\hyphenpenalty=10000\exhyphenpenalty=10000\relax}},
  jd/.style ={turn, draw=fgB3,  fill=fgB0},
  rv/.style ={turn, draw=fgAcc, fill=fgAccBg},
  fnl/.style={turn, draw=fgRule, fill=fgWash, align=center},
]
  \node[jd] (t1) {\bandicon{2}~\textbf{Judge} $\Rightarrow$ band \textbf{2}\\[1.5pt]
    ``both verdicts cite concrete $K$ measurements by dtype and magnitude---enough to ground them'' \,[step 69]};
  \node[rv, below=3.5mm of t1] (t2) {\objicon~\textbf{Reviewer} $\Rightarrow$ \textbf{Revise}\\[1.5pt]
    ``$K_{\mathrm{bwd}}{=}11744$ appears in no probe (observed $10076$, $7960$); a top band needs each verdict grounded in an \textbf{observed} value''};
  \node[jd, below=3.5mm of t2] (t3) {\bandicon{1}~\textbf{Judge} $\Rightarrow$ band \textbf{1}\\[1.5pt]
    ``agreed---$11744$ is only in the submission text, not the probes; not $0$, since other values are real''};
  \node[fnl, below=3.5mm of t3] (t4) {\okicon~\textbf{Reviewer} \textbf{Accept} $\;\Rightarrow\;$ \textbf{final band 1}};

  \draw[fgedge] (t1) -- (t2);
  \draw[fgedge] (t2) -- (t3);
  \draw[fgedge] (t3) -- (t4);
\end{tikzpicture}\\[4pt]
{\small lane: claim grounding (Evidence)}
\end{minipage}%
\begin{tikzpicture}[remember picture, overlay,
  bub/.style={draw=fgRule, line width=0.5pt, fill=none, rounded corners=4pt}]
  \node[fit=(ehdr)(r1)(r4)(esum), inner xsep=2.2mm, inner ysep=1.8mm] (midbox) {};
  \node[fit=(t1)(t4)(ehdr)(esum), inner ysep=1.8mm] (span) {};
  \coordinate (mt)  at (midbox.west |- Elab);
  \coordinate (mnw) at (midbox.west |- span.north);
  \coordinate (mne) at (midbox.east |- span.north);
  \coordinate (mse) at (midbox.east |- span.south);
  \coordinate (msw) at (midbox.west |- span.south);
  \draw[bub] (mnw) -- (mne) -- (mse) -- (msw) -- ([yshift=-2.4mm]mt) [sharp corners]
             -- ([xshift=-3.4mm]mt) -- ([yshift=2.4mm]mt) [rounded corners=4pt] -- cycle;
  \node[fit=(t1)(t4), inner xsep=2.2mm, inner ysep=1.8mm] (lanebox) {};
  \coordinate (lt)  at (lanebox.west |- r1);
  \coordinate (lnw) at (lanebox.west |- span.north);
  \coordinate (lne) at (lanebox.east |- span.north);
  \coordinate (lse) at (lanebox.east |- span.south);
  \coordinate (lsw) at (lanebox.west |- span.south);
  \draw[bub] (lnw) -- (lne) -- (lse) -- (lsw) -- ([yshift=-2.4mm]lt) [sharp corners]
             -- ([xshift=-3.4mm]lt) -- ([yshift=2.4mm]lt) [rounded corners=4pt] -- cycle;
\end{tikzpicture}
\caption{HDS6 on two trajectories for one operator-alignment task (does a ported numerical kernel match its reference?). Left: the six capability scores ($0$--$4$; rings mark $1$--$4$); both trajectories pass the integrity gate, Repair is n/a for this single-submission task, and the two differ most on Evidence ($4.0$ versus $2.0$). Middle: that Evidence score, decomposed into its four subrubrics, each scored $0$--$2$ and weighted $2{:}2{:}2{:}1$, whose weighted mean doubles to $2.0$; the heaviest loss is source integrity at band $0$. Right: the \emph{claim grounding} lane, where the reviewer disputes the judge's top band---a cited backward figure ($K{=}11744$) appears in no probe---so the judge lowers it to $1$ and the reviewer accepts.}
\label{fig:hds6-showcase}
\smallskip
\parbox{\linewidth}{\footnotesize\emph{Claim-grounding band anchors for this task.}\enspace \textbf{0}: the verdict cites no concrete $K$ value or divergence.\enspace \textbf{1}: cites a number but not enough to separate aligned from misaligned (one dtype only).\enspace \textbf{2}: each verdict carries the specific $K$ measurements (dtype\,/\,magnitude\,/\,pass) that ground it.}
\end{figure}
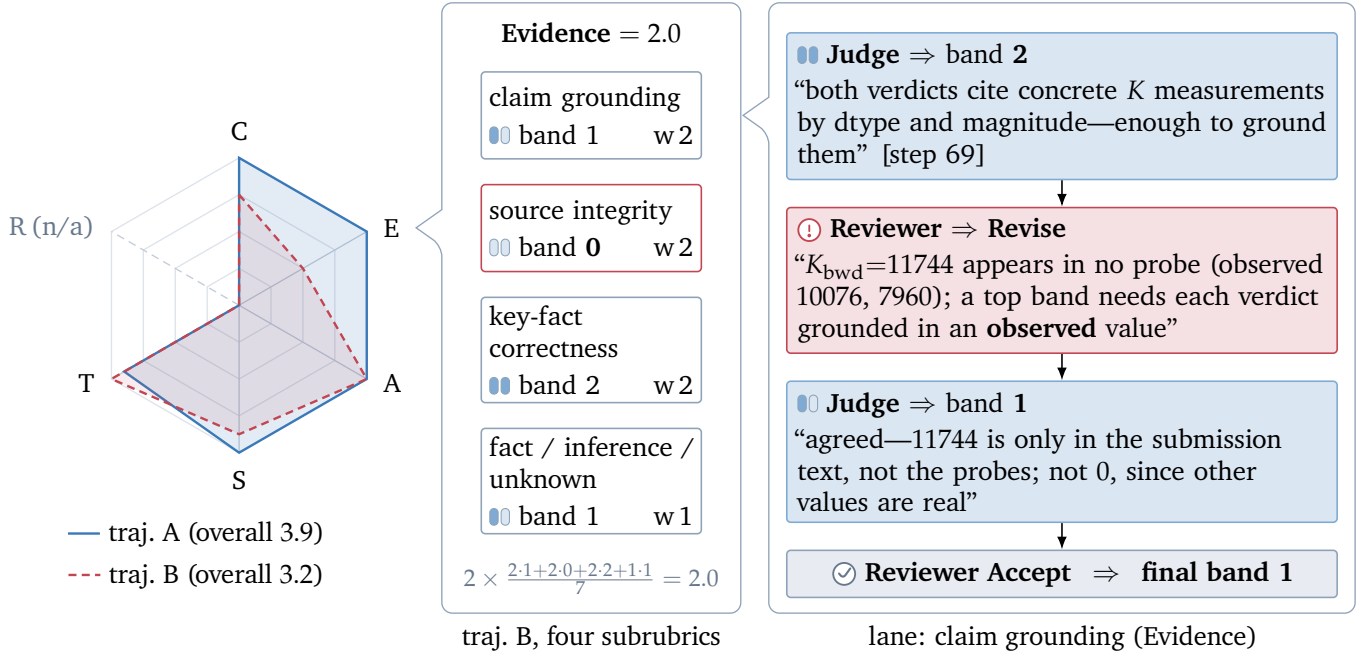

Figure~\ref{fig:hds6-showcase} shows the pieces working together on one such task: two solver trajectories carried through to scoring, with their HDS6 capability profiles and one item lane resolved by the judging agents of Section~\ref{sec:methods-verification}. Constructing an executable environment of this kind from an open-ended research problem is the labor-intensive part of this work, and we treat it as a methodological contribution in its own right: for the problems considered here, building a faithful environment with a trustworthy hidden verifier is at least as difficult as the task the solver is asked to perform.

\subsection{Environments}
\label{sec:methods-environment-definition}

An \emph{environment} is the operational substrate of a benchmark or solver runtime. It may include file systems, databases, APIs, code execution sandboxes, simulators, laboratory interfaces, retrieval systems, hidden labels, and evaluator services. It also specifies resource constraints, such as time limits, compute budgets, tool-call limits, safety boundaries, and data-access rules. In practice an environment is declared by a \emph{manifest}: a full specification of the task, the target outcome, the hidden verifier, the data provided to the solver, and the tools it is permitted to call, together with the resource budgets above. The manifest is the static contract from which the environment's episodes are instantiated with the right configurations of data, tools, verifiers, scoring rules, and other important configuration parameters.

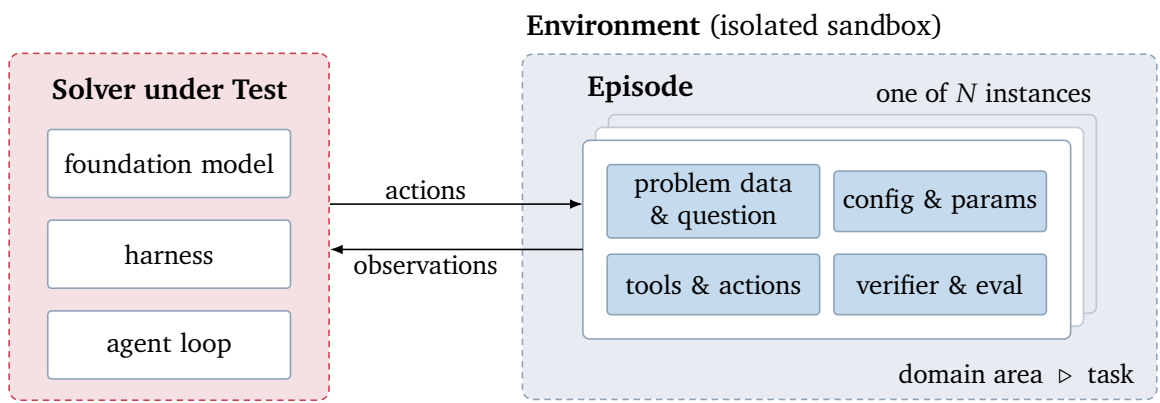
\begin{figure}[t]
\centering
\begin{adjustbox}{max width=\linewidth}
\begin{tikzpicture}[fg,
  fglab/.append style={text=fgInk},
  fgmut/.append style={text=fgInk},
  slot/.style={draw=fgRule, line width=0.6pt, rounded corners=1.5pt,
               align=center, font=\small, inner sep=3pt,
               text width=26mm, minimum height=8mm, fill=fgB1},
]
  \node[slot, fill=white, text width=30mm, minimum height=9mm] (smdl) at (0, 1.05) {foundation model};
  \node[slot, fill=white, text width=30mm, minimum height=9mm] (shrn) at (0,-0.15) {harness};
  \node[slot, fill=white, text width=30mm, minimum height=9mm] (sagt) at (0,-1.35) {agent loop};

  \node[slot] (d1) at (7.2, 0.55) {problem data \& question};
  \node[slot] (d2) at (10.2,0.55) {config \& params};
  \node[slot] (d3) at (7.2,-0.55) {tools \& actions};
  \node[slot] (d4) at (10.2,-0.55) {verifier \& eval};

  \node[fit=(d1)(d2)(d3)(d4), inner sep=3mm, draw=none, fill=none] (epbb) {};
  \coordinate (cardTR) at ([xshift=3.4mm,yshift=3.4mm]epbb.north east);

  \begin{scope}[on background layer]
    \node[draw=fgRule, line width=0.6pt, dash pattern=on 2.4pt off 1.8pt,
          rounded corners=4pt, fill=fgWash,
          fit=(epbb)(cardTR), inner sep=8mm] (env) {};
    \draw[rounded corners=2.5pt, draw=fgRule!60, line width=0.6pt, fill=fgWash]
      ([xshift=3.4mm,yshift=3.4mm]epbb.south west)
      rectangle ([xshift=3.4mm,yshift=3.4mm]epbb.north east);
    \draw[rounded corners=2.5pt, draw=fgRule!60, line width=0.6pt, fill=white]
      ([xshift=1.7mm,yshift=1.7mm]epbb.south west)
      rectangle ([xshift=1.7mm,yshift=1.7mm]epbb.north east);
    \draw[rounded corners=2.5pt, draw=fgRule, line width=0.6pt, fill=white]
      (epbb.south west) rectangle (epbb.north east);

    \coordinate (sutT) at (smdl.center |- env.north);
    \coordinate (sutB) at (smdl.center |- env.south);
    \node[rounded corners=4pt, draw=fgAcc, line width=0.6pt, fill=fgAccBg,
          dash pattern=on 2.4pt off 1.8pt,
          fit=(smdl)(shrn)(sagt)(sutT)(sutB), inner xsep=5mm, inner ysep=0pt] (sut) {};
  \end{scope}

  \node[anchor=north, fgttl] at ([yshift=-2.6mm]sut.north) {Solver under Test};

  \node[anchor=south west, fgttl] at ([yshift=4.4mm]epbb.north west)
       {Episode};
  \node[anchor=south east, fgmut] at ([yshift=1.0mm]cardTR) {one of $N$ instances};
  \node[anchor=south west, fgttl] at ([yshift=1.2mm]env.north west)
       {Environment \normalfont(isolated sandbox)};
  \node[anchor=south east, fgmut] at ([xshift=-2.5mm,yshift=1.6mm]env.south east)
       {domain area $\;\triangleright\;$ task};

  \draw[fgedge] ([yshift=3mm]sut.east) -- ([yshift=3mm]sut.east -| epbb.west);
  \draw[fgedge] ([yshift=-3mm]sut.east -| epbb.west) -- ([yshift=-3mm]sut.east);
  \coordinate (gapc) at ($(sut.east)!0.5!(env.west)$);
  \node[fglab, above] at ([yshift= 3mm]gapc) {actions};
  \node[fglab, below] at ([yshift=-3mm]gapc) {observations};
\end{tikzpicture}
\end{adjustbox}
\caption{Environment, tasks, and episodes. An environment is an \emph{isolated} sandbox with which a Solver under Test (SuT) interacts through actions and observations. Within it, each task or combination of tasks is realized by many episodes, its runnable instances. An episode is the concrete interface between the environment and the solver, and the ground truth stays inside the sandbox and never reaches the solver.}
\label{fig:env-task-episode}
\smallskip
\parbox{\linewidth}{\footnotesize
\textbf{What each episode fixes.}\enspace \emph{Problem data \& question}: the instance's inputs and the criterion for what counts as solved. \emph{Config \& parameters}: the seeds, budgets, and difficulty settings that make the instance concrete. \emph{Tools \& actions}: the action interface the solver may call, and nothing beyond it. \emph{Verifier \& eval}: the hidden grader and scoring rule applied to the final submission. The same task or combination/sequence of tasks is instantiated as many such episodes, so a system is measured across instances rather than on a single case, and each episode's full trajectory is recorded for both outcome and process verification.}
\end{figure}

The defining property of an environment is that it returns observations in response to solver actions. These observations may include tool outputs, execution errors, retrieved documents, simulation results, experimental measurements, verifier feedback, or human-entered data. Because of this, an environment is much more than a single prompt which, even if carefully crafted, provides only static contexts, whereas an environment can produce new information as the solver acts. Figure~\ref{fig:env-task-episode} shows this structure and, in particular, what a single episode exposes to the solver.

\subsubsection{Tasks}
\label{sec:methods-tasks}

A task is a scoped work unit within an environment, defined by its input, its expected output, the tools it permits, its success criteria, and its verification procedures. Tasks are important for two reasons.

First, tasks decompose a complex real-world problem into manageable sub-problems that can be verified where they arise, rather than inferred from a single terminal result. For example, our AAV capsid-design environment decomposes the discovery pipeline into four tasks: predicting whether a capsid variant remains viable, predicting its tissue or cellular tropism, reconstructing its three-dimensional structure, and generating novel, manufacturable, and on-target capsid sequences under a limited experimental-feedback budget. Each task carries its own verifier, allowing the solver's progress and point of failure to be localized within the pipeline rather than collapsed into a single final score.

Additionally, tasks allow the definition of an \emph{opportunity matrix} of HDS6 capabilities, which enables a more targeted evaluation of those capabilities of a discoverative AI system by declaring, for each task, which of the six are in play; Table~\ref{tab:opportunity-matrix} shows this variation across a range of tasks. 
For example, the viability-prediction task exercises \emph{Evidence}, since a viability call must be grounded in sequence and structural features, and \emph{Scope}, since the solver must state the regime in which its prediction holds; yet, being a single graded prediction with no feedback loop, it gives no occasion to measure \emph{Repair}. The sequence-design task of the same environment complements it precisely there: as the solver proposes candidate sequences, checks them against the environment's viability and constraint feedback, and revises the ones that fail, it exercises \emph{Repair} together with \emph{Tools}. Combining the tasks of an environment therefore yields a rather complete picture of a solver system's capability as measured against the six core dimensions, even though no single task exercises or measures all of them. 

\begin{table}[t]
\centering\small
\setlength{\tabcolsep}{7pt}
\renewcommand{\arraystretch}{1.05}
\caption{Opportunity matrices for a range of benchmark tasks: different tasks activate different HDS6 capabilities. Columns are the six TRACES dimensions---\textbf{T}ools, \textbf{R}epair, \textbf{A}lternatives, \textbf{C}oherence, \textbf{E}vidence, \textbf{S}cope; $\bullet$ required, $\circ$ optional, -- not applicable.}
\label{tab:opportunity-matrix}
\begin{tabular}{@{}l@{\hskip 12pt}cccccc@{}}
\toprule
Task & T & R & A & C & E & S \\
\midrule
Solution ranking        & $\circ$   & --        & $\bullet$ & $\bullet$ & $\bullet$ & $\bullet$ \\
Capsid viability        & $\bullet$ & --        & $\bullet$ & $\bullet$ & $\bullet$ & $\bullet$ \\
SWE issue-fixing        & $\bullet$ & $\bullet$ & $\circ$   & $\bullet$ & $\bullet$ & $\bullet$ \\
Corpus deduplication    & $\bullet$ & $\circ$   & $\circ$   & $\circ$   & $\bullet$ & $\bullet$ \\
Speedrun training       & $\bullet$ & $\bullet$ & $\circ$   & $\bullet$ & $\circ$   & $\bullet$ \\
Clinical trial analysis & $\bullet$ & $\bullet$ & $\bullet$ & $\bullet$ & $\bullet$ & $\bullet$ \\
\bottomrule
\end{tabular}
\end{table}

\subsubsection{Episodes}
\label{sec:methods-episodes}

An episode is a concrete solver run on a specific problem instance, and it is the unit in which the solver system under test meets the environment. Within an episode the solver acts and the environment responds---returning observations, charging resources against a budget, and accepting intermediate and final submissions. Episodes are consequently the primary unit of end-to-end evaluation: task-level scores explain performance within an episode, while the episode-level outcome score grades the final result.

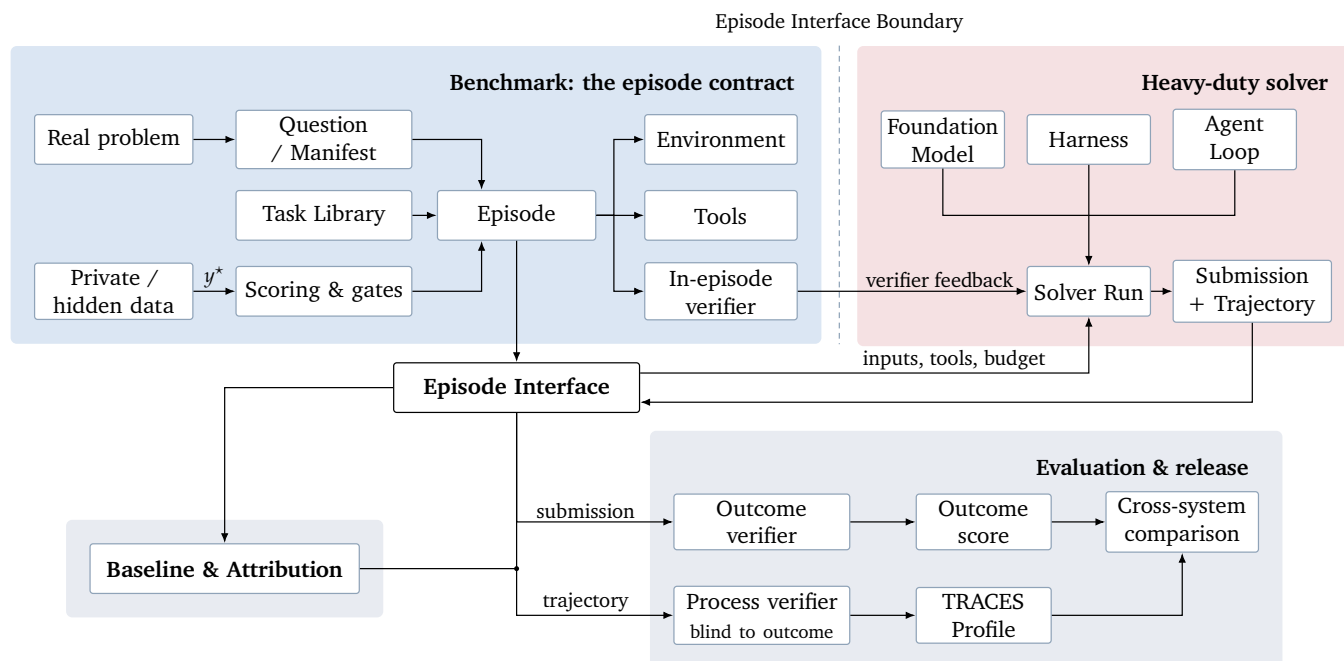
\begin{figure*}[t]
\centering
\begin{adjustbox}{max width=\textwidth}
\begin{tikzpicture}[fg,
  fglab/.append style={text=fgInk},
  fgmut/.append style={text=fgInk},
  box/.style ={draw=fgRule, line width=0.6pt, rounded corners=1.5pt,
               fill=white, align=center, inner sep=3pt,
               text width=25mm, minimum height=8.5mm},
  sm/.style  ={box, text width=19mm},
  smw/.style ={box, text width=21mm},
  wide/.style={box, text width=44mm},
]
  \node[fgttl, anchor=east] (btitle) at (13.22,9.95) {Benchmark: the episode contract};
  \node[box]  (real)   at (1.5,9.0) {Real problem};
  \node[box]  (hidden) at (1.5,6.4) {Private / hidden data};
  \node[box, text width=28mm]  (qman)   at (5.1,9.0) {Question / Manifest};
  \node[box, text width=28mm]  (task)   at (5.1,7.7) {Task Library};
  \node[box, text width=28mm]  (score)  at (5.1,6.4) {Scoring \& gates};
  \node[box]  (epi)    at (8.4,7.7) {Episode};
  \node[box, text width=24mm] (env)  at (11.9,9.0) {Environment};
  \node[box, text width=24mm] (tool) at (11.9,7.7) {Tools};
  \node[box, text width=24mm] (ver)  at (11.9,6.4) {In-episode verifier};

  \node[fgttl, anchor=east] (stitle) at (22.36,9.95) {Heavy-duty solver};
  \node[sm]  (fm)      at (15.7,9.0) {Foundation Model};
  \node[sm]  (harness) at (18.2,9.0) {Harness};
  \node[sm]  (agent)   at (20.7,9.0) {Agent Loop};
  \node[sm]  (run)     at (18.2,6.4) {Solver Run};
  \node[box] (trace)   at (21.0,6.4) {Submission + Trajectory};

  \node[box, text width=40mm, draw=fgInk, line width=0.6pt] (eif)
       at (8.4,4.75) {\textbf{Episode Interface}};

  \node[wide] (base)  at (3.4,1.65) {\textbf{Baseline \& Attribution}};
  \node[fgttl, anchor=east] (etitle) at (21.0,3.4) {Evaluation \& release};
  \node[box, text width=28mm] (overif) at (12.6,2.45) {Outcome verifier};
  \node[box, text width=28mm] (pverif) at (12.6,0.85)
       {Process verifier\\[0.5pt]{\footnotesize blind to outcome}};
  \node[smw] (out)  at (16.4,2.45) {Outcome score};
  \node[smw] (prof) at (16.4,0.85) {TRACES Profile};
  \node[box, text width=24mm] (lead) at (19.8,2.45) {Cross-system comparison};

  \begin{scope}[on background layer]
    \node[fill=fgB0,   rounded corners=3pt, inner sep=4mm,
          fit=(btitle)(real)(hidden)(qman)(task)(score)(epi)(env)(tool)(ver)]
          (bwash) {};
    \node[fill=fgAccBg, rounded corners=3pt, inner sep=4mm,
          fit=(stitle)(fm)(harness)(agent)(run)(trace)] (swash) {};
    \node[fill=fgWash, rounded corners=3pt, inner sep=4mm,
          fit=(etitle)(overif)(pverif)(out)(prof)(lead)] {};
    \node[fill=fgWash, rounded corners=3pt, inner sep=4mm, fit=(base)] {};
  \end{scope}

  \coordinate (bmid) at ($(bwash.east)!0.5!(swash.west)$);
  \draw[draw=fgRule, line width=0.6pt, dash pattern=on 2.8pt off 2.2pt]
       (bmid |- bwash.north) -- (bmid |- bwash.south);
  \node[fgmut, anchor=south] at ([yshift=1.6mm]bmid |- bwash.north)
       {\upshape Episode Interface Boundary};

  \draw[fgedge] (real)   -- (qman);
  \draw[fgedge] (hidden) -- node[fglab, above]{$y^\star$} (score);
  \draw[fgedge] (qman.east)  -| ([xshift=-6mm]epi.north);
  \draw[fgedge] (task)   -- (epi);
  \draw[fgedge] (score.east) -| ([xshift=-6mm]epi.south);
  \draw[fgedge] (epi.east) -- (tool.west);
  \draw[fgedge] (epi.east) -- ++(0.3,0) |- (env.west);
  \draw[fgedge] (epi.east) -- ++(0.3,0) |- (ver.west);

  \draw[draw=fgInk, line width=0.6pt] (fm.south)    |- (18.2,7.7);
  \draw[draw=fgInk, line width=0.6pt] (agent.south) |- (18.2,7.7);
  \draw[fgedge] (harness.south) -- (run.north);
  \draw[fgedge] (run) -- (trace);

  \draw[fgedge] (ver.east) --
        node[fglab, above, pos=0.62]{verifier feedback} (run.west);
  \draw[fgedge] ([yshift= 2.5mm]eif.east) -|
        node[fglab, above, pos=0.35]{inputs, tools, budget} (run.south);
  \draw[fgedge] (trace.south) |- ([yshift=-2.5mm]eif.east);

  \draw[fgedge] (epi.south) -- (eif.north);
  \draw[fgedge] (eif.west)  -| (base.north);

  \draw[fgedge] (eif.south) |-
        node[fglab, above, pos=0.72]{submission} (overif.west);
  \draw[fgedge] (eif.south) |-
        node[fglab, above, pos=0.72]{trajectory} (pverif.west);
  \draw[draw=fgInk, line width=0.6pt] (base.east) -- (8.4,1.65);
  \fill[fgInk] (8.4,1.65) circle (1.2pt);
  \draw[fgedge] (overif) -- (out);
  \draw[fgedge] (pverif) -- (prof);
  \draw[fgedge] (out) -- (lead);
  \draw[fgedge] (prof.east) -| (lead.south);
\end{tikzpicture}
\end{adjustbox}
\caption{The framework across the episode-interface boundary. The benchmark side defines the episode contract; the heavy-duty solver consumes it and returns a submission with its trajectory; the evaluation layer is outcome-first, with the TRACES/HDS6 process profile as a diagnostic. The solver under test and all baselines pass through the same interface, which is what makes attribution to a component of the system, rather than to the environment, meaningful.}
\label{fig:framework}
\end{figure*}

This meeting between solver and environment takes place only through a fixed \emph{episode interface}, and that discipline is what gives the evaluation its meaning (Figure~\ref{fig:framework}). The interface fixes, on the environment's side, the inputs, tools, budget, and submission format the solver is granted, and, on the evaluation side, the hidden verifier, the hard gates, and the outcome metrics that grade what the solver returns; the verifier feedback passed back across it is what makes an episode \emph{repairable}. Two products cross this boundary from the solver to the evaluation layer, and the whole verification framework rests on them: the submission, which outcome verification grades against ground truth the solver never sees, and the \emph{trajectory}, which the blind process verifier of Section~\ref{sec:methods-process-verification} reads. Because the solver under test and every baseline pass through the identical interface, a difference in outcome can be attributed to a specific part of the solver system---its foundation model, harness, agent loop, tool use, or handling of verifier feedback.

\paragraph{The episode record.}
The trajectory is the core of a self-contained episode record: the ordered sequence of attempts, each with its actions and parameters, the resource charged, the remaining budget, and either an observation summary or the error raised, together with the metered model calls, any asynchronous verifier jobs, a snapshot of the final workspace, and the verification result.
Such episode records can then be evaluated by the process verifier, and capability scores as well as diagnostic repair notes can be produced to evaluate and further improve the SuT's capability on the particular tasks.


\section{Results on High-Value Problems}
\label{sec:results}

The following results are early, promising evidence for discoverative AI. Evaluating heavy-duty solvers along different tasks, we find that they surpass the published state of the art on AAV capsid design and gain measurably from a task-specific environment on drug repurposing.
We also find some evidence on the effectiveness of the HDS6 metric, and its induced verify-repair loops. These are encouraging first steps toward the investigations from which genuine discoveries emerge.

\subsection{Adeno-associated virus capsid design}

\subsubsection{Task Formulation}
Adeno-associated virus (AAV) is widely used as a leading gene therapy vector for treating a range of genetic diseases, including inherited retinal disorders, hemophilia, muscular dystrophy, and neurological conditions. Its protein shell, the capsid, determines whether the vector can be manufactured and whether it can efficiently reach target cells. Capsid engineering is challenging because mutations that improve delivery or reduce immune recognition may also prevent the virus from assembling or packaging DNA. Our benchmark captures four key stages of this process: Task 1 predicts whether a capsid variant remains viable~\cite{bryant2021aav,Ogden2019ComprehensiveAC}; Task 2 predicts which tissues a variant targets~\cite{eid2024systematic}; Task 3 predicts the capsid's three-dimensional structure from its sequence; and Task 4 designs a small, diverse set of novel, on-target sequences for laboratory testing~\cite{eid2024systematic,Huang2023TargetingAV}.

Together, these tasks form a realistic discovery pipeline, in which a candidate must be manufacturable, reach the right target, and be novel enough to improve on existing vectors, all under a limited experimental budget. The benchmark therefore evaluates not only whether an AI model can predict protein properties, but whether it can support the sequence of biological decisions required to turn computational designs into experimentally testable AAV candidates.

Rigorous control of data leakage is essential to the validity of any benchmark that aims to measure genuine biological reasoning rather than memorization of the underlying corpora. We therefore built every task on real screening or structural data and enforced out-of-distribution splits across mutational load, species, and structural deposition dates. Ground truth is further protected because it is exposed only through a metered scoring action, rather than as labels the solver can read directly. Full task definitions, metrics, splits, and per-instance result tables are given in Appendix~\ref{app:aav-full}.

\subsubsection{A domain-specific environment improves outcomes}
The leading AI model operating within our domain-specific environment consistently outperformed the same model using the generic Claude Code harness, demonstrating the value of providing models with task-specific data, tools, biological constraints, and executable feedback. Averaged across tasks 1-3, claude-opus-4-8 achieved a score of 0.741 in our environment, substantially outperforming its score of 0.716 with the Claude Code harness (Table~\ref{tab:aav-overview}). The gap was widest on structure prediction (0.657 versus 0.595), the stage that most depends on orchestrating folding engines and validating intermediate results rather than returning a single prediction. The same advantage held on Task 4 (Table~\ref{tab:aav-04}), indicating that the benefit of a domain-specific environment is not confined to any single stage but recurs throughout the capsid-engineering workflow.

\subsubsection{Surpassing the published state of the art}
Building on this domain-specific environment, apodex-1.1 surpassed the previous published state of the art at every stage, from predicting whether a variant survives to designing new candidates from scratch (Table~\ref{tab:aav-01}-\ref{tab:aav-04}). The viability task first asks whether an engineered capsid can assemble and package its genome, the first and rate-limiting filter of any campaign. On it, apodex-1.1 reached an out-of-distribution AUROC of 0.904, above the CAP-PLM capsid language model that marks the published state of the art (0.878)~\cite{Wu2024PredictionOA, guo2026mapping, hsu2022learning, bryant2021aav}. This result establishes that apodex-1.1 can build AI models that reliably identify which candidates are worth carrying forward before any downstream analysis begins.

Beyond viability, two further dimensions determine whether a candidate is useful, the tissues it reaches and its three-dimensional structure. The tropism task predicts which tissues a variant reaches and whether this targeting pattern holds across species. Here, apodex-1.1 scored 0.635, again outperforming the published state of the art (Fit4Function, 0.622)~\cite{eid2024systematic}, demonstrating that apodex-1.1 has developed models that can rank variants by tissue enrichment more accurately than this established expert method. The structure task asks the solver to build the three-dimensional capsid at three levels, the surface loops of a single subunit, the full 60-mer shell, and the capsid in complex with an antibody or receptor, which together form the basis for epitope mapping and receptor-guided design. Averaged over the three levels, apodex-1.1 scored 0.649, above the published state of the art of 0.605, AlphaFold 3~\cite{abramson2024accurate} with template-based symmetry expansion.

The design task, by contrast, requires a fundamentally different, generative capability. The solver must search a vast sequence space, balance viability against specificity and novelty, and prioritize candidates under a limited experimental budget. On this task, apodex-1.1 scored 0.180, surpassing the specialist generative methods AAVGen (0.109), AAVDiff (0.116), and ALICE (0.110)~\cite{GhaffarzadehEsfahani2026AAVGenPE,liu2024aavdiff,guo2026mapping}. This result shows that apodex-1.1 can generate viable, novel, and on-target candidates under a constrained experimental budget. Together, these results suggest that frontier models equipped with an appropriate scientific environment can not only automate existing analytical workflows, but also make higher-quality biological design decisions than established expert approaches.

\subsubsection{Results based on process verification}
The HDS6 process scores exhibit the same overall trend as the outcome-based results, since systems that achieved stronger final biological performance also received higher scores for their scientific problem-solving process. Within the domain-specific environment, stronger-performing models generally showed more coherent hypothesis formation, more effective use of data and tools, better handling of biological constraints, and stronger self-correction. The same pattern holds at both the task and instance level. For instance, on the structure prediction task (Task 3), gpt-5.5 scored only 0.544, 20\% lower than kimi-k3, and its process scores fell to 1.38 on the single-subunit loop instance and 0.94 on the full-shell assembly instance (Table~\ref{tab:aav-03}), due to the fact that gpt-5.5 consistently skips the self-validation step the task requires and submits its first-pass answer instead. Kimi-k3, the strongest solver on structure, earned some of the highest process scores on the same two instances, reaching 3.56 and 3.60, respectively. This consistency suggests that HDS6 can meaningfully characterize the quality of scientific problem solving, rather than merely evaluating the fluency or presentation of the final response.

\begin{table}[t]\centering
\begin{threeparttable}
\caption{Model comparison across the four AAV tasks, ordered by mean result score over Tasks 1--3. For each task, we report the result score and the mean HDS6 process score across instances. All methods are evaluated in our domain-specific environment, except CC Opus-4.8, which uses the Claude Code harness.
}
\label{tab:aav-overview}
\begin{tabular}{l cc cc cc cc c}
\toprule
 & \multicolumn{2}{c}{T1 Viability} & \multicolumn{2}{c}{T2 Tropism} & \multicolumn{2}{c}{T3 Structure} & \multicolumn{2}{c}{T4 Design} & \\
\cmidrule(lr){2-3}\cmidrule(lr){4-5}\cmidrule(lr){6-7}\cmidrule(lr){8-9}
\textbf{solver} & score & HDS6 & score & HDS6 & score & HDS6 & score & HDS6 & \shortstack{\textbf{mean}\\\textbf{T1--T3}} \\
\midrule
  kimi-k3 & \textbf{0.936} & 3.49 & 0.649 & 3.61 & \textbf{0.679} & \textbf{3.68} & \textbf{0.194} & \textbf{3.59} & \textbf{0.755} \\
  claude-opus-5 & 0.936 & 3.46 & \textbf{0.653} & 3.55 & 0.668 & 3.61 & \tnote{$\dagger$} & - & 0.752 \\
  glm-5.2 & 0.932 & 3.27 & 0.644 & \textbf{3.76} & 0.665 & 3.39 & 0.191 & 3.32 & 0.747 \\
  claude-opus-4-8 & 0.931 & \textbf{3.65} & 0.635 & 3.30 & 0.657 & 3.36 & \tnote{$\dagger$} & - & 0.741 \\
  \textbf{apodex-1.1} & 0.904 & 3.36 & 0.635 & 3.37 & 0.649 & 3.52 & 0.180 & 3.23 & 0.729 \\
  CC opus-4-8 (w/o env.)\tnote{*} & 0.921 & - & 0.633 & - & 0.595 & - & \tnote{$\dagger$} & - & 0.716 \\
  deepseek-v4-pro & 0.918 & 3.37 & 0.613 & 2.97 & 0.610 & 2.19 & 0.129 & 2.64 & 0.714 \\
  claude-sonnet-4-6 & 0.922 & 3.45 & 0.628 & 2.77 & 0.575 & 2.73 & 0.094 & 2.97 & 0.709 \\
  gpt-5.5 & 0.909 & 2.99 & 0.600 & 3.57 & 0.544 & 1.22 & 0.142 & 2.24 & 0.684 \\
  qwen3.5-397b & 0.896 & 3.27 & 0.605 & 2.93 & 0.543 & 1.97 & 0.085 & 2.21 & 0.682 \\
  \textbf{apodex-1.0} & 0.884 & 3.23 & 0.518 & 2.90 & 0.526 & 1.84 & 0.072 & 2.34 & 0.643 \\
\midrule
published SOTA\tnote{**} & 0.878 & - & 0.622 & - & 0.605 & - & 0.116 & - & 0.702 \\
\bottomrule
\end{tabular}
\begin{tablenotes}\footnotesize
\item[*]the Claude Code agent running on claude-opus-4-8 without using our domain-specific environment.
\item[**]published SOTA is, per task, the strongest reference method on that task, scored by its mean over instances.
\item[$\dagger$]Opus-series models (claude-opus-5, claude-opus-4-8) declined to produce outputs on Task 4, consistently triggering the models' biosafety refusal behavior.
\end{tablenotes}
\end{threeparttable}
\end{table}

\subsubsection{Datasets and evaluation}
Each task is built from published screens or structures, and is split into instances that each test a different perspective; the exact metrics are defined in Appendix~\ref{app:aav-full}. The viability task draws on the deep-mutagenesis screen of Bryant et al.~\cite{bryant2021aav} and the single-mutation scan of Ogden et al.~\cite{Ogden2019ComprehensiveAC}. It has two instances. The multi-mutation instance tests whether a model trained on lightly mutated variants can predict the viability of heavily mutated ones; the single-mutation instance tests whether it generalizes to single substitutions at positions across the capsid sequence that it has not seen in training. Both are scored by how well the model separates viable from non-viable variants that carry the same number of mutations.

The tropism task uses the Fit4Function screens~\cite{eid2024systematic}, where a seven-residue peptide is inserted into a fixed surface loop of AAV9 and its enrichment in each target tissue is measured. It has three instances: multi-organ tropism in mouse, cross-species transfer to macaque, and human in-vitro assays. Predictions are scored by their correlation with the measured enrichment.

The structure task gives only the amino-acid sequence and asks the solver to build the three-dimensional capsid structure. It has three instances for different structural levels: the variable surface loops of a single subunit, the full 60-mer icosahedral shell, and the complex with an antibody or cell-surface receptor. All target structures were deposited after 30 September 2021, later than the folding engines' training cut-off, so they could not have been memorized during pretraining. Each instance is scored against its experimental reference structure.

The design task asks the solver to generate novel peptides for a target, under a limited query budget. Its seed data and scoring oracles are built from the Fit4Function screens~\cite{eid2024systematic}, and for the receptor instances from LY6A and LY6C1 binding measurements~\cite{Huang2023TargetingAV}. A generated sequence counts only if it is novel, manufacturable, and on target, meaning it is either selectively enriched in the intended organ or selectively binds the intended receptor.

\subsection{Drug Repurposing and Reformulation}

\subsubsection{Task Formulation}

Drug repurposing and reformulation (DRR) does not aim to discover new drugs. Instead, it seeks to identify new therapeutic indications and optimized procedure design for existing approved or clinically characterized drugs, enabling their application to additional diseases and patient populations. In contrast to de novo molecule discovery, DRR can reuse prior pharmacology, safety, manufacturing, and clinical evidence. The task includes not only indication selection, but also biomarker-guided patient stratification, formulation and delivery, combination therapy, and protocol design.

DRR decisions require the integration of heterogeneous and sometimes conflicting evidence. Mechanistic plausibility alone is insufficient because biological hypotheses may not translate into clinical benefit, trial reports may be incomplete, and outcomes depend on dose, safety, population selection, delivery, combination strategy, and study design. We therefore formulate DRR as a prospective pointwise forecasting task. Given a disease and a set of candidate drugs, the system evaluates each drug--disease pair using only information available before a fixed temporal cutoff. For each pair, it outputs a promisingness score $\hat y\in[0,1]$, a confidence estimate $c\in[0,1]$, and an evidence-grounded rationale. The promisingness score represents the predicted position of the pair on the clinical-development and approval spectrum.

Ground-truth labels are constructed from clinical and regulatory outcomes observed after the information cutoff. The primary endpoint evaluates $\hat y$; confidence and rationale support complementary calibration and trajectory analyses. The benchmark contains a 100-pair public development set and a 200-pair hidden test set. Drug and disease identities are normalized before splitting, and outcome-bearing fields from trials used to construct labels are withheld from model inputs. Temporal filtering and pair-level decontamination prevent the same outcome evidence from appearing in both the model context and the evaluation label.

\subsubsection{Scoring Methods}

Let the evaluation set be
$\mathcal{D}=\{(x_i,y_i,\tau_i)\}_{i=1}^{N}$, where $x_i$ is the
drug--disease pair, $y_i\in[0,1]$ is its observed value on the
clinical-outcome scale, and
$\tau_i\in\{\text{exact},\text{lower bound}\}$ specifies how that value
should be interpreted. Exact labels represent terminal outcomes, whereas
lower-bound labels record a stage already attained without asserting that
the program will progress no further. A failed program receives the value of
the highest phase it had previously passed. The model outputs
$\hat y_i\in[0,1]$ without observing either $y_i$ or $\tau_i$. Evaluation
maps the observed outcome to a fixed nonuniform tier scale, applies an
absolute-error loss that distinguishes exact outcomes from lower bounds, and
normalizes the mean score between a knowledge-free random predictor and a
perfect predictor.

\paragraph{Tier scale.}
We represent clinical outcomes using five ordered tiers: preclinical only or
failed in Phase~I, passed Phase~I, passed Phase~II, passed Phase~III, and
approved for the target indication. Rather than imposing equal spacing
between stages, we fix a nonuniform scale informed by the empirical
distribution of concluded programs in a historical corpus. The scale is
specified before evaluation and is not refit for individual systems or runs.
Table~\ref{tab:tier-scale} reports both the tier values and their historical
support.

\begin{table}[t]
\centering
\begin{tabular}{@{}lrrr@{}}
\toprule
Clinical outcome & Tier value & Historical count & Historical fraction \\
\midrule
Preclinical only or failed in Phase~I & 0.00 & 662     & 6.3\%  \\
Passed Phase~I                         & 0.15 & 2{,}048 & 19.6\% \\
Passed Phase~II                        & 0.45 & 2{,}178 & 20.9\% \\
Passed Phase~III                       & 0.65 & 1{,}641 & 15.7\% \\
Approved for the target indication    & 1.00 & 3{,}898 & 37.4\% \\
\bottomrule
\end{tabular}
\caption{Tier-scale values and their support among the 10{,}427 concluded
programs in the historical corpus.}
\label{tab:tier-scale}
\end{table}

\paragraph{Forward-looking absolute-error score.}
Lower-bound labels are used for successful but not-yet-approved programs
and for ongoing programs. The per-item loss is
\begin{equation}
\ell_i=
\begin{cases}
\lvert \hat y_i-y_i\rvert,
    & \tau_i=\text{exact},\\[4pt]
\max(y_i-\hat y_i,\,0),
    & \tau_i=\text{lower bound}.
\end{cases}
\label{eq:per-item-loss}
\end{equation}
Thus, exact outcomes are penalized symmetrically in proportion to the
absolute prediction error, whereas a lower-bound outcome is penalized only
when the prediction falls below the stage already attained. In particular,
predictions above a lower bound incur no loss.

The corresponding per-item quality is
\begin{equation}
q_i=\max(0,\,1-\ell_i).
\label{eq:per-item-quality}
\end{equation}
Because $y_i,\hat y_i\in[0,1]$, the loss lies in $[0,1]$, so the clipping
operation in Eq.~\eqref{eq:per-item-quality} is inactive and
$q_i=1-\ell_i$. The raw continuous score is the plain mean over all
evaluation items, including both exact and lower-bound labels:
\begin{equation}
S=\frac{1}{N}\sum_{i=1}^{N}q_i
 =1-\frac{1}{N}\sum_{i=1}^{N}\ell_i.
\label{eq:prediction-score}
\end{equation}

\paragraph{Normalized continuous score.}
The random-guess anchor is defined using a knowledge-free predictor $R$.
For each item in a predefined benchmark reference set, $R$ draws a prediction
independently from the empirical training-label distribution. The
random-anchor score $S_R$ is its expected quality under the same exact or
lower-bound loss branch used for that reference item. This construction
accounts for both the frequency of the prediction tiers and the mixture of
exact and lower-bound outcomes. The anchor $S_R$ is computed once using $R$ on the development set's labels, and held fixed
for all systems. For the benchmark reported here, $S_R=0.86925$. The oracle
anchor is $S_O=1$, corresponding to zero loss on every item.

The raw score is rescaled through the linear anchor map
\begin{equation}
    S_{\mathrm{norm}}
    =\frac{S-S_R}{S_O-S_R}\times 100.
\label{eq:normalization}
\end{equation}
Accordingly, $S_{\mathrm{norm}}=0$ corresponds to the expected performance
of the random predictor and $S_{\mathrm{norm}}=100$ to perfect prediction;
scores below $0$ indicate performance worse than the random anchor. We call
$S_{\mathrm{norm}}$ the \emph{normalized continuous score} and use it as the
primary outcome metric. Unlike a categorical metric, it retains the magnitude
of prediction errors on the full $[0,1]$ scale.

\paragraph{Categorical accuracy.}
For a complementary categorical measure, each prediction is assigned to its
nearest tier in $\mathcal{A}=\{0.00,0.15,0.45,0.65,1.00\}$. An exact midpoint
is assigned to the lower tier. Let $\tilde y_i\in\mathcal{A}$ denote the tier
assigned to $\hat y_i$. Per-item categorical correctness is
\begin{equation}
m_i=
\begin{cases}
\mathbf{1}[\tilde y_i=y_i],
    & \tau_i=\text{exact},\\[4pt]
\mathbf{1}[\tilde y_i\geq y_i],
    & \tau_i=\text{lower bound}.
\end{cases}
\label{eq:tier-match}
\end{equation}
Thus, terminal outcomes require an exact tier match, whereas ongoing or
successful unapproved programs are counted as correct when the predicted tier
is at least the stage already attained. Missing or invalid predictions count
as incorrect. We report
\begin{equation}
A_{\mathrm{cat}}=\frac{100}{N}\sum_{i=1}^{N}m_i
\label{eq:categorical-accuracy}
\end{equation}
as \emph{categorical accuracy}. This metric is easy to interpret but discards
within-category distances; the normalized continuous score remains the more
fine-grained primary measure.

\subsubsection{Public-100 Benchmark Results}
We use two matched evaluation conditions. In the \emph{no env} condition, the backbone receives the task input and produces a single response without external retrieval, executable tools, or network access. In the \emph{with env} condition, the same backbone can iteratively use an executable DRR environment that provides structured biomedical knowledge, literature-derived evidence, and scientific tools. The backbone model, benchmark items, output schema, and scoring procedure are otherwise held fixed, and neither condition has access to held-out outcomes. Thus, ``environment'' denotes the external evidence and tool layer rather than a separate prediction model.

We evaluate GPT-5.5 and GPT-5.6-sol with medium reasoning effort as the backbones.
As shown in Table~\ref{tab:drr-main-results}, the \emph{with env} condition improves every reported metric for both backbones. 
Agreement between the continuous and categorical metrics indicates that the result is not an artifact of normalization. Because the comparison uses three runs on the public development set, these gains should be interpreted descriptively rather than as definitive estimates of test-set performance.

\begin{table*}[t]
\centering
\small
\caption{Outcome-prediction performance on the 100-pair public development set. The normalized continuous score is the random-anchored 0--100 metric in Eq.~\eqref{eq:normalization}; the raw continuous score is the corresponding mean per-item quality in Eq.~\eqref{eq:prediction-score}. Categorical accuracy is defined in Eq.~\eqref{eq:categorical-accuracy}. Values are mean $\pm$ sample standard deviation over three independent runs. Absolute gain is the with-env mean minus the no-env mean, computed from unrounded values; gains are reported as normalized-score points, raw-score units, and categorical-accuracy percentage points (pp), respectively.}
\label{tab:drr-main-results}
{\setlength{\tabcolsep}{5pt}
\begin{tabular}{@{}llrrr@{}}
\toprule
Backbone & Condition & Normalized continuous score & Raw continuous score & Categorical accuracy \\
\midrule
GPT-5.5 & No env & $53.78\pm3.22$ & $0.93957\pm0.00421$ & $(81.00\pm1.73)\%$ \\
 & With Apodex-env & $56.30\pm1.56$ & $0.94287\pm0.00204$ & $(83.67\pm1.15)\%$ \\
 & \emph{Absolute gain} & $+2.52$ & $+0.00330$ & $+2.67$ pp \\
\midrule
GPT-5.6-sol & No env & $54.21\pm2.20$ & $0.94013\pm0.00287$ & $(80.67\pm1.15)\%$ \\
 & With Apodex-env & $61.81\pm3.92$ & $0.95007\pm0.00512$ & $(85.00\pm1.73)\%$ \\
 & \emph{Absolute gain} & $+7.60$ & $+0.00993$ & $+4.33$ pp \\
\bottomrule
\end{tabular}
}
\end{table*}

\subsubsection{HDS6 Trajectory Analysis}
We use HDS6 to characterize the quality of the scientific reasoning process in addition to final-answer accuracy. HDS6 rates six dimensions on a 0--4 scale: long-horizon state coherence ($C$), evidence fidelity ($E$), hypothesis management ($A$), boundary and failure reasoning ($S$), tool use and execution-state management ($T$), and self-correction under verification ($R$). Scores of 0, 1, 2, 3, and 4 correspond to absent, flawed, insufficient, good, and excellent behavior, respectively. An automated process evaluator scores the recorded reasoning and tool interactions with the held-out outcome hidden. Capability scores retain the rubric's critical-failure caps. A separate integrity gate excludes trajectories that are too incomplete or malformed for reliable process scoring, and a capability marked not rated (NR) is omitted rather than assigned a score of zero. 
Table~\ref{tab:drr-hds6} reports capability means, dispersion, and paired condition gains.

\begin{table*}[t]
\centering
\scriptsize
\caption{HDS6 capability scores on matched drug--disease cases. Condition rows report mean $\pm$ population standard deviation on the 0--4 scale; NR and missing capability ratings are omitted. Absolute gain is the mean paired within-case difference (with env minus no env) among cases rated in both conditions. Repair gains are unavailable because repair was not observable in the no-env trajectories.}
\label{tab:drr-hds6}
{\setlength{\tabcolsep}{3pt}
\begin{tabular*}{\textwidth}{@{\extracolsep{\fill}}llrrrrrr@{}}
\toprule
Backbone & Condition & $C$ & $E$ & $A$ & $S$ & $T$ & $R$ \\
\midrule
GPT-5.5 & No env
  & $1.62\pm0.21$ & $1.52\pm0.32$ & $2.35\pm0.65$ & $2.72\pm0.88$ & $1.00\pm0.00$ & --- \\
 & With Apodex-env
  & $2.76\pm0.72$ & $3.10\pm0.62$ & $2.69\pm0.64$ & $2.96\pm0.65$ & $3.64\pm0.69$ & $3.45\pm0.79$ \\
 & \emph{Absolute gain}
  & $+1.13$ & $+1.58$ & $+0.34$ & $+0.24$ & $+2.64$ & --- \\
\midrule
GPT-5.6-sol & No env
  & $1.67\pm0.18$ & $1.42\pm0.32$ & $2.48\pm0.72$ & $2.85\pm0.79$ & $0.99\pm0.15$ & --- \\
 & With Apodex-env
  & $2.81\pm0.82$ & $3.10\pm0.65$ & $2.72\pm0.62$ & $3.09\pm0.62$ & $3.56\pm0.79$ & $3.83\pm0.35$ \\
 & \emph{Absolute gain}
  & $+1.14$ & $+1.68$ & $+0.25$ & $+0.24$ & $+2.57$ & --- \\
\bottomrule
\end{tabular*}
}
\end{table*}

Across backbones, the environment most strongly improves tool use, evidence fidelity, and long-horizon coherence, while gains in hypothesis management and scope control are smaller. This pattern suggests that the environment primarily helps systems ground claims, preserve state, and execute evidence-seeking workflows. The low no-env tool scores reflect the deliberate absence of executable tools, not failed attempts to use them.

GPT-5.5 and GPT-5.6-sol exhibit similar with-env profiles, indicating that the process-level benefits are stable across the two GPT backbones. 
Repair scores are conditional on an observable correction opportunity and are unavailable for the no-env trajectories, so they should not be treated as directly comparable coverage-matched estimates. The HDS6 findings are descriptive and remain subject to the survivor-selection limitation above.

\subsubsection{Repurposing and Reformulation Procedure Design}
We separately evaluate whether a system can propose a clinically actionable development procedure for a drug--disease pair. The system predicts four structured fields: target population, biomarker strategy, combination therapy, and formulation or delivery. Reference values are derived from registered trial protocols. In the language-model-only condition, these fields and their source records are withheld during prediction. An automated semantic evaluator compares each predicted field with its reference on a 1--5 scale, where 1 denotes an irrelevant or contradictory proposal, 3 denotes a directionally correct but incomplete proposal, and 5 denotes agreement on the key clinical content. Field means are computed over examples with an available reference value, and the macro-average gives equal weight to the four field means.

Table~\ref{tab:drr-procedure} compares a language-model-only condition (LM only) with a knowledge-augmented condition (LM + context). The augmented condition revises the initial model prediction using structured eligibility context retrieved from registered clinical-trial records. This context is available at sufficient coverage only for population specification; the biomarker, combination, and formulation/delivery predictions are therefore identical across conditions. Because the retrieved eligibility record is also the source from which the population reference is derived, the population comparison measures evidence retrieval and synthesis rather than fully prospective procedure generation. Population improves by 0.89 points, increasing the macro-average from 3.72 to 3.94. Together, the outcome, trajectory, and procedure-design evaluations assess three complementary capabilities: selecting promising therapeutic candidates, producing evidence-grounded reasoning, and specifying clinically actionable development strategies.

\begin{table}[t]
\centering
\small
\caption{Drug repurposing and reformulation procedure-design results. Scores are mean semantic-evaluator ratings on a 1--5 scale. LM + context differs from LM only for population, the field for which structured trial context is available.}
\label{tab:drr-procedure}
{\setlength{\tabcolsep}{3pt}
\begin{tabular}{@{}lrrr@{}}
\toprule
Field & LM only & LM + context & $\Delta$ \\
\midrule
Population & 2.89 & 3.78 & $+0.89$ \\
Biomarker & 4.00 & 4.00 & $0.00$ \\
Drug combination & 3.75 & 3.75 & $0.00$ \\
Formulation/delivery & 4.22 & 4.22 & $0.00$ \\
\midrule
Macro-average & 3.72 & 3.94 & $+0.22$ \\
\bottomrule
\end{tabular}
}
\end{table}

\subsection{Large language model problems and ablations}
\label{sec:results-ablation}

Previous sections focused on problems whose verification relies on slow experimental or clinical evidence. This section turns to a different class of heavy-duty tasks: the engineering work underpinning foundation models such as pretraining data curation, filtering, deduplication, inference-serving correctness, reward design, and training-recipe optimization. These tasks are routinely delegated to research engineers because they demand sustained, tool‑mediated, self‑correcting investigation, not a single prompt or inference call, and they already account for much of the labor a heavy‑duty solver would need to absorb in an AI research organization. This environment family also offers a practical advantage for ablations: its outcome verifiers (\emph{e.g.,} held‑out loss, throughput gates, decontamination audits, and bit‑exactness suites) are recomputable on demand, enabling denser and more controlled ablations than biomedical environments allow. We use this LLM environment family to pose two nested attribution questions.

We design two hierarchical ablation layers: (i) \emph{solver-configuration} ablation, which fixes episode interfaces while varying back-end harnesses and foundation models; (ii) \emph{episode-mechanism} ablation, which locks environments and solver setups to toggle individual internal mechanisms on or off. We evaluate two episode-level interventions: initial skill guidance and post-submission verifier-driven repair. This decomposition quantifies performance variance stemming from the solver framework itself versus in-episode or out-episode, process-centric guidance feedback.

\subsubsection{Solver-configuration ablations}
The current result matrix spans 11 evaluated LLM environments, six harnesses, and up to six foundation models on the multi-model harnesses. Two harnesses are model-restricted by construction rather than by choice: Claude Code is a closed-source packaged CLI that exposes no source to instrument, so we drive it only with Opus, and Codex CLI is similarly run only with GPT, since it is tightly coupled to the OpenAI Responses API and is operated as a packaged
binary rather than a modified harness. The remaining harnesses, DeerFlow,
OpenHands~\cite{openhands}, A-Evolve~\cite{aevolve}, and ApodexHarness (our own harness), are model-agnostic by design and are swept across all six models where the environment permits. 

This asymmetry means the matrix cannot be summarized by a single harness-level average without first controlling for which models each harness was paired with, since an uncontrolled average would cause the performance of Claude Code and Codex CLI to be overestimated for never having been tested against weaker models. 
To enable reliable controlled comparisons and fine-grained ablation analyses, we defined several quantifiable inclusion criteria prior to experiment execution to select a pre-specified diagnostic subset of environments: (i) full coverage across all four model-agnostic harnesses; (ii) no severe floor-level scores; (iii) sufficient run-to-run reliability.
Seven of the eleven LLM environments meet all three criteria and are used for the aggregate analyses below: \texttt{LLM-pretrain-data-probe}, \texttt{LLM-rollout-verify}, \texttt{operator-align}, \texttt{rl-recipe}, \texttt{posttrain-data-dedup}, \texttt{pretrain-data-filter}, and \texttt{solution-verifier}.
%
%
All 11 environments are reported in full in Appendix~\ref{app:full-matrix}.

\begin{table}[t]
\centering\small
\caption{Aggregate system-level performance across all tested harness--model configurations on the seven pre-specified diagnostic environments. 
Mean score: average performance of each model across all harnesses it was paired with. 
Best cell: highest score achieved by any harness--model configuration for that model. 
Failed cells: total number of outright failures per model (score of zero or run with no valid output); note that GPT-5.6-sol and Opus-4.8 are evaluated on 35 cells while the remaining four models are evaluated on 28, so these counts are not directly comparable in rate terms across the two groups.
Harness coverage is asymmetric by design: Opus-4.8 and GPT-5.6-sol include dedicated packaged harnesses (Claude Code, Codex CLI) in addition to the four model-agnostic harnesses, while other models are evaluated only on the four universal harnesses. This table summarizes the overall achievable end-to-end performance per model.}
\label{tab:model-ablation}
\vspace{2pt}
\begin{tabular}{@{}l S[table-format=1.3] S[table-format=1.3] S[table-format=1.3] S[table-format=1.3] S[table-format=1.3] S[table-format=1.3]@{}}
\toprule
{Model} & {GPT-5.6-sol} & {Opus-4.8} & {GLM-5.2} & {Kimi-K3} & {DeepSeek-v4-pro} & {Apodex-1.0} \\
\midrule
Mean score    & 0.588 & 0.575 & 0.516 & 0.553 & 0.478 & 0.459 \\
Best cell     & 0.950 & 0.850 & 0.850 & 0.900 & 0.776 & 0.800 \\
{Failed cells} & {0}   & {0}   & {1}   & {1}   & {3}   & {2}   \\
\bottomrule
\end{tabular}
\end{table}

\begin{table}[t]
\centering\small
\caption{Harness performance comparison on the seven pre-specified diagnostic environments, presented for two representative foundation models (Opus-4.8 and GPT-5.6-sol).
Claude Code and Codex CLI are dedicated packaged harnesses tightly coupled to a single model, and are evaluated only on their respective native model.
DeerFlow, OpenHands, A-Evolve, and ApodexHarness are model-agnostic harnesses designed to work across all foundation models; the two representative configurations shown here illustrate how relative harness performance varies with model choice.
``Best cell'' is the highest score the harness reached on the corresponding model across environments.
``Failed cells'' counts environments where the harness scored exactly zero or produced no valid result, out of seven total environments.}
\label{tab:harness-ablation}
\vspace{2pt}
\begin{tabular}{@{}l S[table-format=1.3] S[table-format=1.3] S[table-format=1.3] S[table-format=1.3] S[table-format=1.3] S[table-format=1.3]@{}}
\toprule
{Harness} & {Claude Code} & {Codex} & {OpenHands} & {A-Evolve} & {DeerFlow} & {ApodexHarness} \\
\midrule
Opus-4.8 mean  & 0.583 & {---}  & 0.578 & 0.569 & 0.534 & 0.611 \\
Opus-4.8 best  & 0.800 & {---}  & 0.750 & 0.850 & 0.821 & 0.850 \\
GPT-5.6-sol mean   & {---}  & 0.559 & 0.560 & 0.648 & 0.580 & 0.592 \\
GPT-5.6-sol best   & {---}  & 0.900 & 0.950 & 0.900 & 0.800 & 0.950 \\
\midrule
{Failed cells (Opus-4.8)} & {0} & {---}  & {0} & {0} & {0} & {0} \\
{Failed cells (GPT-5.6-sol)}  & {---}  & {0} & {0} & {0} & {0} & {0} \\
\bottomrule
\end{tabular}
\end{table}

\paragraph{Model performance: aggregate scores and caveats.}
Even the strongest model–harness combinations remain far from solving this environment family. Across all tested configurations, the highest aggregate mean per model is 0.588, as shown in Table~\ref{tab:model-ablation}. This gap is itself a meaningful benchmark for progress, indicating that substantial headroom remains for heavy-duty solvers relative to task-specific score ceilings.

As noted in the table caption, these aggregate scores reflect end-to-end system-level performance across all tested harness pairings, including dedicated packaged solver tools. Harness coverage is intentionally asymmetric, so this table characterizes practical real-world deployment performance per model rather than isolating pure foundation model capability.
To isolate pure foundation model effects with equal harness coverage, we also compute balanced means using only the four model-agnostic harnesses (a fully matched 4$\times$6 matrix across all seven environments). Under this balanced setting, GPT-5.6-sol ranks first (0.595), followed by Opus-4.8 (0.573) and Kimi-K3 (0.553), then GLM-5.2 (0.516), DeepSeek-v4-pro (0.478), and Apodex-1.0 (0.459). Adding each model's dedicated packaged harness in the full end-to-end table changes these two models' means only marginally, so it does not materially change the ordering between them.

Two patterns emerge from the table. First, best-cell scores cluster tightly across models, most exceeding 0.78, even though mean scores and failure counts vary substantially. This suggests that run-to-run consistency and harness compatibility, more than peak reasoning capability, drive much of the mean-score gap. Best-cell values are drawn from an asymmetric set of harness and environment pairings and are sensitive to noise and to how well a given harness matches a given model, so a definitive ceiling comparison requires paired analysis within fixed configurations.
Second, the overall performance frontier remains narrow. Even the strongest configurations leave substantial room for improvement, and harness optimization mainly improves reliability rather than peak performance.

\paragraph{Harness performance: per-model and cross-model comparisons.}
Table~\ref{tab:harness-ablation} shows that no harness dominates once the
model is held fixed: ApodexHarness leads on Opus, A-Evolve leads on GPT, and the ranking of the four model-agnostic harnesses differs substantially between the two model configurations.
Harness effects and model effects are therefore both present, and neither is separable from the other using Claude Code or Codex, since these packaged harnesses are tightly coupled to a single model and cannot provide cross-model comparison. The four model-agnostic harnesses are architecturally compatible with all six models, but based on the two representative configurations tested here, no universal ranking of harness strength holds independent of model choice, and significant model–harness interaction is evident from the reversed rankings alone.
Across all six foundation models, average performance of the four model-agnostic harnesses is led by ApodexHarness (0.548), followed closely by A-Evolve (0.544), then DeerFlow (0.521) and OpenHands (0.503). No single harness is universally optimal, and relative rankings shift substantially across model families.

Across the two representative closed-source models, ApodexHarness stays in the top half of the four model-agnostic harnesses on both; A-Evolve leads on GPT but drops to third on Opus, while OpenHands ranks second on Opus but falls to the bottom of the four model-agnostic harnesses on GPT.
DeerFlow also shows substantial variability, ranking last on Opus but third on GPT. We adopt OpenHands as the primary instrument for the episode-mechanism ablations that follow for two reasons: it is fully open-source and auditable, and its control loop can be directly modified to implement interventions such as verifier-guided repair.

These quantitative patterns are reinforced by qualitative observations that aggregates alone do not reveal.
DeerFlow's shortfall is concentrated in premature termination, narrating a plan rather than executing it, a property of its control policy rather than of the underlying model. Full per-environment breakdowns are reported in Appendix~\ref{app:full-matrix}.

Our ApodexHarness illustrates that harness design effects can persist regardless of the foundation model with which the harness is paired. It outperforms OpenHands when paired with both GPT (0.592 vs. 0.560) and Opus (0.611 vs. 0.578), providing a comparison unconfounded by model choice because both harnesses were tested on the same two models. This pattern indicates that harness design makes a meaningful contribution to performance, though a formal decomposition of main effects and interactions is needed to definitively quantify the relative impacts of harness design versus model choice.

\subsubsection{Episode-mechanism ablations}
We next hold the harness, model, environment, and budget fixed and vary a single episode-level mechanism. We study two such mechanisms in turn: initial skill guidance, supplied at episode initialization and active throughout the trajectory, and verifier guidance, which reviews the work before final submission and gives the solver an opportunity to improve it. 
Note that here the ``verifier guidance'' should \emph{not} be confused with ``verifier-repair'' loops presented in Sec.~\ref{sec:repair-case},
because it is a lightweight in-environment guidance rather than an agentic system that provides diagnostics and repair guidance based on the entire solver trajectory.
Table~\ref{tab:episode-mechanism-ablation} reports the completed comparisons across three harness--model configurations. We use these runs as a focused analysis of mechanism behavior rather than as a broad estimate of average effects across tasks.

\begin{table}[t]
\centering
\begin{threeparttable}
\caption{Episode-mechanism ablations under corresponding run configurations.}
\label{tab:episode-mechanism-ablation}
\footnotesize
\setlength{\tabcolsep}{3.5pt}
\begin{tabularx}{\linewidth}{@{}Xllccc@{}}
\toprule
\multicolumn{3}{c}{Run configuration} & \multicolumn{3}{c}{Outcome score} \\
\cmidrule(lr){1-3}\cmidrule(l){4-6}
Environment & Harness & Model & Baseline & Skill guidance & Verifier guidance \\
\midrule
\texttt{LLM-pretrain-data-probe} & Claude Code & Opus-4.8 & 0.563 & 0.518 \loss{0.045} & \na \\
                                  & OpenHands   & Opus-4.8 & 0.442 & 0.557 \gain{0.115} & 0.507 \gain{0.065} \\
                                  & OpenHands   & GPT-5.6-sol & 0.534 & 0.730 \gain{0.196} & 0.659 \gain{0.125} \\
\addlinespace
\texttt{pretrain-data-filter} & Claude Code & Opus-4.8 & 0.485 & 0.441 \loss{0.044} & \na \\
                              & OpenHands   & Opus-4.8 & 0.446 & 0.488 \gain{0.042} & 0.460 \gain{0.014} \\
                              & OpenHands   & GPT-5.6-sol & 0.215 & 0.542 \gain{0.327} & 0.523 \gain{0.308} \\
\bottomrule
\end{tabularx}
\begin{tablenotes}[flushleft]
\footnotesize
\item[] \textit{Note.} Colored triangles denote descriptive changes from the corresponding latest appendix baseline. Verifier guidance is evaluated only with OpenHands because the intervention requires an instrumentable harness control loop. \texttt{LLM-pretrain-data-probe} cells average six episodes; \texttt{pretrain-data-filter} cells average its two canonical instances. Each instance has one run, so observed run-to-run variance limits causal interpretation of small differences.
\end{tablenotes}
\end{threeparttable}
\end{table}

\paragraph{Skill guidance.}
The injected skill provides generic procedural guidance for a problem class; it is not tailored to any evaluated instance and contains no test items or solution information. Skill guidance is provided at episode initialization and shapes decisions throughout the trajectory rather than only at submission. On \texttt{LLM-pretrain-data-probe}, the OpenHands score increases from 0.442 to 0.557 with Opus-4.8 and from 0.534 to 0.730 with GPT-5.6-sol. The latter is the strongest reported condition and succeeds on all six episodes. By contrast, Claude Code--Opus-4.8 scores 0.518 with skill guidance, compared with its latest appendix baseline of 0.563. The same broad pattern holds on \texttt{pretrain-data-filter}: skill guidance improves both OpenHands configurations, most strongly with GPT-5.6-sol ($+0.327$), but not Claude Code--Opus-4.8 ($-0.044$).

Across both environments, the largest gains occur with OpenHands--GPT-5.6-sol, whereas Claude Code--Opus-4.8 has the highest unassisted baseline among the three configurations and does not improve with the injected skill. This may leave less room for generic guidance to add useful behavior, while differences in how a packaged harness incorporates the guidance may also affect how much of it reaches the trajectory. More broadly, skill guidance can partially compensate for weaknesses at the harness layer: although the unassisted OpenHands configurations start below Claude Code, their skill-guided scores match or exceed the Claude Code baseline on both tasks. Skill effectiveness should therefore be treated as a property of the complete solver configuration, and a well-matched skill can strengthen the overall system by narrowing a harness-level performance gap.

\paragraph{Verifier guidance.}
Verifier guidance instead inspects the work near submission and returns concrete issues for the solver to address before it is scored. The reviewer uses the same model and sees only the episode-visible task specification and transcript, with no access to hidden ground truth, reference answers, evaluation outcomes, or scores.
It is also different from the process verifier to be discussed in Sec.~\ref{sec:repair-case},
which is a much heavier system with stronger models to critique the process and trajectory of a system under testing.
We evaluate this mechanism with OpenHands because its open-source control loop exposes the required intervention point. On \texttt{LLM-pretrain-data-probe}, verifier guidance increases the Opus-4.8 score from 0.442 to 0.507 and the GPT-5.6-sol score from 0.534 to 0.659. On \texttt{pretrain-data-filter}, the completed Opus-4.8 comparison changes from 0.446 to 0.460. The reviewer frequently identifies incomplete estimates and weak overlap measurements on \texttt{LLM-pretrain-data-probe}, indicating that outcome-blind feedback can locate deficiencies relevant to the final score even when it cannot inspect that score directly.

We select skill guidance and verifier guidance because they bracket the episode: skill shapes the trajectory from initialization, while verification intervenes near submission. In every completed OpenHands comparison, the guided score exceeds the corresponding appendix baseline, although the gain varies substantially by environment, model, and intervention. Skill guidance produces the larger gain in the configurations for which both interventions are complete. Together with the different result for Claude Code, this pattern shows that guidance does not contribute a fixed increment independent of the surrounding solver system. The reported differences remain descriptive because each instance has one run and the OpenHands prompt conditions are not fully controlled across the baseline and updated variants.

\subsection{Verification-Driven Repair}
\label{sec:repair-case}

Process verification yields a structured diagnosis rather than only a number, and that diagnosis can be returned to the solver to close a loop from measurement to improvement---a concrete and measurable discovery loop at the heart of discoverative AI. Because the subrubric-based HDS6 assessment assigns every band together with a citation to the trajectory steps that determine it, the assessment records both the location of each deficiency and its severity. From it we synthesize a \emph{repair note}: a compact set of instructions that names the specific steps at fault, states the target band each should reach, and supplies a concrete, actionable correction for each. The note withholds the verified outcome, the hidden ground truth, and the outcome score, so that it guides the process without disclosing the answer. It is then inserted into the solver's prompt and the task is re-run under otherwise identical conditions, and re-scoring the repaired trajectory against the same rubric quantifies the effect of the guidance. Repair is in this sense a fixed procedure rather than a manual intervention, and it improves a trajectory without ever relaxing the isolation between solver and ground truth described in Section~\ref{sec:methods-verification}.

We apply this procedure at scale. Every trajectory the process verifier places below band~3 on the HDS6 aggregate is paired with a repaired re-run of the same task under identical conditions, and the two are scored by the environment's own outcome verifier; the repair delta is the difference between them. Table~\ref{tab:repair-delta} reports the mean delta per environment, over the environments whose outcome verifiers and required services were operational in this deployment. These results are experimental: the repair procedure is applied once to already-collected ``deficient'' trajectories (with HDS6 score < 3.0), and its effect on the environment's own outcome score is measured. 

\begin{table}[t]\centering
\caption{Outcome effect of verification-driven repair on deficient trajectories (HDS6 $<3$). For each environment we report the mean change in the environment's own outcome score between a trajectory and its repaired re-run, the number of trajectories, and how many improved, were unchanged, or declined.}
\label{tab:repair-delta}
\small
\begin{tabular}{lrrrrr}
\toprule
Environment & $\Delta$ outcome & $n$ & Improved & Unchanged & Declined \\
\midrule
\texttt{solution\_verifier}        & $+0.365$ & 37 & 20 & 12 & 5 \\
\texttt{aav\_viability}            & $+0.360$ & 34 & 21 & 7  & 6 \\
\texttt{clinical\_sap\_tlf}        & $+0.271$ & 90 & 41 & 48 & 1 \\
\texttt{aav\_capsid\_enrichment}   & $+0.199$ & 40 & 22 & 5  & 13 \\
\texttt{LLM-pretrain-data-probe}   & $+0.082$ & 51 & 27 & 3  & 21 \\
\texttt{LLM\_rollout\_verifier}    & $+0.048$ & 55 & 27 & 20 & 8 \\
\texttt{posttrain\_data\_dedup}    & $+0.040$ & 51 & 26 & 5  & 20 \\
\texttt{internal\_benchmark}       & $+0.034$ & 28 & 7  & 19 & 2 \\
\texttt{aav\_sequence\_design}     & $+0.010$ & 33 & 9  & 19 & 5 \\
\texttt{pretrain\_data\_filter}    & $-0.077$ & 15 & 4  & 6  & 5 \\
\midrule
\textbf{All environments}          & $\mathbf{+0.155}$ & \textbf{434} & \textbf{204} & \textbf{144} & \textbf{86} \\
\bottomrule
\end{tabular}
\end{table}

The effect is positive in nine of the ten environments and in the aggregate: across 434 deficient trajectories the repaired re-run scores $0.155$ higher on average, with 204 trajectories improving against 86 declining. For trajectories the process verifier marks as deficient, the repaired re-runs improve the \emph{outcome} on average and not merely the recorded process, and they do so from a diagnosis that never consults the outcome it improves. Because the study has no matched comparison group---the same deficient trajectories re-run without a repair note---we cannot fully separate this improvement from ordinary run-to-run variation. 

In the rest of this section, we complement the results in Table \ref{tab:repair-delta} by presenting two case studies on how our HDS6 process verifier improves two concrete trajectories, from two episodes in different environments.

\subsubsection{Case study: an under-specified clinical safety report}

The \texttt{clinical\_sap\_tlf} environment poses a clinical-reporting task: reproduce a treatment-emergent adverse-event safety table from a statistical analysis plan (SAP). The plan is deliberately \emph{under-specified}---it never names the patient population that forms the denominator, nor which adverse-event flag counts---so the solver must itself choose these analysis decisions and, ideally, declare them before computing, exactly the judgment a trial biostatistician is expected to exercise. Outcome scores incorporate both accuracy of the numerically calculated quantities, but also rigor, reproducibility and accountability of such calculations which are part of the report as submitted artifacts.

\begin{figure}[p]
  \centering
  \includegraphics[width=\textwidth,keepaspectratio]{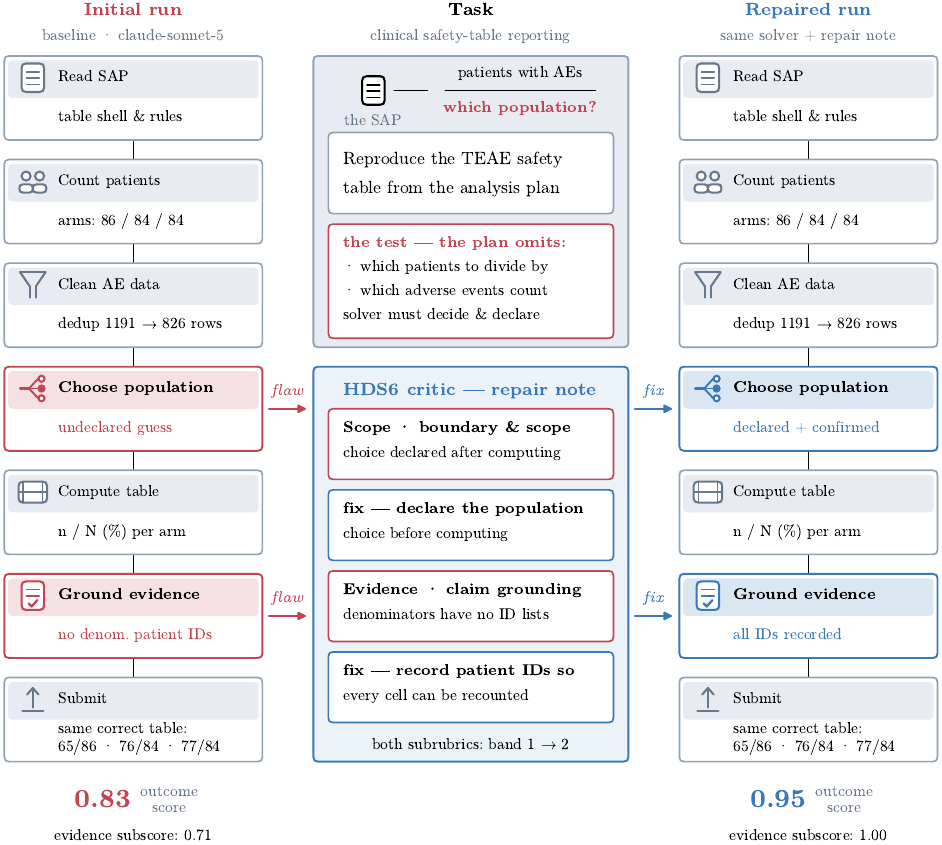}
  \caption{\textbf{A verification--repair episode on \texttt{clinical\_sap\_tlf}}
  (full walkthrough in Section~\ref{sec:repair-case}). The task is to reproduce a
  treatment-emergent adverse-event (TEAE) safety table from the trial's
  statistical analysis plan (SAP); each cell divides the patients with adverse
  events (AEs) by a patient population that the plan deliberately never names ---
  the fraction sketched in the task card. The left and right columns are the same
  solver's two attempts; white steps run identically in both, and the two red
  steps of the first attempt are where the critic's findings land. The centre column states the task
  and shows the outcome-blind HDS6 critique: each red-bordered box is a finding,
  each blue-bordered box the fix it prescribes, and the repair note carries both
  into the second attempt. The two runs submit the identical, correct table
  (bottom row), so the $0.83 \rightarrow 0.95$ score gap lies entirely on the
  evidence/traceability axis ($0.71 \rightarrow 1.00$) which is part of the submitted artifact as reports and evidences of the said calculations.}
  \label{fig:repair-case}
\end{figure}

Figure~\ref{fig:repair-case} traces a matched pair of trajectories from the same solver (\texttt{claude-sonnet-5}) on one such episode: an initial run (left), the outcome-blind process critique of that run (centre), and the repaired re-run it induces (right). Both runs submit the same quantities as outcome, but their reports differ in terms of rigor and accountability, as we will explain in the next paragraph. Reading only the first run's trajectory, the process critic flags two band-1 (of~2) subrubric failures. \texttt{boundary\_and\_scope} (the \textbf{Scope} capability) fires because the solver computed the table \emph{before} declaring its under-specified population choice, leaving the assumptions log as after-the-fact justification; \texttt{claim\_grounding} (the \textbf{Evidence} capability, here the subrubric of Table~\ref{tab:subrubric-example} instantiated for the clinical task) fires because the evidence ledger records the subject IDs behind each numerator but never behind the denominator, so the reported percentages cannot be independently recomputed.

The repair note carries these two findings---and only these---into the re-run. Guided by it, the repaired trajectory declares the population choice up front by interacting with the ``virtual clinician" provided in the environment, and records patient IDs for both the numerators and the denominators in the submitted artifact, so that every cell becomes independently recountable. In terms of result, the first run's submitted artifact lists patient IDs only for the numerators and omits them for the denominators; its deliverable therefore cannot be fully recounted, and the outcome verifier scores it lower---$0.83$ against the repaired run's $0.95$, a gap lying entirely on the evidence/traceability axis ($0.71\rightarrow1.00$). With the missing denominator ID lists supplied, both flagged subrubrics rise to band~2 and the outcome score rises accordingly.

\subsubsection{Case study: a named estimator that was never run}
\label{sec:repair-case-probe}

The \texttt{LLM-pretrain-data-probe} environment poses a corpus-scouting task essential for data collection in LLM pre-training efforts: estimate how many unique high-value documents a collection of sources holds, together with their deduplicated token total, while drawing only small metered samples under a fixed budget. The brief states that naive document counting over-counts, because mirrors and near-duplicates re-serve the same content, and directs the solver to estimate the population by capture--recapture on the cross-source overlap. Because no source can be read in full, its size is never observed and must itself be inferred, so the submitted total rests on an inference the solver has to construct rather than on a quantity it can look up. The failure worth catching in this environment is therefore an estimate that carries the form of that inference without its substance, and it differs from the clinical case of Section~\ref{sec:repair-case} in a way that matters for repair: there the numbers were already correct and the deficiency lay in the audit trail, whereas here an ungrounded link propagates directly into the reported quantity.

\begin{figure}[p]
  \centering
  \includegraphics[width=\textwidth,keepaspectratio]{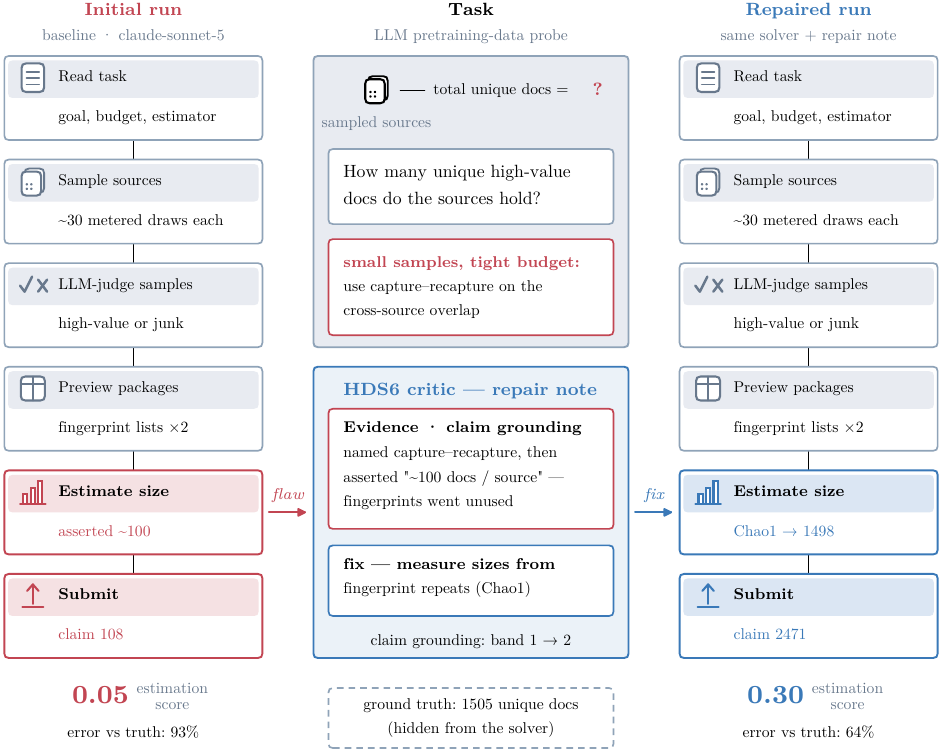}
  \caption{\textbf{A verification--repair episode on \texttt{LLM-pretrain-data-probe}}
  (full walkthrough in Section~\ref{sec:repair-case-probe}). The left and right
  columns are the same solver's two attempts at estimating how many unique
  high-value documents a corpus holds; white steps run identically in both. The
  first attempt asserts a source size of ${\sim}100$ documents and submits $108$;
  the critic's finding (red-bordered box) records that the run named
  capture--recapture while the content-hash fingerprints it had downloaded went
  unused, and the
  prescribed fix (blue-bordered box) drives the repaired run to measure each
  source from its fingerprint repeats (Chao1 $\rightarrow 1498$) and submit
  $2471$. Against the hidden truth of $1505$ documents (dashed box, never shown
  to the solver), the two claims are $93\%$ and $64\%$ off, and the estimation
  score moves $0.05 \rightarrow 0.30$.}
  \label{fig:repair-case-probe}
\end{figure}

Figure~\ref{fig:repair-case-probe} traces a matched pair of trajectories from the same solver (\texttt{claude-sonnet-5}) on one episode of this environment: an initial run (left), the outcome-blind process critique of that run (centre), and the repaired re-run it induces (right). The initial run draws approximately thirty documents from each source and has an LLM-judge rate them, obtaining a high-value share of nine of thirty for \texttt{src\_04}. It then supplies the remaining factor by assertion, taking each source to hold about one hundred documents, and scales the judged share by that constant, summing the per-source products together with a purchased 190-document package to submit $M_{\mathrm{high}} = 108$ with the interval $[75,145]$. The three crawlable sources in fact hold 451, 463 and 1498 documents, so the asserted constant is between four and fifteen times too small and the submitted total falls roughly an order of magnitude below the true value of 1505.

Reading only the initial trajectory, the process critic assigns \texttt{claim\_grounding}---the \textbf{Evidence} capability, and the subrubric of Table~\ref{tab:subrubric-example}---band~1 of~2, recording that the run named capture--recapture while computing a small-sample scale-up and that the content-hash fingerprints it had already downloaded went unused. The middle band, rather than the lowest, is the precise assessment, and the trajectory shows why: the sampling and the judging are genuine and are recorded step by step, and every arithmetic operation from the judged share to the submitted total is sound, so that exactly one link in the chain, the size of the source, rests on no observation. The integrity gate, which alone may consult the verified outcome, passes for both runs, which places the deficiency in grounding rather than in fabrication. This configuration is the one an outcome score cannot resolve: a scalar records that 108 is wrong, whereas the band and its citation identify which input of the estimator was never measured, and that distinction is what makes the diagnosis actionable.

The repair note carries this finding into the re-run, directing the solver above all to use the per-document fingerprints it had already downloaded to measure cross-source overlap rather than to assume a size. Guided by it, the repaired trajectory re-samples each source and tabulates how often individual documents recur; for \texttt{src\_04} it draws 450 times and observes 391 distinct documents, of which 336 appear exactly once and 51 appear exactly twice. Because first sightings dominate recurrences, most of the source remains unseen, and the Chao1 estimator accordingly returns $391 + 336^{2}/(2\cdot 51) \approx 1498$ documents for that source in place of the one hundred previously asserted. Applying the same procedure to each source, discounting cross-source duplicates by a factor of $0.97$ that the run set conservatively from the mirrors it had observed, and adding the purchased package yields $M_{\mathrm{high}} = 2471$ with the interval $[2129,2813]$.

The estimation score rewards how close the reported quantity is to the truth. The initial run's document and token counts are far from the true values, whereas the repaired run's are markedly closer, so its estimation score rises sharply---a clear gain on exactly the axis the critique named. The repaired estimate, while being closer to the truth in relative terms, still overshoots the truth, which places the remaining gap on a calibration axis this note did not target and marks interval-targeted repair as the next increment. The aggregate score improves as well, and its decomposition is instructive: procurement quality is essentially unchanged, but the extra measurement the correction requires consumes budget, so the efficiency term falls even as the estimate improves. 

\section{Related Work}
\label{sec:related}

\paragraph{Knowledge and reasoning benchmarks.}
A large body of work evaluates language models by comparing a single produced answer against a reference key. Broad knowledge and reasoning suites include MMLU~\cite{hendrycks2021mmlu}, BIG-bench~\cite{srivastava2023bigbench}, the AI2 Reasoning Challenge~\cite{clark2018arc}, and the graduate-level GPQA~\cite{rein2023gpqa}, while mathematical ability is measured by GSM8K~\cite{cobbe2021gsm8k} and MATH~\cite{hendrycks2021math}. More recent frontier suites such as FrontierMath~\cite{glazer2024frontiermath} and Humanity's Last Exam~\cite{phan2025hle} raise difficulty sharply while retaining the same answer-checking format. These benchmarks measure what a model knows and whether it can reason in a single pass, but they expose neither tools nor state, and they record nothing of the process by which an answer is reached.

\paragraph{Interactive and agentic benchmarks.}
A second line embeds a model in an executable environment and scores task success. The practice of evaluating agents by having them act in an environment has long-standing roots in reinforcement learning, from the Arcade Learning Environment~\cite{bellemare2013ale} to OpenAI Gym~\cite{brockman2016gym}. This paradigm was later taken up by interactive environments for language agents, including the text-embodied world of ALFWorld~\cite{shridhar2021alfworld}, grounded web interaction in WebShop~\cite{yao2022webshop} and Mind2Web~\cite{deng2023mind2web}, and interactive coding with execution feedback in InterCode~\cite{yang2023intercode}. Software engineering is the most developed domain, spanning HumanEval~\cite{chen2021humaneval}, SWE-bench and its human-validated subset SWE-bench Verified~\cite{jimenez2024swebench,openai2024swebenchverified}, and SWE-Gym~\cite{pan2024swegym}. Broader agentic suites target the web, the operating system, and general tool use, including WebArena~\cite{zhou2023webarena}, VisualWebArena~\cite{koh2024visualwebarena}, OSWorld~\cite{xie2024osworld}, AppWorld~\cite{trivedi2024appworld}, GAIA~\cite{mialon2023gaia}, AgentBench~\cite{liu2023agentbench}, $\tau$-bench~\cite{yao2024taubench}, ToolLLM~\cite{qin2024toolllm}, and Terminal-Bench~\cite{terminalbench2025}, and specialized settings evaluate cybersecurity~\cite{zhang2024cybench} and machine-learning or research engineering, as in MLAgentBench~\cite{huang2024mlagentbench}, MLE-bench~\cite{chan2024mlebench}, RE-Bench~\cite{wijk2024rebench}, and PaperBench~\cite{starace2025paperbench}. Closest to the scientific domains we target are ScienceAgentBench~\cite{chen2024scienceagentbench}, DiscoveryBench~\cite{majumder2024discoverybench}, the interactive science environment ScienceWorld~\cite{wang2022scienceworld}, and LAB-Bench~\cite{laurent2024labbench}. A complementary strand grounds tasks in measurable real-world value: GDPval~\cite{gdpval2025} assembles professional deliverables across dozens of economically significant occupations, SWE-Lancer~\cite{miserendino2025swelancer} draws software tasks from real freelance contracts with associated payouts, and Agents' Last Exam~\cite{sun2026ale} evaluates agents on existing jobs that human professionals perform today, while the time-horizon analysis of Kwa et al.~\cite{kwa2025longtasks} measures the length of tasks that agents complete reliably. These benchmarks establish interactivity and real-world value across an expanding range of domains, and TRACES builds on them by verifying load-bearing intermediate results and reporting a process-quality signal alongside task success, within a single environment--task--episode abstraction that spans biomedical, clinical, and frontier-model-engineering problems.

\paragraph{Agent scaffolds.}
Because we evaluate a model together with the harness that drives it, the scaffolds that turn a model into an agent are directly relevant. General prompting and control strategies include ReAct~\cite{yao2023react}, Tree of Thoughts~\cite{yao2023tot}, Reflexion~\cite{shinn2023reflexion}, and Self-Refine~\cite{madaan2023selfrefine}; tool use is enabled by Toolformer~\cite{schick2023toolformer}; and multi-agent and open-ended frameworks include AutoGen~\cite{wu2023autogen} and Voyager~\cite{wang2023voyager}. In software engineering, agent--computer interfaces such as SWE-agent~\cite{yang2024sweagent} and platforms such as OpenHands~\cite{openhands} package a model into a complete solver. Iterative self-correction has been studied both as a tool-grounded procedure, as in CRITIC~\cite{gou2024critic}, which revises outputs through interaction with external tools, and as an intrinsic capability, where Huang et al.~\cite{huang2024selfcorrect} find that self-correction without external feedback often does not improve, and can degrade, reasoning; these findings motivate the verifier-grounded feedback on which our repair loop depends. Our framework treats the surrounding scaffold as part of the system under evaluation rather than as fixed infrastructure, and it therefore measures a model and its harness jointly.

\paragraph{Automated and {LLM}-based evaluation.}
Using strong models to judge the outputs of others has become standard practice, from pairwise, arena-style judging~\cite{zheng2023judging} to reference-free scoring with G-Eval~\cite{liu2023geval} and open evaluator models such as JudgeLM~\cite{zhu2023judgelm} and Prometheus~2~\cite{kim2024prometheus2}. Rubric- and skill-based protocols, including FLASK~\cite{ye2024flask} and, in medicine, HealthBench~\cite{arora2025healthbench}, decompose quality into explicit criteria. Closest to process evaluation, trained critics such as CriticGPT~\cite{mcaleese2024critics} produce natural-language critiques that surface errors in model-written code and can match or exceed human reviewers at locating them. These judges are typically single-pass, outcome-facing, and aware of the model identity or the reference answer. Our process verifier differs on each count: it is agentic rather than single-pass, it grounds every judgment in cited trajectory evidence, and it is blind both to the verified outcome and to the identity of the solver.

\paragraph{Process supervision and verifiable rewards.}
A related thread supervises the reasoning process rather than only the final answer. Uesato et al.~\cite{uesato2022process} contrasted process- and outcome-based feedback on mathematical problems, and Lightman et al.~\cite{lightman2024verify} then showed that step-level supervision produces markedly stronger verifiers; Math-Shepherd~\cite{wang2024mathshepherd} and generative verifiers~\cite{zhang2024genrm} subsequently reduced the annotation cost of such process reward models, and the reliability of step-level judgments has itself become an object of study, as in ProcessBench~\cite{zheng2024processbench}, which measures how accurately a critic localizes the first erroneous step in a reasoning trace. In parallel, reinforcement learning from verifiable rewards has driven recent reasoning models, including DeepSeek-R1~\cite{guo2025deepseekr1} and open recipes such as T\"ulu~3~\cite{lambert2024tulu3}, building on the verifier-based approach to math word problems of Cobbe et al.~\cite{cobbe2021gsm8k}. Our HDS6 metric extends process evaluation from short arithmetic chains to long, tool-using trajectories, and grades them against task-specific, capability-level rubrics rather than a single scalar reward.

\paragraph{Specification gaming and contamination.}
Our reliance on hidden verifiers and anti-gaming gates connects to work on specification gaming and reward hacking~\cite{amodei2016concrete,skalse2022reward} and to the growing concern over benchmark data contamination~\cite{sainz2023contamination,golchin2024timetravel}. Both literatures motivate verifiers that a solver cannot inspect and outcomes that cannot be reached by shortcut, which are design commitments of the executable environment.

\paragraph{Autonomous scientific discovery.}
Finally, a growing body of systems pursues end-to-end scientific research. Coscientist~\cite{boiko2023coscientist} and ChemCrow~\cite{bran2024chemcrow} couple large language models to laboratory tools and automation to plan and execute chemistry experiments; the AI co-scientist~\cite{gottweis2025aicoscientist} generates and refines novel research hypotheses through multi-agent debate; The AI Scientist~\cite{lu2024aiscientist} carries out machine-learning research end to end; and broader analyses chart the emerging role of biomedical AI agents~\cite{gao2024biomedagents}. More broadly, AI is increasingly central to scientific discovery~\cite{wang2023sciencediscovery}, with concrete results ranging from new algorithms found by evolutionary coding agents~\cite{novikov2025alphaevolve} to the large-scale discovery of stable materials~\cite{merchant2023gnome} and the autonomous laboratories that synthesize them~\cite{szymanski2023alab}. These systems share our long-term goal of open-ended discovery, and Apodex Discovery supplies the reality-based environments, hidden verifiers, and process-level evaluation through which such systems can be measured, compared, and improved on consequential problems.

\section{Conclusion}
\label{sec:conclusion}

Apodex Discovery advances three complementary capabilities for building and evaluating discoverative AI on consequential real-world problems. It provides a systematic process for identifying problems that are both high-value and objectively verifiable; it converts those problems into executable environments in which complete solver systems can act through data, tools, and feedback; and it introduces HDS6 to evaluate the quality of the investigation itself when definitive outcome ground truth is delayed, incomplete, or not yet available. Together, these components shift evaluation from asking whether a model produced the right answer to asking whether a solver conducted a reliable, evidence-grounded, and self-correcting investigation. Alongside this paper we release \textbf{TRACES}, the reality benchmark that instantiates this framework: seventeen executable environments spanning 218 episodes across biomedicine, clinical translation, and frontier-model engineering, all with hidden verifiers and two supporting the full verification--repair loop.

Our initial results show why this broader evaluation unit matters. In adeno-associated virus capsid design, domain-specific environments improved performance across the discovery pipeline, with Apodex exceeding prior human baselines on the most demanding tasks. In drug repurposing, the biomedical environment improved the mean normalized prediction score of GPT-5.5 and GPT-5.6-sol by 2.5 and 7.6 points, respectively, over the same closed-book backbone. Controlled ablations further demonstrate that Apodex Discovery can attribute performance differences to distinct parts of the solver system---including the foundation model, harness, guidance, budget handling, and verifier-driven repair---rather than treating every failed episode as a generic model-capability failure.

\paragraph{Limitations and Extensions.}
The TRACES benchmark currently instantiates seventeen executable environments with hidden verifiers, spanning 218 episodes and realizing a subset of the twenty selected high-value problems across biomedicine, clinical translation, and frontier-model engineering; two of these environments support the full verification--repair loop. This already constitutes a substantial benchmark, although coverage remains limited to this initial subset of the 423-problem registry, with many additional environments under development and expected to be released in the near future. Our process-verification results likewise provide a strong initial demonstration while leaving clear room for further development, to align process verification scores better with outcome scores and demonstrate its effectiveness over an even broader range of applications.

\section{Author Contributions and Acknowledgments}
We gratefully acknowledge Mr. Tianqiao Chen, the project lead of this work, who proposed the concept of the heavy-duty solver, defined the six capabilities of the HDS6 process metric, and shaped the overall architecture of Apodex Discovery. We sincerely thank him for the vision, guidance, and support that made this work possible.

We would like to acknowledge Dr. Yuan Cai, Dr. Chengyu Fu, Dr. Shiyi Zhang, and Harry Zhang for the insightful discussion and contribution.

\paragraph{Core technical contributors.} Sheng Wang, Brian Wang, Bin Feng, Xiaoman Pan, Chenyang An, Felix Liu.

\paragraph{Contributors.} Tangqi Fang, Gongbo Sun, Yueqi Guo, Kailong Wen,
Feng Xing, Yiling Guo, 
Lingfeng Shen, Ning Wang, Handuo Zhang, Feng Chen, Fuchao Yang,
Jiacheng Lin, Siting Li, Zixuan Liu, Xiang Wang, 
Chi Han, Zhenhailong Wang, Kunlun Zhu, Lawrence Zhao, Lidong Bing, David Tan, Bo An, Heng Ji.


\newpage
\bibliographystyle{unsrtnat}
\bibliography{references}


\sisetup{
  table-format = 1.3,
  table-alignment = center,
  detect-weight = true,
}
\newcolumntype{M}{S[table-format=1.3]}

\newcommand{\tabnote}[1]{%
  \par\vspace{4pt}
  \begin{minipage}{0.9\linewidth}\footnotesize #1\end{minipage}%
}

\appendix

\newpage
\section{Domains and questions considered by Apodex Discovery}
\label{app:domains} 
We list below the domains and questions considered by Apodex Discovery.

\begingroup
\footnotesize
\setlength{\tabcolsep}{4pt}
\renewcommand{\arraystretch}{1.15}
\begin{xltabular}{\textwidth}{@{}r@{\hspace{6pt}} Y{1.00} Y{1.00} Y{1.25} Y{0.75}@{}}
\caption{The ten selected domains.}\label{tab:domains}\\
\toprule
\hdr{\#} & \hdr{Domain} & \hdr{Why the value is high} & \hdr{Why it suits a long-chain reasoning agent} & \hdr{Representative players} \\
\midrule
\endfirsthead
\multicolumn{5}{@{}l}{\footnotesize\emph{Table \thetable{} --- continued from previous page}}\\[2pt]
\toprule
\hdr{\#} & \hdr{Domain} & \hdr{Why the value is high} & \hdr{Why it suits a long-chain reasoning agent} & \hdr{Representative players} \\
\midrule
\endhead
\midrule
\multicolumn{5}{r@{}}{\footnotesize\emph{continued on next page}}\\
\endfoot
\bottomrule
\endlastfoot

1 & Bioinformatics, Computational Biology \& Genomic Medicine\newline\sub{Healthcare \& Life Sciences}
  & Diagnosis-to-therapy genomics approaches a \$100B vertical; breakthrough algorithmic biology carries near-Nobel recognition.
  & The agent assembles alignment and variant-calling pipelines, runs them, and verifies output against held-out truth sets; fuses multi-omics layers and reasons over pathways to testable hypotheses with quantified uncertainty.
  & No single incumbent --- an AI-startup-saturated category; pharma R\&D informatics contracts limit the flywheel. \\
\addlinespace[3pt]

2 & Foundation Models for Scientific Reasoning\newline\sub{Software \& Internet}
  & A true scientific-reasoning foundation model would be Turing/Nobel-class \emph{and} a $>$\$100B platform under all of R\&D --- it maxes both axes.
  & The agent drafts a proof, formalises it in Lean and accepts only what the kernel verifies; derives governing equations, discretises them to simulator code, and cross-checks against numerical ground truth.
  & Frontier labs --- the single hottest AI-native battleground; model-swap risk weakens lock-in. \\
\addlinespace[3pt]

3 & Scientific HPC Code Generation \& Optimization\newline\sub{Software \& Internet}
  & Developer-productivity and compute, a \$10--100B opportunity adjacent to the AI-coding boom.
  & The agent ports CUDA to HIP/SYCL and verifies bitwise numerical agreement; applies tiling and vectorisation, measures speedup, and rejects regressions at a CI benchmark gate.
  & Copilot, Cursor and the hyperscalers contest general code generation; the scientific-HPC niche is less contested. \\
\addlinespace[3pt]

4 & Healthcare IT \& Digital Health\newline\sub{Healthcare \& Life Sciences}
  & EHR, telehealth and monitoring form a vertical approaching \$100B with platform-scale winners.
  & Long-horizon reasoning over longitudinal records, with outcomes that arrive as data rather than as opinion.
  & Epic and the incumbent EHR platforms. \\
\addlinespace[3pt]

5 & Weather-Driven Commodity \& Energy Market Forecasting\newline\sub{Agriculture \& Food ($+$ Energy \& Environment)}
  & Grain and soft commodities each exceed \$100B, dominated by a few majors controlling sourcing, storage and logistics; energy trading supports a $>$\$100B opportunity.
  & Physical-world forecasting where the verifier is the official report that lands weeks later --- a genuine held-out future, not a stored key.
  & The ABCD grain majors; concentrated merchant and roaster franchises. \\
\addlinespace[3pt]

6 & Structural \& Civil Engineering Services\newline\sub{Real Estate \& Construction}
  & A \$10--50B global vertical, squarely in the \$10--100B band.
  & Code compliance is machine-checkable: layouts can be generated, then verified against the standard and against seismic constraints.
  & Legacy CAD incumbents; concentrated engineering firms make the buyer reachable. \\
\addlinespace[3pt]

7 & Chip Design \& Verification (EDA)\newline\sub{Hardware \& Semiconductors}
  & A \$10--100B concentrated, high-margin tooling industry essential to all chip design; the fabless model it serves is a $>$\$100B market.
  & The whole design pipeline runs inside the tools and is verified \emph{in silico} before fabrication --- a fast, exact loop over a hard search.
  & Synopsys and Cadence in signoff tooling; TI and ADI in analog; Nvidia, Qualcomm, Broadcom, AMD as fabless designers. \\
\addlinespace[3pt]

8 & Market Data \& Financial Indices\newline\sub{Financial Services}
  & Oligopolistic, extremely high-margin franchises whose aggregate addressable revenue exceeds \$100B.
  & Claims are settled by the market itself on a fixed horizon; out-of-sample tracking is native to the domain.
  & Bloomberg, S\&P, MSCI. \\
\addlinespace[3pt]

9 & Materials Discovery, Design \& Qualification\newline\sub{Materials \& Chemicals}
  & The materials-discovery prize, with recognition on the scale of the Materials Genome effort, over a \$10--100B-plus advanced-materials market.
  & Inverse design closes a loop: propose a structure, predict the property, and qualify against measurement.
  & Crowded with AI-native entrants; DeepMind's GNoME contests it at hyperscaler scale. \\
\addlinespace[3pt]

10 & Industrial Automation \& Control\newline\sub{Industrials \& Manufacturing}
  & PLC, DCS and MES software and hardware form a vertical approaching \$100B --- critical infrastructure with no public fame.
  & Control problems are formalisable and simulatable, with plant telemetry as the verifier.
  & Siemens, Rockwell, Emerson, Schneider. \\
\addlinespace[3pt]
\end{xltabular}
\endgroup


\begingroup
\footnotesize
\setlength{\tabcolsep}{4pt}
\renewcommand{\arraystretch}{1.15}
\begin{xltabular}{\textwidth}{@{}r@{\hspace{6pt}} Y{1.05} Y{1.30} Y{1.00} Y{0.65}@{}}
\caption{The twenty-question registry.}\label{tab:registry}\\
\toprule
\hdr{\#} & \hdr{Question} & \hdr{What the solver must do} & \hdr{How the answer is checked} & \hdr{Buyer} \\
\midrule
\endfirsthead
\multicolumn{5}{@{}l}{\footnotesize\emph{Table \thetable{} --- continued from previous page}}\\[2pt]
\toprule
\hdr{\#} & \hdr{Question} & \hdr{What the solver must do} & \hdr{How the answer is checked} & \hdr{Buyer} \\
\midrule
\endhead
\midrule
\multicolumn{5}{r@{}}{\footnotesize\emph{continued on next page}}\\
\endfoot
\bottomrule
\endlastfoot

1 & LLM training recipe design\newline\sub{Frontier model R\&D}
  & Design and optimise pretrain / midtrain / posttrain recipes --- architecture, optimiser, schedule, RL, distillation --- under a compute constraint.
  & Must beat expert-tuned baselines at matched compute.
  & Frontier and enterprise model builders \\
\addlinespace[3pt]

2 & Training data curation \& decontamination\newline\sub{Frontier model R\&D}
  & Discover, filter, construct and decontaminate training and benchmark data at scale, under budget and throughput limits.
  & Downstream quality at fixed budget; contamination measurably removed.
  & Frontier and enterprise model builders \\
\addlinespace[3pt]

3 & LLM training \& serving infrastructure\newline\sub{Frontier model R\&D}
  & Build and verify reliable, high-throughput training and serving infrastructure across models and hardware.
  & Bit-exact reproduction and measured throughput across targets.
  & Model builders; compute providers \\
\addlinespace[3pt]

4 & Clinical trial statistical programming\newline\sub{Healthcare \& Life Sciences}
  & Generate SAP-faithful trial analysis code and top-line tables, listings and figures.
  & Cell-level audit lineage back to the protocol; fidelity to the statistical analysis plan.
  & Pharma biostatistics; CROs \\
\addlinespace[3pt]

5 & Clinical trial safety \& efficacy signal detection\newline\sub{Healthcare \& Life Sciences}
  & Auto-generate and stress-test safety and efficacy hypotheses from accruing trial data.
  & Signals validated against post-cutoff outcomes.
  & Pharma; medical monitors; CROs \\
\addlinespace[3pt]

6 & Drug repurposing and reformulation\newline\sub{Healthcare \& Life Sciences}
  & Rank repurposing candidates for a target disease using only evidence visible before a cutoff.
  & Enrichment of the top-$K$ against post-cutoff clinical and real-world success.
  & Pharma; biotech; disease foundations \\
\addlinespace[3pt]

7 & Causal drug-target validation\newline\sub{Healthcare \& Life Sciences}
  & Decide whether a candidate gene target is a causal driver or an associated passenger, and name the falsifying experiment.
  & Held-out genetics and perturbation outcomes; known target successes and failures.
  & Pharma discovery; biotech \\
\addlinespace[3pt]

8 & Non-consensus investment thesis generation\newline\sub{Financial Services}
  & Surface overlooked evidence, non-consensus theses, and bull / base / bear scenarios for deep-tech decisions.
  & Historical time-split replay, shadow portfolio, out-of-sample thesis tracking.
  & Asset managers; venture and growth investors \\
\addlinespace[3pt]

9 & Crop yield shortfall forecasting\newline\sub{Agriculture \& Food}
  & Predict localised crop-yield shortfalls sixty days ahead of official reports, from satellite, soil and climate data.
  & The official report, when it lands.
  & Agricultural traders; food processors; insurers \\
\addlinespace[3pt]

10 & Code-compliant structural design\newline\sub{Real Estate \& Construction}
  & Generate structural and foundation layouts from a floor plan, minimising embodied carbon under seismic constraints.
  & Automated code-compliance check; carbon and seismic constraints evaluated.
  & Engineering and design firms; developers \\
\addlinespace[3pt]

11 & Cell therapy manufacturing optimization\newline\sub{Healthcare \& Life Sciences}
  & Optimise per-batch manufacturing parameters to raise potency and post-infusion persistence.
  & Measured batch potency and persistence after infusion.
  & Cell-therapy manufacturers; CDMOs \\
\addlinespace[3pt]

12 & AAV capsid assessment and design\newline\sub{Healthcare \& Life Sciences}
  & Design organ-specific AAV capsid variants end-to-end from public sequence--function data.
  & Unseen variants under a leak-proof hold-out.
  & Gene-therapy developers \\
\addlinespace[3pt]

13 & Biological age \& mortality risk prediction\newline\sub{Healthcare \& Life Sciences}
  & Estimate biological age and mortality risk from multimodal health data --- records, imaging, labs.
  & Unseen held-out cohorts.
  & Health systems; insurers; longevity clinics \\
\addlinespace[3pt]

14 & iPSC-cardiomyocyte differentiation modeling\newline\sub{Healthcare \& Life Sciences}
  & Model iPSC-to-cardiomyocyte differentiation trajectories and the effect of disease mutations.
  & Unseen donors and unseen mutations.
  & Cardiac drug discovery; academic labs \\
\addlinespace[3pt]

15 & Rare disease diagnosis from whole-genome data\newline\sub{Healthcare \& Life Sciences}
  & Produce structured diagnostic reports from raw whole-genome or whole-exome data and match patients to trial eligibility.
  & Confirmed diagnoses; eligibility verified against trial criteria.
  & Rare-disease centres; patient organisations \\
\addlinespace[3pt]

16 & Metabolic treatment response prediction\newline\sub{Healthcare \& Life Sciences}
  & Predict response and non-response, and rank the best twelve-month interventions per patient.
  & Twelve-month weight, HbA1c, discontinuation and rebound outcomes.
  & Payers; health systems; pharma \\
\addlinespace[3pt]

17 & Preclinical-to-human translation prediction\newline\sub{Healthcare \& Life Sciences}
  & Predict whether a preclinical result will translate, and attribute the cross-species causal gap when it does not.
  & Paired preclinical and human outcomes on held-out drugs.
  & Pharma R\&D; biotech \\
\addlinespace[3pt]

18 & Near-term health deterioration prediction\newline\sub{Healthcare \& Life Sciences}
  & Predict near-term health decline from longitudinal personal data and recommend safe next actions.
  & Observed near-term decline and escalation outcomes.
  & Health systems; payers; care providers \\
\addlinespace[3pt]

19 & Work-state reconstruction \& open-loop detection\newline\sub{Software \& Internet}
  & Reconstruct active work state across a company's tools and detect unresolved commitments, decisions and risks.
  & Whether flagged items were genuinely open, and whether the loop closed.
  & Enterprises \\
\addlinespace[3pt]

20 & Material-event detection from real-time news\newline\sub{Software \& Internet}
  & Monitor real-time signals for events that materially change a user's world model, with alternative readings.
  & Whether flagged events proved material, judged after the fact.
  & Enterprises; investors; analysts \\
\addlinespace[3pt]
\end{xltabular}
\endgroup

\newpage

\section{Adeno-associated virus capsid design full results}
\label{app:aav-full}

\paragraph{Task~1: Viability.}Each variant is classified as viable (it assembles and packages its genome) or not, from its amino-acid sequence alone. The two instances test complementary extrapolation, a multi-mutation instance that extrapolates along mutational load~\cite{bryant2021aav}, and a single-mutation instance tiled across the whole capsid sequence that extrapolates across sequence position and mutation type~\cite{Ogden2019ComprehensiveAC}. Because viability is strongly confounded with mutational load, a predictor that merely counts mutations already scores well; the primary metric is therefore the AUROC computed separately within each exact mutation count and then averaged, isolating whether a model ranks survivors above non-survivors among equally mutated variants (chance 0.5). The multi-mutation instance trains on variants with at most 12 mutations and tests on those with 13 or more. It has 139,268 training and 133,433 test variants, plus 20,816 behind the validation oracle. The single-mutation instance trains on substitutions and tests on insertions and deletions. It has 15,409 training and 13,891 test variants, plus 1,544 behind the oracle.

\paragraph{Task~2: Tropism.}
A seven-amino-acid peptide inserted into a fixed surface loop of the AAV9 capsid is scored for its ability to reach each target tissue, against the measured enrichment of each peptide in that tissue from the Fit4Function screens~\cite{eid2024systematic}. Three instances of increasing difficulty are reported, multi-organ tropism in mouse, cross-species transfer from mouse and in-vitro data to macaque, and human in-vitro assays. The primary metric is the rank correlation between predicted and measured enrichment (Spearman for the mouse and cross-species instances, mean per-assay Pearson for the human in-vitro instance), The mouse instance draws 73,089 training and 15,679 validation peptides at random from one library and is scored on 6,032 held-out ones. Every test peptide there is at Hamming distance at least 2 from every training peptide. The human in-vitro instance splits its own library the same way: 66,476 training, 14,244 validation and 14,244 test peptides. The cross-species instance trains on 50,235 mouse and in-vitro peptides and is scored on 13,319 peptides from a separate macaque library, with 12,589 more for validation.

\paragraph{Task~3: Structure.}
Given only the amino-acid sequence, a solver must produce the three-dimensional capsid structure by orchestrating a provided set of folding engines and analysis tools inside a network-isolated environment. Three structural levels are scored, the variable surface loops of the single subunit, by per-residue positional accuracy on the loop atoms; the full 60-mer icosahedral shell, by a geometric-accuracy term and a physical-plausibility term; and the capsid in complex with an antibody or receptor, by an interface-docking quality~\cite{Basu2016DockQAQ} and the fraction of recovered native interface contacts. Leakage is controlled by a temporal split (only structures deposited after 30~September~2021, past each folding engine's training cut-off), sequence-similarity filtering, and no network access. The loop and shell levels score the same 12 post-cutoff capsids. Each is under 95\% identical to its closest match among the 62 pre-cutoff structures released as a reference pool. The complex level scores 15 targets: 8 with an antibody and 7 with the AAVR receptor.

\paragraph{Task~4: Design.}
A solver generates up to 1000 novel peptides for a target. It starts from a labeled seed set: 2,000 training and 500 validation peptides on each tissue instance, 188,000 and 10,000 on each receptor instance. It may train surrogate models on that set. It may also query a scoring oracle, but on at most 100 sequences, a tenth of what it submits, so the oracle cannot be brute-forced. A design counts only if it is simultaneously novel (at least two substitutions from every training peptide), manufacturable (passing a viability gate), and on target. On-target is judged by rank, for the tissue instances (brain, spinal cord, heart), a design qualifies if its predicted organ selectivity ranks in the top 1\%, 2\%, or 5\% of all sequences, and the instance score is the pass rate averaged over these thresholds; for the receptor instances (LY6A, LY6C1), if a trained classifier judges it specific. The oracles are trained on real fitness data, the Fit4Function screens~\cite{eid2024systematic} and LY6A/LY6C1 binding measurements~\cite{Huang2023TargetingAV}.

\begin{table}[H]\centering
\caption{Viability prediction. Tested on two held-out instances, with validation labels hidden and
reachable only through a metered scoring action. The score is out-of-distribution ranking
accuracy, the area under the ROC curve for separating viable from non-viable variants,
averaged over different number of mutations. Scores are means over valid repeats.}
\label{tab:aav-01}
\begin{tabular}{l cc cc c}
\toprule
 & \multicolumn{2}{c}{\textbf{Multi-mutation} } & \multicolumn{2}{c}{\textbf{Single-mutation} } & \\
\cmidrule(lr){2-3}\cmidrule(lr){4-5}
\textbf{model} & score & HDS6 & score & HDS6 & \textbf{mean} \\
\midrule
  kimi-k3 & \best{0.962} & 3.48 & 0.910 & 3.51 & 0.936 \\
  claude-opus-5 & 0.959 & 3.43 & \best{0.913} & 3.49 & 0.936 \\
  glm-5.2 & 0.960 & 3.37 & 0.905 & 3.18 & 0.932 \\
  claude-opus-4-8 & 0.951 & \best{3.62} & 0.912 & \best{3.68} & 0.931 \\
  claude-sonnet-4-6 & 0.952 & 3.36 & 0.893 & 3.55 & 0.922 \\
  Claude Code (opus 4.8) & 0.950 & {---} & 0.891 & {---} & 0.921 \\
  deepseek-v4-pro & 0.941 & 3.33 & 0.894 & 3.41 & 0.918 \\
  gpt-5.5 & 0.927 & 2.85 & 0.891 & 3.14 & 0.909 \\
  \textbf{apodex-1.1} & 0.931 & 3.27 & 0.877 & 3.44 & 0.904 \\
  qwen3.5-397b & 0.890 & 3.24 & 0.902 & 3.31 & 0.896 \\
  \textbf{apodex-1.0} & 0.905 & 3.17 & 0.863 & 3.30 & 0.884 \\
\midrule
CAP-PLM~\cite{Wu2024PredictionOA} (published SOTA)     & 0.897 & {---} & 0.859 & {---} & 0.878 \\
ALICE-GB~\cite{guo2026mapping}           & 0.885 & {---} & 0.696 & {---} & 0.791 \\
augmented ridge~\cite{hsu2022learning}   & 0.874 & {---} & 0.848 & {---} & 0.861 \\
logistic additive~\cite{bryant2021aav}  & 0.888 & {---} & 0.852 & {---} & 0.870 \\
\bottomrule
\end{tabular}
\end{table}

\begin{table}[H]\centering
\caption{Tropism prediction. Seven-amino-acid inserts in a fixed surface loop of the AAV9 capsid, scored by
the agreement between predicted and measured per-target enrichment across mouse multi-organ
tropism (A), cross-species transfer from mouse and in-vitro data to macaque (B), and human in-vitro assays (C). Scores are per-target means over valid repeats.}
\label{tab:aav-02}
\begin{tabular}{l cc cc cc c}
\toprule
 & \multicolumn{2}{c}{\textbf{A $\cdot$ Mouse tropism}} & \multicolumn{2}{c}{\textbf{B $\cdot$ Cross-species}} & \multicolumn{2}{c}{\textbf{C $\cdot$ In-vitro cell}} & \\
\cmidrule(lr){2-3}\cmidrule(lr){4-5}\cmidrule(lr){6-7}
\textbf{solver} & score & HDS6 & score & HDS6 & score & HDS6 & \textbf{mean} \\
\midrule
  claude-opus-5 & \best{0.562} & 3.65 & \best{0.638} & 3.10 & \best{0.760} & \best{3.90} & 0.653 \\
  kimi-k3 & 0.559 & 3.70 & 0.637 & 3.63 & 0.751 & 3.51 & 0.649 \\
  glm-5.2 & 0.553 & \best{3.73} & 0.626 & \best{3.84} & 0.752 & 3.72 & 0.644 \\
  claude-opus-4-8 & 0.548 & 3.70 & 0.619 & 3.40 & 0.740 & 2.80 & 0.635 \\
  \textbf{apodex-1.1} & 0.539 & 3.16 & 0.621 & 3.57 & 0.746 & 3.37 & 0.635 \\
  Claude Code (opus 4.8) & 0.545 & {---} & 0.617 & {---} & 0.738 & {---} & 0.633 \\
  claude-sonnet-4-6 & 0.549 & 3.42 & 0.610 & 1.91 & 0.725 & 2.97 & 0.628 \\
  deepseek-v4-pro & 0.508 & 3.23 & 0.602 & 2.81 & 0.728 & 2.88 & 0.613 \\
  qwen3.5-397b & 0.500 & 2.98 & 0.600 & 2.84 & 0.716 & 2.96 & 0.605 \\
  gpt-5.5 & 0.534 & 3.56 & 0.575 & 3.75 & 0.691 & 3.39 & 0.600 \\
  \textbf{apodex-1.0} & 0.525 & 2.43 & 0.540 & 2.64 & 0.489 & 3.62 & 0.518 \\
\midrule
F4F\cite{eid2024systematic} (published SOTA) & 0.532 & {---} & 0.614 & {---} & 0.719 & {---} & 0.622 \\
\bottomrule
\end{tabular}
\end{table}

\begin{table}[H]\centering
\begin{threeparttable}
\caption{Structure prediction. The variable surface loops of a single subunit, the full 60-mer
icosahedral shell, and the capsid in complex with an antibody or receptor. Scores measure agreement with the reference
structure, per-residue positional accuracy for the surface loops and the shell, and interface quality
for the complex. Scores are means over valid repeats.}
\label{tab:aav-03}
\begin{tabular}{l cc cc cc c}
\toprule
 & \multicolumn{2}{c}{\textbf{Loop}} & \multicolumn{2}{c}{\textbf{Assembly}} & \multicolumn{2}{c}{\textbf{Complex}} & \\
\cmidrule(lr){2-3}\cmidrule(lr){4-5}\cmidrule(lr){6-7}
\textbf{solver} & score & HDS6 & score & HDS6 & score & HDS6 & \textbf{mean} \\
\midrule
  kimi-k3 & 0.924 & \best{3.56} & 0.754 & 3.60 & \best{0.359} & \best{3.89} & 0.679 \\
  claude-opus-5 & 0.913 & 3.43 & 0.768 & 3.79 & 0.322 & 3.60 & 0.668 \\
  glm-5.2 & 0.922 & 3.31 & 0.736 & 3.53 & 0.336 & 3.33 & 0.665 \\
  claude-opus-4-8 & 0.927 & 3.28 & \best{0.769} & \best{3.82} & 0.275 & 2.99 & 0.657 \\
  \textbf{apodex-1.1} & \best{0.928} & 3.47 & 0.742 & 3.60 & 0.277 & 3.49 & 0.649 \\
  deepseek-v4-pro & 0.927 & 2.50 & 0.749 & 2.94 & 0.154 & 1.13 & 0.610 \\
  Claude Code (opus 4.8) & 0.889 & {---} & 0.660 & {---} & 0.236 & {---} & 0.595 \\
  claude-sonnet-4-6 & 0.915 & 2.69 & 0.662 & 3.42 & 0.148 & 2.09 & 0.575 \\
  gpt-5.5\tnote{$\ddagger$} & 0.927 & 1.38 & 0.490 & 0.94 & 0.216 & 1.35 & 0.544 \\
  qwen3.5-397b & 0.927 & 1.29 & 0.555 & 3.18 & 0.148 & 1.45 & 0.543 \\
  \textbf{apodex-1.0} & 0.927 & 1.53 & 0.504 & 3.06 & 0.148 & 0.93 & 0.526 \\
\midrule
AlphaFold 3~\cite{abramson2024accurate} + symmetry expansion\tnote{$*$ $\S$}  & 0.927 & {---} & 0.740 & {---} & 0.148 & {---} & 0.605 \\
Boltz-1~\cite{wohlwend2025boltz} + symmetry expansion\tnote{$*$}  & 0.867 & {---} & 0.695 & {---} & 0.074 & {---} & 0.545 \\
AlphaFold 2~\cite{Jumper2021HighlyAP} + symmetry expansion\tnote{$*$}  & 0.901 & {---} & 0.732 & {---} & 0.073 & {---} & 0.569 \\
CapBuild~\cite{Loeb2025CompleteNA}             & {---} & {---} & 0.593 & {---} & {---} & {---} & {---} \\
Template docking~\cite{meng2023comdock}    & {---} & {---} & {---} & {---} & 0.252 & {---} & {---} \\
\bottomrule
\end{tabular}
\begin{tablenotes}\footnotesize
\item[$\ddagger$] gpt-5.5 has low HDS6 scores across all three instances, it consistently skips the self-validation step the task requires and submits the cheapest answer. Loop: folds every target with one default engine without checking or comparison. Assembly: copies the highest sequence-identity template's coordinates verbatim for every target. Complex: hardcodes outcome with severe collapses.
\item[$*$] Folding models cannot predict the 60-mer shell directly. For the assembly instance we fold the single subunit with the named engine and expand it onto the icosahedral operators of a fixed AAV capsid template.
\item[$\S$] published SOTA.
\end{tablenotes}
\end{threeparttable}
\end{table}

\begin{table}[H]\centering
\begin{threeparttable}
\caption{Generative design. Designing novel AAV9 capsid peptides under a limited scoring budget,
across three tissue instances (brain, spinal cord, heart) and two receptor instances (LY6A, LY6C1).
A design counts as a hit if it is novel, manufacturable, and its predicted selectivity for the target ranks in
the top of all sequences; the score is the fraction of the 1000 designs that are
hits. All three reference methods are our own reproductions under the identical episode budget
(1000 submitted designs, 100 scoring queries), differing only in the generator.}
\label{tab:aav-04}
\small
\setlength{\tabcolsep}{4.5pt}
\begin{tabular}{l cc cc cc cc cc c}
\toprule
 & \multicolumn{6}{c}{\textbf{Tissue selectivity}} & \multicolumn{4}{c}{\textbf{Receptor specificity}} & \\
\cmidrule(lr){2-7}\cmidrule(lr){8-11}
 & \multicolumn{2}{c}{brain} & \multicolumn{2}{c}{spinalcord} & \multicolumn{2}{c}{heart} & \multicolumn{2}{c}{LY6A} & \multicolumn{2}{c}{LY6C1} & \textbf{mean} \\
\cmidrule(lr){2-3}\cmidrule(lr){4-5}\cmidrule(lr){6-7}\cmidrule(lr){8-9}\cmidrule(lr){10-11}
\textbf{model} & score & HDS6 & score & HDS6 & score & HDS6 & score & HDS6 & score & HDS6 & \\
\midrule
kimi-k3            & 0.195 & 3.28 & \best{0.302} & \best{3.70} & 0.223 & 3.75 & \best{0.132} & \best{3.53} & 0.117 & \best{3.71} & 0.194 \\
glm-5.2            & 0.249 & \best{3.73} & 0.265 & 3.57 & \best{0.242} & 3.42 & 0.082 & 2.62 & 0.117 & 3.28 & 0.191 \\
\textbf{apodex-1.1} & \best{0.259} & 3.26 & 0.265 & 3.50 & 0.220 & \best{3.81} & 0.035 & 3.21 & \best{0.122} & 2.36 & 0.180 \\
gpt-5.5            & 0.117 & 1.94 & 0.265 & 2.33 & 0.192 & 2.55 & 0.073 & 2.53 & 0.061 & 1.85 & 0.142 \\
deepseek-v4-pro    & 0.112 & 2.91 & 0.257 & 3.03 & 0.204 & 3.06 & 0.038 & 1.61 & 0.032 & 2.58 & 0.129 \\
claude-sonnet-4-6  & 0.078 & 2.95 & 0.201 & 3.02 & 0.108 & 3.00 & 0.041 & 3.35 & 0.043 & 2.55 & 0.094 \\
qwen3.5-397b       & 0.066 & 2.58 & 0.175 & 1.67 & 0.099 & 2.28 & 0.054 & 2.25 & 0.030 & 2.25 & 0.085 \\
Claude Code (sonnet 4-6) & 0.077 & {---} & 0.187 & {---} & 0.100 & {---} & 0.035 & {---} & 0.002 & {---} & 0.080 \\
\textbf{apodex-1.0} & 0.025 & 2.37 & 0.228 & 2.94 & 0.062 & 2.78 & 0.039 & 1.67 & 0.008 & 1.92 & 0.072 \\
\midrule
AAVGen\cite{GhaffarzadehEsfahani2026AAVGenPE} & 0.104 & {---} & 0.247 & {---} & 0.160 & {---} & 0.026 & {---} & 0.009 & {---} & 0.109 \\
AAVDiff\cite{liu2024aavdiff}\tnote{$\S$} & 0.113 & {---} & 0.267 & {---} & 0.177 & {---} & 0.021 & {---} & 0.003 & {---} & 0.116 \\
ALICE\cite{guo2026mapping} & 0.093 & {---} & 0.247 & {---} & 0.168 & {---} & 0.024 & {---} & 0.019 & {---} & 0.110 \\
\bottomrule
\end{tabular}
\begin{tablenotes}\footnotesize
\item[$\S$] published SOTA.
\end{tablenotes}
\end{threeparttable}
\end{table}

\newpage

\section{Full harness \texorpdfstring{$\times$}{x} model result matrix}
\label{app:full-matrix}

For each of the 11 evaluated LLM environments, Tables~\ref{tab:app-01}--\ref{tab:app-11}
report every harness--model score underlying Section~\ref{sec:results-ablation}.
Rows are harnesses, columns are foundation models (Opus-4.8, GPT-5.6-sol,
DeepSeek-v4-pro, GLM-5.2, Kimi-k3, and Apodex-1.0 (397B)); a blank cell means the
configuration was not run. Superscript letters flag a non-capability outcome
(harness bug, infrastructure fault, refusal, or gate failure), explained in a
note beneath the table. Claude Code and Codex CLI are run on a single model each (Opus-4.8 and GPT-5.6-sol respectively) by construction, not by omission.

Of the 11 LLM environments (Tables~\ref{tab:app-01}--\ref{tab:app-11}), seven are
used for the main-text aggregates in Section~\ref{sec:results-ablation}. The remaining
four are excluded for failing one of the pre-specified inclusion criteria:
\texttt{nanogpt-speedrun} exhibits severe floor-level scores across nearly all
configurations; \texttt{swe-juice} has substantial
harness--model coverage gaps; \texttt{model-weakness-internal-benchmark}
shows incomplete or non-compliant submissions across several harness--model cells; and \texttt{llm-determinism} is dominated by a single-model effect rather than a harness effect. All four environments are nonetheless reported in full below.



\bigskip

\begin{table}[H]
\caption{Harness--model results on \texttt{LLM-pretrain-data-probe}, an environment that uses
capture--recapture population estimation to probe the coverage and poison sources of a pretraining
corpus. Each score is the mean over six episodes (arXiv / bioRxiv / medRxiv, at two coverage rungs),
combining the accuracy of the population estimate with decoy-source detection, procurement outcome
and budget efficiency, gated on corpus contamination.}
\label{tab:app-01}
\centering\small
\begin{tabular}{@{}l M M M M M M@{}}
\toprule
Harness & {Opus-4.8} & {GPT-5.6-sol} & {DeepSeek-v4-pro} & {GLM-5.2} & {Kimi-K3} & {Apodex-1.0} \\
\midrule
Claude Code   & 0.563 & {---} & {---} & {---} & {---} & {---} \\
Codex CLI     & {---} & 0.294$^{a}$ & {---} & {---} & {---} & {---} \\
DeerFlow      & 0.491 & 0.601 & 0.447 & 0.454 & 0.351 & 0.395 \\
OpenHands     & 0.442 & 0.534 & 0.444 & 0.432 & 0.379 & 0.459 \\
A-Evolve      & 0.553 & 0.585 & 0.506 & 0.567 & 0.609 & 0.451 \\
ApodexHarness & 0.553 & 0.506 & 0.552 & 0.385 & 0.601 & 0.546 \\
\bottomrule
\end{tabular}
\tabnote{$^{a}$ Codex scores 0 on three episodes: its tight shell-loop probing exhausts the
environment's per-episode request limiter, which also carries its model traffic. It averages 0.589
on the other three.}
\end{table}

\bigskip

\begin{table}[H]
\caption{Harness--model results on \texttt{nanogpt-speedrun}, an environment that trains nanoGPT to reach a target validation loss as quickly as possible. A submission is scored only if it simultaneously passes both a loss gate and a speed gate; otherwise the score is zero. Each cell is the mean over the environment's three episodes (\texttt{main}, \texttt{fw\_slice\_b}, and \texttt{fineweb\_edu}).}
\label{tab:app-02}
\centering\small
\begin{tabular}{@{}l M M M M M M@{}}
\toprule
Harness & {Opus-4.8} & {GPT-5.6-sol} & {DeepSeek-v4-pro} & {GLM-5.2} & {Kimi-K3} & {Apodex-1.0} \\
\midrule
Claude Code   & 0.009 & {---} & {---} & {---} & {---} & {---} \\
Codex CLI     & {---} & 0.014 & {---} & {---} & {---} & {---} \\
DeerFlow      & 0.062 & 0.000$^{b}$ & 0.109 & 0.000$^{ab}$ & 0.005 & 0.000$^{b}$ \\
OpenHands     & 0.123 & 0.147 & 0.005 & 0.056$^{a}$ & 0.033 & 0.025$^{a}$ \\
A-Evolve      & 0.025 & 0.005 & 0.004 & 0.018 & 0.000$^{b}$ & 0.037 \\
ApodexHarness & 0.016 & 0.000$^{ac}$ & 0.000$^{c}$ & 0.108 & 0.076 & 0.000$^{b}$ \\
\bottomrule
\end{tabular}
\tabnote{$^{a}$ Mean over two episodes; the third returned no result.\quad
$^{b}$ Zero from mixed per-episode causes; where the loss gate was met, the speed gate was missed by under 13\,s.\quad
$^{c}$ No episode produced a submittable training run.}
\end{table}

\bigskip

\begin{table}[H]
\caption{Harness--model results on \texttt{pretrain-data-filter}, an environment that trains a
multi-head quality scorer to filter pretraining data. Scores combine filtering quality, measured per
dimension against a reference by Pearson correlation, with a throughput gate that zeroes the score if
not met, and are averaged over the environment's two canonical episodes.}
\label{tab:app-03}
\centering\small
\begin{tabular}{@{}l M M M M M M@{}}
\toprule
Harness & {Opus-4.8} & {GPT-5.6-sol} & {DeepSeek-v4-pro} & {GLM-5.2} & {Kimi-K3} & {Apodex-1.0} \\
\midrule
Claude Code   & 0.485 & {---} & {---} & {---} & {---} & {---} \\
Codex CLI     & {---} & 0.497 & {---} & {---} & {---} & {---} \\
DeerFlow      & 0.234$^{b}$ & 0.491 & 0.000$^{a}$ & 0.217$^{b}$ & 0.416 & 0.000$^{a}$ \\
OpenHands     & 0.446 & 0.215$^{b}$ & 0.000$^{a}$ & 0.223$^{b}$ & 0.000$^{c}$ & 0.224$^{b}$ \\
A-Evolve      & 0.457 & 0.532 & 0.244$^{b}$ & 0.000$^{a}$ & 0.254$^{b}$ & 0.000$^{a}$ \\
ApodexHarness & 0.462 & 0.429 & 0.000$^{a}$ & 0.241$^{b}$ & 0.154$^{b}$ & 0.354 \\
\bottomrule
\end{tabular}
\tabnote{$^{a}$ Zero in both episodes: the submission missed a hard gate --- most often it was not
self-contained, loading a backbone from a path the offline scoring sandbox does not mount.\quad
$^{b}$ Zero in one of the two episodes, so the mean is roughly half the working episode's score.\quad
$^{c}$ One episode only; the other returned no result (harness SDK fault) and is excluded.}
\end{table}

\bigskip

\begin{table}[H]
\caption{Harness--model results on \texttt{operator-align}, an environment that runs TransformerEngine on a GPU to detect misaligned or mismatched numerical operators. Scores are the pooled outcome $F_1 \times (0.5 + 0.5 \cdot \text{trigger\_rate})$ over all 33 instances: detections are pooled across instances before scoring, never averaged per instance.}
\label{tab:app-04}
\centering\small
\begin{tabular}{@{}l M M M M M M@{}}
\toprule
Harness & {Opus-4.8} & {GPT-5.6-sol} & {DeepSeek-v4-pro} & {GLM-5.2} & {Kimi-K3} & {Apodex-1.0} \\
\midrule
Claude Code   & 0.722 & {---} & {---} & {---} & {---} & {---} \\
Codex CLI     & {---} & 0.491 & {---} & {---} & {---} & {---} \\
DeerFlow      & 0.663 & 0.659 & 0.659 & 0.675 & 0.725 & 0.714 \\
OpenHands     & 0.740 & 0.409 & 0.592 & 0.631 & 0.646 & 0.716 \\
A-Evolve      & 0.722 & 0.662 & 0.628 & 0.712 & 0.720 & 0.641 \\
ApodexHarness & 0.727 & 0.734 & 0.657 & 0.653 & 0.743 & 0.607 \\
\bottomrule
\end{tabular}
\end{table}

\bigskip

\begin{table}[H]
\caption{Harness--model results on \texttt{model-weakness-internal-benchmark}, an environment for building STEM-domain internal benchmark suites targeting model weaknesses. Submitted evaluation harnesses are scored against an internal reference suite on a 0--1 scale reflecting benchmark quality. Each entry averages scores against two held-out reference models, Qwen-3.5-35B and Qwen-3-30B.}
\label{tab:app-05}
\centering\small
\begin{tabular}{@{}l M M M M M M@{}}
\toprule
Harness & {Opus-4.8} & {GPT-5.6-sol} & {DeepSeek-v4-pro} & {GLM-5.2} & {Kimi-K3} & {Apodex-1.0} \\
\midrule
Claude Code   & 0.160 & {---} & {---} & {---} & {---} & {---} \\
Codex CLI     & {---} & 0.097 & {---} & {---} & {---} & {---} \\
DeerFlow      & 0.000$^{a}$ & 0.138 & 0.088 & 0.085 & 0.000$^{b}$ & 0.020 \\
OpenHands     & 0.089 & 0.195 & 0.044 & 0.044 & 0.000$^{b}$ & 0.057 \\
A-Evolve      & 0.065 & 0.117 & 0.048 & 0.046 & 0.031 & 0.078 \\
ApodexHarness & 0.355 & 0.102 & 0.070 & 0.108 & 0.000$^{b}$ & 0.068 \\
\bottomrule
\end{tabular}
\tabnote{$^{a}$ Timed out due to a slow multi-round loop in one of the two settings.\quad
$^{b}$ Did not follow the submission protocol.}
\end{table}

\bigskip

\begin{table}[H]
\caption{Harness--model results on the \texttt{rl-recipe} environment. This environment assesses
submitted reinforcement-learning recipes via held-out performance improvement over a baseline
recipe. Reported scores use a process-imputed proxy outcome derived from HDS6 metrics (imputation
correlation with the true held-out outcome: $r=0.979$), combined with anti-hacking penalties and
cost terms, normalized to a 0--1 scale. Entries average the environment's two canonical episodes
(\texttt{dapo\_math}, \texttt{ifbench}); Codex CLI is \texttt{dapo\_math} only, as its
\texttt{ifbench} episode submitted no recipe. Since the true held-out outcome spans just
0.626--0.715, these scores are dominated by the anti-hacking term.}
\label{tab:app-06}
\centering\small
\begin{tabular}{@{}l M M M M M M@{}}
\toprule
Harness & {Opus-4.8} & {GPT-5.6-sol} & {DeepSeek-v4-pro} & {GLM-5.2} & {Kimi-K3} & {Apodex-1.0} \\
\midrule
Claude Code   & 0.751 & {---} & {---} & {---} & {---} & {---} \\
Codex CLI     & {---} & 0.800 & {---} & {---} & {---} & {---} \\
DeerFlow      & 0.821 & 0.791 & 0.748 & 0.724 & 0.789 & 0.731 \\
OpenHands     & 0.747 & 0.797 & 0.776 & 0.805 & 0.732 & 0.732 \\
A-Evolve      & 0.751 & 0.779 & 0.709 & 0.742 & 0.793 & 0.708 \\
ApodexHarness & 0.741 & 0.804 & 0.749 & 0.732 & 0.807 & 0.697 \\
\bottomrule
\end{tabular}
\end{table}

\bigskip

\begin{table}[H]
\caption{Harness--model results on \texttt{posttrain-data-dedup}, an environment that deduplicates post-training data and removes contamination from a held-out test set. Scores combine deduplication $F_1$, decontamination leak recall, and an over-removal rate. Each entry averages three instances of differing difficulty.}
\label{tab:app-07}
\centering\small
\begin{tabular}{@{}l M M M M M M@{}}
\toprule
Harness & {Opus-4.8} & {GPT-5.6-sol} & {DeepSeek-v4-pro} & {GLM-5.2} & {Kimi-K3} & {Apodex-1.0} \\
\midrule
Claude Code   & 0.386 & {---} & {---} & {---} & {---} & {---} \\
Codex CLI     & {---} & 0.296 & {---} & {---} & {---} & {---} \\
DeerFlow      & 0.304 & 0.269 & 0.310 & 0.360 & 0.266 & 0.319 \\
OpenHands     & 0.233 & 0.272 & 0.249 & 0.273 & 0.309 & 0.273 \\
A-Evolve      & 0.307 & 0.338 & 0.309 & 0.416 & 0.402 & 0.319 \\
ApodexHarness & 0.316 & 0.304 & 0.279 & 0.252 & 0.234 & 0.301 \\
\bottomrule
\end{tabular}
\end{table}

\bigskip

\begin{table}[H]
\caption{Harness--model results on \texttt{swe-juice}, an environment that fine-tunes Qwen3-8B so its reasoning length is controllable by a reasoning-effort knob without losing agentic SWE issue-resolution ability. Scores combine effort-to-length controllability with the issue-resolution rate over hidden GitHub issues.}
\label{tab:app-08}
\centering\small
\begin{tabular}{@{}l M M M M M M@{}}
\toprule
Harness & {Opus-4.8} & {GPT-5.6-sol} & {DeepSeek-v4-pro} & {GLM-5.2} & {Kimi-K3} & {Apodex-1.0} \\
\midrule
Claude Code   & 0.000$^{a}$ & {---} & {---} & {---} & {---} & {---} \\
Codex CLI     & {---} & 0.210 & {---} & {---} & {---} & {---} \\
DeerFlow      & 0.362 & 0.215 & 0.000$^{b}$ & 0.462 & 0.399 & 0.000$^{c}$ \\
OpenHands     & 0.096 & 0.090 & 0.244 & {---}$^{d}$ & {---}$^{d}$ & 0.000 \\
A-Evolve      & 0.493 & {---}$^{e}$ & 0.243 & 0.095 & 0.000 & 0.243 \\
ApodexHarness & 0.245 & 0.246 & 0.000$^{f}$ & 0.000$^{g}$ & 0.244 & 0.248 \\
\bottomrule
\end{tabular}
\tabnote{$^{a}$ No submission.\quad $^{b}$ Harness bug killed training.\quad
$^{c}$ Kept exploring, training never started.\quad
$^{d}$ Model--harness compatibility fault.\quad
$^{e}$ OpenAI refused to run (safety flag).\quad
$^{f}$ ApodexHarness's history truncation drops the reasoning-content field, causing a 400 from the model API.\quad
$^{g}$ Refused on content-policy grounds.}
\end{table}

\bigskip

\begin{table}[H]
\caption{Harness--model results on \texttt{solution-verifier}, an environment that verifies whether a proposed code-repair patch is correct across pick (candidate selection) and rank (candidate ordering) instances drawn from OpenSWE and ACR. Scores are the binary correctness of the pick/rank verdict, averaged over all instances for every harness--model pair. Cells where the solver engaged but never submitted are scored zero, per the environment's submission rule.}
\label{tab:app-09}
\centering\small
\begin{tabular}{@{}l M M M M M M@{}}
\toprule
Harness & {Opus-4.8} & {GPT-5.6-sol} & {DeepSeek-v4-pro} & {GLM-5.2} & {Kimi-K3} & {Apodex-1.0} \\
\midrule
Claude Code   & 0.80 & {---} & {---} & {---} & {---} & {---} \\
Codex CLI     & {---} & 0.90 & {---} & {---} & {---} & {---} \\
DeerFlow      & 0.80 & 0.80 & 0.75 & 0.65 & 0.90 & 0.80 \\
OpenHands     & 0.75 & 0.95 & 0.75 & 0.60 & 0.80 & 0.50 \\
A-Evolve      & 0.85 & 0.90 & 0.75 & 0.80 & 0.75 & 0.50 \\
ApodexHarness & 0.85 & 0.95 & 0.70 & 0.85 & 0.90 & 0.60 \\
\bottomrule
\end{tabular}
\end{table}

\bigskip

\begin{table}[H]
\caption{Harness--model results on \texttt{LLM-rollout-verify}, an environment that builds a
three-way verifier for math solutions (correct / incorrect / ill-posed): the agent authors a
\texttt{judge\_one} function and the environment runs it over the corpus. Scores are macro-$F_1$
on a hidden held-out split, discounted logarithmically for output-token cost per question.}
\label{tab:app-10}
\centering\small
\begin{tabular}{@{}l M M M M M M@{}}
\toprule
Harness & {Opus-4.8} & {GPT-5.6-sol} & {DeepSeek-v4-pro} & {GLM-5.2} & {Kimi-K3} & {Apodex-1.0} \\
\midrule
Claude Code   & 0.375 & {---} & {---} & {---} & {---} & {---} \\
Codex CLI     & {---} & 0.636 & {---} & {---} & {---} & {---} \\
DeerFlow      & 0.424 & 0.452 & 0.347 & 0.477 & 0.531 & 0.312 \\
OpenHands     & 0.687 & 0.742 & 0.359 & 0.441 & 0.490 & 0.310 \\
A-Evolve      & 0.341 & 0.740 & 0.354 & 0.577 & 0.547 & 0.313 \\
ApodexHarness & 0.625 & 0.416 & 0.506 & 0.550 & 0.625 & 0.343 \\
\bottomrule
\end{tabular}
\end{table}

\bigskip

\begin{table}[H]
\caption{Harness--model results on \texttt{llm-determinism}: an environment that finds and fixes sources of non-determinism in a pinned SGLang deployment, graded against a hidden bit-exactness suite with a no-regression check as a hard gate. Each cell is the \textit{best-of-n} over that cell's repeats. A dash is a cell that was not run.}
\centering\small
\setlength{\tabcolsep}{3pt}
\begin{tabular}{@{}l c c c c c c@{}}
\toprule
Harness & {Opus-4.8} & {GPT-5.6-sol} & {DeepSeek-v4-pro} & {GLM-5.2} & {Kimi-K3} & {Apodex-1.0} \\
\midrule
Claude Code & 0.050 & --- & --- & --- & --- & --- \\
Codex CLI & --- & 0.480$^{a}$ & --- & 0.000$^{a}$ & --- & --- \\
DeerFlow & 0.000$^{b}$ & 0.000$^{b}$ & 0.000$^{b}$ & 0.000$^{b}$ & 0.000$^{b}$ & 0.000$^{b}$ \\
OpenHands & 0.000$^{a}$ & 0.856 & 0.000$^{b}$ & 0.000$^{a}$ & 0.000$^{b}$ & 0.000$^{b}$ \\
A-Evolve & 0.000$^{b}$ & 0.100 & 0.067 & 0.000$^{b}$ & 0.100 & 0.000$^{b}$ \\
ApodexHarness & 0.000$^{a}$ & 0.856 & 0.000$^{a}$ & 0.820 & 0.856 & 0.100 \\
\bottomrule
\end{tabular}
\label{tab:app-11}
\tabnote{Why a cell is zero: 8 of the 23 recorded zeros are genuine, the rest are artifacts, so an unannotated zero would not mean what it appears to. None is caused by a confirmed regression --- the no-regression gate fired five times across 72 episodes and no firing was ever confirmed. $^{a}$~Genuine: the run completed and satisfied none of the required fixes. $^{b}$~Harness artifact --- no scorable submission ever reached the world: early stopping or a tool-frequency guard (DeerFlow), an iteration or wall-clock cap.}
\end{table}

\bigskip

\begin{table}[H]
\caption{Harness--model results on \texttt{clinical-sap-tlf}, an environment that, given a
statistical analysis plan and real CDISC ADaM data, reproduces the pre-specified tables,
listings, and figures for a clinical trial, on its TEAE-summary, MMRM-efficacy, and
time-to-event tracks. The score is the rigour of the documented statistical judgement, gated
to zero unless every pre-specified number is reproduced exactly and every audit gate passes.
Tracks use different statistical methods and are never averaged.}
\label{tab:app-12}
\centering\small
\begin{tabular}{@{}l M M M@{}}
\toprule
Harness (model) & {teae} & {mmrm} & {tte} \\
\midrule
Claude Code (Opus-4.8)      & 0.000$^{a}$ & 0.000$^{b}$ & 0.304 \\
\addlinespace
Codex CLI (GPT-5.6-sol)     & 0.670 & 0.437 & 0.406 \\
\addlinespace
DeerFlow (Opus-4.8)         & 0.953 & 0.964 & 0.979 \\
DeerFlow (GPT-5.6-sol)      & 0.982 & 0.622 & 0.911 \\
DeerFlow (DeepSeek-v4-pro)  & 0.000$^{e}$ & 0.000$^{e}$ & 0.000$^{e}$ \\
DeerFlow (Kimi-k3)          & 0.773 & 0.000$^{b}$ & 0.979 \\
DeerFlow (GLM-5.2)          & 0.947 & 0.973 & 0.954 \\
\addlinespace
OpenHands (Opus-4.8)        & 1.000 & 0.676 & 1.000 \\
OpenHands (GPT-5.6-sol)     & 1.000 & 0.796 & 0.911 \\
OpenHands (DeepSeek-v4-pro) & \multicolumn{3}{c}{{---}$^{f}$} \\
OpenHands (Kimi-k3)         & 0.941 & 0.911 & 1.000 \\
OpenHands (GLM-5.2)         & 0.947 & 0.000$^{c}$ & 1.000 \\
\addlinespace
A-Evolve (Opus-4.8)         & 0.667 & 0.000$^{d}$ & 0.656 \\
A-Evolve (GPT-5.6-sol)      & 0.659 & 0.410 & 0.389 \\
A-Evolve (DeepSeek-v4-pro)  & 0.000$^{a}$ & 0.000$^{b}$ & 0.653 \\
A-Evolve (Kimi-k3)          & 0.674 & 0.000$^{a}$ & 0.400 \\
A-Evolve (GLM-5.2)          & 0.649 & 0.000$^{b}$ & 0.000$^{a}$ \\
\bottomrule
\end{tabular}
\tabnote{48 cells, reported after eight environment and contract fixes that removed thirteen
spurious zeros. A zero is usually an audit-gate zero on a table that is numerically correct.\quad
$^{a}$ Hard-gate violation: figures reported without supporting executed code, an undeclared
assumption, or submitted code that does not re-run in a clean container.\quad
$^{b}$ Five of seven numbers matched; the committed reference p-values carry an implicit
Dunnett adjustment the plan never specifies, and the gate admits no partial credit --- six
configurations stop at exactly this signature.\quad
$^{c}$ One of seven numbers matched.\quad
$^{d}$ Ran 35 actions but never submitted.\quad
$^{e}$ Harness-side defect: its client drops the \texttt{reasoning\_content} the model
requires to be echoed back, killing the run at the first action; the same model completes all
three tracks under A-Evolve.\quad
$^{f}$ Not runnable for the same reason, so the pair produced no episode.}
\end{table}

\bigskip

\begin{table}[H]
\caption{Harness--model results on \texttt{drug-repurposing-reformulation} (drr), an environment that scores
a submitted promisingness and confidence estimate for a drug--disease pair against held-out
clinical outcomes, evaluated by ranking and accuracy of the repurposing prediction. Each cell
is the mean per-episode calibrated score over the 100 public instances, one run per instance.}
\label{tab:app-13}
\centering\small
\begin{tabular}{@{}l M M M M M@{}}
\toprule
Harness & {Opus-4.8} & {GPT-5.6-sol} & {DeepSeek-v4-pro} & {Kimi-k3} & {GLM-5.2} \\
\midrule
Claude Code & 0.942 & {---} & {---} & {---} & {---} \\
Codex CLI   & {---} & 0.963 & {---} & {---} & {---} \\
DeerFlow    & 0.933 & 0.914 & 0.888 & 0.907 & 0.899 \\
OpenHands   & 0.865 & 0.926 & 0.916 & 0.956 & 0.934 \\
A-Evolve    & 0.970 & 0.884 & 0.914 & 0.912 & 0.948 \\
\bottomrule
\end{tabular}
\tabnote{1700 episodes. Balanced-grid mean 0.9178; spread across harnesses 0.0176, across
models 0.0212. Cells include submission-failure zeros (per 100 instances, in column order:
DeerFlow 2/5/5/4/4, OpenHands 10/4/3/0/2, A-Evolve 0/8/3/4/0; Claude Code 2, Codex 0).
Excluding those zeros the harness spread collapses to 0.0098 and the ordering flips, so this
environment separates submission reliability rather than judgement quality --- GPT has the
highest judgement quality (0.9625 excluding zeros) but the most submission failures. Instance
difficulty dominates: the between-instance standard deviation (0.133) is about four times the
between-cell one (0.034). This is the per-episode calibrated score, not the official drep
normalized/combined metric.}
\end{table}

\bigskip

\begin{table}[H]
\caption{Harness--model results on AAV capsid enrichment, an environment that trains a
sequence-to-enrichment model for AAV9 588-loop 7-mer variants, on three generalisation tracks:
mouse multi-organ tropism, cross-species transfer to macaque liver, and human liver-cell lines
\emph{in vitro}. Scores are the per-assay Pearson correlation between predicted and true
enrichment, aggregated into a track total; tracks are reported separately and never averaged.}
\label{tab:app-14}
\centering\small
\begin{tabular}{@{}l M M M@{}}
\toprule
Harness (model) & {mouse} & {macaque} & {in vitro} \\
\midrule
Claude Code (Opus-4.8)      & 0.680 & 0.577 & 0.756 \\
\addlinespace
Codex CLI (GPT-5.6-sol)     & 0.583 & 0.523 & 0.754 \\
\addlinespace
DeerFlow (Opus-4.8)         & 0.540 & 0.485 & 0.731 \\
DeerFlow (GPT-5.6-sol)      & 0.602 & 0.580 & 0.742 \\
DeerFlow (DeepSeek-v4-pro)  & 0.549 & 0.501 & 0.000$^{a}$ \\
DeerFlow (Kimi-k3)          & 0.698 & 0.561 & 0.741 \\
DeerFlow (GLM-5.2)          & 0.633 & 0.320 & 0.727 \\
\addlinespace
OpenHands (Opus-4.8)        & 0.575 & 0.617 & 0.748 \\
OpenHands (GPT-5.6-sol)     & 0.409 & 0.631 & 0.762 \\
OpenHands (DeepSeek-v4-pro) & 0.502 & 0.585 & 0.707 \\
OpenHands (Kimi-k3)         & 0.515 & 0.494 & 0.750 \\
OpenHands (GLM-5.2)         & 0.517 & 0.589 & 0.747 \\
\addlinespace
A-Evolve (Opus-4.8)         & \multicolumn{3}{c}{{---}$^{b}$} \\
A-Evolve (GPT-5.6-sol)      & 0.628 & 0.472 & 0.749 \\
A-Evolve (DeepSeek-v4-pro)  & 0.371 & 0.594 & 0.649 \\
A-Evolve (Kimi-k3)          & 0.468 & 0.568 & 0.751 \\
A-Evolve (GLM-5.2)          & 0.516 & 0.633 & 0.612 \\
\bottomrule
\end{tabular}
\tabnote{Reference points per track: k-mer lookup floor 0.182 / 0.052 / 0.000; Fit4Function
paper reproduction 0.532 / 0.614 / 0.719. Every scored cell clears the k-mer floor and
most reach or exceed the paper reproduction; the cross-species macaque track is the
discriminating one. GPT's first-pass timeouts on this host-bash world were recovered by a
low-concurrency re-run, confirming a submission-discipline rather than a capability limit.\quad
$^{a}$ Grounding failure: the agent stopped to ask the user instead of acting, and stronger
prompting did not recover it --- the only unrecovered cell in the sweep.\quad
$^{b}$ A-Evolve's provider layer rejects non-OpenAI model names, so all three cells are
unavailable by construction rather than failures.}
\end{table}

\bigskip

\begin{table}[H]
\caption{Harness--model results on AAV capsid structure, an environment that predicts
three-dimensional structure from sequence for the VR-loop monomer, the full 60-mer capsid
assembly, and the VP--ligand complex instances of the task. Scores compare the predicted
structure against a time-cut-off held-out reference; the complex instance is ungated.}
\label{tab:app-15}
\centering\small
\begin{tabular}{@{}l M M M@{}}
\toprule
Harness (model) & {loop} & {assembly} & {complex} \\
\midrule
Claude Code (Opus-4.8)      & 0.929 & 0.692 & 0.136 \\
\addlinespace
Codex CLI (GPT-5.6-sol)     & 0.927 & 0.774 & 0.379 \\
\addlinespace
DeerFlow (Opus-4.8)         & 0.927 & 0.000$^{a}$ & 0.163 \\
DeerFlow (GPT-5.6-sol)      & 0.927 & 0.664 & 0.163 \\
DeerFlow (DeepSeek-v4-pro)  & 0.927 & 0.000$^{b}$ & 0.000$^{c}$ \\
DeerFlow (Kimi-k3)          & 0.927 & 0.000$^{d}$ & 0.000$^{c}$ \\
DeerFlow (GLM-5.2)          & 0.928$^{e}$ & 0.000$^{d}$ & 0.163 \\
\addlinespace
OpenHands (Opus-4.8)        & 0.927 & 0.668 & 0.120 \\
OpenHands (GPT-5.6-sol)     & 0.930 & 0.819 & 0.342 \\
OpenHands (DeepSeek-v4-pro) & 0.927 & 0.000$^{b}$ & 0.163 \\
OpenHands (Kimi-k3)         & 0.925 & 0.000$^{d}$ & 0.017 \\
OpenHands (GLM-5.2)         & 0.927 & 0.709 & 0.163 \\
\addlinespace
A-Evolve (Opus-4.8)         & \multicolumn{3}{c}{{---}$^{f}$} \\
A-Evolve (GPT-5.6-sol)      & 0.930 & 0.591 & 0.128 \\
A-Evolve (DeepSeek-v4-pro)  & 0.927 & 0.704 & 0.136 \\
A-Evolve (Kimi-k3)          & 0.927 & 0.000$^{d}$ & 0.365 \\
A-Evolve (GLM-5.2)          & 0.927 & 0.634 & 0.163 \\
\bottomrule
\end{tabular}
\tabnote{Metrics: loop = VR-loop lDDT on the low-homology ($<$80\%) stratum, gated on beating
the nearest-template baseline (lDDT 0.709) with coverage $\geq 0.9$; assembly =
median$(\exp(-\text{RMSD}/5\,\text{\AA})\times\text{stability})$ under clash gating, scaled by
the fraction of the 60-mer actually built; complex = mean over an antibody (8 targets) and a
receptor (7 targets) probe of $0.5\cdot\text{DockQ}+0.5\cdot\text{interface Fnat}$, with no
pass/fail gate. Each cell is the best of up to three attempts; no leakage fingerprint
triggered in any of the 79 episodes. The loop track is saturated --- 11 of the 16 scored cells
sit at 0.927 and the submitted structures are byte-identical across harnesses, because the
default fold path is deterministic and cached; six cells likewise sit at exactly 0.163 on complex,
the floor for submitting the default fold unchanged. Assembly is the only strongly
discriminating axis in this round.\quad
$^{a}$ The scorer found no submission to grade.\quad
$^{b}$ All 60 chains were placed but whole-capsid RMSD is 100--155\,\AA, so the RMSD score is
zero.\quad
$^{c}$ All 15 complexes were graded but DockQ and interface Fnat are both zero --- verified to
be a real failure, not a silently swallowed export error.\quad
$^{d}$ None of the 60-mer was built.\quad
$^{e}$ Score is fine but only 5 of 12 targets were folded (coverage 0.417), so the
correctness gate fails.\quad
$^{f}$ A-Evolve's provider layer rejects non-OpenAI model names.}
\end{table}

\bigskip

\begin{table}[H]
\caption{Harness--model results on AAV sequence design, an environment that designs AAV9
588-loop 7-mer peptides optimized for targets such as receptor specificity or tissue tropism,
for the brain, heart, spinal cord, Ly6a, and Ly6c1 targets. Tropism targets are scored as the
mean of the pass rates at the per-organ top-5\%, top-2\%, and top-1\% selectivity bars,
combined with a viability gate; receptor targets as the specific pass rate. Opus refuses the
task on Claude Code and DeerFlow.}
\label{tab:app-16}
\centering\small
\begin{tabular}{@{}l M M M M M@{}}
\toprule
Harness (model) & {brain} & {heart} & {spinalcord} & {ly6a} & {ly6c1} \\
\midrule
Claude Code (Opus-4.8)      & \multicolumn{5}{c}{REFUSE (all targets)} \\
\addlinespace
Codex CLI (GPT-5.6-sol)     & 0.119 & 0.290 & 0.415 & 0.070 & 0.135 \\
\addlinespace
DeerFlow (Opus-4.8)         & \multicolumn{5}{c}{REFUSE (all targets)} \\
DeerFlow (GPT-5.6-sol)      & 0.059$^{a}$ & 0.076$^{a}$ & 0.221 & 0.062 & 0.000$^{b}$ \\
DeerFlow (DeepSeek-v4-pro)  & 0.025 & 0.051 & 0.138 & 0.010 & 0.013 \\
DeerFlow (Kimi-k3)          & 0.040 & 0.102 & 0.241 & 0.021 & 0.015 \\
DeerFlow (GLM-5.2)          & 0.090$^{a}$ & 0.120$^{a}$ & 0.220 & 0.088 & 0.056 \\
\addlinespace
OpenHands (Opus-4.8)        & \multicolumn{5}{c}{{---}$^{c}$} \\
OpenHands (GPT-5.6-sol)     & 0.185 & 0.153 & 0.392 & 0.132 & 0.175 \\
OpenHands (DeepSeek-v4-pro) & 0.075 & 0.150 & 0.269 & 0.080 & 0.021 \\
OpenHands (Kimi-k3)         & 0.100 & 0.095 & 0.214 & 0.076 & 0.038 \\
OpenHands (GLM-5.2)         & 0.109 & 0.193 & 0.357 & 0.066 & 0.001 \\
\addlinespace
A-Evolve (GPT-5.6-sol)      & 0.122 & 0.174 & 0.349 & 0.141 & 0.088 \\
A-Evolve (DeepSeek-v4-pro)  & 0.060 & 0.091 & 0.174 & 0.025 & 0.073 \\
A-Evolve (Kimi-k3)          & 0.047 & 0.075 & 0.148 & 0.066 & 0.051 \\
A-Evolve (GLM-5.2)          & 0.136 & 0.151 & 0.091 & 0.087 & 0.059 \\
\bottomrule
\end{tabular}
\tabnote{Each tropism column is the mean of that organ's two evaluation splits (same metric
and scale); 84 Track-A cells in total, best attempt per cell. Generative reference baselines
under the same three-bar rule: aavdiff 0.113 / 0.177 / 0.267, aavgen 0.104 / 0.160 / 0.247 and
alice 0.093 / 0.168 / 0.247 for brain / heart / spinal cord; only 4 of 14 evaluated configurations beat aavdiff on the
six-instance mean. The strictest bar is counting-noise dominated (median 24 qualifying designs
at the 2\% bar and 2 at the 1\% bar out of a 1000-design budget), which is why the reported
score averages the three bars. A-Evolve cannot be run with Opus (its provider layer rejects
non-OpenAI model names).\quad
$^{a}$ Average includes one split that scored 0.000 because nothing was submitted.\quad
$^{b}$ Isolated incident, not reproduced on other targets under the same configuration.\quad
$^{c}$ Opus was not part of the sweep these three-bar scores were computed from; the earlier
Opus figures were obtained under the superseded single top-2\% bar and are not comparable.}
\end{table}

\bigskip

\begin{table}[H]
\caption{Harness--model results on AAV viability, an environment that predicts from
protein-sequence features whether an AAV capsid variant is viable, meaning it can assemble and
package its genome, on two orthogonal difficulty axes: dataset (Bryant 2021 substitutions
vs.\ Ogden 2019 insertions and deletions) and validation-set visibility (labelled, or a hidden
oracle reachable only through a metered query). Scores are prediction accuracy against hidden
labels; the Bryant settings are saturated near ceiling, with limited discrimination among
configurations.}
\label{tab:app-17}
\centering\small
\begin{tabular}{@{}l M M M M@{}}
\toprule
 & \multicolumn{2}{c}{Bryant} & \multicolumn{2}{c}{Ogden} \\
Harness (model) & {labelled} & {oracle} & {labelled} & {oracle} \\
\midrule
Claude Code (Opus-4.8)  & 0.947 & 0.968 & {ref}$^{a}$ & {ref}$^{a}$ \\
\addlinespace
Codex CLI (GPT-5.6-sol) & 0.980 & 0.955 & 0.928 & 0.903 \\
\addlinespace
DeerFlow (Opus-4.8)     & 0.960 & 0.952 & 0.908 & 0.862 \\
DeerFlow (GPT-5.6-sol)  & 0.976 & 0.948 & 0.936 & 0.854 \\
\addlinespace
OpenHands (Opus-4.8)    & 0.982 & 0.957 & 0.867 & 0.851 \\
OpenHands (GPT-5.6-sol) & 0.982 & 0.943 & 0.940 & 0.913 \\
\addlinespace
A-Evolve (Opus-4.8)     & \multicolumn{4}{c}{{---}$^{c}$} \\
A-Evolve (GPT-5.6-sol)  & 0.986 & 0.970 & 0.930 & {ref}$^{b}$ \\
\bottomrule
\end{tabular}
\tabnote{Metric: mean per-distance AUROC, with the passing threshold set by the environment's
additive baseline (Bryant 0.885, Ogden 0.851; literature reference cap-PLM/ESM2
$\approx 0.897$ on Bryant). All 33 scored cells pass and the Bryant settings broadly exceed
the literature reference; the difficulty gradient Ogden $>$ Bryant and oracle $>$ labelled is
visible in both scores and pass rates. Eleven first-pass zeros were concurrency artefacts and
recovered on a low-concurrency re-run; capability failures: zero. Effort is uncorrelated with
score (10 to 1970 actions). DeepSeek-v4-pro, Kimi-k3 and GLM-5.2 were not run on this
environment. \textsc{ref} = deterministic provider-side safety refusal on the Ogden
capsid-engineering framing, harness-specific and not a capability result.\quad
$^{a}$ Anthropic usage-policy refusal (one action, never submitted).\quad
$^{b}$ OpenAI \texttt{invalid\_prompt}, reproduced at three concurrency levels.\quad
$^{c}$ A-Evolve's provider layer rejects non-OpenAI model names.}
\end{table}

\clearpage
\section{Drug Repurposing and Reformulation Trace Example}

\label{app:traces-cases}

This appendix reports one evaluation episode end to end: the full recorded reasoning trace, and
the full process score with every sub-rubric shown.

\begin{table}[htbp]
\centering\small
\caption{Episode summary. The reference column is never visible to the judging panel.}
\label{tab:case-task}
\begin{tabular}{@{}l l@{}}
\toprule
\textbf{Field} & \textbf{Value} \\
\midrule
Drug--disease pair            & Dostarlimab $\rightarrow$ Stage II Breast Cancer \\
System score                  & \textbf{0.3} \\
System confidence             & 0.48 \\
Held-out reference            & \textbf{0.15} \ \ (lower bound; see below) \\
Reference clinical phase      & Phase 2 \\
Reference outcome             & Ongoing \\
\textbf{Benchmark score}          & \textbf{1.0} \ \ (error \textbf{0.0}) \\
Integrity gate                & \textbf{PASS} \\
Recorded steps                & 18 \ \ (9 reasoning, 8 tool calls, 1 submission) \\
Key claims submitted          & 7 \\
\bottomrule
\end{tabular}
\end{table}

\subsection{The task as given}
\label{sec:task-as-given}

Before any reasoning, this is the complete brief the system received --- the enforced
offline-evaluation policy, the analyst role, the required workflow, the evidence-grounding rules,
and the output contract it had to satisfy. It is reproduced verbatim.

\begin{taskbox}{TASK INPUT \textbullet\ BRIEF GIVEN TO THE SYSTEM}{\footnotesize \noindent OFFLINE EVALUATION MODE (\allowbreak{}ENFORCED)\allowbreak{}\par
\addvspace{3pt}
\noindent Do not use the open internet to look up this pair'\allowbreak{}s recent outcome:\allowbreak{} no general web search,\allowbreak{} no fetching arbitrary URLs,\allowbreak{} no browsing news,\allowbreak{} and no directly hitting public services (\allowbreak{}e.\allowbreak{}g.\allowbreak{} ClinicalTrials.\allowbreak{}gov,\allowbreak{} PubMed,\allowbreak{} preprint servers)\allowbreak{} for the trial'\allowbreak{}s result.\allowbreak{} The held-\allowbreak{}out outcome is recent,\allowbreak{} so retrieving it that way would invalidate the evaluation;\allowbreak{} those paths are blocked.\allowbreak{}\par
\addvspace{3pt}
\noindent This does NOT restrict the sanctioned environment.\allowbreak{} Only capabilities listed by the selected bundle manifest and bound by the active access adapter are allowed.\allowbreak{} That includes tool\_\allowbreak{}A'\allowbreak{}s Stage-\allowbreak{}1 \texttt{provider\_\allowbreak{}A} (\allowbreak{}live path-\allowbreak{}finding over the structured bio-\allowbreak{}graph)\allowbreak{} and any resident inference-\allowbreak{}service interfaces explicitly exposed by that bundle;\allowbreak{} copied documentation alone never grants another capability.\allowbreak{}\par
\addvspace{3pt}
\noindent You are a senior drug-\allowbreak{}repurposing analyst acting as an autonomous research agent.\allowbreak{}\par
\addvspace{3pt}
\noindent You assess drug-\allowbreak{}repurposing candidates.\allowbreak{} For each drug-\allowbreak{}disease pair you are asked about,\allowbreak{} you estimate how far that pair is likely to progress along the FDA clinical-\allowbreak{}approval spectrum,\allowbreak{} and how confident you are in that estimate.\allowbreak{}\par
\addvspace{3pt}
\noindent Treat this as a genuine discovery task.\allowbreak{} Start with one bounded tool\_\allowbreak{}A Stage-\allowbreak{}1 call using \texttt{-\allowbreak{}-\allowbreak{}tier extra -\allowbreak{}-\allowbreak{}output <\allowbreak{}file>\allowbreak{} -\allowbreak{}-\allowbreak{}summary};\allowbreak{} its summary is the canonical and sufficient decision surface.\allowbreak{} Do not inspect the artifact/\allowbreak{}filesystem or dump \texttt{coarse.\allowbreak{}provider\_\allowbreak{}results}/\allowbreak{}\texttt{fused\_\allowbreak{}graph}.\allowbreak{} Run a supplied \texttt{detail\_\allowbreak{}command} once only for a load-\allowbreak{}bearing page marked \texttt{truncated\_\allowbreak{}for\_\allowbreak{}summary:\allowbreak{} true}.\allowbreak{} An absent page is an evidence limitation,\allowbreak{} not a missing field.\allowbreak{}\par
\addvspace{3pt}
\noindent A node budget is each provider'\allowbreak{}s frontier-\allowbreak{}expansion budget,\allowbreak{} not its returned-\allowbreak{}node count.\allowbreak{} A \texttt{provider\_\allowbreak{}A} \texttt{partial} is usable bounded coverage-\allowbreak{}-\allowbreak{}-\allowbreak{}not a provider failure or negative evidence-\allowbreak{}-\allowbreak{}-\allowbreak{}when \texttt{error} is null,\allowbreak{} \texttt{timed\_\allowbreak{}out} is false,\allowbreak{} and \texttt{budget\_\allowbreak{}exhausted} is true.\allowbreak{} Answer when resolved anchors have grounded connections.\allowbreak{} Otherwise,\allowbreak{} decision-\allowbreak{}critical coverage permits one next-\allowbreak{}tier rerun into a new file;\allowbreak{} stable IDs or budget exhaustion alone do not.\allowbreak{} Stage 2 is optional:\allowbreak{} use one suggested \texttt{tool\_\allowbreak{}A expand} route only if it could change the score or confidence,\allowbreak{} never by rote or through bulk \texttt{-\allowbreak{}-\allowbreak{}stage2 auto}.\allowbreak{} Never repeat a tier,\allowbreak{} widen an error/\allowbreak{}timeout,\allowbreak{} rebuild/\allowbreak{}install providers,\allowbreak{} or use \texttt{bash -\allowbreak{}lc},\allowbreak{} pipelines,\allowbreak{} or retry loops.\allowbreak{} Preserve conflicts and operational limitations in your conclusion.\allowbreak{}\par
\addvspace{3pt}
\noindent Check the operating manual on how to interact with the environment below (\allowbreak{}in the <\allowbreak{}manual\_\allowbreak{}B>\allowbreak{} section)\allowbreak{}.\allowbreak{}\par
\addvspace{3pt}
\noindent Definition of required scores for output\par
\addvspace{3pt}
\addvspace{3pt}
\noindent PROMISINGNESS (\allowbreak{}a number in [\allowbreak{}0,\allowbreak{} 1]\allowbreak{})\allowbreak{} is the pair'\allowbreak{}s predicted position on the clinical-\allowbreak{}approval spectrum.\allowbreak{} It is scored against the pair'\allowbreak{}s real,\allowbreak{} held-\allowbreak{}out clinical outcome and is also used to rank candidates,\allowbreak{} so calibrate it to reflect true relative promise rather than enthusiasm.\allowbreak{}\par
\addvspace{3pt}
\noindent CONFIDENCE (\allowbreak{}a number in [\allowbreak{}0,\allowbreak{} 1]\allowbreak{})\allowbreak{} is your estimate of prediction quality:\allowbreak{} how close your promisingness score is to the true held-\allowbreak{}out value.\allowbreak{} It is scored as 1 -\allowbreak{} (\allowbreak{}confidence -\allowbreak{} quality)\allowbreak{}\textasciicircum{\allowbreak{}}\allowbreak{}2,\allowbreak{} where quality =\allowbreak{} 1 -\allowbreak{} prediction\_\allowbreak{}error\textasciicircum{\allowbreak{}}\allowbreak{}2.\allowbreak{} The optimal strategy is to set confidence equal to your expected prediction quality.\allowbreak{} Report low confidence when evidence is thin or conflicting.\allowbreak{}\par
\addvspace{3pt}
\noindent Anchor your promisingness estimate to the tier scale:\allowbreak{}\par
\addvspace{3pt}
\noindent Reference points (\allowbreak{}how the held-\allowbreak{}out outcome maps to the target score)\allowbreak{}:\allowbreak{} 0.\allowbreak{}00  preclinical only (\allowbreak{}or failed@Ph1)\allowbreak{} 0.\allowbreak{}15  passed Phase 1 (\allowbreak{}safety established)\allowbreak{} 0.\allowbreak{}45  passed Phase 2 (\allowbreak{}efficacy signal)\allowbreak{} 0.\allowbreak{}65  passed Phase 3 (\allowbreak{}pivotal)\allowbreak{} or off-\allowbreak{}label established 1.\allowbreak{}00  FDA approved on-\allowbreak{}label A Phase-\allowbreak{}3 success (\allowbreak{}\textasciitilde{\allowbreak{}}\allowbreak{}0.\allowbreak{}65)\allowbreak{} is the strongest single predictor of approval;\allowbreak{} a full approval scores 1.\allowbreak{}00.\allowbreak{} Interpolate between points.\allowbreak{}\par
\addvspace{3pt}
\noindent Scoring rules:\allowbreak{}\par
\begin{itemize}[leftmargin=1.2em,itemsep=2pt,topsep=2pt,parsep=0pt]
\item A successful but not-\allowbreak{}yet-\allowbreak{}approved Phase-\allowbreak{}1,\allowbreak{} Phase-\allowbreak{}2,\allowbreak{} or Phase-\allowbreak{}3 result is a LOWER BOUND on eventual program potential,\allowbreak{} not a terminal point target.\allowbreak{} Predicting below the phase actually passed is penalised;\allowbreak{} predicting above it is fine when the evidence supports further progression.\allowbreak{}
\item Genuinely terminal outcomes use an exact target.\allowbreak{} Failing at a later phase caps the credit at the last passed phase -\allowbreak{}-\allowbreak{}-\allowbreak{} e.\allowbreak{}g.\allowbreak{} a Phase-\allowbreak{}3 failure scores exactly 0.\allowbreak{}45 (\allowbreak{}Phase 2 was passed)\allowbreak{},\allowbreak{} not 0.\allowbreak{}65.\allowbreak{}
\item Ongoing trials should score at LEAST the tier of the last phase they passed.\allowbreak{} A drug currently in Phase 3 has already passed Phase 2,\allowbreak{} so its score should be >\allowbreak{}=\allowbreak{} 0.\allowbreak{}45.\allowbreak{} Predicting below that threshold is penalised;\allowbreak{} predicting above it is fine (\allowbreak{}you are estimating its eventual outcome)\allowbreak{}.\allowbreak{}
\item Interpolate between tier points when the evidence suggests an intermediate position.\allowbreak{}
\end{itemize}
\addvspace{3pt}
\noindent Evidence grounding (\allowbreak{}this is graded -\allowbreak{}-\allowbreak{}-\allowbreak{} cite like a RAG system)\allowbreak{}:\allowbreak{}\par
\begin{itemize}[leftmargin=1.2em,itemsep=2pt,topsep=2pt,parsep=0pt]
\item Ground every load-\allowbreak{}bearing claim in a SPECIFIC source you actually retrieved from the environment -\allowbreak{}-\allowbreak{}-\allowbreak{} a PubMed id (\allowbreak{}PMID:\allowbreak{}\#\#\#\#\#\#\#\#)\allowbreak{},\allowbreak{} a trial registry id (\allowbreak{}NCT\#\#\#\#\#\#\#\#)\allowbreak{},\allowbreak{} a knowledge-\allowbreak{}graph edge/\allowbreak{}id,\allowbreak{} a database record (\allowbreak{}e.\allowbreak{}g.\allowbreak{} ChEMBL /\allowbreak{} Open Targets /\allowbreak{} drug label)\allowbreak{},\allowbreak{} or the exact tool + query you ran.\allowbreak{} Prefer sources you pulled from the tools over parametric memory.\allowbreak{}
\item For each such claim,\allowbreak{} keep the SUPPORTING SNIPPET -\allowbreak{}-\allowbreak{}-\allowbreak{} the specific retrieved finding (\allowbreak{}a quoted line,\allowbreak{} a value,\allowbreak{} an edge)\allowbreak{} that backs it -\allowbreak{}-\allowbreak{}-\allowbreak{} so the claim is self-\allowbreak{}justifying and independently checkable.\allowbreak{}
\item You will report these as a structured \texttt{key\_\allowbreak{}claims} list in the output contract (\allowbreak{}claim + source + support)\allowbreak{}.\allowbreak{} A downstream verifier checks whether each claim'\allowbreak{}s cited source actually supports it,\allowbreak{} so uncited or unsupported claims lower the score -\allowbreak{}-\allowbreak{}-\allowbreak{} cite precisely,\allowbreak{} quote faithfully,\allowbreak{} do not invent sources.\allowbreak{}
\item If a claim rests only on model knowledge with no retrieved source,\allowbreak{} set its source to "\allowbreak{}parametric"\allowbreak{},\allowbreak{} leave \texttt{support} empty,\allowbreak{} and lower your confidence.\allowbreak{}
\end{itemize}
\addvspace{3pt}
\addvspace{3pt}
\noindent TASK:\allowbreak{} Assess the drug-\allowbreak{}repurposing potential of the following drug-\allowbreak{}disease pair.\allowbreak{}\par
\addvspace{3pt}
\noindent drug:\allowbreak{} Dostarlimab disease:\allowbreak{} Stage II Breast Cancer drug\_\allowbreak{}id:\allowbreak{} <\allowbreak{}drug\_\allowbreak{}id>\allowbreak{} disease\_\allowbreak{}id:\allowbreak{} <\allowbreak{}disease\_\allowbreak{}id>\allowbreak{}\par
\addvspace{3pt}
\noindent Investigate the pair carefully,\allowbreak{} using the available tools if helpful.\allowbreak{} Write a comprehensive report,\allowbreak{} then provide a promisingness score,\allowbreak{} a confidence score,\allowbreak{} and a concise rationale,\allowbreak{} following the output contract exactly.\allowbreak{}\par
\addvspace{3pt}
\noindent Output contract\par
\addvspace{3pt}
\noindent Write your full,\allowbreak{} comprehensive report first.\allowbreak{} Then,\allowbreak{} as the VERY LAST thing in your response,\allowbreak{} emit exactly ONE fenced code block tagged \texttt{dreg-\allowbreak{}result} containing a single JSON object with exactly these keys:\allowbreak{}\par
\addvspace{3pt}
\begin{payload}\{\allowbreak{}\\
~~"\allowbreak{}promisingness\_\allowbreak{}score"\allowbreak{}:\allowbreak{} <\allowbreak{}number between 0 and 1>\allowbreak{},\allowbreak{}\\
~~"\allowbreak{}confidence\_\allowbreak{}score"\allowbreak{}:\allowbreak{} <\allowbreak{}number between 0 and 1>\allowbreak{},\allowbreak{}\\
~~"\allowbreak{}rationale"\allowbreak{}:\allowbreak{} "\allowbreak{}<\allowbreak{}concise justification:\allowbreak{} key evidence,\allowbreak{} mechanism,\allowbreak{} and what drove the score;\allowbreak{} 3-\allowbreak{}8 sentences>\allowbreak{}"\allowbreak{},\allowbreak{}\\
~~"\allowbreak{}key\_\allowbreak{}claims"\allowbreak{}:\allowbreak{} [\allowbreak{}\\
~~~~\{\allowbreak{}\\
~~~~~~"\allowbreak{}claim"\allowbreak{}:\allowbreak{} "\allowbreak{}<\allowbreak{}one load-\allowbreak{}bearing factual claim that drove your score>\allowbreak{}"\allowbreak{},\allowbreak{}\\
~~~~~~"\allowbreak{}source"\allowbreak{}:\allowbreak{} "\allowbreak{}<\allowbreak{}the specific retrieved source:\allowbreak{} PMID:\allowbreak{}\#\#\#\#\#\#\#\# |\allowbreak{} NCT\#\#\#\#\#\#\#\# |\allowbreak{} a KG edge/\allowbreak{}id |\allowbreak{} a DB record e.\allowbreak{}g.\allowbreak{} ChEMBL:\allowbreak{}.\allowbreak{}.\allowbreak{}.\allowbreak{} /\allowbreak{} OpenTargets:\allowbreak{}.\allowbreak{}.\allowbreak{}.\allowbreak{} /\allowbreak{} drug label |\allowbreak{} tool:\allowbreak{}<\allowbreak{}name>\allowbreak{}+<\allowbreak{}query>\allowbreak{} |\allowbreak{} '\allowbreak{}parametric'\allowbreak{} if only model knowledge>\allowbreak{}"\allowbreak{},\allowbreak{}\\
~~~~~~"\allowbreak{}support"\allowbreak{}:\allowbreak{} "\allowbreak{}<\allowbreak{}the exact retrieved finding/\allowbreak{}quote that backs this claim;\allowbreak{} empty string if source is parametric>\allowbreak{}"\allowbreak{}\\
~~~~\}\allowbreak{}\\
~~]\allowbreak{}\\
\}\allowbreak{}\end{payload}
\addvspace{3pt}
\noindent Rules:\allowbreak{}\par
\begin{itemize}[leftmargin=1.2em,itemsep=2pt,topsep=2pt,parsep=0pt]
\item The block MUST be valid JSON (\allowbreak{}double-\allowbreak{}quoted keys and string values,\allowbreak{} no trailing commas,\allowbreak{} no comments)\allowbreak{}.\allowbreak{} Escape any quotes/\allowbreak{}newlines inside strings.\allowbreak{}
\item \texttt{promisingness\_\allowbreak{}score} and \texttt{confidence\_\allowbreak{}score} are plain numbers in [\allowbreak{}0,\allowbreak{} 1]\allowbreak{}.\allowbreak{}
\item \texttt{key\_\allowbreak{}claims}:\allowbreak{} 3-\allowbreak{}8 entries covering the MAIN drivers of your score.\allowbreak{} Each claim must carry the specific \texttt{source} you retrieved it from and the \texttt{support} snippet that backs it (\allowbreak{}see "\allowbreak{}Evidence grounding"\allowbreak{} above)\allowbreak{}.\allowbreak{} A downstream verifier scores whether each claim'\allowbreak{}s cited source supports it -\allowbreak{}-\allowbreak{}-\allowbreak{} so cite precisely,\allowbreak{} quote faithfully,\allowbreak{} and never invent a source.\allowbreak{} Use "\allowbreak{}parametric"\allowbreak{} only when you truly have no retrieved evidence,\allowbreak{} and lower \texttt{confidence\_\allowbreak{}score} accordingly.\allowbreak{}
\item Emit the block only once,\allowbreak{} at the end.\allowbreak{} Do not wrap it in extra prose after it.\allowbreak{}
\end{itemize}
\addvspace{3pt}
\addvspace{3pt}
\noindent the execution environment transport adapter\par
\addvspace{3pt}
\noindent This evaluation uses the execution environment actions.\allowbreak{} Do not finish with only the fenced JSON described above.\allowbreak{} Submit the same canonical JSON object with \texttt{act submit -\allowbreak{}-\allowbreak{}params '\allowbreak{}\{\allowbreak{}"\allowbreak{}submission"\allowbreak{}:\allowbreak{} \{\allowbreak{}.\allowbreak{}.\allowbreak{}.\allowbreak{}\}\allowbreak{}\}\allowbreak{}'\allowbreak{}}.\allowbreak{} Use \texttt{act domain-\allowbreak{}tool} for the bounded the domain toolkit tools documented in <\allowbreak{}toolkit\_\allowbreak{}dir>\allowbreak{}.\allowbreak{}\par}\end{taskbox}\begin{taskbox}{TASK INPUT \textbullet\ REQUIRED OUTPUT SCHEMA}\begin{payload}\{\allowbreak{}\\
~~"\allowbreak{}promisingness\_\allowbreak{}score"\allowbreak{}:\allowbreak{} "\allowbreak{}number in [\allowbreak{}0,\allowbreak{}1]\allowbreak{}"\allowbreak{},\allowbreak{}\\
~~"\allowbreak{}confidence\_\allowbreak{}score"\allowbreak{}:\allowbreak{} "\allowbreak{}number in [\allowbreak{}0,\allowbreak{}1]\allowbreak{}"\allowbreak{},\allowbreak{}\\
~~"\allowbreak{}rationale"\allowbreak{}:\allowbreak{} "\allowbreak{}string"\allowbreak{},\allowbreak{}\\
~~"\allowbreak{}key\_\allowbreak{}claims"\allowbreak{}:\allowbreak{} "\allowbreak{}list of \{\allowbreak{}claim,\allowbreak{} source,\allowbreak{} support\}\allowbreak{}"\allowbreak{}\\
\}\allowbreak{}\end{payload}\end{taskbox}\begin{taskbox}{TASK INPUT \textbullet\ EVALUATION CONTRACT}\begin{payload}the domain toolkit-\allowbreak{}eval/\allowbreak{}v1\end{payload}\end{taskbox}\begin{taskbox}{TASK INPUT \textbullet\ SCORING SCALE}\begin{payload}\{\allowbreak{}\\
~~"\allowbreak{}name"\allowbreak{}:\allowbreak{} "\allowbreak{}approval-\allowbreak{}spectrum-\allowbreak{}v2"\allowbreak{},\allowbreak{}\\
~~"\allowbreak{}values"\allowbreak{}:\allowbreak{} [\allowbreak{}\\
~~~~0.\allowbreak{}0,\allowbreak{}\\
~~~~0.\allowbreak{}15,\allowbreak{}\\
~~~~0.\allowbreak{}45,\allowbreak{}\\
~~~~0.\allowbreak{}65,\allowbreak{}\\
~~~~1.\allowbreak{}0\\
~~]\allowbreak{}\\
\}\allowbreak{}\end{payload}\end{taskbox}

\subsection{Full reasoning trace}

Every recorded step is shown. Reasoning turns are grey, tool calls blue, the failed call red, and
the submission gold. Tool calls are shown verbatim, with the returned payload trimmed only for page width. Model
reasoning turns are listed but their content is withheld: the judging panel never sees the
system's private chain of thought, only the actions and outputs it produced.

\begin{stepR}{REASONING \textbullet\ STEP 1 \textbullet\ Orient on the episode brief and output contract}
\stepsub{ACTION EMITTED}\begin{payload}read  \{\allowbreak{}"\allowbreak{}path"\allowbreak{}:\allowbreak{}"\allowbreak{}<\allowbreak{}manual\_\allowbreak{}C>\allowbreak{}"\allowbreak{}\}\allowbreak{}\end{payload}
\end{stepR}
\begin{stepR}{REASONING \textbullet\ STEP 2 \textbullet\ Decide to read the manuals before retrieving anything}
\stepsub{ACTION EMITTED}\begin{payload}bash  \{\allowbreak{}"\allowbreak{}command"\allowbreak{}:\allowbreak{}"\allowbreak{}act inspect-\allowbreak{}task"\allowbreak{},\allowbreak{}"\allowbreak{}timeout"\allowbreak{}:\allowbreak{}120\}\allowbreak{}\end{payload}
\end{stepR}
\begin{stepT}{TOOL CALL \textbullet\ STEP 3 \textbullet\ Inspect the canonical task record}
\stepsub{RETURNED}\noindent{\footnotesize\itshape The \texttt{description} field is the task brief reproduced in Section~\ref{sec:task-as-given} above and is omitted here to avoid duplication; every other field is shown.}\par\stepsub{RETURNED}\begin{payload}\{\allowbreak{}\\
~~"\allowbreak{}task"\allowbreak{}:\allowbreak{} "\allowbreak{}pointwise-\allowbreak{}promisingness"\allowbreak{},\allowbreak{}\\
~~"\allowbreak{}drug"\allowbreak{}:\allowbreak{} "\allowbreak{}Dostarlimab"\allowbreak{},\allowbreak{}\\
~~"\allowbreak{}disease"\allowbreak{}:\allowbreak{} "\allowbreak{}Stage II Breast Cancer"\allowbreak{},\allowbreak{}\\
~~"\allowbreak{}drug\_\allowbreak{}id"\allowbreak{}:\allowbreak{} "\allowbreak{}<\allowbreak{}drug\_\allowbreak{}id>\allowbreak{}"\allowbreak{},\allowbreak{}\\
~~"\allowbreak{}disease\_\allowbreak{}id"\allowbreak{}:\allowbreak{} "\allowbreak{}<\allowbreak{}disease\_\allowbreak{}id>\allowbreak{}"\allowbreak{},\allowbreak{}\\
~~"\allowbreak{}submission\_\allowbreak{}schema"\allowbreak{}:\allowbreak{} \{\allowbreak{}\\
~~~~"\allowbreak{}promisingness\_\allowbreak{}score"\allowbreak{}:\allowbreak{} "\allowbreak{}number in [\allowbreak{}0,\allowbreak{}1]\allowbreak{}"\allowbreak{},\allowbreak{}\\
~~~~"\allowbreak{}confidence\_\allowbreak{}score"\allowbreak{}:\allowbreak{} "\allowbreak{}number in [\allowbreak{}0,\allowbreak{}1]\allowbreak{}"\allowbreak{},\allowbreak{}\\
~~~~"\allowbreak{}rationale"\allowbreak{}:\allowbreak{} "\allowbreak{}string"\allowbreak{},\allowbreak{}\\
~~~~"\allowbreak{}key\_\allowbreak{}claims"\allowbreak{}:\allowbreak{} "\allowbreak{}list of \{\allowbreak{}claim,\allowbreak{} source,\allowbreak{} support\}\allowbreak{}"\allowbreak{}\\
~~\}\allowbreak{},\allowbreak{}\\
~~"\allowbreak{}tier\_\allowbreak{}scale"\allowbreak{}:\allowbreak{} \{\allowbreak{}\\
~~~~"\allowbreak{}name"\allowbreak{}:\allowbreak{} "\allowbreak{}approval-\allowbreak{}spectrum-\allowbreak{}v2"\allowbreak{},\allowbreak{}\\
~~~~"\allowbreak{}values"\allowbreak{}:\allowbreak{} [\allowbreak{}\\
~~~~~~0.\allowbreak{}0,\allowbreak{}\\
~~~~~~0.\allowbreak{}15,\allowbreak{}\\
~~~~~~0.\allowbreak{}45,\allowbreak{}\\
~~~~~~0.\allowbreak{}65,\allowbreak{}\\
~~~~~~1.\allowbreak{}0\\
~~~~]\allowbreak{}\\
~~\}\allowbreak{},\allowbreak{}\\
~~"\allowbreak{}eval\_\allowbreak{}contract"\allowbreak{}:\allowbreak{} "\allowbreak{}the domain toolkit-\allowbreak{}eval/\allowbreak{}v1"\allowbreak{},\allowbreak{}\\
~~"\allowbreak{}workdir"\allowbreak{}:\allowbreak{} "\allowbreak{}<\allowbreak{}workdir>\allowbreak{}"\allowbreak{},\allowbreak{}\\
~~"\allowbreak{}staged\_\allowbreak{}files"\allowbreak{}:\allowbreak{} [\allowbreak{}\\
~~~~"\allowbreak{}instance.\allowbreak{}json"\allowbreak{},\allowbreak{}\\
~~~~"\allowbreak{}toolkit/\allowbreak{}"\allowbreak{}\\
~~]\allowbreak{},\allowbreak{}\\
~~"\allowbreak{}toolkit"\allowbreak{}:\allowbreak{} \{\allowbreak{}\\
~~~~"\allowbreak{}access\_\allowbreak{}adapter"\allowbreak{}:\allowbreak{} "\allowbreak{}<\allowbreak{}manual\_\allowbreak{}A>\allowbreak{}"\allowbreak{},\allowbreak{}\\
~~~~"\allowbreak{}operating\_\allowbreak{}manual"\allowbreak{}:\allowbreak{} "\allowbreak{}<\allowbreak{}manual\_\allowbreak{}B>\allowbreak{}"\allowbreak{},\allowbreak{}\\
~~~~"\allowbreak{}skills\_\allowbreak{}dir"\allowbreak{}:\allowbreak{} "\allowbreak{}<\allowbreak{}skills\_\allowbreak{}dir>\allowbreak{}"\allowbreak{},\allowbreak{}\\
~~~~"\allowbreak{}manifest"\allowbreak{}:\allowbreak{} "\allowbreak{}<\allowbreak{}manifest>\allowbreak{}"\allowbreak{}\\
~~\}\allowbreak{},\allowbreak{}\\
~~"\allowbreak{}note"\allowbreak{}:\allowbreak{} "\allowbreak{}Bootstrap complete.\allowbreak{} Next read toolkit.\allowbreak{}access\_\allowbreak{}adapter,\allowbreak{} then toolkit.\allowbreak{}operating\_\allowbreak{}manual;\allowbreak{} follow them and submit the canonical the domain toolkit fields.\allowbreak{}"\allowbreak{}\\
\}\allowbreak{}\end{payload}
\end{stepT}
\begin{stepR}{REASONING \textbullet\ STEP 4 \textbullet\ Register the enforced offline-evaluation policy}
\stepsub{ACTION EMITTED}\begin{payload}bash  \{\allowbreak{}"\allowbreak{}command"\allowbreak{}:\allowbreak{}"\allowbreak{}act bash -\allowbreak{}-\allowbreak{}cmd '\allowbreak{}cat <\allowbreak{}manual\_\allowbreak{}A>\allowbreak{}'\allowbreak{}"\allowbreak{},\allowbreak{}"\allowbreak{}timeout"\allowbreak{}:\allowbreak{}120\}\allowbreak{}\\
bash  \{\allowbreak{}"\allowbreak{}command"\allowbreak{}:\allowbreak{}"\allowbreak{}act bash -\allowbreak{}-\allowbreak{}cmd '\allowbreak{}cat <\allowbreak{}manual\_\allowbreak{}B>\allowbreak{}'\allowbreak{}"\allowbreak{},\allowbreak{}"\allowbreak{}timeout"\allowbreak{}:\allowbreak{}120\}\allowbreak{}\end{payload}
\end{stepR}
\begin{stepT}{TOOL CALL \textbullet\ STEP 5 \textbullet\ Read the environment access manual}
\stepsub{CALL}\begin{payload}cat <\allowbreak{}manual\_\allowbreak{}A>\allowbreak{}\end{payload}\stepsub{RETURNED}{\footnotesize \noindent exit\_\allowbreak{}code:\allowbreak{} 0 stdout:\allowbreak{}\par
\par\addvspace{5pt}\noindent{\normalsize\bfseries\sffamily Access adapter -\allowbreak{}-\allowbreak{}-\allowbreak{} the execution environment (\allowbreak{}isolated SUT)\allowbreak{}}\par\addvspace{2pt}
\addvspace{3pt}
\noindent This file owns only the execution environment execution mechanics.\allowbreak{} [\allowbreak{}\texttt{.\allowbreak{}.\allowbreak{}/\allowbreak{}<\allowbreak{}manual\_\allowbreak{}B>\allowbreak{}}]\allowbreak{}(\allowbreak{}.\allowbreak{}.\allowbreak{}/\allowbreak{}<\allowbreak{}manual\_\allowbreak{}B>\allowbreak{})\allowbreak{} owns the the domain toolkit retrieval and inference policy.\allowbreak{}\par
\addvspace{3pt}
\par\addvspace{5pt}\noindent{\small\bfseries\sffamily Actions}\par\addvspace{2pt}
\addvspace{3pt}
\noindent Every action is metered and recorded:\allowbreak{}\par
\addvspace{3pt}
\begin{itemize}[leftmargin=1.2em,itemsep=2pt,topsep=2pt,parsep=0pt]
\item \texttt{act inspect-\allowbreak{}task} returns the canonical task,\allowbreak{} visible instance,\allowbreak{} and output schema.\allowbreak{}
\item \texttt{act bash -\allowbreak{}-\allowbreak{}cmd '\allowbreak{}<\allowbreak{}command>\allowbreak{}'\allowbreak{}} runs notes or small computations in the isolated the sandbox.\allowbreak{}
\item \texttt{act domain-\allowbreak{}tool -\allowbreak{}-\allowbreak{}params '\allowbreak{}<\allowbreak{}json>\allowbreak{}'\allowbreak{}} invokes a structured,\allowbreak{} allowlisted the domain toolkit capability on the trusted tool host.\allowbreak{}
\item \texttt{act submit -\allowbreak{}-\allowbreak{}params '\allowbreak{}\{\allowbreak{}"\allowbreak{}submission"\allowbreak{}:\allowbreak{} \{\allowbreak{}.\allowbreak{}.\allowbreak{}.\allowbreak{}\}\allowbreak{}\}\allowbreak{}'\allowbreak{}} records the final answer and ends the episode.\allowbreak{}
\end{itemize}
\addvspace{3pt}
\par\addvspace{5pt}\noindent{\small\bfseries\sffamily the sandbox and paths}\par\addvspace{2pt}
\addvspace{3pt}
\begin{itemize}[leftmargin=1.2em,itemsep=2pt,topsep=2pt,parsep=0pt]
\item \texttt{bash} works only under \texttt{<\allowbreak{}workdir>\allowbreak{}}.\allowbreak{} The supplied files are \texttt{<\allowbreak{}instance\_\allowbreak{}file>\allowbreak{}} and \texttt{<\allowbreak{}toolkit\_\allowbreak{}dir>\allowbreak{}};\allowbreak{} files you create are ephemeral.\allowbreak{}
\item The the sandbox has no network,\allowbreak{} host mounts,\allowbreak{} the domain toolkit checkout,\allowbreak{} benchmark data,\allowbreak{} hidden reference,\allowbreak{} host tool workspace,\allowbreak{} Docker socket,\allowbreak{} or host GPU devices.\allowbreak{}
\item Never use a host session,\allowbreak{} mounted-\allowbreak{}data,\allowbreak{} checkout,\allowbreak{} or the execution environment episode path.\allowbreak{}
\item Persist the evaluated answer with \texttt{submit},\allowbreak{} not a local file.\allowbreak{}
\end{itemize}
\addvspace{3pt}
\par\addvspace{5pt}\noindent{\small\bfseries\sffamily the domain toolkit tool boundary}\par\addvspace{2pt}
\addvspace{3pt}
\noindent tool\_\allowbreak{}A requests have this envelope:\allowbreak{}\par
\addvspace{3pt}
\begin{payload}act domain-\allowbreak{}tool -\allowbreak{}-\allowbreak{}params \textbackslash{\allowbreak{}}\allowbreak{}\\
~~'\allowbreak{}\{\allowbreak{}"\allowbreak{}tool"\allowbreak{}:\allowbreak{}"\allowbreak{}tool\_\allowbreak{}A"\allowbreak{},\allowbreak{}"\allowbreak{}command"\allowbreak{}:\allowbreak{}"\allowbreak{}tool\_\allowbreak{}A .\allowbreak{}.\allowbreak{}.\allowbreak{}"\allowbreak{}\}\allowbreak{}'\allowbreak{}\end{payload}
\addvspace{3pt}
\noindent inference-\allowbreak{}service requests have this envelope:\allowbreak{}\par
\addvspace{3pt}
\begin{payload}act domain-\allowbreak{}tool -\allowbreak{}-\allowbreak{}params \textbackslash{\allowbreak{}}\allowbreak{}\\
~~'\allowbreak{}\{\allowbreak{}"\allowbreak{}tool"\allowbreak{}:\allowbreak{}"\allowbreak{}tool\_\allowbreak{}B"\allowbreak{},\allowbreak{}"\allowbreak{}operation"\allowbreak{}:\allowbreak{}"\allowbreak{}nim"\allowbreak{},\allowbreak{}"\allowbreak{}arguments"\allowbreak{}:\allowbreak{}\{\allowbreak{}"\allowbreak{}method"\allowbreak{}:\allowbreak{}"\allowbreak{}GET"\allowbreak{},\allowbreak{}"\allowbreak{}port"\allowbreak{}:\allowbreak{}8000,\allowbreak{}"\allowbreak{}path"\allowbreak{}:\allowbreak{}"\allowbreak{}/\allowbreak{}openapi.\allowbreak{}json"\allowbreak{}\}\allowbreak{}\}\allowbreak{}'\allowbreak{}\end{payload}
\addvspace{3pt}
\begin{itemize}[leftmargin=1.2em,itemsep=2pt,topsep=2pt,parsep=0pt]
\item Tool output paths must be virtual paths below \texttt{<\allowbreak{}artifact>\allowbreak{}}.\allowbreak{} The adapter maps them to this episode'\allowbreak{}s private host-\allowbreak{}side workspace.\allowbreak{}
\item \texttt{bash} cannot read tool artifacts.\allowbreak{} Use the returned summary or a policy-\allowbreak{}permitted inspection request through \texttt{domain-\allowbreak{}tool}.\allowbreak{}
\item Only capabilities in \texttt{<\allowbreak{}manifest>\allowbreak{}} are callable.\allowbreak{}
\item Requests are parsed into fixed arguments.\allowbreak{} No shell,\allowbreak{} SSH,\allowbreak{} pipelines,\allowbreak{} command substitution,\allowbreak{} arbitrary executable,\allowbreak{} URL,\allowbreak{} or filesystem path reaches the tool host.\allowbreak{} inference-\allowbreak{}service calls are limited to the configured local port range.\allowbreak{}
\end{itemize}
\addvspace{3pt}
\noindent \{\allowbreak{} "\allowbreak{}stderr"\allowbreak{}:\allowbreak{} "\allowbreak{}"\allowbreak{} \}\allowbreak{}\par}
\end{stepT}
\begin{stepT}{TOOL CALL \textbullet\ STEP 6 \textbullet\ Read the domain toolkit manual}
\stepsub{CALL}\begin{payload}cat <\allowbreak{}manual\_\allowbreak{}B>\allowbreak{}\end{payload}\stepsub{RETURNED}{\footnotesize \noindent exit\_\allowbreak{}code:\allowbreak{} 0 stdout:\allowbreak{}\par
\par\addvspace{5pt}\noindent{\normalsize\bfseries\sffamily the domain toolkit biomedical toolkit in the execution environment}\par\addvspace{2pt}
\addvspace{3pt}
\noindent This file owns only the domain toolkit'\allowbreak{}s retrieval and inference workflow.\allowbreak{} The the execution environment bootstrap brief owns episode order;\allowbreak{} [\allowbreak{}\texttt{access/\allowbreak{}<\allowbreak{}manual\_\allowbreak{}A>\allowbreak{}}]\allowbreak{}(\allowbreak{}access/\allowbreak{}<\allowbreak{}manual\_\allowbreak{}A>\allowbreak{})\allowbreak{} owns action syntax,\allowbreak{} the sandbox isolation,\allowbreak{} virtual artifact mapping,\allowbreak{} and submission mechanics.\allowbreak{} No skill document is required for this focused surface.\allowbreak{}\par
\addvspace{3pt}
\par\addvspace{5pt}\noindent{\small\bfseries\sffamily Available tools}\par\addvspace{2pt}
\addvspace{3pt}
\noindent The catalog below is generated from the domain toolkit'\allowbreak{}s benchmark tool registry.\allowbreak{} Ground-\allowbreak{}truth and verifier tools are never included.\allowbreak{}\par
\addvspace{3pt}
\par\addvspace{5pt}\noindent{\footnotesize\bfseries\sffamily tool\_\allowbreak{}A -\allowbreak{}-\allowbreak{}-\allowbreak{} Coarse-\allowbreak{}to-\allowbreak{}fine biomedical pathfinding across broad KGs and typed specialist providers.\allowbreak{}}\par\addvspace{2pt}
\addvspace{3pt}
\par\addvspace{5pt}\noindent{\footnotesize\bfseries\sffamily tool\_\allowbreak{}B -\allowbreak{}-\allowbreak{}-\allowbreak{} Vendored NVIDIA tool\_\allowbreak{}B -\allowbreak{}-\allowbreak{}-\allowbreak{} inference-\allowbreak{}service skills,\allowbreak{} open models,\allowbreak{} GPU libraries.\allowbreak{}}\par\addvspace{2pt}
\addvspace{3pt}
\par\addvspace{5pt}\noindent{\small\bfseries\sffamily Required tool\_\allowbreak{}A workflow}\par\addvspace{2pt}
\addvspace{3pt}
\noindent tool\_\allowbreak{}A is the single coarse-\allowbreak{}to-\allowbreak{}fine entry point for connecting drugs,\allowbreak{} diseases,\allowbreak{} genes,\allowbreak{} proteins,\allowbreak{} pathways,\allowbreak{} and mechanisms.\allowbreak{}\par
\addvspace{3pt}
\noindent Start with one bounded Stage-\allowbreak{}1 search per pair:\allowbreak{}\par
\addvspace{3pt}
\begin{payload}act domain-\allowbreak{}tool -\allowbreak{}-\allowbreak{}params '\allowbreak{}\{\allowbreak{}\\
~~"\allowbreak{}tool"\allowbreak{}:\allowbreak{} "\allowbreak{}tool\_\allowbreak{}A"\allowbreak{},\allowbreak{}\\
~~"\allowbreak{}command"\allowbreak{}:\allowbreak{} "\allowbreak{}tool\_\allowbreak{}A connect "\allowbreak{}<\allowbreak{}drug>\allowbreak{}"\allowbreak{} "\allowbreak{}<\allowbreak{}disease>\allowbreak{}"\allowbreak{} -\allowbreak{}-\allowbreak{}tier extra -\allowbreak{}-\allowbreak{}timeout 240 -\allowbreak{}-\allowbreak{}output <\allowbreak{}artifact\_\allowbreak{}1>\allowbreak{} -\allowbreak{}-\allowbreak{}summary"\allowbreak{}\\
\}\allowbreak{}'\allowbreak{}\end{payload}
\addvspace{3pt}
\noindent The returned \texttt{-\allowbreak{}-\allowbreak{}summary} projection is the canonical decision surface.\allowbreak{} Inspect \texttt{resolved\_\allowbreak{}entities},\allowbreak{} \texttt{providers\_\allowbreak{}run},\allowbreak{} \texttt{candidate\_\allowbreak{}paths},\allowbreak{} \texttt{connections},\allowbreak{} \texttt{load\_\allowbreak{}bearing\_\allowbreak{}evidence},\allowbreak{} \texttt{relationship\_\allowbreak{}pages},\allowbreak{} \texttt{expansion\_\allowbreak{}candidates},\allowbreak{} \texttt{stage2},\allowbreak{} and \texttt{report} there.\allowbreak{} Do not request or dump \texttt{coarse.\allowbreak{}provider\_\allowbreak{}results} or \texttt{fused\_\allowbreak{}graph}.\allowbreak{}\par
\addvspace{3pt}
\noindent Only when a load-\allowbreak{}bearing relationship page is marked \texttt{truncated\_\allowbreak{}for\_\allowbreak{}summary:\allowbreak{} true},\allowbreak{} send its supplied \texttt{detail\_\allowbreak{}command} through \texttt{domain-\allowbreak{}tool} once.\allowbreak{} That command must use an existing virtual artifact:\allowbreak{}\par
\addvspace{3pt}
\begin{payload}act domain-\allowbreak{}tool -\allowbreak{}-\allowbreak{}params '\allowbreak{}\{\allowbreak{}\\
~~"\allowbreak{}tool"\allowbreak{}:\allowbreak{} "\allowbreak{}tool\_\allowbreak{}A"\allowbreak{},\allowbreak{}\\
~~"\allowbreak{}command"\allowbreak{}:\allowbreak{} "\allowbreak{}tool\_\allowbreak{}A inspect <\allowbreak{}artifact\_\allowbreak{}1>\allowbreak{} /\allowbreak{}json/\allowbreak{}pointer"\allowbreak{}\\
\}\allowbreak{}'\allowbreak{}\end{payload}
\addvspace{3pt}
\noindent A provider'\allowbreak{}s node budget is its frontier-\allowbreak{}expansion budget,\allowbreak{} not its returned-\allowbreak{}node count.\allowbreak{} A \texttt{provider\_\allowbreak{}A} \texttt{partial} result is usable bounded coverage when \texttt{error} is null,\allowbreak{} \texttt{timed\_\allowbreak{}out} is false,\allowbreak{} and \texttt{budget\_\allowbreak{}exhausted} is true.\allowbreak{} It is not negative evidence.\allowbreak{}\par
\addvspace{3pt}
\noindent If correctly resolved anchors already have grounded connections,\allowbreak{} answer from the summary.\allowbreak{} Otherwise,\allowbreak{} decision-\allowbreak{}critical coverage permits one next-\allowbreak{}tier rerun to a new \texttt{<\allowbreak{}artifact>\allowbreak{}} path.\allowbreak{} Never repeat a tier or widen a timed-\allowbreak{}out/\allowbreak{}error run.\allowbreak{}\par
\addvspace{3pt}
\noindent Stage 2 is optional.\allowbreak{} Use one route from \texttt{expansion\_\allowbreak{}candidates[\allowbreak{}]\allowbreak{}.\allowbreak{}compatible\_\allowbreak{}routes} only when the extra evidence could change the conclusion or confidence:\allowbreak{}\par
\addvspace{3pt}
\begin{payload}act domain-\allowbreak{}tool -\allowbreak{}-\allowbreak{}params '\allowbreak{}\{\allowbreak{}\\
~~"\allowbreak{}tool"\allowbreak{}:\allowbreak{} "\allowbreak{}tool\_\allowbreak{}A"\allowbreak{},\allowbreak{}\\
~~"\allowbreak{}command"\allowbreak{}:\allowbreak{} "\allowbreak{}tool\_\allowbreak{}A expand "\allowbreak{}<\allowbreak{}typed A>\allowbreak{}"\allowbreak{} "\allowbreak{}<\allowbreak{}typed B>\allowbreak{}"\allowbreak{} -\allowbreak{}-\allowbreak{}provider <\allowbreak{}provider\_\allowbreak{}id>\allowbreak{} -\allowbreak{}-\allowbreak{}algorithm <\allowbreak{}algorithm\_\allowbreak{}id>\allowbreak{} -\allowbreak{}-\allowbreak{}a-\allowbreak{}type <\allowbreak{}type>\allowbreak{} -\allowbreak{}-\allowbreak{}b-\allowbreak{}type <\allowbreak{}type>\allowbreak{} -\allowbreak{}-\allowbreak{}timeout 120 -\allowbreak{}-\allowbreak{}output <\allowbreak{}artifact\_\allowbreak{}5>\allowbreak{} -\allowbreak{}-\allowbreak{}summary"\allowbreak{}\\
\}\allowbreak{}'\allowbreak{}\end{payload}
\addvspace{3pt}
\noindent Do not invent a route when none is suggested.\allowbreak{} Do not use bulk \texttt{-\allowbreak{}-\allowbreak{}stage2 auto},\allowbreak{} shell wrappers,\allowbreak{} pipelines,\allowbreak{} retry loops,\allowbreak{} provider installation,\allowbreak{} or provider build commands during an evaluation.\allowbreak{}\par
\addvspace{3pt}
\par\addvspace{5pt}\noindent{\small\bfseries\sffamily Optional tool\_\allowbreak{}B inference-\allowbreak{}service inference}\par\addvspace{2pt}
\addvspace{3pt}
\noindent The the execution environment contract supports structured GET/\allowbreak{}POST inference against these resident tool-\allowbreak{}host endpoints:\allowbreak{}\par
\addvspace{3pt}
\noindent |\allowbreak{} inference-\allowbreak{}service |\allowbreak{} Port |\allowbreak{} Inference path |\allowbreak{} |\allowbreak{}-\allowbreak{}-\allowbreak{}-\allowbreak{}|\allowbreak{}-\allowbreak{}-\allowbreak{}-\allowbreak{}:\allowbreak{}|\allowbreak{}-\allowbreak{}-\allowbreak{}-\allowbreak{}|\allowbreak{} |\allowbreak{} <\allowbreak{}model\_\allowbreak{}endpoint>\allowbreak{} |\allowbreak{} 8000 |\allowbreak{} \texttt{POST <\allowbreak{}endpoint\_\allowbreak{}path>\allowbreak{}<\allowbreak{}model\_\allowbreak{}endpoint>\allowbreak{}/\allowbreak{}predict} |\allowbreak{} |\allowbreak{} <\allowbreak{}model\_\allowbreak{}endpoint>\allowbreak{} |\allowbreak{} 8001 |\allowbreak{} \texttt{POST <\allowbreak{}endpoint\_\allowbreak{}path>\allowbreak{}<\allowbreak{}model\_\allowbreak{}endpoint>\allowbreak{}/\allowbreak{}generate} |\allowbreak{} |\allowbreak{} <\allowbreak{}model\_\allowbreak{}endpoint>\allowbreak{} |\allowbreak{} 8002 |\allowbreak{} \texttt{POST /\allowbreak{}molecular-\allowbreak{}docking/\allowbreak{}<\allowbreak{}model\_\allowbreak{}endpoint>\allowbreak{}/\allowbreak{}generate} |\allowbreak{} |\allowbreak{} <\allowbreak{}model\_\allowbreak{}endpoint>\allowbreak{} |\allowbreak{} 8003 |\allowbreak{} \texttt{POST <\allowbreak{}endpoint\_\allowbreak{}path>\allowbreak{}<\allowbreak{}model\_\allowbreak{}endpoint>\allowbreak{}/\allowbreak{}predict} |\allowbreak{} |\allowbreak{} <\allowbreak{}model\_\allowbreak{}endpoint>\allowbreak{} |\allowbreak{} 8004 |\allowbreak{} \texttt{POST /\allowbreak{}generate} |\allowbreak{} |\allowbreak{} <\allowbreak{}model\_\allowbreak{}endpoint>\allowbreak{} |\allowbreak{} 8005 |\allowbreak{} \texttt{POST /\allowbreak{}generate} |\allowbreak{} |\allowbreak{} <\allowbreak{}model\_\allowbreak{}endpoint>\allowbreak{} |\allowbreak{} 8006 |\allowbreak{} \texttt{POST <\allowbreak{}endpoint\_\allowbreak{}path>\allowbreak{}} |\allowbreak{} |\allowbreak{} <\allowbreak{}model\_\allowbreak{}endpoint>\allowbreak{} |\allowbreak{} 8007 |\allowbreak{} \texttt{POST <\allowbreak{}endpoint\_\allowbreak{}path>\allowbreak{}<\allowbreak{}model\_\allowbreak{}endpoint>\allowbreak{}/\allowbreak{}predict-\allowbreak{}structure-\allowbreak{}from-\allowbreak{}msa-\allowbreak{}and-\allowbreak{}template} |\allowbreak{} |\allowbreak{} <\allowbreak{}model\_\allowbreak{}endpoint>\allowbreak{} |\allowbreak{} 8008 |\allowbreak{} \texttt{POST <\allowbreak{}endpoint\_\allowbreak{}path>\allowbreak{}<\allowbreak{}model\_\allowbreak{}endpoint>\allowbreak{}/\allowbreak{}predict} |\allowbreak{} |\allowbreak{} <\allowbreak{}model\_\allowbreak{}endpoint>\allowbreak{} |\allowbreak{} 8009 |\allowbreak{} \texttt{POST <\allowbreak{}endpoint\_\allowbreak{}path>\allowbreak{}<\allowbreak{}model\_\allowbreak{}endpoint>\allowbreak{}/\allowbreak{}msa-\allowbreak{}search/\allowbreak{}predict} |\allowbreak{}\par
\addvspace{3pt}
\noindent Call a inference-\allowbreak{}service through the structured adapter:\allowbreak{}\par
\addvspace{3pt}
\begin{payload}act domain-\allowbreak{}tool -\allowbreak{}-\allowbreak{}params '\allowbreak{}\{\allowbreak{}\\
~~"\allowbreak{}tool"\allowbreak{}:\allowbreak{} "\allowbreak{}tool\_\allowbreak{}B"\allowbreak{},\allowbreak{}\\
~~"\allowbreak{}operation"\allowbreak{}:\allowbreak{} "\allowbreak{}nim"\allowbreak{},\allowbreak{}\\
~~"\allowbreak{}arguments"\allowbreak{}:\allowbreak{} \{\allowbreak{}\\
~~~~"\allowbreak{}method"\allowbreak{}:\allowbreak{} "\allowbreak{}POST"\allowbreak{},\allowbreak{}\\
~~~~"\allowbreak{}port"\allowbreak{}:\allowbreak{} 8003,\allowbreak{}\\
~~~~"\allowbreak{}path"\allowbreak{}:\allowbreak{} "\allowbreak{}<\allowbreak{}endpoint\_\allowbreak{}path>\allowbreak{}<\allowbreak{}model\_\allowbreak{}endpoint>\allowbreak{}/\allowbreak{}predict"\allowbreak{},\allowbreak{}\\
~~~~"\allowbreak{}json"\allowbreak{}:\allowbreak{} \{\allowbreak{}\}\allowbreak{}\\
~~\}\allowbreak{}\\
\}\allowbreak{}'\allowbreak{}\end{payload}
\addvspace{3pt}
\noindent The port must be 8000–8009,\allowbreak{} the path must be local and absolute,\allowbreak{} and the body must be JSON.\allowbreak{} If the exact deployed request schema is needed,\allowbreak{} send a structured \texttt{GET} to \texttt{/\allowbreak{}openapi.\allowbreak{}json} on that inference-\allowbreak{}service'\allowbreak{}s assigned port.\allowbreak{} Do not run inference-\allowbreak{}service setup,\allowbreak{} \texttt{docker login},\allowbreak{} \texttt{docker run},\allowbreak{} readiness loops,\allowbreak{} or direct \texttt{curl};\allowbreak{} deployment health is controlled by the operator.\allowbreak{}\par
\addvspace{3pt}
\noindent \{\allowbreak{} "\allowbreak{}stderr"\allowbreak{}:\allowbreak{} "\allowbreak{}"\allowbreak{} \}\allowbreak{}\par}
\end{stepT}
\begin{stepR}{REASONING \textbullet\ STEP 7 \textbullet\ Form the retrieval plan}
\stepsub{MODEL NARRATION}\noindent{\footnotesize\itshape I have the operating rules.\allowbreak{} I'\allowbreak{}m going to run the required bounded tool\_\allowbreak{}A Stage-\allowbreak{}1 search and base the decision on its summary unless it exposes a specific truncated page that needs one permitted detail lookup.\allowbreak{}}\par
\end{stepR}
\begin{stepT}{TOOL CALL \textbullet\ STEP 8 \textbullet\ Bounded pairwise search for the assigned pair}
\stepsub{CALL}\begin{payload}tool\_\allowbreak{}A connect "\allowbreak{}Dostarlimab"\allowbreak{} "\allowbreak{}Stage II Breast Cancer"\allowbreak{} -\allowbreak{}-\allowbreak{}tier extra -\allowbreak{}-\allowbreak{}timeout 240 -\allowbreak{}-\allowbreak{}output <\allowbreak{}artifact\_\allowbreak{}1>\allowbreak{} -\allowbreak{}-\allowbreak{}summary\end{payload}\stepsub{RETURNED}{\footnotesize \noindent exit\_\allowbreak{}code:\allowbreak{} 0  tool:\allowbreak{} tool\_\allowbreak{}A  contract\_\allowbreak{}version:\allowbreak{} the domain toolkit-\allowbreak{}tools/\allowbreak{}v1 stdout:\allowbreak{} \{\allowbreak{}"\allowbreak{}output\_\allowbreak{}path"\allowbreak{}:\allowbreak{} "\allowbreak{}<\allowbreak{}artifact\_\allowbreak{}1>\allowbreak{}"\allowbreak{},\allowbreak{} "\allowbreak{}summary"\allowbreak{}:\allowbreak{} \{\allowbreak{}"\allowbreak{}schema\_\allowbreak{}version"\allowbreak{}:\allowbreak{} "\allowbreak{}tool\_\allowbreak{}A/\allowbreak{}v2"\allowbreak{},\allowbreak{} "\allowbreak{}stage"\allowbreak{}:\allowbreak{} "\allowbreak{}stage1"\allowbreak{},\allowbreak{} "\allowbreak{}workflow"\allowbreak{}:\allowbreak{} \{\allowbreak{}"\allowbreak{}review\_\allowbreak{}before\_\allowbreak{}next\_\allowbreak{}call"\allowbreak{}:\allowbreak{} true,\allowbreak{} "\allowbreak{}stage1\_\allowbreak{}follow\_\allowbreak{}up"\allowbreak{}:\allowbreak{} "\allowbreak{}Optionally query one decision-\allowbreak{}relevant related\_\allowbreak{}page\_\allowbreak{}hint for more Stage-\allowbreak{}1 literature context.\allowbreak{}"\allowbreak{},\allowbreak{} "\allowbreak{}stage1\_\allowbreak{}follow\_\allowbreak{}up\_\allowbreak{}command\_\allowbreak{}template"\allowbreak{}:\allowbreak{} "\allowbreak{}tool\_\allowbreak{}A connect <\allowbreak{}hint.\allowbreak{}title>\allowbreak{} -\allowbreak{}-\allowbreak{}a-\allowbreak{}id <\allowbreak{}hint.\allowbreak{}uid>\allowbreak{} -\allowbreak{}-\allowbreak{}stage2 suggest -\allowbreak{}-\allowbreak{}output <\allowbreak{}new-\allowbreak{}stage1-\allowbreak{}path>\allowbreak{} -\allowbreak{}-\allowbreak{}summary"\allowbreak{},\allowbreak{} "\allowbreak{}stage2"\allowbreak{}:\allowbreak{} "\allowbreak{}Only after reviewing Stage 1,\allowbreak{} optionally run one advertised ready expansion route when a deeper connection could change the conclusion.\allowbreak{}"\allowbreak{}\}\allowbreak{},\allowbreak{} "\allowbreak{}artifact\_\allowbreak{}inspection"\allowbreak{}:\allowbreak{} \{\allowbreak{}"\allowbreak{}default"\allowbreak{}:\allowbreak{} "\allowbreak{}not\_\allowbreak{}needed"\allowbreak{},\allowbreak{} "\allowbreak{}reason"\allowbreak{}:\allowbreak{} "\allowbreak{}This projection already contains provider status,\allowbreak{} selected paths,\allowbreak{} load-\allowbreak{}bearing evidence,\allowbreak{} anchor and relationship pages,\allowbreak{} related-\allowbreak{}page hints,\allowbreak{} conflicts,\allowbreak{} and optional expansion candidates.\allowbreak{}"\allowbreak{},\allowbreak{} "\allowbreak{}exception"\allowbreak{}:\allowbreak{} "\allowbreak{}Run only an item'\allowbreak{}s supplied detail\_\allowbreak{}command,\allowbreak{} once,\allowbreak{} when its truncated\_\allowbreak{}for\_\allowbreak{}summary flag is true and the omitted text is load-\allowbreak{}bearing.\allowbreak{} An absent relationship page is not a missing field and never authorizes artifact or filesystem search.\allowbreak{}"\allowbreak{},\allowbreak{} "\allowbreak{}provider\_\allowbreak{}results\_\allowbreak{}shape"\allowbreak{}:\allowbreak{} "\allowbreak{}coarse.\allowbreak{}provider\_\allowbreak{}results is an object keyed by canonical provider ID,\allowbreak{} not a list.\allowbreak{}"\allowbreak{}\}\allowbreak{},\allowbreak{} "\allowbreak{}query"\allowbreak{}:\allowbreak{} \{\allowbreak{}"\allowbreak{}id"\allowbreak{}:\allowbreak{} "\allowbreak{}query-\allowbreak{}<\allowbreak{}record>\allowbreak{}"\allowbreak{},\allowbreak{} "\allowbreak{}text"\allowbreak{}:\allowbreak{} null,\allowbreak{} "\allowbreak{}entities"\allowbreak{}:\allowbreak{} [\allowbreak{}"\allowbreak{}Dostarlimab"\allowbreak{},\allowbreak{} "\allowbreak{}Stage II Breast Cancer"\allowbreak{}]\allowbreak{},\allowbreak{} "\allowbreak{}input\_\allowbreak{}entities"\allowbreak{}:\allowbreak{} [\allowbreak{}\{\allowbreak{}"\allowbreak{}id"\allowbreak{}:\allowbreak{} "\allowbreak{}Dostarlimab"\allowbreak{},\allowbreak{} "\allowbreak{}name"\allowbreak{}:\allowbreak{} "\allowbreak{}Dostarlimab"\allowbreak{},\allowbreak{} "\allowbreak{}type"\allowbreak{}:\allowbreak{} "\allowbreak{}unknown"\allowbreak{},\allowbreak{} "\allowbreak{}species"\allowbreak{}:\allowbreak{} null,\allowbreak{} "\allowbreak{}aliases"\allowbreak{}:\allowbreak{} [\allowbreak{}]\allowbreak{},\allowbreak{} "\allowbreak{}resolution"\allowbreak{}:\allowbreak{} \{\allowbreak{}"\allowbreak{}source"\allowbreak{}:\allowbreak{} "\allowbreak{}caller"\allowbreak{},\allowbreak{} "\allowbreak{}structured\_\allowbreak{}input"\allowbreak{}:\allowbreak{} true\}\allowbreak{}\}\allowbreak{},\allowbreak{} \{\allowbreak{}"\allowbreak{}id"\allowbreak{}:\allowbreak{} "\allowbreak{}Stage II Breast Cancer"\allowbreak{},\allowbreak{} "\allowbreak{}name"\allowbreak{}:\allowbreak{} "\allowbreak{}Stage II Breast Cancer"\allowbreak{},\allowbreak{} "\allowbreak{}type"\allowbreak{}:\allowbreak{} "\allowbreak{}unknown"\allowbreak{},\allowbreak{} "\allowbreak{}species"\allowbreak{}:\allowbreak{} null,\allowbreak{} "\allowbreak{}aliases"\allowbreak{}:\allowbreak{} [\allowbreak{}]\allowbreak{},\allowbreak{} "\allowbreak{}resolution"\allowbreak{}:\allowbreak{} \{\allowbreak{}"\allowbreak{}source"\allowbreak{}:\allowbreak{} "\allowbreak{}caller"\allowbreak{},\allowbreak{} "\allowbreak{}structured\_\allowbreak{}input"\allowbreak{}:\allowbreak{} true\}\allowbreak{}\}\allowbreak{}]\allowbreak{},\allowbreak{} "\allowbreak{}operation"\allowbreak{}:\allowbreak{} "\allowbreak{}connect"\allowbreak{},\allowbreak{} "\allowbreak{}anchor\_\allowbreak{}quality"\allowbreak{}:\allowbreak{} \{\allowbreak{}"\allowbreak{}trusted"\allowbreak{}:\allowbreak{} false,\allowbreak{} "\allowbreak{}unresolved"\allowbreak{}:\allowbreak{} [\allowbreak{}\{\allowbreak{}"\allowbreak{}id"\allowbreak{}:\allowbreak{} "\allowbreak{}<\allowbreak{}local\_\allowbreak{}id>\allowbreak{}"\allowbreak{},\allowbreak{} "\allowbreak{}name"\allowbreak{}:\allowbreak{} "\allowbreak{}stage ii breast cancer"\allowbreak{},\allowbreak{} "\allowbreak{}type"\allowbreak{}:\allowbreak{} "\allowbreak{}unknown"\allowbreak{},\allowbreak{} "\allowbreak{}species"\allowbreak{}:\allowbreak{} null,\allowbreak{} "\allowbreak{}aliases"\allowbreak{}:\allowbreak{} [\allowbreak{}"\allowbreak{}Stage II Breast Cancer"\allowbreak{}]\allowbreak{},\allowbreak{} "\allowbreak{}resolution"\allowbreak{}:\allowbreak{} \{\allowbreak{}"\allowbreak{}source"\allowbreak{}:\allowbreak{} "\allowbreak{}surface\_\allowbreak{}form"\allowbreak{},\allowbreak{} "\allowbreak{}ambiguous"\allowbreak{}:\allowbreak{} true,\allowbreak{} "\allowbreak{}providers"\allowbreak{}:\allowbreak{} [\allowbreak{}"\allowbreak{}provider\_\allowbreak{}A"\allowbreak{}]\allowbreak{},\allowbreak{} "\allowbreak{}provider\_\allowbreak{}ids"\allowbreak{}:\allowbreak{} \{\allowbreak{}"\allowbreak{}provider\_\allowbreak{}A"\allowbreak{}:\allowbreak{} "\allowbreak{}stage ii breast cancer"\allowbreak{}\}\allowbreak{},\allowbreak{} "\allowbreak{}is\_\allowbreak{}anchor"\allowbreak{}:\allowbreak{} true,\allowbreak{} "\allowbreak{}input\_\allowbreak{}aliases"\allowbreak{}:\allowbreak{} [\allowbreak{}"\allowbreak{}Stage II Breast Cancer"\allowbreak{}]\allowbreak{}\}\allowbreak{}\}\allowbreak{}]\allowbreak{}\}\allowbreak{}\}\allowbreak{},\allowbreak{} "\allowbreak{}resolved\_\allowbreak{}entities"\allowbreak{}:\allowbreak{} [\allowbreak{}\{\allowbreak{}"\allowbreak{}id"\allowbreak{}:\allowbreak{} "\allowbreak{}<\allowbreak{}record\_\allowbreak{}1>\allowbreak{}"\allowbreak{},\allowbreak{} "\allowbreak{}name"\allowbreak{}:\allowbreak{} "\allowbreak{}DOSTARLIMAB"\allowbreak{},\allowbreak{} "\allowbreak{}type"\allowbreak{}:\allowbreak{} "\allowbreak{}drug"\allowbreak{},\allowbreak{} "\allowbreak{}species"\allowbreak{}:\allowbreak{} null,\allowbreak{} "\allowbreak{}aliases"\allowbreak{}:\allowbreak{} [\allowbreak{}"\allowbreak{}dostarlimab"\allowbreak{},\allowbreak{} "\allowbreak{}Dostarlimab"\allowbreak{}]\allowbreak{},\allowbreak{} "\allowbreak{}resolution"\allowbreak{}:\allowbreak{} \{\allowbreak{}"\allowbreak{}source"\allowbreak{}:\allowbreak{} "\allowbreak{}provider\_\allowbreak{}canonical\_\allowbreak{}id"\allowbreak{},\allowbreak{} "\allowbreak{}ambiguous"\allowbreak{}:\allowbreak{} false,\allowbreak{} "\allowbreak{}providers"\allowbreak{}:\allowbreak{} [\allowbreak{}"\allowbreak{}provider\_\allowbreak{}A"\allowbreak{},\allowbreak{} "\allowbreak{}provider\_\allowbreak{}B"\allowbreak{}]\allowbreak{},\allowbreak{} "\allowbreak{}provider\_\allowbreak{}ids"\allowbreak{}:\allowbreak{} \{\allowbreak{}"\allowbreak{}provider\_\allowbreak{}A"\allowbreak{}:\allowbreak{} "\allowbreak{}DOSTARLIMAB"\allowbreak{},\allowbreak{} "\allowbreak{}provider\_\allowbreak{}B"\allowbreak{}:\allowbreak{} "\allowbreak{}<\allowbreak{}record\_\allowbreak{}1>\allowbreak{}"\allowbreak{}\}\allowbreak{},\allowbreak{} "\allowbreak{}is\_\allowbreak{}anchor"\allowbreak{}:\allowbreak{} true,\allowbreak{} "\allowbreak{}previous\_\allowbreak{}ids"\allowbreak{}:\allowbreak{} [\allowbreak{}"\allowbreak{}<\allowbreak{}local\_\allowbreak{}id>\allowbreak{}"\allowbreak{}]\allowbreak{},\allowbreak{} "\allowbreak{}input\_\allowbreak{}aliases"\allowbreak{}:\allowbreak{} [\allowbreak{}"\allowbreak{}Dostarlimab"\allowbreak{}]\allowbreak{}\}\allowbreak{}\}\allowbreak{},\allowbreak{} \{\allowbreak{}"\allowbreak{}id"\allowbreak{}:\allowbreak{} "\allowbreak{}<\allowbreak{}local\_\allowbreak{}id>\allowbreak{}"\allowbreak{},\allowbreak{} "\allowbreak{}name"\allowbreak{}:\allowbreak{} "\allowbreak{}stage ii breast cancer"\allowbreak{},\allowbreak{} "\allowbreak{}type"\allowbreak{}:\allowbreak{} "\allowbreak{}unknown"\allowbreak{},\allowbreak{} "\allowbreak{}species"\allowbreak{}:\allowbreak{} null,\allowbreak{} "\allowbreak{}aliases"\allowbreak{}:\allowbreak{} [\allowbreak{}"\allowbreak{}Stage II Breast Cancer"\allowbreak{}]\allowbreak{},\allowbreak{} "\allowbreak{}resolution"\allowbreak{}:\allowbreak{} \{\allowbreak{}"\allowbreak{}source"\allowbreak{}:\allowbreak{} "\allowbreak{}surface\_\allowbreak{}form"\allowbreak{},\allowbreak{} "\allowbreak{}ambiguous"\allowbreak{}:\allowbreak{} true,\allowbreak{} "\allowbreak{}providers"\allowbreak{}:\allowbreak{} [\allowbreak{}"\allowbreak{}provider\_\allowbreak{}A"\allowbreak{}]\allowbreak{},\allowbreak{} "\allowbreak{}provider\_\allowbreak{}ids"\allowbreak{}:\allowbreak{} \{\allowbreak{}"\allowbreak{}provider\_\allowbreak{}A"\allowbreak{}:\allowbreak{} "\allowbreak{}stage ii breast cancer"\allowbreak{}\}\allowbreak{},\allowbreak{} "\allowbreak{}is\_\allowbreak{}anchor"\allowbreak{}:\allowbreak{} true,\allowbreak{} "\allowbreak{}input\_\allowbreak{}aliases"\allowbreak{}:\allowbreak{} [\allowbreak{}"\allowbreak{}Stage II Breast Cancer"\allowbreak{}]\allowbreak{}\}\allowbreak{}\}\allowbreak{}]\allowbreak{},\allowbreak{} "\allowbreak{}providers\_\allowbreak{}run"\allowbreak{}:\allowbreak{} [\allowbreak{}\{\allowbreak{}"\allowbreak{}provider\_\allowbreak{}id"\allowbreak{}:\allowbreak{} "\allowbreak{}provider\_\allowbreak{}A"\allowbreak{},\allowbreak{} "\allowbreak{}status"\allowbreak{}:\allowbreak{} "\allowbreak{}partial"\allowbreak{},\allowbreak{} "\allowbreak{}algorithm\_\allowbreak{}status"\allowbreak{}:\allowbreak{} "\allowbreak{}partial"\allowbreak{},\allowbreak{} "\allowbreak{}algorithm\_\allowbreak{}id"\allowbreak{}:\allowbreak{} "\allowbreak{}weighted\_\allowbreak{}best\_\allowbreak{}first"\allowbreak{},\allowbreak{} "\allowbreak{}elapsed\_\allowbreak{}ms"\allowbreak{}:\allowbreak{} 71465.\allowbreak{}406,\allowbreak{} "\allowbreak{}timed\_\allowbreak{}out"\allowbreak{}:\allowbreak{} false,\allowbreak{} "\allowbreak{}budget\_\allowbreak{}exhausted"\allowbreak{}:\allowbreak{} true,\allowbreak{} "\allowbreak{}expansions"\allowbreak{}:\allowbreak{} 150,\allowbreak{} "\allowbreak{}frontier\_\allowbreak{}remaining"\allowbreak{}:\allowbreak{} 3197,\allowbreak{} "\allowbreak{}early\_\allowbreak{}stopped\_\allowbreak{}on\_\allowbreak{}connection"\allowbreak{}:\allowbreak{} false,\allowbreak{} "\allowbreak{}effective\_\allowbreak{}limits"\allowbreak{}:\allowbreak{} \{\allowbreak{}"\allowbreak{}max\_\allowbreak{}expansions"\allowbreak{}:\allowbreak{} 150,\allowbreak{} "\allowbreak{}neighbor\_\allowbreak{}limit"\allowbreak{}:\allowbreak{} 5,\allowbreak{} "\allowbreak{}max\_\allowbreak{}depth"\allowbreak{}:\allowbreak{} 5,\allowbreak{} "\allowbreak{}max\_\allowbreak{}hops"\allowbreak{}:\allowbreak{} 7,\allowbreak{} "\allowbreak{}top\_\allowbreak{}k"\allowbreak{}:\allowbreak{} 5,\allowbreak{} "\allowbreak{}timeout\_\allowbreak{}seconds"\allowbreak{}:\allowbreak{} 240\}\allowbreak{},\allowbreak{} "\allowbreak{}warnings"\allowbreak{}:\allowbreak{} [\allowbreak{}"\allowbreak{}algorithm expansion budget exhausted after 150 frontier expansions;\allowbreak{} unvisited frontier remains"\allowbreak{}]\allowbreak{},\allowbreak{} "\allowbreak{}error"\allowbreak{}:\allowbreak{} null\}\allowbreak{},\allowbreak{} \{\allowbreak{}"\allowbreak{}provider\_\allowbreak{}id"\allowbreak{}:\allowbreak{} "\allowbreak{}provider\_\allowbreak{}B"\allowbreak{},\allowbreak{} "\allowbreak{}status"\allowbreak{}:\allowbreak{} "\allowbreak{}partial"\allowbreak{},\allowbreak{} "\allowbreak{}algorithm\_\allowbreak{}status"\allowbreak{}:\allowbreak{} "\allowbreak{}partial"\allowbreak{},\allowbreak{} "\allowbreak{}algorithm\_\allowbreak{}id"\allowbreak{}:\allowbreak{} "\allowbreak{}weighted\_\allowbreak{}best\_\allowbreak{}first"\allowbreak{},\allowbreak{} "\allowbreak{}elapsed\_\allowbreak{}ms"\allowbreak{}:\allowbreak{} 11516.\allowbreak{}748,\allowbreak{} "\allowbreak{}timed\_\allowbreak{}out"\allowbreak{}:\allowbreak{} false,\allowbreak{} "\allowbreak{}budget\_\allowbreak{}exhausted"\allowbreak{}:\allowbreak{} true,\allowbreak{} "\allowbreak{}expansions"\allowbreak{}:\allowbreak{} 150,\allowbreak{} "\allowbreak{}frontier\_\allowbreak{}remaining"\allowbreak{}:\allowbreak{} 739,\allowbreak{} "\allowbreak{}early\_\allowbreak{}stopped\_\allowbreak{}on\_\allowbreak{}connection"\allowbreak{}:\allowbreak{} false,\allowbreak{} "\allowbreak{}effective\_\allowbreak{}limits"\allowbreak{}:\allowbreak{} \{\allowbreak{}"\allowbreak{}max\_\allowbreak{}expansions"\allowbreak{}:\allowbreak{} 150,\allowbreak{} "\allowbreak{}neighbor\_\allowbreak{}limit"\allowbreak{}:\allowbreak{} 5,\allowbreak{} "\allowbreak{}max\_\allowbreak{}depth"\allowbreak{}:\allowbreak{} 5,\allowbreak{} "\allowbreak{}max\_\allowbreak{}hops"\allowbreak{}:\allowbreak{} 7,\allowbreak{} "\allowbreak{}top\_\allowbreak{}k"\allowbreak{}:\allowbreak{} 5,\allowbreak{} "\allowbreak{}timeout\_\allowbreak{}seconds"\allowbreak{}:\allowbreak{} 240\}\allowbreak{},\allowbreak{} "\allowbreak{}warnings"\allowbreak{}:\allowbreak{} [\allowbreak{}"\allowbreak{}algorithm expansion budget exhausted after 150 frontier expansions;\allowbreak{} unvisited frontier remains"\allowbreak{}]\allowbreak{},\allowbreak{} "\allowbreak{}error"\allowbreak{}:\allowbreak{} null\}\allowbreak{}]\allowbreak{},\allowbreak{} "\allowbreak{}projection\_\allowbreak{}counts"\allowbreak{}:\allowbreak{} \{\allowbreak{}"\allowbreak{}candidate\_\allowbreak{}paths"\allowbreak{}:\allowbreak{} \{\allowbreak{}"\allowbreak{}shown"\allowbreak{}:\allowbreak{} 0,\allowbreak{} "\allowbreak{}total"\allowbreak{}:\allowbreak{} 0\}\allowbreak{},\allowbreak{} "\allowbreak{}connections"\allowbreak{}:\allowbreak{} \{\allowbreak{}"\allowbreak{}shown"\allowbreak{}:\allowbreak{} 0,\allowbreak{} "\allowbreak{}total"\allowbreak{}:\allowbreak{} 0\}\allowbreak{},\allowbreak{} "\allowbreak{}load\_\allowbreak{}bearing\_\allowbreak{}evidence"\allowbreak{}:\allowbreak{} \{\allowbreak{}"\allowbreak{}shown"\allowbreak{}:\allowbreak{} 0,\allowbreak{} "\allowbreak{}total\_\allowbreak{}selected\_\allowbreak{}edge\_\allowbreak{}ids"\allowbreak{}:\allowbreak{} 0\}\allowbreak{},\allowbreak{} "\allowbreak{}relationship\_\allowbreak{}pages"\allowbreak{}:\allowbreak{} \{\allowbreak{}"\allowbreak{}shown"\allowbreak{}:\allowbreak{} 0,\allowbreak{} "\allowbreak{}total"\allowbreak{}:\allowbreak{} 0,\allowbreak{} "\allowbreak{}direct\_\allowbreak{}for\_\allowbreak{}anchor\_\allowbreak{}pair\_\allowbreak{}available"\allowbreak{}:\allowbreak{} false\}\allowbreak{},\allowbreak{} "\allowbreak{}anchor\_\allowbreak{}pages"\allowbreak{}:\allowbreak{} \{\allowbreak{}"\allowbreak{}shown"\allowbreak{}:\allowbreak{} 1,\allowbreak{} "\allowbreak{}resolved\_\allowbreak{}anchors"\allowbreak{}:\allowbreak{} 1\}\allowbreak{},\allowbreak{} "\allowbreak{}related\_\allowbreak{}page\_\allowbreak{}hints"\allowbreak{}:\allowbreak{} \{\allowbreak{}"\allowbreak{}shown"\allowbreak{}:\allowbreak{} 7,\allowbreak{} "\allowbreak{}total"\allowbreak{}:\allowbreak{} 21\}\allowbreak{},\allowbreak{} "\allowbreak{}expansion\_\allowbreak{}candidates"\allowbreak{}:\allowbreak{} \{\allowbreak{}"\allowbreak{}shown"\allowbreak{}:\allowbreak{} 0,\allowbreak{} "\allowbreak{}total"\allowbreak{}:\allowbreak{} 0\}\allowbreak{}\}\allowbreak{},\allowbreak{} "\allowbreak{}candidate\_\allowbreak{}paths"\allowbreak{}:\allowbreak{} [\allowbreak{}]\allowbreak{},\allowbreak{} "\allowbreak{}expansion\_\allowbreak{}candidates"\allowbreak{}:\allowbreak{} [\allowbreak{}]\allowbreak{},\allowbreak{} "\allowbreak{}connections"\allowbreak{}:\allowbreak{} [\allowbreak{}]\allowbreak{},\allowbreak{} "\allowbreak{}load\_\allowbreak{}bearing\_\allowbreak{}evidence"\allowbreak{}:\allowbreak{} [\allowbreak{}]\allowbreak{},\allowbreak{} "\allowbreak{}anchor\_\allowbreak{}pages"\allowbreak{}:\allowbreak{} [\allowbreak{}\{\allowbreak{}"\allowbreak{}uid"\allowbreak{}:\allowbreak{} "\allowbreak{}<\allowbreak{}record\_\allowbreak{}1>\allowbreak{}"\allowbreak{},\allowbreak{} "\allowbreak{}title"\allowbreak{}:\allowbreak{} "\allowbreak{}dostarlimab"\allowbreak{},\allowbreak{} "\allowbreak{}type"\allowbreak{}:\allowbreak{} "\allowbreak{}drug"\allowbreak{},\allowbreak{} "\allowbreak{}path"\allowbreak{}:\allowbreak{} "\allowbreak{}drugs/\allowbreak{}dostarlimab.\allowbreak{}md"\allowbreak{},\allowbreak{} "\allowbreak{}n\_\allowbreak{}papers"\allowbreak{}:\allowbreak{} 238,\allowbreak{} "\allowbreak{}pmids"\allowbreak{}:\allowbreak{} [\allowbreak{}"\allowbreak{}39282229"\allowbreak{},\allowbreak{} "\allowbreak{}41904276"\allowbreak{},\allowbreak{} "\allowbreak{}35892341"\allowbreak{},\allowbreak{} "\allowbreak{}36005013"\allowbreak{},\allowbreak{} "\allowbreak{}0"\allowbreak{},\allowbreak{} "\allowbreak{}41811823"\allowbreak{},\allowbreak{} "\allowbreak{}36762991"\allowbreak{},\allowbreak{} "\allowbreak{}38365286"\allowbreak{},\allowbreak{} "\allowbreak{}38605944"\allowbreak{}]\allowbreak{},\allowbreak{} "\allowbreak{}text"\allowbreak{}:\allowbreak{} "\allowbreak{}-\allowbreak{}-\allowbreak{}-\allowbreak{} uid:\allowbreak{} <\allowbreak{}record\_\allowbreak{}1>\allowbreak{} type:\allowbreak{} drug name:\allowbreak{} "\allowbreak{}dostarlimab"\allowbreak{} aliases:\allowbreak{} [\allowbreak{}"\allowbreak{}Dostarlimab"\allowbreak{},\allowbreak{} "\allowbreak{}Dostarlimab (\allowbreak{}Jemperli)\allowbreak{}"\allowbreak{}]\allowbreak{} n\_\allowbreak{}papers:\allowbreak{} 238 n\_\allowbreak{}claims:\allowbreak{} 388 n\_\allowbreak{}claims\_\allowbreak{}used:\allowbreak{} 13 grounded:\allowbreak{} false synth\_\allowbreak{}model:\allowbreak{} "\allowbreak{}glm-\allowbreak{}5.\allowbreak{}2"\allowbreak{} prompt\_\allowbreak{}template\_\allowbreak{}version:\allowbreak{} "\allowbreak{}entity\_\allowbreak{}page\_\allowbreak{}v1"\allowbreak{} updated:\allowbreak{} 2026-\allowbreak{}07-\allowbreak{}06 tags:\allowbreak{} [\allowbreak{}"\allowbreak{}drug"\allowbreak{}]\allowbreak{}\par
\addvspace{3pt}
\addvspace{3pt}
\par\addvspace{5pt}\noindent{\normalsize\bfseries\sffamily dostarlimab}\par\addvspace{2pt}
\addvspace{3pt}
\par\addvspace{5pt}\noindent{\small\bfseries\sffamily Overview}\par\addvspace{2pt}
\noindent Dostarlimab is an FDA-\allowbreak{}approved anti-\allowbreak{}PD-\allowbreak{}1/\allowbreak{}PD-\allowbreak{}L1 antibody widely used in cancer treatment [\allowbreak{}ref-\allowbreak{}9]\allowbreak{},\allowbreak{} [\allowbreak{}ref-\allowbreak{}11]\allowbreak{}.\allowbreak{} It is an IgG4 isotype designed to avoid tumor-\allowbreak{}reactive T cell depletion,\allowbreak{} with little to no binding to Fc or complement protein C1q [\allowbreak{}ref-\allowbreak{}7]\allowbreak{}.\allowbreak{} Its pharmacokinetic profile is characterized by a 2-\allowbreak{}compartment model with time-\allowbreak{}dependent linear elimination,\allowbreak{} a mean terminal elimination half-\allowbreak{}life of 25.\allowbreak{}4 days,\allowbreak{} a mean clearance of 0.\allowbreak{}007 L/\allowbreak{}h,\allowbreak{} and a steady state volume of distribution of 5.\allowbreak{}3 L [\allowbreak{}ref-\allowbreak{}1]\allowbreak{},\allowbreak{} [\allowbreak{}ref-\allowbreak{}2]\allowbreak{}.\allowbreak{} As a tumor-\allowbreak{}agnostic therapy based on genetic changes,\allowbreak{} dostarlimab is highly relevant for drug repurposing across multiple histologies [\allowbreak{}ref-\allowbreak{}10]\allowbreak{}.\allowbreak{}\par
\addvspace{3pt}
\par\addvspace{5pt}\noindent{\small\bfseries\sffamily Repurposing directions}\par\addvspace{2pt}
\addvspace{3pt}
\par\addvspace{5pt}\noindent{\footnotesize\bfseries\sffamily Strongly supported directions}\par\addvspace{2pt}
\noindent Dostarlimab is strongly supported as a treatment for [\allowbreak{}[\allowbreak{}dostarlimab $\rightarrow$ cancer]\allowbreak{}]\allowbreak{},\allowbreak{} with high-\allowbreak{}quality clinical trial and review evidence plus in vivo animal data indicating beneficial"\allowbreak{},\allowbreak{} "\allowbreak{}truncated\_\allowbreak{}for\_\allowbreak{}summary"\allowbreak{}:\allowbreak{} true,\allowbreak{} "\allowbreak{}full\_\allowbreak{}text\_\allowbreak{}artifact\_\allowbreak{}json\_\allowbreak{}pointer"\allowbreak{}:\allowbreak{} "\allowbreak{}/\allowbreak{}coarse/\allowbreak{}provider\_\allowbreak{}results/\allowbreak{}provider\_\allowbreak{}B/\allowbreak{}pages/\allowbreak{}0/\allowbreak{}text"\allowbreak{},\allowbreak{} "\allowbreak{}detail\_\allowbreak{}command"\allowbreak{}:\allowbreak{} "\allowbreak{}tool\_\allowbreak{}A inspect <\allowbreak{}artifact\_\allowbreak{}1>\allowbreak{} /\allowbreak{}coarse/\allowbreak{}provider\_\allowbreak{}results/\allowbreak{}provider\_\allowbreak{}B/\allowbreak{}pages/\allowbreak{}0/\allowbreak{}text"\allowbreak{}\}\allowbreak{}]\allowbreak{},\allowbreak{} "\allowbreak{}relationship\_\allowbreak{}pages"\allowbreak{}:\allowbreak{} [\allowbreak{}]\allowbreak{},\allowbreak{} "\allowbreak{}related\_\allowbreak{}page\_\allowbreak{}hints"\allowbreak{}:\allowbreak{} [\allowbreak{}\{\allowbreak{}"\allowbreak{}uid"\allowbreak{}:\allowbreak{} "\allowbreak{}<\allowbreak{}record>\allowbreak{}"\allowbreak{},\allowbreak{} "\allowbreak{}title"\allowbreak{}:\allowbreak{} "\allowbreak{}dostarlimab $\rightarrow$ endometrial cancer"\allowbreak{},\allowbreak{} "\allowbreak{}path"\allowbreak{}:\allowbreak{} "\allowbreak{}hypotheses/\allowbreak{}dostarlimab $\rightarrow$ endometrial cancer.\allowbreak{}md"\allowbreak{},\allowbreak{} "\allowbreak{}n\_\allowbreak{}papers"\allowbreak{}:\allowbreak{} 29,\allowbreak{} "\allowbreak{}type"\allowbreak{}:\allowbreak{} "\allowbreak{}hypothesis"\allowbreak{},\allowbreak{} "\allowbreak{}follow\_\allowbreak{}up\_\allowbreak{}stage"\allowbreak{}:\allowbreak{} "\allowbreak{}stage1"\allowbreak{}\}\allowbreak{},\allowbreak{} \{\allowbreak{}"\allowbreak{}uid"\allowbreak{}:\allowbreak{} "\allowbreak{}<\allowbreak{}record\_\allowbreak{}2>\allowbreak{}"\allowbreak{},\allowbreak{} "\allowbreak{}title"\allowbreak{}:\allowbreak{} "\allowbreak{}PD-\allowbreak{}1"\allowbreak{},\allowbreak{} "\allowbreak{}path"\allowbreak{}:\allowbreak{} "\allowbreak{}targets/\allowbreak{}PD-\allowbreak{}1.\allowbreak{}md"\allowbreak{},\allowbreak{} "\allowbreak{}n\_\allowbreak{}papers"\allowbreak{}:\allowbreak{} 12098,\allowbreak{} "\allowbreak{}type"\allowbreak{}:\allowbreak{} "\allowbreak{}target"\allowbreak{},\allowbreak{} "\allowbreak{}follow\_\allowbreak{}up\_\allowbreak{}stage"\allowbreak{}:\allowbreak{} "\allowbreak{}stage1"\allowbreak{}\}\allowbreak{},\allowbreak{} \{\allowbreak{}"\allowbreak{}uid"\allowbreak{}:\allowbreak{} "\allowbreak{}<\allowbreak{}record\_\allowbreak{}1>\allowbreak{}"\allowbreak{},\allowbreak{} "\allowbreak{}title"\allowbreak{}:\allowbreak{} "\allowbreak{}FC"\allowbreak{},\allowbreak{} "\allowbreak{}path"\allowbreak{}:\allowbreak{} "\allowbreak{}drugs/\allowbreak{}FC.\allowbreak{}md"\allowbreak{},\allowbreak{} "\allowbreak{}n\_\allowbreak{}papers"\allowbreak{}:\allowbreak{} 20,\allowbreak{} "\allowbreak{}type"\allowbreak{}:\allowbreak{} "\allowbreak{}drug"\allowbreak{},\allowbreak{} "\allowbreak{}follow\_\allowbreak{}up\_\allowbreak{}stage"\allowbreak{}:\allowbreak{} "\allowbreak{}stage1"\allowbreak{}\}\allowbreak{},\allowbreak{} \{\allowbreak{}"\allowbreak{}uid"\allowbreak{}:\allowbreak{} "\allowbreak{}<\allowbreak{}record>\allowbreak{}"\allowbreak{},\allowbreak{} "\allowbreak{}title"\allowbreak{}:\allowbreak{} "\allowbreak{}dostarlimab $\rightarrow$ locally advanced rectal cancer"\allowbreak{},\allowbreak{} "\allowbreak{}path"\allowbreak{}:\allowbreak{} "\allowbreak{}hypotheses/\allowbreak{}dostarlimab $\rightarrow$ locally advanced rectal cancer.\allowbreak{}md"\allowbreak{},\allowbreak{} "\allowbreak{}n\_\allowbreak{}papers"\allowbreak{}:\allowbreak{} 12,\allowbreak{} "\allowbreak{}type"\allowbreak{}:\allowbreak{} "\allowbreak{}hypothesis"\allowbreak{},\allowbreak{} "\allowbreak{}follow\_\allowbreak{}up\_\allowbreak{}stage"\allowbreak{}:\allowbreak{} "\allowbreak{}stage1"\allowbreak{}\}\allowbreak{},\allowbreak{} \{\allowbreak{}"\allowbreak{}uid"\allowbreak{}:\allowbreak{} "\allowbreak{}<\allowbreak{}record\_\allowbreak{}2>\allowbreak{}"\allowbreak{},\allowbreak{} "\allowbreak{}title"\allowbreak{}:\allowbreak{} "\allowbreak{}PD-\allowbreak{}L1"\allowbreak{},\allowbreak{} "\allowbreak{}path"\allowbreak{}:\allowbreak{} "\allowbreak{}targets/\allowbreak{}PD-\allowbreak{}L1.\allowbreak{}md"\allowbreak{},\allowbreak{} "\allowbreak{}n\_\allowbreak{}papers"\allowbreak{}:\allowbreak{} 10947,\allowbreak{} "\allowbreak{}type"\allowbreak{}:\allowbreak{} "\allowbreak{}target"\allowbreak{},\allowbreak{} "\allowbreak{}follow\_\allowbreak{}up\_\allowbreak{}stage"\allowbreak{}:\allowbreak{} "\allowbreak{}stage1"\allowbreak{}\}\allowbreak{},\allowbreak{} \{\allowbreak{}"\allowbreak{}uid"\allowbreak{}:\allowbreak{} "\allowbreak{}<\allowbreak{}record>\allowbreak{}"\allowbreak{},\allowbreak{} "\allowbreak{}title"\allowbreak{}:\allowbreak{} "\allowbreak{}dostarlimab $\rightarrow$ rectal cancer"\allowbreak{},\allowbreak{} "\allowbreak{}path"\allowbreak{}:\allowbreak{} "\allowbreak{}hypotheses/\allowbreak{}dostarlimab $\rightarrow$ rectal cancer.\allowbreak{}md"\allowbreak{},\allowbreak{} "\allowbreak{}n\_\allowbreak{}papers"\allowbreak{}:\allowbreak{} 11,\allowbreak{} "\allowbreak{}type"\allowbreak{}:\allowbreak{} "\allowbreak{}hypothesis"\allowbreak{},\allowbreak{} "\allowbreak{}follow\_\allowbreak{}up\_\allowbreak{}stage"\allowbreak{}:\allowbreak{} "\allowbreak{}stage1"\allowbreak{}\}\allowbreak{},\allowbreak{} \{\allowbreak{}"\allowbreak{}uid"\allowbreak{}:\allowbreak{} "\allowbreak{}<\allowbreak{}record\_\allowbreak{}2>\allowbreak{}"\allowbreak{},\allowbreak{} "\allowbreak{}title"\allowbreak{}:\allowbreak{} "\allowbreak{}programmed cell death protein 1 (\allowbreak{}PD-\allowbreak{}1)\allowbreak{}"\allowbreak{},\allowbreak{} "\allowbreak{}path"\allowbreak{}:\allowbreak{} "\allowbreak{}targets/\allowbreak{}programmed cell death protein 1 (\allowbreak{}PD-\allowbreak{}1)\allowbreak{}.\allowbreak{}md"\allowbreak{},\allowbreak{} "\allowbreak{}n\_\allowbreak{}papers"\allowbreak{}:\allowbreak{} 195,\allowbreak{} "\allowbreak{}type"\allowbreak{}:\allowbreak{} "\allowbreak{}target"\allowbreak{},\allowbreak{} "\allowbreak{}follow\_\allowbreak{}up\_\allowbreak{}stage"\allowbreak{}:\allowbreak{} "\allowbreak{}stage1"\allowbreak{}\}\allowbreak{}]\allowbreak{},\allowbreak{} "\allowbreak{}stage2"\allowbreak{}:\allowbreak{} \{\allowbreak{}"\allowbreak{}policy"\allowbreak{}:\allowbreak{} "\allowbreak{}suggest"\allowbreak{},\allowbreak{} "\allowbreak{}optional"\allowbreak{}:\allowbreak{} true,\allowbreak{} "\allowbreak{}suggested\_\allowbreak{}expansions"\allowbreak{}:\allowbreak{} 0,\allowbreak{} "\allowbreak{}requested\_\allowbreak{}routes"\allowbreak{}:\allowbreak{} 0,\allowbreak{} "\allowbreak{}guidance"\allowbreak{}:\allowbreak{} "\allowbreak{}Stage 1 is complete.\allowbreak{} The calling agent may answer now or request one or more compatible expansion routes only when a load-\allowbreak{}bearing weak,\allowbreak{} missing,\allowbreak{} or conflicting link could change the conclusion.\allowbreak{}"\allowbreak{}\}\allowbreak{},\allowbreak{} "\allowbreak{}report"\allowbreak{}:\allowbreak{} \{\allowbreak{}"\allowbreak{}summary"\allowbreak{}:\allowbreak{} "\allowbreak{}No complete connection path was reconstructed for Dostarlimab $\leftrightarrow$ Stage II Breast Cancer.\allowbreak{} Provider results and unresolved gaps are retained for diagnosis.\allowbreak{}"\allowbreak{},\allowbreak{} "\allowbreak{}support\_\allowbreak{}grade"\allowbreak{}:\allowbreak{} "\allowbreak{}incomplete"\allowbreak{},\allowbreak{} "\allowbreak{}conflicts"\allowbreak{}:\allowbreak{} [\allowbreak{}]\allowbreak{},\allowbreak{} "\allowbreak{}conflict\_\allowbreak{}count"\allowbreak{}:\allowbreak{} 0,\allowbreak{} "\allowbreak{}unresolved\_\allowbreak{}gaps"\allowbreak{}:\allowbreak{} [\allowbreak{}\{\allowbreak{}"\allowbreak{}gap\_\allowbreak{}id"\allowbreak{}:\allowbreak{} null,\allowbreak{} "\allowbreak{}candidate\_\allowbreak{}path\_\allowbreak{}id"\allowbreak{}:\allowbreak{} null,\allowbreak{} "\allowbreak{}subject"\allowbreak{}:\allowbreak{} "\allowbreak{}<\allowbreak{}local\_\allowbreak{}id>\allowbreak{}"\allowbreak{},\allowbreak{} "\allowbreak{}object"\allowbreak{}:\allowbreak{} null,\allowbreak{} "\allowbreak{}required\_\allowbreak{}provider"\allowbreak{}:\allowbreak{} null,\allowbreak{} "\allowbreak{}reason"\allowbreak{}:\allowbreak{} "\allowbreak{}unresolved\_\allowbreak{}or\_\allowbreak{}local\_\allowbreak{}query\_\allowbreak{}anchor"\allowbreak{}\}\allowbreak{}]\allowbreak{},\allowbreak{} "\allowbreak{}unresolved\_\allowbreak{}gap\_\allowbreak{}count"\allowbreak{}:\allowbreak{} 1,\allowbreak{} "\allowbreak{}optional\_\allowbreak{}expansions"\allowbreak{}:\allowbreak{} \{\allowbreak{}"\allowbreak{}policy"\allowbreak{}:\allowbreak{} "\allowbreak{}suggest"\allowbreak{},\allowbreak{} "\allowbreak{}suggested"\allowbreak{}:\allowbreak{} 0,\allowbreak{} "\allowbreak{}requested"\allowbreak{}:\allowbreak{} 0\}\allowbreak{},\allowbreak{} "\allowbreak{}diagnostics"\allowbreak{}:\allowbreak{} \{\allowbreak{}"\allowbreak{}non\_\allowbreak{}load\_\allowbreak{}bearing\_\allowbreak{}conflict\_\allowbreak{}count"\allowbreak{}:\allowbreak{} 245\}\allowbreak{},\allowbreak{} "\allowbreak{}interpretation"\allowbreak{}:\allowbreak{} "\allowbreak{}Provider-\allowbreak{}native scores are shown separately and were not averaged.\allowbreak{} A support grade describes evidence structure,\allowbreak{} not treatment efficacy.\allowbreak{}"\allowbreak{}\}\allowbreak{},\allowbreak{} "\allowbreak{}graph\_\allowbreak{}warnings"\allowbreak{}:\allowbreak{} [\allowbreak{}]\allowbreak{}\}\allowbreak{}\}\allowbreak{}\par
\addvspace{3pt}
\noindent \{\allowbreak{} "\allowbreak{}stderr"\allowbreak{}:\allowbreak{} "\allowbreak{}"\allowbreak{} \}\allowbreak{}\par}
\end{stepT}
\begin{stepR}{REASONING \textbullet\ STEP 9 \textbullet\ Read the negative result; note the unresolved disease anchor}
\stepsub{REASONING SUMMARY}\noindent{\footnotesize \textbf{Deciding truncated load-\allowbreak{}bearing detail inclusion}}\par\stepsub{MODEL NARRATION}\noindent{\footnotesize\itshape Stage-\allowbreak{}1 found no direct reconstructed connection,\allowbreak{} but the drug anchor page is truncated and is the only available support for mechanism/\allowbreak{}status,\allowbreak{} so I'\allowbreak{}m using the one permitted detail lookup for that page.\allowbreak{}}\par
\end{stepR}
\begin{stepT}{TOOL CALL \textbullet\ STEP 10 \textbullet\ Field-read the retrieved drug page}
\stepsub{CALL}\begin{payload}tool\_\allowbreak{}A inspect <\allowbreak{}artifact\_\allowbreak{}1>\allowbreak{} /\allowbreak{}coarse/\allowbreak{}provider\_\allowbreak{}results/\allowbreak{}provider\_\allowbreak{}B/\allowbreak{}pages/\allowbreak{}0/\allowbreak{}text\end{payload}\stepsub{RETURNED}{\footnotesize \noindent exit\_\allowbreak{}code:\allowbreak{} 0  tool:\allowbreak{} tool\_\allowbreak{}A  contract\_\allowbreak{}version:\allowbreak{} the domain toolkit-\allowbreak{}tools/\allowbreak{}v1 stdout:\allowbreak{} \{\allowbreak{}"\allowbreak{}schema\_\allowbreak{}version"\allowbreak{}:\allowbreak{} "\allowbreak{}tool\_\allowbreak{}A/\allowbreak{}field-\allowbreak{}v1"\allowbreak{},\allowbreak{} "\allowbreak{}artifact\_\allowbreak{}path"\allowbreak{}:\allowbreak{} "\allowbreak{}<\allowbreak{}artifact\_\allowbreak{}1>\allowbreak{}"\allowbreak{},\allowbreak{} "\allowbreak{}json\_\allowbreak{}pointer"\allowbreak{}:\allowbreak{} "\allowbreak{}/\allowbreak{}coarse/\allowbreak{}provider\_\allowbreak{}results/\allowbreak{}provider\_\allowbreak{}B/\allowbreak{}pages/\allowbreak{}0/\allowbreak{}text"\allowbreak{},\allowbreak{} "\allowbreak{}value\_\allowbreak{}kind"\allowbreak{}:\allowbreak{} "\allowbreak{}str"\allowbreak{},\allowbreak{} "\allowbreak{}value"\allowbreak{}:\allowbreak{} "\allowbreak{}-\allowbreak{}-\allowbreak{}-\allowbreak{} uid:\allowbreak{} <\allowbreak{}record\_\allowbreak{}1>\allowbreak{} type:\allowbreak{} drug name:\allowbreak{} "\allowbreak{}dostarlimab"\allowbreak{} aliases:\allowbreak{} [\allowbreak{}"\allowbreak{}Dostarlimab"\allowbreak{},\allowbreak{} "\allowbreak{}Dostarlimab (\allowbreak{}Jemperli)\allowbreak{}"\allowbreak{}]\allowbreak{} n\_\allowbreak{}papers:\allowbreak{} 238 n\_\allowbreak{}claims:\allowbreak{} 388 n\_\allowbreak{}claims\_\allowbreak{}used:\allowbreak{} 13 grounded:\allowbreak{} false synth\_\allowbreak{}model:\allowbreak{} "\allowbreak{}glm-\allowbreak{}5.\allowbreak{}2"\allowbreak{} prompt\_\allowbreak{}template\_\allowbreak{}version:\allowbreak{} "\allowbreak{}entity\_\allowbreak{}page\_\allowbreak{}v1"\allowbreak{} updated:\allowbreak{} 2026-\allowbreak{}07-\allowbreak{}06 tags:\allowbreak{} [\allowbreak{}"\allowbreak{}drug"\allowbreak{}]\allowbreak{}\par
\addvspace{3pt}
\addvspace{3pt}
\par\addvspace{5pt}\noindent{\normalsize\bfseries\sffamily dostarlimab}\par\addvspace{2pt}
\addvspace{3pt}
\par\addvspace{5pt}\noindent{\small\bfseries\sffamily Overview}\par\addvspace{2pt}
\noindent Dostarlimab is an FDA-\allowbreak{}approved anti-\allowbreak{}PD-\allowbreak{}1/\allowbreak{}PD-\allowbreak{}L1 antibody widely used in cancer treatment [\allowbreak{}ref-\allowbreak{}9]\allowbreak{},\allowbreak{} [\allowbreak{}ref-\allowbreak{}11]\allowbreak{}.\allowbreak{} It is an IgG4 isotype designed to avoid tumor-\allowbreak{}reactive T cell depletion,\allowbreak{} with little to no binding to Fc or complement protein C1q [\allowbreak{}ref-\allowbreak{}7]\allowbreak{}.\allowbreak{} Its pharmacokinetic profile is characterized by a 2-\allowbreak{}compartment model with time-\allowbreak{}dependent linear elimination,\allowbreak{} a mean terminal elimination half-\allowbreak{}life of 25.\allowbreak{}4 days,\allowbreak{} a mean clearance of 0.\allowbreak{}007 L/\allowbreak{}h,\allowbreak{} and a steady state volume of distribution of 5.\allowbreak{}3 L [\allowbreak{}ref-\allowbreak{}1]\allowbreak{},\allowbreak{} [\allowbreak{}ref-\allowbreak{}2]\allowbreak{}.\allowbreak{} As a tumor-\allowbreak{}agnostic therapy based on genetic changes,\allowbreak{} dostarlimab is highly relevant for drug repurposing across multiple histologies [\allowbreak{}ref-\allowbreak{}10]\allowbreak{}.\allowbreak{}\par
\addvspace{3pt}
\par\addvspace{5pt}\noindent{\small\bfseries\sffamily Repurposing directions}\par\addvspace{2pt}
\addvspace{3pt}
\par\addvspace{5pt}\noindent{\footnotesize\bfseries\sffamily Strongly supported directions}\par\addvspace{2pt}
\noindent Dostarlimab is strongly supported as a treatment for [\allowbreak{}[\allowbreak{}dostarlimab $\rightarrow$ cancer]\allowbreak{}]\allowbreak{},\allowbreak{} with high-\allowbreak{}quality clinical trial and review evidence plus in vivo animal data indicating beneficial outcomes [\allowbreak{}ref-\allowbreak{}3]\allowbreak{}.\allowbreak{} It is strongly supported for [\allowbreak{}[\allowbreak{}dostarlimab $\rightarrow$ advanced solid tumors]\allowbreak{}]\allowbreak{},\allowbreak{} [\allowbreak{}[\allowbreak{}dostarlimab $\rightarrow$ solid tumors]\allowbreak{}]\allowbreak{},\allowbreak{} [\allowbreak{}[\allowbreak{}dostarlimab $\rightarrow$ dMMR solid tumors]\allowbreak{}]\allowbreak{},\allowbreak{} and [\allowbreak{}[\allowbreak{}dostarlimab $\rightarrow$ dMMR recurrent or advanced solid tumors]\allowbreak{}]\allowbreak{},\allowbreak{} with tissue-\allowbreak{}agnostic FDA approval for mismatch repair-\allowbreak{}deficient tumors [\allowbreak{}ref-\allowbreak{}10]\allowbreak{}.\allowbreak{}\par
\addvspace{3pt}
\noindent For endometrial malignancies,\allowbreak{} the drug is strongly supported across multiple specific presentations:\allowbreak{} [\allowbreak{}[\allowbreak{}dostarlimab $\rightarrow$ endometrial cancer]\allowbreak{}]\allowbreak{},\allowbreak{} [\allowbreak{}[\allowbreak{}dostarlimab $\rightarrow$ advanced endometrial cancer]\allowbreak{}]\allowbreak{},\allowbreak{} [\allowbreak{}[\allowbreak{}dostarlimab $\rightarrow$ advanced or recurrent EC]\allowbreak{}]\allowbreak{},\allowbreak{} [\allowbreak{}[\allowbreak{}dostarlimab $\rightarrow$ advanced EC]\allowbreak{}]\allowbreak{},\allowbreak{} [\allowbreak{}[\allowbreak{}dostarlimab $\rightarrow$ dMMR endometrial cancer]\allowbreak{}]\allowbreak{},\allowbreak{} [\allowbreak{}[\allowbreak{}dostarlimab $\rightarrow$ dMMR recurrent or advanced endometrial cancer]\allowbreak{}]\allowbreak{},\allowbreak{} [\allowbreak{}[\allowbreak{}dostarlimab $\rightarrow$ recurrent or advanced endometrial cancer]\allowbreak{}]\allowbreak{},\allowbreak{} and [\allowbreak{}[\allowbreak{}dostarlimab $\rightarrow$ primary advanced or recurrent endometrial cancer]\allowbreak{}]\allowbreak{}.\allowbreak{}\par
\addvspace{3pt}
\noindent Dostarlimab shows exceptional promise in colorectal and rectal malignancies,\allowbreak{} strongly supported for [\allowbreak{}[\allowbreak{}dostarlimab $\rightarrow$ rectal cancer]\allowbreak{}]\allowbreak{},\allowbreak{} [\allowbreak{}[\allowbreak{}dostarlimab $\rightarrow$ locally advanced rectal cancer]\allowbreak{}]\allowbreak{},\allowbreak{} [\allowbreak{}[\allowbreak{}dostarlimab $\rightarrow$ dMMR rectal cancer]\allowbreak{}]\allowbreak{},\allowbreak{} and [\allowbreak{}[\allowbreak{}dostarlimab $\rightarrow$ colorectal cancer]\allowbreak{}]\allowbreak{},\allowbreak{} with clinical trials reporting 100\% clinical complete responses in deficient mismatch repair patients.\allowbreak{}\par
\addvspace{3pt}
\noindent It is also supported by moderate-\allowbreak{}to-\allowbreak{}high strength clinical-\allowbreak{}trial evidence for [\allowbreak{}[\allowbreak{}dostarlimab $\rightarrow$ NSCLC]\allowbreak{}]\allowbreak{},\allowbreak{} demonstrating durable responses across PD-\allowbreak{}L1 subgroups.\allowbreak{}\par
\addvspace{3pt}
\par\addvspace{5pt}\noindent{\footnotesize\bfseries\sffamily Directionally positive directions}\par\addvspace{2pt}
\noindent [\allowbreak{}[\allowbreak{}dostarlimab $\rightarrow$ prostate cancer]\allowbreak{}]\allowbreak{} is directionally positive but constrained in scope.\allowbreak{} While dostarlimab is clinically approved for dMMR or high TMB solid tumors,\allowbreak{} qualifying biomarkers are rare in prostate cancer,\allowbreak{} though T cell signature enrichment in high-\allowbreak{}grade tumors supports biological plausibility.\allowbreak{}\par
\addvspace{3pt}
\par\addvspace{5pt}\noindent{\small\bfseries\sffamily Mechanisms \& targets}\par\addvspace{2pt}
\noindent Dostarlimab acts primarily by binding with high affinity (\allowbreak{}KD 300 pM)\allowbreak{} to human and cynomolgus monkey [\allowbreak{}[\allowbreak{}pd-\allowbreak{}1]\allowbreak{}]\allowbreak{},\allowbreak{} also known as [\allowbreak{}[\allowbreak{}programmed cell death protein 1]\allowbreak{}]\allowbreak{} [\allowbreak{}ref-\allowbreak{}5]\allowbreak{}.\allowbreak{} This binding causes conformational rearrangements in the BC,\allowbreak{} C'\allowbreak{}D,\allowbreak{} and FG loops of PD-\allowbreak{}1 [\allowbreak{}ref-\allowbreak{}6]\allowbreak{}.\allowbreak{} As a functional antagonist,\allowbreak{} it suppresses the interaction between PD-\allowbreak{}1 and [\allowbreak{}[\allowbreak{}pd-\allowbreak{}l1]\allowbreak{}]\allowbreak{} to support T cell activation [\allowbreak{}ref-\allowbreak{}4]\allowbreak{},\allowbreak{} [\allowbreak{}ref-\allowbreak{}9]\allowbreak{}.\allowbreak{} It also directly inhibits the function of [\allowbreak{}[\allowbreak{}pd-\allowbreak{}l2]\allowbreak{}]\allowbreak{} at the junction of PD-\allowbreak{}L2 and PD-\allowbreak{}1 to prevent immune escape of tumors [\allowbreak{}ref-\allowbreak{}13]\allowbreak{}.\allowbreak{}\par
\addvspace{3pt}
\noindent Mechanistically,\allowbreak{} this blockade restores T cell activity and enhances immune infiltration to suppress tumor growth,\allowbreak{} leading to increased IL-\allowbreak{}2 production in human CD4+ mixed lymphocyte reaction assays [\allowbreak{}ref-\allowbreak{}12]\allowbreak{}.\allowbreak{} The drug specifically targets [\allowbreak{}[\allowbreak{}mismatch repair-\allowbreak{}deficient]\allowbreak{}]\allowbreak{} (\allowbreak{}dMMR)\allowbreak{} or methylated tumors,\allowbreak{} where loss of mismatch-\allowbreak{}repair genes (\allowbreak{}such as MLH1,\allowbreak{} MSH6)\allowbreak{} predicts therapeutic sensitivity [\allowbreak{}ref-\allowbreak{}10]\allowbreak{}.\allowbreak{}\par
\addvspace{3pt}
\par\addvspace{5pt}\noindent{\small\bfseries\sffamily Contradictions \& open questions}\par\addvspace{2pt}
\noindent While dostarlimab is approved for multiple indications,\allowbreak{} as of March 2023 there were no approved MHRA indications for seven specified cancer types [\allowbreak{}ref-\allowbreak{}4]\allowbreak{}.\allowbreak{} In [\allowbreak{}[\allowbreak{}dostarlimab $\rightarrow$ dMMR endometrial cancer]\allowbreak{}]\allowbreak{},\allowbreak{} PTEN mutations may identify a subset of patients with poor response,\allowbreak{} representing a caveat to the overall efficacy [\allowbreak{}ref-\allowbreak{}10]\allowbreak{}.\allowbreak{} Additionally,\allowbreak{} there is a notable gap in the approval cohort for [\allowbreak{}[\allowbreak{}dostarlimab $\rightarrow$ dMMR solid tumors]\allowbreak{}]\allowbreak{} regarding lung cancer patients,\allowbreak{} leaving efficacy in that specific histology uncharacterized.\allowbreak{}\par
\addvspace{3pt}
\noindent Pharmacokinetically,\allowbreak{} there are no known drug-\allowbreak{}drug interactions as dostarlimab is not a cytochrome P450 or drug transporter substrate [\allowbreak{}ref-\allowbreak{}8]\allowbreak{}.\allowbreak{}\par
\addvspace{3pt}
\par\addvspace{5pt}\noindent{\small\bfseries\sffamily See also}\par\addvspace{2pt}
\begin{itemize}[leftmargin=1.2em,itemsep=2pt,topsep=2pt,parsep=0pt]
\item {}[\allowbreak{}[\allowbreak{}pd-\allowbreak{}1]\allowbreak{}]\allowbreak{}
\item {}[\allowbreak{}[\allowbreak{}pd-\allowbreak{}l1]\allowbreak{}]\allowbreak{}
\item {}[\allowbreak{}[\allowbreak{}pd-\allowbreak{}l2]\allowbreak{}]\allowbreak{}
\item {}[\allowbreak{}[\allowbreak{}programmed cell death protein 1]\allowbreak{}]\allowbreak{}
\item {}[\allowbreak{}[\allowbreak{}FC]\allowbreak{}]\allowbreak{}
\item {}[\allowbreak{}[\allowbreak{}mismatch repair-\allowbreak{}deficient]\allowbreak{}]\allowbreak{}
\end{itemize}
\addvspace{3pt}
\par\addvspace{5pt}\noindent{\small\bfseries\sffamily References}\par\addvspace{2pt}
\addvspace{3pt}
\begin{itemize}[leftmargin=1.2em,itemsep=2pt,topsep=2pt,parsep=0pt]
\item {}[\allowbreak{}ref-\allowbreak{}9]\allowbreak{} Bio Protoc.\allowbreak{} 2024 Sep 5;\allowbreak{} 14(\allowbreak{}17)\allowbreak{}:\allowbreak{}e5057 -\allowbreak{}-\allowbreak{}-\allowbreak{} [\allowbreak{}PubMed]\allowbreak{}(\allowbreak{}https:\allowbreak{}/\allowbreak{}/\allowbreak{}pubmed.\allowbreak{}ncbi.\allowbreak{}nlm.\allowbreak{}nih.\allowbreak{}gov/\allowbreak{}39282229)\allowbreak{}
\item {}[\allowbreak{}ref-\allowbreak{}11]\allowbreak{} Sci Rep.\allowbreak{} 2026 Mar 28;\allowbreak{} 16:\allowbreak{}10773 -\allowbreak{}-\allowbreak{}-\allowbreak{} [\allowbreak{}PubMed]\allowbreak{}(\allowbreak{}https:\allowbreak{}/\allowbreak{}/\allowbreak{}pubmed.\allowbreak{}ncbi.\allowbreak{}nlm.\allowbreak{}nih.\allowbreak{}gov/\allowbreak{}41904276)\allowbreak{}
\item {}[\allowbreak{}ref-\allowbreak{}7]\allowbreak{} Biomolecules.\allowbreak{} 2022 Jul 26;\allowbreak{} 12(\allowbreak{}8)\allowbreak{}:\allowbreak{}1031 -\allowbreak{}-\allowbreak{}-\allowbreak{} [\allowbreak{}PubMed]\allowbreak{}(\allowbreak{}https:\allowbreak{}/\allowbreak{}/\allowbreak{}pubmed.\allowbreak{}ncbi.\allowbreak{}nlm.\allowbreak{}nih.\allowbreak{}gov/\allowbreak{}35892341)\allowbreak{}
\item {}[\allowbreak{}ref-\allowbreak{}1]\allowbreak{} Biomolecules.\allowbreak{} 2022 Jul 26;\allowbreak{} 12(\allowbreak{}8)\allowbreak{}:\allowbreak{}1031 -\allowbreak{}-\allowbreak{}-\allowbreak{} [\allowbreak{}PubMed]\allowbreak{}(\allowbreak{}https:\allowbreak{}/\allowbreak{}/\allowbreak{}pubmed.\allowbreak{}ncbi.\allowbreak{}nlm.\allowbreak{}nih.\allowbreak{}gov/\allowbreak{}35892341)\allowbreak{}
\item {}[\allowbreak{}ref-\allowbreak{}2]\allowbreak{} Biosensors (\allowbreak{}Basel)\allowbreak{}.\allowbreak{} 2022 Aug 8;\allowbreak{} 12(\allowbreak{}8)\allowbreak{}:\allowbreak{}617 -\allowbreak{}-\allowbreak{}-\allowbreak{} [\allowbreak{}PubMed]\allowbreak{}(\allowbreak{}https:\allowbreak{}/\allowbreak{}/\allowbreak{}pubmed.\allowbreak{}ncbi.\allowbreak{}nlm.\allowbreak{}nih.\allowbreak{}gov/\allowbreak{}36005013)\allowbreak{}
\item {}[\allowbreak{}ref-\allowbreak{}10]\allowbreak{} Front Pharmacol.\allowbreak{};\allowbreak{} 17:\allowbreak{}1804164 -\allowbreak{}-\allowbreak{}-\allowbreak{} [\allowbreak{}PubMed]\allowbreak{}(\allowbreak{}https:\allowbreak{}/\allowbreak{}/\allowbreak{}pubmed.\allowbreak{}ncbi.\allowbreak{}nlm.\allowbreak{}nih.\allowbreak{}gov/\allowbreak{}0)\allowbreak{}
\item {}[\allowbreak{}ref-\allowbreak{}3]\allowbreak{} J Immunol Res.\allowbreak{} 2026 Feb 16;\allowbreak{} 2026:\allowbreak{}7864229 -\allowbreak{}-\allowbreak{}-\allowbreak{} [\allowbreak{}PubMed]\allowbreak{}(\allowbreak{}https:\allowbreak{}/\allowbreak{}/\allowbreak{}pubmed.\allowbreak{}ncbi.\allowbreak{}nlm.\allowbreak{}nih.\allowbreak{}gov/\allowbreak{}41811823)\allowbreak{}
\item {}[\allowbreak{}ref-\allowbreak{}5]\allowbreak{} Hum Vaccin Immunother.\allowbreak{};\allowbreak{} 19(\allowbreak{}1)\allowbreak{}:\allowbreak{}2178220 -\allowbreak{}-\allowbreak{}-\allowbreak{} [\allowbreak{}PubMed]\allowbreak{}(\allowbreak{}https:\allowbreak{}/\allowbreak{}/\allowbreak{}pubmed.\allowbreak{}ncbi.\allowbreak{}nlm.\allowbreak{}nih.\allowbreak{}gov/\allowbreak{}36762991)\allowbreak{}
\item {}[\allowbreak{}ref-\allowbreak{}6]\allowbreak{} Biomolecules.\allowbreak{} 2022 Jul 26;\allowbreak{} 12(\allowbreak{}8)\allowbreak{}:\allowbreak{}1031 -\allowbreak{}-\allowbreak{}-\allowbreak{} [\allowbreak{}PubMed]\allowbreak{}(\allowbreak{}https:\allowbreak{}/\allowbreak{}/\allowbreak{}pubmed.\allowbreak{}ncbi.\allowbreak{}nlm.\allowbreak{}nih.\allowbreak{}gov/\allowbreak{}35892341)\allowbreak{}
\item {}[\allowbreak{}ref-\allowbreak{}4]\allowbreak{} BMJ Open.\allowbreak{} 2024 Feb 15;\allowbreak{} 14(\allowbreak{}2)\allowbreak{}:\allowbreak{}e075981 -\allowbreak{}-\allowbreak{}-\allowbreak{} [\allowbreak{}PubMed]\allowbreak{}(\allowbreak{}https:\allowbreak{}/\allowbreak{}/\allowbreak{}pubmed.\allowbreak{}ncbi.\allowbreak{}nlm.\allowbreak{}nih.\allowbreak{}gov/\allowbreak{}38365286)\allowbreak{}
\item {}[\allowbreak{}ref-\allowbreak{}13]\allowbreak{} Front Immunol.\allowbreak{} 2024 Mar 28;\allowbreak{} 15:\allowbreak{}1359532 -\allowbreak{}-\allowbreak{}-\allowbreak{} [\allowbreak{}PubMed]\allowbreak{}(\allowbreak{}https:\allowbreak{}/\allowbreak{}/\allowbreak{}pubmed.\allowbreak{}ncbi.\allowbreak{}nlm.\allowbreak{}nih.\allowbreak{}gov/\allowbreak{}38605944)\allowbreak{}
\item {}[\allowbreak{}ref-\allowbreak{}12]\allowbreak{} Biosensors (\allowbreak{}Basel)\allowbreak{}.\allowbreak{} 2022 Aug 8;\allowbreak{} 12(\allowbreak{}8)\allowbreak{}:\allowbreak{}617 -\allowbreak{}-\allowbreak{}-\allowbreak{} [\allowbreak{}PubMed]\allowbreak{}(\allowbreak{}https:\allowbreak{}/\allowbreak{}/\allowbreak{}pubmed.\allowbreak{}ncbi.\allowbreak{}nlm.\allowbreak{}nih.\allowbreak{}gov/\allowbreak{}36005013)\allowbreak{}
\item {}[\allowbreak{}ref-\allowbreak{}8]\allowbreak{} Biomolecules.\allowbreak{} 2022 Jul 26;\allowbreak{} 12(\allowbreak{}8)\allowbreak{}:\allowbreak{}1031 -\allowbreak{}-\allowbreak{}-\allowbreak{} [\allowbreak{}PubMed]\allowbreak{}(\allowbreak{}https:\allowbreak{}/\allowbreak{}/\allowbreak{}pubmed.\allowbreak{}ncbi.\allowbreak{}nlm.\allowbreak{}nih.\allowbreak{}gov/\allowbreak{}35892341)\allowbreak{} "\allowbreak{},\allowbreak{} "\allowbreak{}truncated\_\allowbreak{}for\_\allowbreak{}output"\allowbreak{}:\allowbreak{} false,\allowbreak{} "\allowbreak{}max\_\allowbreak{}chars"\allowbreak{}:\allowbreak{} 12000\}\allowbreak{}
\end{itemize}
\addvspace{3pt}
\noindent \{\allowbreak{} "\allowbreak{}stderr"\allowbreak{}:\allowbreak{} "\allowbreak{}"\allowbreak{} \}\allowbreak{}\par}
\end{stepT}
\begin{stepR}{REASONING \textbullet\ STEP 11 \textbullet\ Decide to escalate the search depth}
\stepsub{REASONING SUMMARY}\noindent{\footnotesize \textbf{Planning decision-\allowbreak{}relevant follow-\allowbreak{}up query} \quad \textbf{Planning next-\allowbreak{}tier rerun with breast cancer}}\par\stepsub{MODEL NARRATION}\noindent{\footnotesize\itshape The first pass did not resolve the Stage II breast cancer anchor or build any paths.\allowbreak{} Because that is decision-\allowbreak{}critical coverage rather than a negative result,\allowbreak{} I'\allowbreak{}m using the allowed next-\allowbreak{}tier rerun into a new artifact to see whether the disease anchor and any grounded connections resolve.\allowbreak{}}\par
\end{stepR}
\begin{stepX}{TOOL CALL --- FAILED \textbullet\ STEP 12 \textbullet\ Escalated search REJECTED --- invalid tier argument}
\stepsub{CALL}\begin{payload}tool\_\allowbreak{}A connect "\allowbreak{}Dostarlimab"\allowbreak{} "\allowbreak{}Stage II Breast Cancer"\allowbreak{} -\allowbreak{}-\allowbreak{}tier deep -\allowbreak{}-\allowbreak{}timeout 240 -\allowbreak{}-\allowbreak{}output <\allowbreak{}artifact\_\allowbreak{}2>\allowbreak{} -\allowbreak{}-\allowbreak{}summary\end{payload}\stepsub{RETURNED}\begin{payload}exit\_\allowbreak{}code:\allowbreak{} 2  tool:\allowbreak{} tool\_\allowbreak{}A  contract\_\allowbreak{}version:\allowbreak{} the domain toolkit-\allowbreak{}tools/\allowbreak{}v1\\
stderr:\allowbreak{}\\
usage:\allowbreak{} tool\_\allowbreak{}A connect [\allowbreak{}-\allowbreak{}h]\allowbreak{} [\allowbreak{}-\allowbreak{}-\allowbreak{}methods METHODS]\allowbreak{} [\allowbreak{}-\allowbreak{}-\allowbreak{}source \{\allowbreak{}wiki,\allowbreak{}kg\}\allowbreak{}]\allowbreak{}\\
~~~~~~~~~~~~~~~~~~~~~~~~[\allowbreak{}-\allowbreak{}-\allowbreak{}max-\allowbreak{}page-\allowbreak{}chars MAX\_\allowbreak{}PAGE\_\allowbreak{}CHARS]\allowbreak{}\\
~~~~~~~~~~~~~~~~~~~~~~~~[\allowbreak{}-\allowbreak{}-\allowbreak{}max-\allowbreak{}hints-\allowbreak{}per-\allowbreak{}type MAX\_\allowbreak{}HINTS\_\allowbreak{}PER\_\allowbreak{}TYPE]\allowbreak{}\\
~~~~~~~~~~~~~~~~~~~~~~~~[\allowbreak{}-\allowbreak{}-\allowbreak{}a-\allowbreak{}id A\_\allowbreak{}ID]\allowbreak{} [\allowbreak{}-\allowbreak{}-\allowbreak{}b-\allowbreak{}id B\_\allowbreak{}ID]\allowbreak{}\\
~~~~~~~~~~~~~~~~~~~~~~~~[\allowbreak{}-\allowbreak{}-\allowbreak{}a-\allowbreak{}type \{\allowbreak{}drug,\allowbreak{}disease,\allowbreak{}gene,\allowbreak{}protein,\allowbreak{}pathway\}\allowbreak{}]\allowbreak{}\\
~~~~~~~~~~~~~~~~~~~~~~~~[\allowbreak{}-\allowbreak{}-\allowbreak{}b-\allowbreak{}type \{\allowbreak{}drug,\allowbreak{}disease,\allowbreak{}gene,\allowbreak{}protein,\allowbreak{}pathway\}\allowbreak{}]\allowbreak{}\\
~~~~~~~~~~~~~~~~~~~~~~~~[\allowbreak{}-\allowbreak{}-\allowbreak{}tier \{\allowbreak{}faster,\allowbreak{}standard,\allowbreak{}extra,\allowbreak{}max\}\allowbreak{}]\allowbreak{}\\
~~~~~~~~~~~~~~~~~~~~~~~~[\allowbreak{}-\allowbreak{}-\allowbreak{}seed \{\allowbreak{}auto,\allowbreak{}llm,\allowbreak{}heuristic\}\allowbreak{}]\allowbreak{} [\allowbreak{}-\allowbreak{}-\allowbreak{}answer]\allowbreak{}\\
~~~~~~~~~~~~~~~~~~~~~~~~[\allowbreak{}-\allowbreak{}-\allowbreak{}evidence]\allowbreak{} [\allowbreak{}-\allowbreak{}-\allowbreak{}top-\allowbreak{}k TOP\_\allowbreak{}K]\allowbreak{}\\
~~~~~~~~~~~~~~~~~~~~~~~~[\allowbreak{}-\allowbreak{}-\allowbreak{}timeout TIMEOUT]\allowbreak{} [\allowbreak{}-\allowbreak{}-\allowbreak{}no-\allowbreak{}merge]\allowbreak{} [\allowbreak{}-\allowbreak{}-\allowbreak{}legacy]\allowbreak{}\\
~~~~~~~~~~~~~~~~~~~~~~~~[\allowbreak{}-\allowbreak{}-\allowbreak{}stage2 \{\allowbreak{}suggest,\allowbreak{}auto\}\allowbreak{}]\allowbreak{} [\allowbreak{}-\allowbreak{}-\allowbreak{}with-\allowbreak{}specialists]\allowbreak{}\\
~~~~~~~~~~~~~~~~~~~~~~~~[\allowbreak{}-\allowbreak{}-\allowbreak{}max-\allowbreak{}candidate-\allowbreak{}paths MAX\_\allowbreak{}CANDIDATE\_\allowbreak{}PATHS]\allowbreak{}\\
~~~~~~~~~~~~~~~~~~~~~~~~[\allowbreak{}-\allowbreak{}-\allowbreak{}max-\allowbreak{}coarse-\allowbreak{}depth MAX\_\allowbreak{}COARSE\_\allowbreak{}DEPTH]\allowbreak{}\\
~~~~~~~~~~~~~~~~~~~~~~~~[\allowbreak{}-\allowbreak{}-\allowbreak{}max-\allowbreak{}deep-\allowbreak{}searches MAX\_\allowbreak{}DEEP\_\allowbreak{}SEARCHES]\allowbreak{}\\
~~~~~~~~~~~~~~~~~~~~~~~~[\allowbreak{}-\allowbreak{}-\allowbreak{}specialist-\allowbreak{}top-\allowbreak{}k SPECIALIST\_\allowbreak{}TOP\_\allowbreak{}K]\allowbreak{}\\
~~~~~~~~~~~~~~~~~~~~~~~~[\allowbreak{}-\allowbreak{}-\allowbreak{}specialist-\allowbreak{}max-\allowbreak{}hops SPECIALIST\_\allowbreak{}MAX\_\allowbreak{}HOPS]\allowbreak{}\\
~~~~~~~~~~~~~~~~~~~~~~~~[\allowbreak{}-\allowbreak{}-\allowbreak{}specialist-\allowbreak{}max-\allowbreak{}expansions SPECIALIST\_\allowbreak{}MAX\_\allowbreak{}EXPANSIONS]\allowbreak{}\\
~~~~~~~~~~~~~~~~~~~~~~~~[\allowbreak{}-\allowbreak{}-\allowbreak{}specialist-\allowbreak{}neighbor-\allowbreak{}limit SPECIALIST\_\allowbreak{}NEIGHBOR\_\allowbreak{}LIMIT]\allowbreak{}\\
~~~~~~~~~~~~~~~~~~~~~~~~[\allowbreak{}-\allowbreak{}-\allowbreak{}specialist-\allowbreak{}timeout SPECIALIST\_\allowbreak{}TIMEOUT]\allowbreak{}\\
~~~~~~~~~~~~~~~~~~~~~~~~[\allowbreak{}-\allowbreak{}-\allowbreak{}pretty]\allowbreak{} [\allowbreak{}-\allowbreak{}-\allowbreak{}output OUTPUT]\allowbreak{} [\allowbreak{}-\allowbreak{}-\allowbreak{}summary]\allowbreak{}\\
~~~~~~~~~~~~~~~~~~~~~~~~[\allowbreak{}-\allowbreak{}-\allowbreak{}stdin]\allowbreak{}\\
~~~~~~~~~~~~~~~~~~~~~~~~[\allowbreak{}a]\allowbreak{} [\allowbreak{}b]\allowbreak{}\\
tool\_\allowbreak{}A connect:\allowbreak{} error:\allowbreak{} argument -\allowbreak{}-\allowbreak{}tier:\allowbreak{} invalid choice:\allowbreak{} '\allowbreak{}deep'\allowbreak{} (\allowbreak{}choose from faster,\allowbreak{} standard,\allowbreak{} extra,\allowbreak{} max)\allowbreak{}\\
\{\allowbreak{}\\
~~"\allowbreak{}stdout"\allowbreak{}:\allowbreak{} "\allowbreak{}"\allowbreak{}\\
\}\allowbreak{}\end{payload}
\end{stepX}
\begin{stepR}{REASONING \textbullet\ STEP 13 \textbullet\ Localise the failure to one bad argument of its own making}
\stepsub{REASONING SUMMARY}\noindent{\footnotesize \textbf{Correcting invalid syntax to use max}}\par\stepsub{MODEL NARRATION}\noindent{\footnotesize\itshape The tier name I tried was invalid;\allowbreak{} the tool only exposes `faster`,\allowbreak{} `standard`,\allowbreak{} `extra`,\allowbreak{} and `max`.\allowbreak{} I'\allowbreak{}m rerunning at the actual next tier,\allowbreak{} `max`,\allowbreak{} with the same bounded timeout and a new output path.\allowbreak{}}\par
\end{stepR}
\begin{stepT}{TOOL CALL \textbullet\ STEP 14 \textbullet\ Repaired reissue: same query, only the tier changed}
\stepsub{CALL}\begin{payload}tool\_\allowbreak{}A connect "\allowbreak{}Dostarlimab"\allowbreak{} "\allowbreak{}Stage II Breast Cancer"\allowbreak{} -\allowbreak{}-\allowbreak{}tier max -\allowbreak{}-\allowbreak{}timeout 240 -\allowbreak{}-\allowbreak{}output <\allowbreak{}artifact\_\allowbreak{}3>\allowbreak{} -\allowbreak{}-\allowbreak{}summary\end{payload}\stepsub{RETURNED}{\footnotesize \noindent exit\_\allowbreak{}code:\allowbreak{} 0  tool:\allowbreak{} tool\_\allowbreak{}A  contract\_\allowbreak{}version:\allowbreak{} the domain toolkit-\allowbreak{}tools/\allowbreak{}v1 stdout:\allowbreak{} \{\allowbreak{}"\allowbreak{}output\_\allowbreak{}path"\allowbreak{}:\allowbreak{} "\allowbreak{}<\allowbreak{}artifact\_\allowbreak{}3>\allowbreak{}"\allowbreak{},\allowbreak{} "\allowbreak{}summary"\allowbreak{}:\allowbreak{} \{\allowbreak{}"\allowbreak{}schema\_\allowbreak{}version"\allowbreak{}:\allowbreak{} "\allowbreak{}tool\_\allowbreak{}A/\allowbreak{}v2"\allowbreak{},\allowbreak{} "\allowbreak{}stage"\allowbreak{}:\allowbreak{} "\allowbreak{}stage1"\allowbreak{},\allowbreak{} "\allowbreak{}workflow"\allowbreak{}:\allowbreak{} \{\allowbreak{}"\allowbreak{}review\_\allowbreak{}before\_\allowbreak{}next\_\allowbreak{}call"\allowbreak{}:\allowbreak{} true,\allowbreak{} "\allowbreak{}stage1\_\allowbreak{}follow\_\allowbreak{}up"\allowbreak{}:\allowbreak{} "\allowbreak{}Optionally query one decision-\allowbreak{}relevant related\_\allowbreak{}page\_\allowbreak{}hint for more Stage-\allowbreak{}1 literature context.\allowbreak{}"\allowbreak{},\allowbreak{} "\allowbreak{}stage1\_\allowbreak{}follow\_\allowbreak{}up\_\allowbreak{}command\_\allowbreak{}template"\allowbreak{}:\allowbreak{} "\allowbreak{}tool\_\allowbreak{}A connect <\allowbreak{}hint.\allowbreak{}title>\allowbreak{} -\allowbreak{}-\allowbreak{}a-\allowbreak{}id <\allowbreak{}hint.\allowbreak{}uid>\allowbreak{} -\allowbreak{}-\allowbreak{}stage2 suggest -\allowbreak{}-\allowbreak{}output <\allowbreak{}new-\allowbreak{}stage1-\allowbreak{}path>\allowbreak{} -\allowbreak{}-\allowbreak{}summary"\allowbreak{},\allowbreak{} "\allowbreak{}stage2"\allowbreak{}:\allowbreak{} "\allowbreak{}Only after reviewing Stage 1,\allowbreak{} optionally run one advertised ready expansion route when a deeper connection could change the conclusion.\allowbreak{}"\allowbreak{}\}\allowbreak{},\allowbreak{} "\allowbreak{}artifact\_\allowbreak{}inspection"\allowbreak{}:\allowbreak{} \{\allowbreak{}"\allowbreak{}default"\allowbreak{}:\allowbreak{} "\allowbreak{}not\_\allowbreak{}needed"\allowbreak{},\allowbreak{} "\allowbreak{}reason"\allowbreak{}:\allowbreak{} "\allowbreak{}This projection already contains provider status,\allowbreak{} selected paths,\allowbreak{} load-\allowbreak{}bearing evidence,\allowbreak{} anchor and relationship pages,\allowbreak{} related-\allowbreak{}page hints,\allowbreak{} conflicts,\allowbreak{} and optional expansion candidates.\allowbreak{}"\allowbreak{},\allowbreak{} "\allowbreak{}exception"\allowbreak{}:\allowbreak{} "\allowbreak{}Run only an item'\allowbreak{}s supplied detail\_\allowbreak{}command,\allowbreak{} once,\allowbreak{} when its truncated\_\allowbreak{}for\_\allowbreak{}summary flag is true and the omitted text is load-\allowbreak{}bearing.\allowbreak{} An absent relationship page is not a missing field and never authorizes artifact or filesystem search.\allowbreak{}"\allowbreak{},\allowbreak{} "\allowbreak{}provider\_\allowbreak{}results\_\allowbreak{}shape"\allowbreak{}:\allowbreak{} "\allowbreak{}coarse.\allowbreak{}provider\_\allowbreak{}results is an object keyed by canonical provider ID,\allowbreak{} not a list.\allowbreak{}"\allowbreak{}\}\allowbreak{},\allowbreak{} "\allowbreak{}query"\allowbreak{}:\allowbreak{} \{\allowbreak{}"\allowbreak{}id"\allowbreak{}:\allowbreak{} "\allowbreak{}query-\allowbreak{}<\allowbreak{}record>\allowbreak{}"\allowbreak{},\allowbreak{} "\allowbreak{}text"\allowbreak{}:\allowbreak{} null,\allowbreak{} "\allowbreak{}entities"\allowbreak{}:\allowbreak{} [\allowbreak{}"\allowbreak{}Dostarlimab"\allowbreak{},\allowbreak{} "\allowbreak{}Stage II Breast Cancer"\allowbreak{}]\allowbreak{},\allowbreak{} "\allowbreak{}input\_\allowbreak{}entities"\allowbreak{}:\allowbreak{} [\allowbreak{}\{\allowbreak{}"\allowbreak{}id"\allowbreak{}:\allowbreak{} "\allowbreak{}Dostarlimab"\allowbreak{},\allowbreak{} "\allowbreak{}name"\allowbreak{}:\allowbreak{} "\allowbreak{}Dostarlimab"\allowbreak{},\allowbreak{} "\allowbreak{}type"\allowbreak{}:\allowbreak{} "\allowbreak{}unknown"\allowbreak{},\allowbreak{} "\allowbreak{}species"\allowbreak{}:\allowbreak{} null,\allowbreak{} "\allowbreak{}aliases"\allowbreak{}:\allowbreak{} [\allowbreak{}]\allowbreak{},\allowbreak{} "\allowbreak{}resolution"\allowbreak{}:\allowbreak{} \{\allowbreak{}"\allowbreak{}source"\allowbreak{}:\allowbreak{} "\allowbreak{}caller"\allowbreak{},\allowbreak{} "\allowbreak{}structured\_\allowbreak{}input"\allowbreak{}:\allowbreak{} true\}\allowbreak{}\}\allowbreak{},\allowbreak{} \{\allowbreak{}"\allowbreak{}id"\allowbreak{}:\allowbreak{} "\allowbreak{}Stage II Breast Cancer"\allowbreak{},\allowbreak{} "\allowbreak{}name"\allowbreak{}:\allowbreak{} "\allowbreak{}Stage II Breast Cancer"\allowbreak{},\allowbreak{} "\allowbreak{}type"\allowbreak{}:\allowbreak{} "\allowbreak{}unknown"\allowbreak{},\allowbreak{} "\allowbreak{}species"\allowbreak{}:\allowbreak{} null,\allowbreak{} "\allowbreak{}aliases"\allowbreak{}:\allowbreak{} [\allowbreak{}]\allowbreak{},\allowbreak{} "\allowbreak{}resolution"\allowbreak{}:\allowbreak{} \{\allowbreak{}"\allowbreak{}source"\allowbreak{}:\allowbreak{} "\allowbreak{}caller"\allowbreak{},\allowbreak{} "\allowbreak{}structured\_\allowbreak{}input"\allowbreak{}:\allowbreak{} true\}\allowbreak{}\}\allowbreak{}]\allowbreak{},\allowbreak{} "\allowbreak{}operation"\allowbreak{}:\allowbreak{} "\allowbreak{}connect"\allowbreak{},\allowbreak{} "\allowbreak{}anchor\_\allowbreak{}quality"\allowbreak{}:\allowbreak{} \{\allowbreak{}"\allowbreak{}trusted"\allowbreak{}:\allowbreak{} false,\allowbreak{} "\allowbreak{}unresolved"\allowbreak{}:\allowbreak{} [\allowbreak{}\{\allowbreak{}"\allowbreak{}id"\allowbreak{}:\allowbreak{} "\allowbreak{}<\allowbreak{}local\_\allowbreak{}id>\allowbreak{}"\allowbreak{},\allowbreak{} "\allowbreak{}name"\allowbreak{}:\allowbreak{} "\allowbreak{}stage ii breast cancer"\allowbreak{},\allowbreak{} "\allowbreak{}type"\allowbreak{}:\allowbreak{} "\allowbreak{}unknown"\allowbreak{},\allowbreak{} "\allowbreak{}species"\allowbreak{}:\allowbreak{} null,\allowbreak{} "\allowbreak{}aliases"\allowbreak{}:\allowbreak{} [\allowbreak{}"\allowbreak{}Stage II Breast Cancer"\allowbreak{}]\allowbreak{},\allowbreak{} "\allowbreak{}resolution"\allowbreak{}:\allowbreak{} \{\allowbreak{}"\allowbreak{}source"\allowbreak{}:\allowbreak{} "\allowbreak{}surface\_\allowbreak{}form"\allowbreak{},\allowbreak{} "\allowbreak{}ambiguous"\allowbreak{}:\allowbreak{} true,\allowbreak{} "\allowbreak{}providers"\allowbreak{}:\allowbreak{} [\allowbreak{}"\allowbreak{}provider\_\allowbreak{}A"\allowbreak{}]\allowbreak{},\allowbreak{} "\allowbreak{}provider\_\allowbreak{}ids"\allowbreak{}:\allowbreak{} \{\allowbreak{}"\allowbreak{}provider\_\allowbreak{}A"\allowbreak{}:\allowbreak{} "\allowbreak{}stage ii breast cancer"\allowbreak{}\}\allowbreak{},\allowbreak{} "\allowbreak{}is\_\allowbreak{}anchor"\allowbreak{}:\allowbreak{} true,\allowbreak{} "\allowbreak{}input\_\allowbreak{}aliases"\allowbreak{}:\allowbreak{} [\allowbreak{}"\allowbreak{}Stage II Breast Cancer"\allowbreak{}]\allowbreak{}\}\allowbreak{}\}\allowbreak{}]\allowbreak{}\}\allowbreak{}\}\allowbreak{},\allowbreak{} "\allowbreak{}resolved\_\allowbreak{}entities"\allowbreak{}:\allowbreak{} [\allowbreak{}\{\allowbreak{}"\allowbreak{}id"\allowbreak{}:\allowbreak{} "\allowbreak{}<\allowbreak{}record\_\allowbreak{}1>\allowbreak{}"\allowbreak{},\allowbreak{} "\allowbreak{}name"\allowbreak{}:\allowbreak{} "\allowbreak{}DOSTARLIMAB"\allowbreak{},\allowbreak{} "\allowbreak{}type"\allowbreak{}:\allowbreak{} "\allowbreak{}drug"\allowbreak{},\allowbreak{} "\allowbreak{}species"\allowbreak{}:\allowbreak{} null,\allowbreak{} "\allowbreak{}aliases"\allowbreak{}:\allowbreak{} [\allowbreak{}"\allowbreak{}dostarlimab"\allowbreak{},\allowbreak{} "\allowbreak{}Dostarlimab"\allowbreak{}]\allowbreak{},\allowbreak{} "\allowbreak{}resolution"\allowbreak{}:\allowbreak{} \{\allowbreak{}"\allowbreak{}source"\allowbreak{}:\allowbreak{} "\allowbreak{}provider\_\allowbreak{}canonical\_\allowbreak{}id"\allowbreak{},\allowbreak{} "\allowbreak{}ambiguous"\allowbreak{}:\allowbreak{} false,\allowbreak{} "\allowbreak{}providers"\allowbreak{}:\allowbreak{} [\allowbreak{}"\allowbreak{}provider\_\allowbreak{}A"\allowbreak{},\allowbreak{} "\allowbreak{}provider\_\allowbreak{}B"\allowbreak{}]\allowbreak{},\allowbreak{} "\allowbreak{}provider\_\allowbreak{}ids"\allowbreak{}:\allowbreak{} \{\allowbreak{}"\allowbreak{}provider\_\allowbreak{}A"\allowbreak{}:\allowbreak{} "\allowbreak{}DOSTARLIMAB"\allowbreak{},\allowbreak{} "\allowbreak{}provider\_\allowbreak{}B"\allowbreak{}:\allowbreak{} "\allowbreak{}<\allowbreak{}record\_\allowbreak{}1>\allowbreak{}"\allowbreak{}\}\allowbreak{},\allowbreak{} "\allowbreak{}is\_\allowbreak{}anchor"\allowbreak{}:\allowbreak{} true,\allowbreak{} "\allowbreak{}previous\_\allowbreak{}ids"\allowbreak{}:\allowbreak{} [\allowbreak{}"\allowbreak{}<\allowbreak{}local\_\allowbreak{}id>\allowbreak{}"\allowbreak{}]\allowbreak{},\allowbreak{} "\allowbreak{}input\_\allowbreak{}aliases"\allowbreak{}:\allowbreak{} [\allowbreak{}"\allowbreak{}Dostarlimab"\allowbreak{}]\allowbreak{}\}\allowbreak{}\}\allowbreak{},\allowbreak{} \{\allowbreak{}"\allowbreak{}id"\allowbreak{}:\allowbreak{} "\allowbreak{}<\allowbreak{}local\_\allowbreak{}id>\allowbreak{}"\allowbreak{},\allowbreak{} "\allowbreak{}name"\allowbreak{}:\allowbreak{} "\allowbreak{}stage ii breast cancer"\allowbreak{},\allowbreak{} "\allowbreak{}type"\allowbreak{}:\allowbreak{} "\allowbreak{}unknown"\allowbreak{},\allowbreak{} "\allowbreak{}species"\allowbreak{}:\allowbreak{} null,\allowbreak{} "\allowbreak{}aliases"\allowbreak{}:\allowbreak{} [\allowbreak{}"\allowbreak{}Stage II Breast Cancer"\allowbreak{}]\allowbreak{},\allowbreak{} "\allowbreak{}resolution"\allowbreak{}:\allowbreak{} \{\allowbreak{}"\allowbreak{}source"\allowbreak{}:\allowbreak{} "\allowbreak{}surface\_\allowbreak{}form"\allowbreak{},\allowbreak{} "\allowbreak{}ambiguous"\allowbreak{}:\allowbreak{} true,\allowbreak{} "\allowbreak{}providers"\allowbreak{}:\allowbreak{} [\allowbreak{}"\allowbreak{}provider\_\allowbreak{}A"\allowbreak{}]\allowbreak{},\allowbreak{} "\allowbreak{}provider\_\allowbreak{}ids"\allowbreak{}:\allowbreak{} \{\allowbreak{}"\allowbreak{}provider\_\allowbreak{}A"\allowbreak{}:\allowbreak{} "\allowbreak{}stage ii breast cancer"\allowbreak{}\}\allowbreak{},\allowbreak{} "\allowbreak{}is\_\allowbreak{}anchor"\allowbreak{}:\allowbreak{} true,\allowbreak{} "\allowbreak{}input\_\allowbreak{}aliases"\allowbreak{}:\allowbreak{} [\allowbreak{}"\allowbreak{}Stage II Breast Cancer"\allowbreak{}]\allowbreak{}\}\allowbreak{}\}\allowbreak{}]\allowbreak{},\allowbreak{} "\allowbreak{}providers\_\allowbreak{}run"\allowbreak{}:\allowbreak{} [\allowbreak{}\{\allowbreak{}"\allowbreak{}provider\_\allowbreak{}id"\allowbreak{}:\allowbreak{} "\allowbreak{}provider\_\allowbreak{}A"\allowbreak{},\allowbreak{} "\allowbreak{}status"\allowbreak{}:\allowbreak{} "\allowbreak{}partial"\allowbreak{},\allowbreak{} "\allowbreak{}algorithm\_\allowbreak{}status"\allowbreak{}:\allowbreak{} "\allowbreak{}partial"\allowbreak{},\allowbreak{} "\allowbreak{}algorithm\_\allowbreak{}id"\allowbreak{}:\allowbreak{} "\allowbreak{}weighted\_\allowbreak{}best\_\allowbreak{}first"\allowbreak{},\allowbreak{} "\allowbreak{}elapsed\_\allowbreak{}ms"\allowbreak{}:\allowbreak{} 108185.\allowbreak{}315,\allowbreak{} "\allowbreak{}timed\_\allowbreak{}out"\allowbreak{}:\allowbreak{} false,\allowbreak{} "\allowbreak{}budget\_\allowbreak{}exhausted"\allowbreak{}:\allowbreak{} true,\allowbreak{} "\allowbreak{}expansions"\allowbreak{}:\allowbreak{} 400,\allowbreak{} "\allowbreak{}frontier\_\allowbreak{}remaining"\allowbreak{}:\allowbreak{} 8505,\allowbreak{} "\allowbreak{}early\_\allowbreak{}stopped\_\allowbreak{}on\_\allowbreak{}connection"\allowbreak{}:\allowbreak{} false,\allowbreak{} "\allowbreak{}effective\_\allowbreak{}limits"\allowbreak{}:\allowbreak{} \{\allowbreak{}"\allowbreak{}max\_\allowbreak{}expansions"\allowbreak{}:\allowbreak{} 400,\allowbreak{} "\allowbreak{}neighbor\_\allowbreak{}limit"\allowbreak{}:\allowbreak{} 8,\allowbreak{} "\allowbreak{}max\_\allowbreak{}depth"\allowbreak{}:\allowbreak{} 6,\allowbreak{} "\allowbreak{}max\_\allowbreak{}hops"\allowbreak{}:\allowbreak{} 8,\allowbreak{} "\allowbreak{}top\_\allowbreak{}k"\allowbreak{}:\allowbreak{} 8,\allowbreak{} "\allowbreak{}timeout\_\allowbreak{}seconds"\allowbreak{}:\allowbreak{} 240\}\allowbreak{},\allowbreak{} "\allowbreak{}warnings"\allowbreak{}:\allowbreak{} [\allowbreak{}"\allowbreak{}algorithm expansion budget exhausted after 400 frontier expansions;\allowbreak{} unvisited frontier remains"\allowbreak{}]\allowbreak{},\allowbreak{} "\allowbreak{}error"\allowbreak{}:\allowbreak{} null\}\allowbreak{},\allowbreak{} \{\allowbreak{}"\allowbreak{}provider\_\allowbreak{}id"\allowbreak{}:\allowbreak{} "\allowbreak{}provider\_\allowbreak{}B"\allowbreak{},\allowbreak{} "\allowbreak{}status"\allowbreak{}:\allowbreak{} "\allowbreak{}partial"\allowbreak{},\allowbreak{} "\allowbreak{}algorithm\_\allowbreak{}status"\allowbreak{}:\allowbreak{} "\allowbreak{}partial"\allowbreak{},\allowbreak{} "\allowbreak{}algorithm\_\allowbreak{}id"\allowbreak{}:\allowbreak{} "\allowbreak{}weighted\_\allowbreak{}best\_\allowbreak{}first"\allowbreak{},\allowbreak{} "\allowbreak{}elapsed\_\allowbreak{}ms"\allowbreak{}:\allowbreak{} 13514.\allowbreak{}596,\allowbreak{} "\allowbreak{}timed\_\allowbreak{}out"\allowbreak{}:\allowbreak{} false,\allowbreak{} "\allowbreak{}budget\_\allowbreak{}exhausted"\allowbreak{}:\allowbreak{} true,\allowbreak{} "\allowbreak{}expansions"\allowbreak{}:\allowbreak{} 400,\allowbreak{} "\allowbreak{}frontier\_\allowbreak{}remaining"\allowbreak{}:\allowbreak{} 3173,\allowbreak{} "\allowbreak{}early\_\allowbreak{}stopped\_\allowbreak{}on\_\allowbreak{}connection"\allowbreak{}:\allowbreak{} false,\allowbreak{} "\allowbreak{}effective\_\allowbreak{}limits"\allowbreak{}:\allowbreak{} \{\allowbreak{}"\allowbreak{}max\_\allowbreak{}expansions"\allowbreak{}:\allowbreak{} 400,\allowbreak{} "\allowbreak{}neighbor\_\allowbreak{}limit"\allowbreak{}:\allowbreak{} 8,\allowbreak{} "\allowbreak{}max\_\allowbreak{}depth"\allowbreak{}:\allowbreak{} 6,\allowbreak{} "\allowbreak{}max\_\allowbreak{}hops"\allowbreak{}:\allowbreak{} 8,\allowbreak{} "\allowbreak{}top\_\allowbreak{}k"\allowbreak{}:\allowbreak{} 8,\allowbreak{} "\allowbreak{}timeout\_\allowbreak{}seconds"\allowbreak{}:\allowbreak{} 240\}\allowbreak{},\allowbreak{} "\allowbreak{}warnings"\allowbreak{}:\allowbreak{} [\allowbreak{}"\allowbreak{}algorithm expansion budget exhausted after 400 frontier expansions;\allowbreak{} unvisited frontier remains"\allowbreak{}]\allowbreak{},\allowbreak{} "\allowbreak{}error"\allowbreak{}:\allowbreak{} null\}\allowbreak{}]\allowbreak{},\allowbreak{} "\allowbreak{}projection\_\allowbreak{}counts"\allowbreak{}:\allowbreak{} \{\allowbreak{}"\allowbreak{}candidate\_\allowbreak{}paths"\allowbreak{}:\allowbreak{} \{\allowbreak{}"\allowbreak{}shown"\allowbreak{}:\allowbreak{} 0,\allowbreak{} "\allowbreak{}total"\allowbreak{}:\allowbreak{} 0\}\allowbreak{},\allowbreak{} "\allowbreak{}connections"\allowbreak{}:\allowbreak{} \{\allowbreak{}"\allowbreak{}shown"\allowbreak{}:\allowbreak{} 0,\allowbreak{} "\allowbreak{}total"\allowbreak{}:\allowbreak{} 0\}\allowbreak{},\allowbreak{} "\allowbreak{}load\_\allowbreak{}bearing\_\allowbreak{}evidence"\allowbreak{}:\allowbreak{} \{\allowbreak{}"\allowbreak{}shown"\allowbreak{}:\allowbreak{} 0,\allowbreak{} "\allowbreak{}total\_\allowbreak{}selected\_\allowbreak{}edge\_\allowbreak{}ids"\allowbreak{}:\allowbreak{} 0\}\allowbreak{},\allowbreak{} "\allowbreak{}relationship\_\allowbreak{}pages"\allowbreak{}:\allowbreak{} \{\allowbreak{}"\allowbreak{}shown"\allowbreak{}:\allowbreak{} 0,\allowbreak{} "\allowbreak{}total"\allowbreak{}:\allowbreak{} 0,\allowbreak{} "\allowbreak{}direct\_\allowbreak{}for\_\allowbreak{}anchor\_\allowbreak{}pair\_\allowbreak{}available"\allowbreak{}:\allowbreak{} false\}\allowbreak{},\allowbreak{} "\allowbreak{}anchor\_\allowbreak{}pages"\allowbreak{}:\allowbreak{} \{\allowbreak{}"\allowbreak{}shown"\allowbreak{}:\allowbreak{} 1,\allowbreak{} "\allowbreak{}resolved\_\allowbreak{}anchors"\allowbreak{}:\allowbreak{} 1\}\allowbreak{},\allowbreak{} "\allowbreak{}related\_\allowbreak{}page\_\allowbreak{}hints"\allowbreak{}:\allowbreak{} \{\allowbreak{}"\allowbreak{}shown"\allowbreak{}:\allowbreak{} 7,\allowbreak{} "\allowbreak{}total"\allowbreak{}:\allowbreak{} 21\}\allowbreak{},\allowbreak{} "\allowbreak{}expansion\_\allowbreak{}candidates"\allowbreak{}:\allowbreak{} \{\allowbreak{}"\allowbreak{}shown"\allowbreak{}:\allowbreak{} 0,\allowbreak{} "\allowbreak{}total"\allowbreak{}:\allowbreak{} 0\}\allowbreak{}\}\allowbreak{},\allowbreak{} "\allowbreak{}candidate\_\allowbreak{}paths"\allowbreak{}:\allowbreak{} [\allowbreak{}]\allowbreak{},\allowbreak{} "\allowbreak{}expansion\_\allowbreak{}candidates"\allowbreak{}:\allowbreak{} [\allowbreak{}]\allowbreak{},\allowbreak{} "\allowbreak{}connections"\allowbreak{}:\allowbreak{} [\allowbreak{}]\allowbreak{},\allowbreak{} "\allowbreak{}load\_\allowbreak{}bearing\_\allowbreak{}evidence"\allowbreak{}:\allowbreak{} [\allowbreak{}]\allowbreak{},\allowbreak{} "\allowbreak{}anchor\_\allowbreak{}pages"\allowbreak{}:\allowbreak{} [\allowbreak{}\{\allowbreak{}"\allowbreak{}uid"\allowbreak{}:\allowbreak{} "\allowbreak{}<\allowbreak{}record\_\allowbreak{}1>\allowbreak{}"\allowbreak{},\allowbreak{} "\allowbreak{}title"\allowbreak{}:\allowbreak{} "\allowbreak{}dostarlimab"\allowbreak{},\allowbreak{} "\allowbreak{}type"\allowbreak{}:\allowbreak{} "\allowbreak{}drug"\allowbreak{},\allowbreak{} "\allowbreak{}path"\allowbreak{}:\allowbreak{} "\allowbreak{}drugs/\allowbreak{}dostarlimab.\allowbreak{}md"\allowbreak{},\allowbreak{} "\allowbreak{}n\_\allowbreak{}papers"\allowbreak{}:\allowbreak{} 238,\allowbreak{} "\allowbreak{}pmids"\allowbreak{}:\allowbreak{} [\allowbreak{}"\allowbreak{}39282229"\allowbreak{},\allowbreak{} "\allowbreak{}41904276"\allowbreak{},\allowbreak{} "\allowbreak{}35892341"\allowbreak{},\allowbreak{} "\allowbreak{}36005013"\allowbreak{},\allowbreak{} "\allowbreak{}0"\allowbreak{},\allowbreak{} "\allowbreak{}41811823"\allowbreak{},\allowbreak{} "\allowbreak{}36762991"\allowbreak{},\allowbreak{} "\allowbreak{}38365286"\allowbreak{},\allowbreak{} "\allowbreak{}38605944"\allowbreak{}]\allowbreak{},\allowbreak{} "\allowbreak{}text"\allowbreak{}:\allowbreak{} "\allowbreak{}-\allowbreak{}-\allowbreak{}-\allowbreak{} uid:\allowbreak{} <\allowbreak{}record\_\allowbreak{}1>\allowbreak{} type:\allowbreak{} drug name:\allowbreak{} "\allowbreak{}dostarlimab"\allowbreak{} aliases:\allowbreak{} [\allowbreak{}"\allowbreak{}Dostarlimab"\allowbreak{},\allowbreak{} "\allowbreak{}Dostarlimab (\allowbreak{}Jemperli)\allowbreak{}"\allowbreak{}]\allowbreak{} n\_\allowbreak{}papers:\allowbreak{} 238 n\_\allowbreak{}claims:\allowbreak{} 388 n\_\allowbreak{}claims\_\allowbreak{}used:\allowbreak{} 13 grounded:\allowbreak{} false synth\_\allowbreak{}model:\allowbreak{} "\allowbreak{}glm-\allowbreak{}5.\allowbreak{}2"\allowbreak{} prompt\_\allowbreak{}template\_\allowbreak{}version:\allowbreak{} "\allowbreak{}entity\_\allowbreak{}page\_\allowbreak{}v1"\allowbreak{} updated:\allowbreak{} 2026-\allowbreak{}07-\allowbreak{}06 tags:\allowbreak{} [\allowbreak{}"\allowbreak{}drug"\allowbreak{}]\allowbreak{}\par
\addvspace{3pt}
\addvspace{3pt}
\par\addvspace{5pt}\noindent{\normalsize\bfseries\sffamily dostarlimab}\par\addvspace{2pt}
\addvspace{3pt}
\par\addvspace{5pt}\noindent{\small\bfseries\sffamily Overview}\par\addvspace{2pt}
\noindent Dostarlimab is an FDA-\allowbreak{}approved anti-\allowbreak{}PD-\allowbreak{}1/\allowbreak{}PD-\allowbreak{}L1 antibody widely used in cancer treatment [\allowbreak{}ref-\allowbreak{}9]\allowbreak{},\allowbreak{} [\allowbreak{}ref-\allowbreak{}11]\allowbreak{}.\allowbreak{} It is an IgG4 isotype designed to avoid tumor-\allowbreak{}reactive T cell depletion,\allowbreak{} with little to no binding to Fc or complement protein C1q [\allowbreak{}ref-\allowbreak{}7]\allowbreak{}.\allowbreak{} Its pharmacokinetic profile is characterized by a 2-\allowbreak{}compartment model with time-\allowbreak{}dependent linear elimination,\allowbreak{} a mean terminal elimination half-\allowbreak{}life of 25.\allowbreak{}4 days,\allowbreak{} a mean clearance of 0.\allowbreak{}007 L/\allowbreak{}h,\allowbreak{} and a steady state volume of distribution of 5.\allowbreak{}3 L [\allowbreak{}ref-\allowbreak{}1]\allowbreak{},\allowbreak{} [\allowbreak{}ref-\allowbreak{}2]\allowbreak{}.\allowbreak{} As a tumor-\allowbreak{}agnostic therapy based on genetic changes,\allowbreak{} dostarlimab is highly relevant for drug repurposing across multiple histologies [\allowbreak{}ref-\allowbreak{}10]\allowbreak{}.\allowbreak{}\par
\addvspace{3pt}
\par\addvspace{5pt}\noindent{\small\bfseries\sffamily Repurposing directions}\par\addvspace{2pt}
\addvspace{3pt}
\par\addvspace{5pt}\noindent{\footnotesize\bfseries\sffamily Strongly supported directions}\par\addvspace{2pt}
\noindent Dostarlimab is strongly supported as a treatment for [\allowbreak{}[\allowbreak{}dostarlimab $\rightarrow$ cancer]\allowbreak{}]\allowbreak{},\allowbreak{} with high-\allowbreak{}quality clinical trial and review evidence plus in vivo animal data indicating beneficial"\allowbreak{},\allowbreak{} "\allowbreak{}truncated\_\allowbreak{}for\_\allowbreak{}summary"\allowbreak{}:\allowbreak{} true,\allowbreak{} "\allowbreak{}full\_\allowbreak{}text\_\allowbreak{}artifact\_\allowbreak{}json\_\allowbreak{}pointer"\allowbreak{}:\allowbreak{} "\allowbreak{}/\allowbreak{}coarse/\allowbreak{}provider\_\allowbreak{}results/\allowbreak{}provider\_\allowbreak{}B/\allowbreak{}pages/\allowbreak{}0/\allowbreak{}text"\allowbreak{},\allowbreak{} "\allowbreak{}detail\_\allowbreak{}command"\allowbreak{}:\allowbreak{} "\allowbreak{}tool\_\allowbreak{}A inspect <\allowbreak{}artifact\_\allowbreak{}3>\allowbreak{} /\allowbreak{}coarse/\allowbreak{}provider\_\allowbreak{}results/\allowbreak{}provider\_\allowbreak{}B/\allowbreak{}pages/\allowbreak{}0/\allowbreak{}text"\allowbreak{}\}\allowbreak{}]\allowbreak{},\allowbreak{} "\allowbreak{}relationship\_\allowbreak{}pages"\allowbreak{}:\allowbreak{} [\allowbreak{}]\allowbreak{},\allowbreak{} "\allowbreak{}related\_\allowbreak{}page\_\allowbreak{}hints"\allowbreak{}:\allowbreak{} [\allowbreak{}\{\allowbreak{}"\allowbreak{}uid"\allowbreak{}:\allowbreak{} "\allowbreak{}<\allowbreak{}record>\allowbreak{}"\allowbreak{},\allowbreak{} "\allowbreak{}title"\allowbreak{}:\allowbreak{} "\allowbreak{}dostarlimab $\rightarrow$ endometrial cancer"\allowbreak{},\allowbreak{} "\allowbreak{}path"\allowbreak{}:\allowbreak{} "\allowbreak{}hypotheses/\allowbreak{}dostarlimab $\rightarrow$ endometrial cancer.\allowbreak{}md"\allowbreak{},\allowbreak{} "\allowbreak{}n\_\allowbreak{}papers"\allowbreak{}:\allowbreak{} 29,\allowbreak{} "\allowbreak{}type"\allowbreak{}:\allowbreak{} "\allowbreak{}hypothesis"\allowbreak{},\allowbreak{} "\allowbreak{}follow\_\allowbreak{}up\_\allowbreak{}stage"\allowbreak{}:\allowbreak{} "\allowbreak{}stage1"\allowbreak{}\}\allowbreak{},\allowbreak{} \{\allowbreak{}"\allowbreak{}uid"\allowbreak{}:\allowbreak{} "\allowbreak{}<\allowbreak{}record\_\allowbreak{}2>\allowbreak{}"\allowbreak{},\allowbreak{} "\allowbreak{}title"\allowbreak{}:\allowbreak{} "\allowbreak{}PD-\allowbreak{}1"\allowbreak{},\allowbreak{} "\allowbreak{}path"\allowbreak{}:\allowbreak{} "\allowbreak{}targets/\allowbreak{}PD-\allowbreak{}1.\allowbreak{}md"\allowbreak{},\allowbreak{} "\allowbreak{}n\_\allowbreak{}papers"\allowbreak{}:\allowbreak{} 12098,\allowbreak{} "\allowbreak{}type"\allowbreak{}:\allowbreak{} "\allowbreak{}target"\allowbreak{},\allowbreak{} "\allowbreak{}follow\_\allowbreak{}up\_\allowbreak{}stage"\allowbreak{}:\allowbreak{} "\allowbreak{}stage1"\allowbreak{}\}\allowbreak{},\allowbreak{} \{\allowbreak{}"\allowbreak{}uid"\allowbreak{}:\allowbreak{} "\allowbreak{}<\allowbreak{}record\_\allowbreak{}1>\allowbreak{}"\allowbreak{},\allowbreak{} "\allowbreak{}title"\allowbreak{}:\allowbreak{} "\allowbreak{}FC"\allowbreak{},\allowbreak{} "\allowbreak{}path"\allowbreak{}:\allowbreak{} "\allowbreak{}drugs/\allowbreak{}FC.\allowbreak{}md"\allowbreak{},\allowbreak{} "\allowbreak{}n\_\allowbreak{}papers"\allowbreak{}:\allowbreak{} 20,\allowbreak{} "\allowbreak{}type"\allowbreak{}:\allowbreak{} "\allowbreak{}drug"\allowbreak{},\allowbreak{} "\allowbreak{}follow\_\allowbreak{}up\_\allowbreak{}stage"\allowbreak{}:\allowbreak{} "\allowbreak{}stage1"\allowbreak{}\}\allowbreak{},\allowbreak{} \{\allowbreak{}"\allowbreak{}uid"\allowbreak{}:\allowbreak{} "\allowbreak{}<\allowbreak{}record>\allowbreak{}"\allowbreak{},\allowbreak{} "\allowbreak{}title"\allowbreak{}:\allowbreak{} "\allowbreak{}dostarlimab $\rightarrow$ locally advanced rectal cancer"\allowbreak{},\allowbreak{} "\allowbreak{}path"\allowbreak{}:\allowbreak{} "\allowbreak{}hypotheses/\allowbreak{}dostarlimab $\rightarrow$ locally advanced rectal cancer.\allowbreak{}md"\allowbreak{},\allowbreak{} "\allowbreak{}n\_\allowbreak{}papers"\allowbreak{}:\allowbreak{} 12,\allowbreak{} "\allowbreak{}type"\allowbreak{}:\allowbreak{} "\allowbreak{}hypothesis"\allowbreak{},\allowbreak{} "\allowbreak{}follow\_\allowbreak{}up\_\allowbreak{}stage"\allowbreak{}:\allowbreak{} "\allowbreak{}stage1"\allowbreak{}\}\allowbreak{},\allowbreak{} \{\allowbreak{}"\allowbreak{}uid"\allowbreak{}:\allowbreak{} "\allowbreak{}<\allowbreak{}record\_\allowbreak{}2>\allowbreak{}"\allowbreak{},\allowbreak{} "\allowbreak{}title"\allowbreak{}:\allowbreak{} "\allowbreak{}PD-\allowbreak{}L1"\allowbreak{},\allowbreak{} "\allowbreak{}path"\allowbreak{}:\allowbreak{} "\allowbreak{}targets/\allowbreak{}PD-\allowbreak{}L1.\allowbreak{}md"\allowbreak{},\allowbreak{} "\allowbreak{}n\_\allowbreak{}papers"\allowbreak{}:\allowbreak{} 10947,\allowbreak{} "\allowbreak{}type"\allowbreak{}:\allowbreak{} "\allowbreak{}target"\allowbreak{},\allowbreak{} "\allowbreak{}follow\_\allowbreak{}up\_\allowbreak{}stage"\allowbreak{}:\allowbreak{} "\allowbreak{}stage1"\allowbreak{}\}\allowbreak{},\allowbreak{} \{\allowbreak{}"\allowbreak{}uid"\allowbreak{}:\allowbreak{} "\allowbreak{}<\allowbreak{}record>\allowbreak{}"\allowbreak{},\allowbreak{} "\allowbreak{}title"\allowbreak{}:\allowbreak{} "\allowbreak{}dostarlimab $\rightarrow$ rectal cancer"\allowbreak{},\allowbreak{} "\allowbreak{}path"\allowbreak{}:\allowbreak{} "\allowbreak{}hypotheses/\allowbreak{}dostarlimab $\rightarrow$ rectal cancer.\allowbreak{}md"\allowbreak{},\allowbreak{} "\allowbreak{}n\_\allowbreak{}papers"\allowbreak{}:\allowbreak{} 11,\allowbreak{} "\allowbreak{}type"\allowbreak{}:\allowbreak{} "\allowbreak{}hypothesis"\allowbreak{},\allowbreak{} "\allowbreak{}follow\_\allowbreak{}up\_\allowbreak{}stage"\allowbreak{}:\allowbreak{} "\allowbreak{}stage1"\allowbreak{}\}\allowbreak{},\allowbreak{} \{\allowbreak{}"\allowbreak{}uid"\allowbreak{}:\allowbreak{} "\allowbreak{}<\allowbreak{}record\_\allowbreak{}2>\allowbreak{}"\allowbreak{},\allowbreak{} "\allowbreak{}title"\allowbreak{}:\allowbreak{} "\allowbreak{}programmed cell death protein 1 (\allowbreak{}PD-\allowbreak{}1)\allowbreak{}"\allowbreak{},\allowbreak{} "\allowbreak{}path"\allowbreak{}:\allowbreak{} "\allowbreak{}targets/\allowbreak{}programmed cell death protein 1 (\allowbreak{}PD-\allowbreak{}1)\allowbreak{}.\allowbreak{}md"\allowbreak{},\allowbreak{} "\allowbreak{}n\_\allowbreak{}papers"\allowbreak{}:\allowbreak{} 195,\allowbreak{} "\allowbreak{}type"\allowbreak{}:\allowbreak{} "\allowbreak{}target"\allowbreak{},\allowbreak{} "\allowbreak{}follow\_\allowbreak{}up\_\allowbreak{}stage"\allowbreak{}:\allowbreak{} "\allowbreak{}stage1"\allowbreak{}\}\allowbreak{}]\allowbreak{},\allowbreak{} "\allowbreak{}stage2"\allowbreak{}:\allowbreak{} \{\allowbreak{}"\allowbreak{}policy"\allowbreak{}:\allowbreak{} "\allowbreak{}suggest"\allowbreak{},\allowbreak{} "\allowbreak{}optional"\allowbreak{}:\allowbreak{} true,\allowbreak{} "\allowbreak{}suggested\_\allowbreak{}expansions"\allowbreak{}:\allowbreak{} 0,\allowbreak{} "\allowbreak{}requested\_\allowbreak{}routes"\allowbreak{}:\allowbreak{} 0,\allowbreak{} "\allowbreak{}guidance"\allowbreak{}:\allowbreak{} "\allowbreak{}Stage 1 is complete.\allowbreak{} The calling agent may answer now or request one or more compatible expansion routes only when a load-\allowbreak{}bearing weak,\allowbreak{} missing,\allowbreak{} or conflicting link could change the conclusion.\allowbreak{}"\allowbreak{}\}\allowbreak{},\allowbreak{} "\allowbreak{}report"\allowbreak{}:\allowbreak{} \{\allowbreak{}"\allowbreak{}summary"\allowbreak{}:\allowbreak{} "\allowbreak{}No complete connection path was reconstructed for Dostarlimab $\leftrightarrow$ Stage II Breast Cancer.\allowbreak{} Provider results and unresolved gaps are retained for diagnosis.\allowbreak{}"\allowbreak{},\allowbreak{} "\allowbreak{}support\_\allowbreak{}grade"\allowbreak{}:\allowbreak{} "\allowbreak{}incomplete"\allowbreak{},\allowbreak{} "\allowbreak{}conflicts"\allowbreak{}:\allowbreak{} [\allowbreak{}]\allowbreak{},\allowbreak{} "\allowbreak{}conflict\_\allowbreak{}count"\allowbreak{}:\allowbreak{} 0,\allowbreak{} "\allowbreak{}unresolved\_\allowbreak{}gaps"\allowbreak{}:\allowbreak{} [\allowbreak{}\{\allowbreak{}"\allowbreak{}gap\_\allowbreak{}id"\allowbreak{}:\allowbreak{} null,\allowbreak{} "\allowbreak{}candidate\_\allowbreak{}path\_\allowbreak{}id"\allowbreak{}:\allowbreak{} null,\allowbreak{} "\allowbreak{}subject"\allowbreak{}:\allowbreak{} "\allowbreak{}<\allowbreak{}local\_\allowbreak{}id>\allowbreak{}"\allowbreak{},\allowbreak{} "\allowbreak{}object"\allowbreak{}:\allowbreak{} null,\allowbreak{} "\allowbreak{}required\_\allowbreak{}provider"\allowbreak{}:\allowbreak{} null,\allowbreak{} "\allowbreak{}reason"\allowbreak{}:\allowbreak{} "\allowbreak{}unresolved\_\allowbreak{}or\_\allowbreak{}local\_\allowbreak{}query\_\allowbreak{}anchor"\allowbreak{}\}\allowbreak{}]\allowbreak{},\allowbreak{} "\allowbreak{}unresolved\_\allowbreak{}gap\_\allowbreak{}count"\allowbreak{}:\allowbreak{} 1,\allowbreak{} "\allowbreak{}optional\_\allowbreak{}expansions"\allowbreak{}:\allowbreak{} \{\allowbreak{}"\allowbreak{}policy"\allowbreak{}:\allowbreak{} "\allowbreak{}suggest"\allowbreak{},\allowbreak{} "\allowbreak{}suggested"\allowbreak{}:\allowbreak{} 0,\allowbreak{} "\allowbreak{}requested"\allowbreak{}:\allowbreak{} 0\}\allowbreak{},\allowbreak{} "\allowbreak{}diagnostics"\allowbreak{}:\allowbreak{} \{\allowbreak{}"\allowbreak{}non\_\allowbreak{}load\_\allowbreak{}bearing\_\allowbreak{}conflict\_\allowbreak{}count"\allowbreak{}:\allowbreak{} 1398\}\allowbreak{},\allowbreak{} "\allowbreak{}interpretation"\allowbreak{}:\allowbreak{} "\allowbreak{}Provider-\allowbreak{}native scores are shown separately and were not averaged.\allowbreak{} A support grade describes evidence structure,\allowbreak{} not treatment efficacy.\allowbreak{}"\allowbreak{}\}\allowbreak{},\allowbreak{} "\allowbreak{}graph\_\allowbreak{}warnings"\allowbreak{}:\allowbreak{} [\allowbreak{}]\allowbreak{}\}\allowbreak{}\}\allowbreak{}\par
\addvspace{3pt}
\noindent \{\allowbreak{} "\allowbreak{}stderr"\allowbreak{}:\allowbreak{} "\allowbreak{}"\allowbreak{} \}\allowbreak{}\par}
\end{stepT}
\begin{stepR}{REASONING \textbullet\ STEP 15 \textbullet\ Direct pair exhausted; pivot to the mechanism target}
\stepsub{REASONING SUMMARY}\noindent{\footnotesize \textbf{Planning related page queries} \quad \textbf{Deciding PD-\allowbreak{}L1 page query}}\par\stepsub{MODEL NARRATION}\noindent{\footnotesize\itshape The max-\allowbreak{}tier rerun still leaves the disease unresolved and shows no candidate paths.\allowbreak{} One related target page could affect confidence in biological plausibility,\allowbreak{} so I'\allowbreak{}m querying the PD-\allowbreak{}L1 hint once for any literature context connecting the checkpoint axis to breast cancer.\allowbreak{}}\par
\end{stepR}
\begin{stepT}{TOOL CALL \textbullet\ STEP 16 \textbullet\ Follow the tool-suggested related-entity hint}
\stepsub{CALL}\begin{payload}tool\_\allowbreak{}A connect "\allowbreak{}PD-\allowbreak{}L1"\allowbreak{} -\allowbreak{}-\allowbreak{}a-\allowbreak{}id <\allowbreak{}record\_\allowbreak{}2>\allowbreak{} -\allowbreak{}-\allowbreak{}stage2 suggest -\allowbreak{}-\allowbreak{}output <\allowbreak{}artifact\_\allowbreak{}4>\allowbreak{} -\allowbreak{}-\allowbreak{}summary\end{payload}\stepsub{RETURNED}{\footnotesize \noindent exit\_\allowbreak{}code:\allowbreak{} 0  tool:\allowbreak{} tool\_\allowbreak{}A  contract\_\allowbreak{}version:\allowbreak{} the domain toolkit-\allowbreak{}tools/\allowbreak{}v1 stdout:\allowbreak{} \{\allowbreak{}"\allowbreak{}output\_\allowbreak{}path"\allowbreak{}:\allowbreak{} "\allowbreak{}<\allowbreak{}artifact\_\allowbreak{}4>\allowbreak{}"\allowbreak{},\allowbreak{} "\allowbreak{}summary"\allowbreak{}:\allowbreak{} \{\allowbreak{}"\allowbreak{}schema\_\allowbreak{}version"\allowbreak{}:\allowbreak{} "\allowbreak{}tool\_\allowbreak{}A/\allowbreak{}v2"\allowbreak{},\allowbreak{} "\allowbreak{}stage"\allowbreak{}:\allowbreak{} "\allowbreak{}stage1"\allowbreak{},\allowbreak{} "\allowbreak{}workflow"\allowbreak{}:\allowbreak{} \{\allowbreak{}"\allowbreak{}review\_\allowbreak{}before\_\allowbreak{}next\_\allowbreak{}call"\allowbreak{}:\allowbreak{} true,\allowbreak{} "\allowbreak{}stage1\_\allowbreak{}follow\_\allowbreak{}up"\allowbreak{}:\allowbreak{} "\allowbreak{}Optionally query one decision-\allowbreak{}relevant related\_\allowbreak{}page\_\allowbreak{}hint for more Stage-\allowbreak{}1 literature context.\allowbreak{}"\allowbreak{},\allowbreak{} "\allowbreak{}stage1\_\allowbreak{}follow\_\allowbreak{}up\_\allowbreak{}command\_\allowbreak{}template"\allowbreak{}:\allowbreak{} "\allowbreak{}tool\_\allowbreak{}A connect <\allowbreak{}hint.\allowbreak{}title>\allowbreak{} -\allowbreak{}-\allowbreak{}a-\allowbreak{}id <\allowbreak{}hint.\allowbreak{}uid>\allowbreak{} -\allowbreak{}-\allowbreak{}stage2 suggest -\allowbreak{}-\allowbreak{}output <\allowbreak{}new-\allowbreak{}stage1-\allowbreak{}path>\allowbreak{} -\allowbreak{}-\allowbreak{}summary"\allowbreak{},\allowbreak{} "\allowbreak{}stage2"\allowbreak{}:\allowbreak{} "\allowbreak{}Only after reviewing Stage 1,\allowbreak{} optionally run one advertised ready expansion route when a deeper connection could change the conclusion.\allowbreak{}"\allowbreak{}\}\allowbreak{},\allowbreak{} "\allowbreak{}artifact\_\allowbreak{}inspection"\allowbreak{}:\allowbreak{} \{\allowbreak{}"\allowbreak{}default"\allowbreak{}:\allowbreak{} "\allowbreak{}not\_\allowbreak{}needed"\allowbreak{},\allowbreak{} "\allowbreak{}reason"\allowbreak{}:\allowbreak{} "\allowbreak{}This projection already contains provider status,\allowbreak{} selected paths,\allowbreak{} load-\allowbreak{}bearing evidence,\allowbreak{} anchor and relationship pages,\allowbreak{} related-\allowbreak{}page hints,\allowbreak{} conflicts,\allowbreak{} and optional expansion candidates.\allowbreak{}"\allowbreak{},\allowbreak{} "\allowbreak{}exception"\allowbreak{}:\allowbreak{} "\allowbreak{}Run only an item'\allowbreak{}s supplied detail\_\allowbreak{}command,\allowbreak{} once,\allowbreak{} when its truncated\_\allowbreak{}for\_\allowbreak{}summary flag is true and the omitted text is load-\allowbreak{}bearing.\allowbreak{} An absent relationship page is not a missing field and never authorizes artifact or filesystem search.\allowbreak{}"\allowbreak{},\allowbreak{} "\allowbreak{}provider\_\allowbreak{}results\_\allowbreak{}shape"\allowbreak{}:\allowbreak{} "\allowbreak{}coarse.\allowbreak{}provider\_\allowbreak{}results is an object keyed by canonical provider ID,\allowbreak{} not a list.\allowbreak{}"\allowbreak{}\}\allowbreak{},\allowbreak{} "\allowbreak{}query"\allowbreak{}:\allowbreak{} \{\allowbreak{}"\allowbreak{}id"\allowbreak{}:\allowbreak{} "\allowbreak{}query-\allowbreak{}<\allowbreak{}record>\allowbreak{}"\allowbreak{},\allowbreak{} "\allowbreak{}text"\allowbreak{}:\allowbreak{} null,\allowbreak{} "\allowbreak{}entities"\allowbreak{}:\allowbreak{} [\allowbreak{}"\allowbreak{}PD-\allowbreak{}L1"\allowbreak{}]\allowbreak{},\allowbreak{} "\allowbreak{}input\_\allowbreak{}entities"\allowbreak{}:\allowbreak{} [\allowbreak{}\{\allowbreak{}"\allowbreak{}id"\allowbreak{}:\allowbreak{} "\allowbreak{}<\allowbreak{}record\_\allowbreak{}2>\allowbreak{}"\allowbreak{},\allowbreak{} "\allowbreak{}name"\allowbreak{}:\allowbreak{} "\allowbreak{}PD-\allowbreak{}L1"\allowbreak{},\allowbreak{} "\allowbreak{}type"\allowbreak{}:\allowbreak{} "\allowbreak{}unknown"\allowbreak{},\allowbreak{} "\allowbreak{}species"\allowbreak{}:\allowbreak{} null,\allowbreak{} "\allowbreak{}aliases"\allowbreak{}:\allowbreak{} [\allowbreak{}]\allowbreak{},\allowbreak{} "\allowbreak{}resolution"\allowbreak{}:\allowbreak{} \{\allowbreak{}"\allowbreak{}source"\allowbreak{}:\allowbreak{} "\allowbreak{}caller"\allowbreak{},\allowbreak{} "\allowbreak{}structured\_\allowbreak{}input"\allowbreak{}:\allowbreak{} true\}\allowbreak{}\}\allowbreak{}]\allowbreak{},\allowbreak{} "\allowbreak{}operation"\allowbreak{}:\allowbreak{} "\allowbreak{}discover"\allowbreak{},\allowbreak{} "\allowbreak{}anchor\_\allowbreak{}quality"\allowbreak{}:\allowbreak{} \{\allowbreak{}"\allowbreak{}trusted"\allowbreak{}:\allowbreak{} true,\allowbreak{} "\allowbreak{}unresolved"\allowbreak{}:\allowbreak{} [\allowbreak{}]\allowbreak{}\}\allowbreak{}\}\allowbreak{},\allowbreak{} "\allowbreak{}resolved\_\allowbreak{}entities"\allowbreak{}:\allowbreak{} [\allowbreak{}\{\allowbreak{}"\allowbreak{}id"\allowbreak{}:\allowbreak{} "\allowbreak{}<\allowbreak{}record\_\allowbreak{}2>\allowbreak{}"\allowbreak{},\allowbreak{} "\allowbreak{}name"\allowbreak{}:\allowbreak{} "\allowbreak{}PD-\allowbreak{}L1"\allowbreak{},\allowbreak{} "\allowbreak{}type"\allowbreak{}:\allowbreak{} "\allowbreak{}gene"\allowbreak{},\allowbreak{} "\allowbreak{}species"\allowbreak{}:\allowbreak{} "\allowbreak{}Homo sapiens"\allowbreak{},\allowbreak{} "\allowbreak{}aliases"\allowbreak{}:\allowbreak{} [\allowbreak{}]\allowbreak{},\allowbreak{} "\allowbreak{}resolution"\allowbreak{}:\allowbreak{} \{\allowbreak{}"\allowbreak{}source"\allowbreak{}:\allowbreak{} "\allowbreak{}provider\_\allowbreak{}canonical\_\allowbreak{}id"\allowbreak{},\allowbreak{} "\allowbreak{}ambiguous"\allowbreak{}:\allowbreak{} false,\allowbreak{} "\allowbreak{}providers"\allowbreak{}:\allowbreak{} [\allowbreak{}"\allowbreak{}provider\_\allowbreak{}B"\allowbreak{}]\allowbreak{},\allowbreak{} "\allowbreak{}provider\_\allowbreak{}ids"\allowbreak{}:\allowbreak{} \{\allowbreak{}"\allowbreak{}provider\_\allowbreak{}B"\allowbreak{}:\allowbreak{} "\allowbreak{}<\allowbreak{}record\_\allowbreak{}2>\allowbreak{}"\allowbreak{}\}\allowbreak{},\allowbreak{} "\allowbreak{}is\_\allowbreak{}anchor"\allowbreak{}:\allowbreak{} true,\allowbreak{} "\allowbreak{}input\_\allowbreak{}aliases"\allowbreak{}:\allowbreak{} [\allowbreak{}"\allowbreak{}PD-\allowbreak{}L1"\allowbreak{}]\allowbreak{}\}\allowbreak{}\}\allowbreak{}]\allowbreak{},\allowbreak{} "\allowbreak{}providers\_\allowbreak{}run"\allowbreak{}:\allowbreak{} [\allowbreak{}\{\allowbreak{}"\allowbreak{}provider\_\allowbreak{}id"\allowbreak{}:\allowbreak{} "\allowbreak{}provider\_\allowbreak{}A"\allowbreak{},\allowbreak{} "\allowbreak{}status"\allowbreak{}:\allowbreak{} "\allowbreak{}unresolved"\allowbreak{},\allowbreak{} "\allowbreak{}algorithm\_\allowbreak{}status"\allowbreak{}:\allowbreak{} "\allowbreak{}unresolved"\allowbreak{},\allowbreak{} "\allowbreak{}algorithm\_\allowbreak{}id"\allowbreak{}:\allowbreak{} "\allowbreak{}weighted\_\allowbreak{}best\_\allowbreak{}first"\allowbreak{},\allowbreak{} "\allowbreak{}elapsed\_\allowbreak{}ms"\allowbreak{}:\allowbreak{} 11354.\allowbreak{}726,\allowbreak{} "\allowbreak{}timed\_\allowbreak{}out"\allowbreak{}:\allowbreak{} false,\allowbreak{} "\allowbreak{}budget\_\allowbreak{}exhausted"\allowbreak{}:\allowbreak{} false,\allowbreak{} "\allowbreak{}expansions"\allowbreak{}:\allowbreak{} 0,\allowbreak{} "\allowbreak{}frontier\_\allowbreak{}remaining"\allowbreak{}:\allowbreak{} 0,\allowbreak{} "\allowbreak{}early\_\allowbreak{}stopped\_\allowbreak{}on\_\allowbreak{}connection"\allowbreak{}:\allowbreak{} false,\allowbreak{} "\allowbreak{}effective\_\allowbreak{}limits"\allowbreak{}:\allowbreak{} \{\allowbreak{}"\allowbreak{}max\_\allowbreak{}expansions"\allowbreak{}:\allowbreak{} 150,\allowbreak{} "\allowbreak{}neighbor\_\allowbreak{}limit"\allowbreak{}:\allowbreak{} 5,\allowbreak{} "\allowbreak{}max\_\allowbreak{}depth"\allowbreak{}:\allowbreak{} 5,\allowbreak{} "\allowbreak{}max\_\allowbreak{}hops"\allowbreak{}:\allowbreak{} 7,\allowbreak{} "\allowbreak{}top\_\allowbreak{}k"\allowbreak{}:\allowbreak{} 5,\allowbreak{} "\allowbreak{}timeout\_\allowbreak{}seconds"\allowbreak{}:\allowbreak{} 240\}\allowbreak{},\allowbreak{} "\allowbreak{}warnings"\allowbreak{}:\allowbreak{} [\allowbreak{}]\allowbreak{},\allowbreak{} "\allowbreak{}error"\allowbreak{}:\allowbreak{} null\}\allowbreak{},\allowbreak{} \{\allowbreak{}"\allowbreak{}provider\_\allowbreak{}id"\allowbreak{}:\allowbreak{} "\allowbreak{}provider\_\allowbreak{}B"\allowbreak{},\allowbreak{} "\allowbreak{}status"\allowbreak{}:\allowbreak{} "\allowbreak{}partial"\allowbreak{},\allowbreak{} "\allowbreak{}algorithm\_\allowbreak{}status"\allowbreak{}:\allowbreak{} "\allowbreak{}partial"\allowbreak{},\allowbreak{} "\allowbreak{}algorithm\_\allowbreak{}id"\allowbreak{}:\allowbreak{} "\allowbreak{}weighted\_\allowbreak{}best\_\allowbreak{}first"\allowbreak{},\allowbreak{} "\allowbreak{}elapsed\_\allowbreak{}ms"\allowbreak{}:\allowbreak{} 11074.\allowbreak{}581,\allowbreak{} "\allowbreak{}timed\_\allowbreak{}out"\allowbreak{}:\allowbreak{} false,\allowbreak{} "\allowbreak{}budget\_\allowbreak{}exhausted"\allowbreak{}:\allowbreak{} true,\allowbreak{} "\allowbreak{}expansions"\allowbreak{}:\allowbreak{} 150,\allowbreak{} "\allowbreak{}frontier\_\allowbreak{}remaining"\allowbreak{}:\allowbreak{} 638,\allowbreak{} "\allowbreak{}early\_\allowbreak{}stopped\_\allowbreak{}on\_\allowbreak{}connection"\allowbreak{}:\allowbreak{} false,\allowbreak{} "\allowbreak{}effective\_\allowbreak{}limits"\allowbreak{}:\allowbreak{} \{\allowbreak{}"\allowbreak{}max\_\allowbreak{}expansions"\allowbreak{}:\allowbreak{} 150,\allowbreak{} "\allowbreak{}neighbor\_\allowbreak{}limit"\allowbreak{}:\allowbreak{} 5,\allowbreak{} "\allowbreak{}max\_\allowbreak{}depth"\allowbreak{}:\allowbreak{} 5,\allowbreak{} "\allowbreak{}max\_\allowbreak{}hops"\allowbreak{}:\allowbreak{} 7,\allowbreak{} "\allowbreak{}top\_\allowbreak{}k"\allowbreak{}:\allowbreak{} 5,\allowbreak{} "\allowbreak{}timeout\_\allowbreak{}seconds"\allowbreak{}:\allowbreak{} 240\}\allowbreak{},\allowbreak{} "\allowbreak{}warnings"\allowbreak{}:\allowbreak{} [\allowbreak{}"\allowbreak{}algorithm expansion budget exhausted after 150 frontier expansions;\allowbreak{} unvisited frontier remains"\allowbreak{}]\allowbreak{},\allowbreak{} "\allowbreak{}error"\allowbreak{}:\allowbreak{} null\}\allowbreak{}]\allowbreak{},\allowbreak{} "\allowbreak{}projection\_\allowbreak{}counts"\allowbreak{}:\allowbreak{} \{\allowbreak{}"\allowbreak{}candidate\_\allowbreak{}paths"\allowbreak{}:\allowbreak{} \{\allowbreak{}"\allowbreak{}shown"\allowbreak{}:\allowbreak{} 0,\allowbreak{} "\allowbreak{}total"\allowbreak{}:\allowbreak{} 0\}\allowbreak{},\allowbreak{} "\allowbreak{}connections"\allowbreak{}:\allowbreak{} \{\allowbreak{}"\allowbreak{}shown"\allowbreak{}:\allowbreak{} 0,\allowbreak{} "\allowbreak{}total"\allowbreak{}:\allowbreak{} 0\}\allowbreak{},\allowbreak{} "\allowbreak{}load\_\allowbreak{}bearing\_\allowbreak{}evidence"\allowbreak{}:\allowbreak{} \{\allowbreak{}"\allowbreak{}shown"\allowbreak{}:\allowbreak{} 0,\allowbreak{} "\allowbreak{}total\_\allowbreak{}selected\_\allowbreak{}edge\_\allowbreak{}ids"\allowbreak{}:\allowbreak{} 0\}\allowbreak{},\allowbreak{} "\allowbreak{}relationship\_\allowbreak{}pages"\allowbreak{}:\allowbreak{} \{\allowbreak{}"\allowbreak{}shown"\allowbreak{}:\allowbreak{} 0,\allowbreak{} "\allowbreak{}total"\allowbreak{}:\allowbreak{} 0,\allowbreak{} "\allowbreak{}direct\_\allowbreak{}for\_\allowbreak{}anchor\_\allowbreak{}pair\_\allowbreak{}available"\allowbreak{}:\allowbreak{} false\}\allowbreak{},\allowbreak{} "\allowbreak{}anchor\_\allowbreak{}pages"\allowbreak{}:\allowbreak{} \{\allowbreak{}"\allowbreak{}shown"\allowbreak{}:\allowbreak{} 1,\allowbreak{} "\allowbreak{}resolved\_\allowbreak{}anchors"\allowbreak{}:\allowbreak{} 1\}\allowbreak{},\allowbreak{} "\allowbreak{}related\_\allowbreak{}page\_\allowbreak{}hints"\allowbreak{}:\allowbreak{} \{\allowbreak{}"\allowbreak{}shown"\allowbreak{}:\allowbreak{} 8,\allowbreak{} "\allowbreak{}total"\allowbreak{}:\allowbreak{} 25\}\allowbreak{},\allowbreak{} "\allowbreak{}expansion\_\allowbreak{}candidates"\allowbreak{}:\allowbreak{} \{\allowbreak{}"\allowbreak{}shown"\allowbreak{}:\allowbreak{} 0,\allowbreak{} "\allowbreak{}total"\allowbreak{}:\allowbreak{} 0\}\allowbreak{}\}\allowbreak{},\allowbreak{} "\allowbreak{}candidate\_\allowbreak{}paths"\allowbreak{}:\allowbreak{} [\allowbreak{}]\allowbreak{},\allowbreak{} "\allowbreak{}expansion\_\allowbreak{}candidates"\allowbreak{}:\allowbreak{} [\allowbreak{}]\allowbreak{},\allowbreak{} "\allowbreak{}connections"\allowbreak{}:\allowbreak{} [\allowbreak{}]\allowbreak{},\allowbreak{} "\allowbreak{}load\_\allowbreak{}bearing\_\allowbreak{}evidence"\allowbreak{}:\allowbreak{} [\allowbreak{}]\allowbreak{},\allowbreak{} "\allowbreak{}anchor\_\allowbreak{}pages"\allowbreak{}:\allowbreak{} [\allowbreak{}\{\allowbreak{}"\allowbreak{}uid"\allowbreak{}:\allowbreak{} "\allowbreak{}<\allowbreak{}record\_\allowbreak{}2>\allowbreak{}"\allowbreak{},\allowbreak{} "\allowbreak{}title"\allowbreak{}:\allowbreak{} "\allowbreak{}PD-\allowbreak{}L1"\allowbreak{},\allowbreak{} "\allowbreak{}type"\allowbreak{}:\allowbreak{} "\allowbreak{}target"\allowbreak{},\allowbreak{} "\allowbreak{}path"\allowbreak{}:\allowbreak{} "\allowbreak{}targets/\allowbreak{}PD-\allowbreak{}L1.\allowbreak{}md"\allowbreak{},\allowbreak{} "\allowbreak{}n\_\allowbreak{}papers"\allowbreak{}:\allowbreak{} 10947,\allowbreak{} "\allowbreak{}pmids"\allowbreak{}:\allowbreak{} [\allowbreak{}"\allowbreak{}36969168"\allowbreak{},\allowbreak{} "\allowbreak{}38565806"\allowbreak{},\allowbreak{} "\allowbreak{}27764774"\allowbreak{},\allowbreak{} "\allowbreak{}41693025"\allowbreak{},\allowbreak{} "\allowbreak{}32509787"\allowbreak{},\allowbreak{} "\allowbreak{}41059490"\allowbreak{},\allowbreak{} "\allowbreak{}36564129"\allowbreak{},\allowbreak{} "\allowbreak{}34759534"\allowbreak{},\allowbreak{} "\allowbreak{}40251364"\allowbreak{},\allowbreak{} "\allowbreak{}38254708"\allowbreak{},\allowbreak{} "\allowbreak{}39962074"\allowbreak{},\allowbreak{} "\allowbreak{}36231138"\allowbreak{},\allowbreak{} "\allowbreak{}33014787"\allowbreak{}]\allowbreak{},\allowbreak{} "\allowbreak{}text"\allowbreak{}:\allowbreak{} "\allowbreak{}-\allowbreak{}-\allowbreak{}-\allowbreak{} uid:\allowbreak{} <\allowbreak{}record\_\allowbreak{}2>\allowbreak{} type:\allowbreak{} target name:\allowbreak{} "\allowbreak{}PD-\allowbreak{}L1"\allowbreak{} aliases:\allowbreak{} [\allowbreak{}"\allowbreak{}Pd-\allowbreak{}l1"\allowbreak{},\allowbreak{} "\allowbreak{}PD-\allowbreak{}L1 (\allowbreak{}combined positivity score)\allowbreak{}"\allowbreak{},\allowbreak{} "\allowbreak{}PD-\allowbreak{}L1 (\allowbreak{}CD274)\allowbreak{}"\allowbreak{},\allowbreak{} "\allowbreak{}Pd-\allowbreak{}L1"\allowbreak{},\allowbreak{} "\allowbreak{}pd-\allowbreak{}l1"\allowbreak{},\allowbreak{} "\allowbreak{}PD-\allowbreak{}L1 (\allowbreak{}CPS=\allowbreak{}105)\allowbreak{}"\allowbreak{},\allowbreak{} "\allowbreak{}PD-\allowbreak{}L1 (\allowbreak{}programmed death-\allowbreak{}ligand 1)\allowbreak{}"\allowbreak{},\allowbreak{} "\allowbreak{}Pd-\allowbreak{}l1 (\allowbreak{}Cd274)\allowbreak{}"\allowbreak{},\allowbreak{} "\allowbreak{}PD-\allowbreak{}L1 (\allowbreak{}B7-\allowbreak{}H1)\allowbreak{}"\allowbreak{}]\allowbreak{} n\_\allowbreak{}papers:\allowbreak{} 10947 n\_\allowbreak{}claims:\allowbreak{} 20406 n\_\allowbreak{}claims\_\allowbreak{}used:\allowbreak{} 400 grounded:\allowbreak{} false synth\_\allowbreak{}model:\allowbreak{} "\allowbreak{}glm-\allowbreak{}5.\allowbreak{}2"\allowbreak{} prompt\_\allowbreak{}template\_\allowbreak{}version:\allowbreak{} "\allowbreak{}entity\_\allowbreak{}page\_\allowbreak{}v1"\allowbreak{} updated:\allowbreak{} 2026-\allowbreak{}07-\allowbreak{}06 tags:\allowbreak{} [\allowbreak{}"\allowbreak{}target"\allowbreak{}]\allowbreak{}\par
\addvspace{3pt}
\addvspace{3pt}
\par\addvspace{5pt}\noindent{\normalsize\bfseries\sffamily PD-\allowbreak{}L1}\par\addvspace{2pt}
\addvspace{3pt}
\par\addvspace{5pt}\noindent{\small\bfseries\sffamily Overview}\par\addvspace{2pt}
\noindent PD-\allowbreak{}L1 (\allowbreak{}programmed death-\allowbreak{}ligand 1)\allowbreak{},\allowbreak{} also known as CD274 or B7-\allowbreak{}H1,\allowbreak{} is an immune checkpoint protein and a critical target in oncology and drug repurposing.\allowbreak{} PD-\allowbreak{}L1 expression (\allowbreak{}measured via combined positivity score,\allowbreak{} tumor proportion score,\allowbreak{} or tumor cell/\allowbreak{}immune cell algorithms)\allowbreak{} frequently predicts survival and response to immune checkpoint inhibitors across multiple advanced malignancies [\allowbreak{}ref-\allowbreak{}248-\allowbreak{}257,\allowbreak{} ref-\allowbreak{}277,\allowbreak{} ref-\allowbreak{}294,\allowbreak{} ref-\allowbreak{}297]\allowbreak{}.\allowbreak{} While its primary clinical application is in cancer therapy,\allowbreak{} non-\allowbreak{}traditional repurposing approaches-\allowbreak{}-\allowbreak{}-\allowbreak{}such as managing sepsis with BMS-\allowbreak{}936559 [\allowbreak{}ref-\allowbreak{}298]\allowbreak{} or using 89Zr-\allowbreak{}labelled atezolizumab as an imaging tracer to predict clinical response [\allowbreak{}ref-\allowbreak{}220]\allowbreak{}-\allowbreak{}-\allowbreak{}-\allowbreak{}highlight its expanding therapeutic footprint.\allowbreak{}\par
\addvspace{3pt}
\par\addvspace{5pt}\noindent{\small\bfseries\sffamily Repurposing directions}\par\addvspace{2pt}
\noindent PD-\allowbreak{}L1 blockade de"\allowbreak{},\allowbreak{} "\allowbreak{}truncated\_\allowbreak{}for\_\allowbreak{}summary"\allowbreak{}:\allowbreak{} true,\allowbreak{} "\allowbreak{}full\_\allowbreak{}text\_\allowbreak{}artifact\_\allowbreak{}json\_\allowbreak{}pointer"\allowbreak{}:\allowbreak{} "\allowbreak{}/\allowbreak{}coarse/\allowbreak{}provider\_\allowbreak{}results/\allowbreak{}provider\_\allowbreak{}B/\allowbreak{}pages/\allowbreak{}0/\allowbreak{}text"\allowbreak{},\allowbreak{} "\allowbreak{}detail\_\allowbreak{}command"\allowbreak{}:\allowbreak{} "\allowbreak{}tool\_\allowbreak{}A inspect <\allowbreak{}artifact\_\allowbreak{}4>\allowbreak{} /\allowbreak{}coarse/\allowbreak{}provider\_\allowbreak{}results/\allowbreak{}provider\_\allowbreak{}B/\allowbreak{}pages/\allowbreak{}0/\allowbreak{}text"\allowbreak{}\}\allowbreak{}]\allowbreak{},\allowbreak{} "\allowbreak{}relationship\_\allowbreak{}pages"\allowbreak{}:\allowbreak{} [\allowbreak{}]\allowbreak{},\allowbreak{} "\allowbreak{}related\_\allowbreak{}page\_\allowbreak{}hints"\allowbreak{}:\allowbreak{} [\allowbreak{}\{\allowbreak{}"\allowbreak{}uid"\allowbreak{}:\allowbreak{} "\allowbreak{}<\allowbreak{}record>\allowbreak{}"\allowbreak{},\allowbreak{} "\allowbreak{}title"\allowbreak{}:\allowbreak{} "\allowbreak{}atezolizumab $\rightarrow$ triple-\allowbreak{}negative breast cancer"\allowbreak{},\allowbreak{} "\allowbreak{}path"\allowbreak{}:\allowbreak{} "\allowbreak{}hypotheses/\allowbreak{}atezolizumab $\rightarrow$ triple-\allowbreak{}negative breast cancer.\allowbreak{}md"\allowbreak{},\allowbreak{} "\allowbreak{}n\_\allowbreak{}papers"\allowbreak{}:\allowbreak{} 221,\allowbreak{} "\allowbreak{}type"\allowbreak{}:\allowbreak{} "\allowbreak{}hypothesis"\allowbreak{},\allowbreak{} "\allowbreak{}follow\_\allowbreak{}up\_\allowbreak{}stage"\allowbreak{}:\allowbreak{} "\allowbreak{}stage1"\allowbreak{}\}\allowbreak{},\allowbreak{} \{\allowbreak{}"\allowbreak{}uid"\allowbreak{}:\allowbreak{} "\allowbreak{}<\allowbreak{}record\_\allowbreak{}2>\allowbreak{}"\allowbreak{},\allowbreak{} "\allowbreak{}title"\allowbreak{}:\allowbreak{} "\allowbreak{}IFNγ"\allowbreak{},\allowbreak{} "\allowbreak{}path"\allowbreak{}:\allowbreak{} "\allowbreak{}targets/\allowbreak{}IFNγ.\allowbreak{}md"\allowbreak{},\allowbreak{} "\allowbreak{}n\_\allowbreak{}papers"\allowbreak{}:\allowbreak{} 777,\allowbreak{} "\allowbreak{}type"\allowbreak{}:\allowbreak{} "\allowbreak{}target"\allowbreak{},\allowbreak{} "\allowbreak{}follow\_\allowbreak{}up\_\allowbreak{}stage"\allowbreak{}:\allowbreak{} "\allowbreak{}stage1"\allowbreak{}\}\allowbreak{},\allowbreak{} \{\allowbreak{}"\allowbreak{}uid"\allowbreak{}:\allowbreak{} "\allowbreak{}disease:\allowbreak{}<\allowbreak{}record>\allowbreak{}"\allowbreak{},\allowbreak{} "\allowbreak{}title"\allowbreak{}:\allowbreak{} "\allowbreak{}cancer"\allowbreak{},\allowbreak{} "\allowbreak{}path"\allowbreak{}:\allowbreak{} "\allowbreak{}diseases/\allowbreak{}cancer.\allowbreak{}md"\allowbreak{},\allowbreak{} "\allowbreak{}n\_\allowbreak{}papers"\allowbreak{}:\allowbreak{} 71479,\allowbreak{} "\allowbreak{}type"\allowbreak{}:\allowbreak{} "\allowbreak{}disease"\allowbreak{},\allowbreak{} "\allowbreak{}follow\_\allowbreak{}up\_\allowbreak{}stage"\allowbreak{}:\allowbreak{} "\allowbreak{}stage1"\allowbreak{}\}\allowbreak{},\allowbreak{} \{\allowbreak{}"\allowbreak{}uid"\allowbreak{}:\allowbreak{} "\allowbreak{}<\allowbreak{}record\_\allowbreak{}1>\allowbreak{}"\allowbreak{},\allowbreak{} "\allowbreak{}title"\allowbreak{}:\allowbreak{} "\allowbreak{}pembrolizumab"\allowbreak{},\allowbreak{} "\allowbreak{}path"\allowbreak{}:\allowbreak{} "\allowbreak{}drugs/\allowbreak{}pembrolizumab.\allowbreak{}md"\allowbreak{},\allowbreak{} "\allowbreak{}n\_\allowbreak{}papers"\allowbreak{}:\allowbreak{} 8116,\allowbreak{} "\allowbreak{}type"\allowbreak{}:\allowbreak{} "\allowbreak{}drug"\allowbreak{},\allowbreak{} "\allowbreak{}follow\_\allowbreak{}up\_\allowbreak{}stage"\allowbreak{}:\allowbreak{} "\allowbreak{}stage1"\allowbreak{}\}\allowbreak{},\allowbreak{} \{\allowbreak{}"\allowbreak{}uid"\allowbreak{}:\allowbreak{} "\allowbreak{}mechanism:\allowbreak{}<\allowbreak{}record>\allowbreak{}"\allowbreak{},\allowbreak{} "\allowbreak{}title"\allowbreak{}:\allowbreak{} "\allowbreak{}PD-\allowbreak{}L1 blockade"\allowbreak{},\allowbreak{} "\allowbreak{}path"\allowbreak{}:\allowbreak{} "\allowbreak{}mechanisms/\allowbreak{}PD-\allowbreak{}L1 blockade.\allowbreak{}md"\allowbreak{},\allowbreak{} "\allowbreak{}n\_\allowbreak{}papers"\allowbreak{}:\allowbreak{} 88,\allowbreak{} "\allowbreak{}type"\allowbreak{}:\allowbreak{} "\allowbreak{}mechanism"\allowbreak{},\allowbreak{} "\allowbreak{}follow\_\allowbreak{}up\_\allowbreak{}stage"\allowbreak{}:\allowbreak{} "\allowbreak{}stage1"\allowbreak{}\}\allowbreak{},\allowbreak{} \{\allowbreak{}"\allowbreak{}uid"\allowbreak{}:\allowbreak{} "\allowbreak{}<\allowbreak{}record>\allowbreak{}"\allowbreak{},\allowbreak{} "\allowbreak{}title"\allowbreak{}:\allowbreak{} "\allowbreak{}atezolizumab $\rightarrow$ hepatocellular carcinoma"\allowbreak{},\allowbreak{} "\allowbreak{}path"\allowbreak{}:\allowbreak{} "\allowbreak{}hypotheses/\allowbreak{}atezolizumab $\rightarrow$ hepatocellular carcinoma.\allowbreak{}md"\allowbreak{},\allowbreak{} "\allowbreak{}n\_\allowbreak{}papers"\allowbreak{}:\allowbreak{} 145,\allowbreak{} "\allowbreak{}type"\allowbreak{}:\allowbreak{} "\allowbreak{}hypothesis"\allowbreak{},\allowbreak{} "\allowbreak{}follow\_\allowbreak{}up\_\allowbreak{}stage"\allowbreak{}:\allowbreak{} "\allowbreak{}stage1"\allowbreak{}\}\allowbreak{},\allowbreak{} \{\allowbreak{}"\allowbreak{}uid"\allowbreak{}:\allowbreak{} "\allowbreak{}<\allowbreak{}record\_\allowbreak{}1>\allowbreak{}"\allowbreak{},\allowbreak{} "\allowbreak{}title"\allowbreak{}:\allowbreak{} "\allowbreak{}nivolumab"\allowbreak{},\allowbreak{} "\allowbreak{}path"\allowbreak{}:\allowbreak{} "\allowbreak{}drugs/\allowbreak{}nivolumab.\allowbreak{}md"\allowbreak{},\allowbreak{} "\allowbreak{}n\_\allowbreak{}papers"\allowbreak{}:\allowbreak{} 6147,\allowbreak{} "\allowbreak{}type"\allowbreak{}:\allowbreak{} "\allowbreak{}drug"\allowbreak{},\allowbreak{} "\allowbreak{}follow\_\allowbreak{}up\_\allowbreak{}stage"\allowbreak{}:\allowbreak{} "\allowbreak{}stage1"\allowbreak{}\}\allowbreak{},\allowbreak{} \{\allowbreak{}"\allowbreak{}uid"\allowbreak{}:\allowbreak{} "\allowbreak{}<\allowbreak{}record>\allowbreak{}"\allowbreak{},\allowbreak{} "\allowbreak{}title"\allowbreak{}:\allowbreak{} "\allowbreak{}nivolumab $\rightarrow$ renal cell carcinoma"\allowbreak{},\allowbreak{} "\allowbreak{}path"\allowbreak{}:\allowbreak{} "\allowbreak{}hypotheses/\allowbreak{}nivolumab $\rightarrow$ renal cell carcinoma.\allowbreak{}md"\allowbreak{},\allowbreak{} "\allowbreak{}n\_\allowbreak{}papers"\allowbreak{}:\allowbreak{} 71,\allowbreak{} "\allowbreak{}type"\allowbreak{}:\allowbreak{} "\allowbreak{}hypothesis"\allowbreak{},\allowbreak{} "\allowbreak{}follow\_\allowbreak{}up\_\allowbreak{}stage"\allowbreak{}:\allowbreak{} "\allowbreak{}stage1"\allowbreak{}\}\allowbreak{}]\allowbreak{},\allowbreak{} "\allowbreak{}stage2"\allowbreak{}:\allowbreak{} \{\allowbreak{}"\allowbreak{}policy"\allowbreak{}:\allowbreak{} "\allowbreak{}suggest"\allowbreak{},\allowbreak{} "\allowbreak{}optional"\allowbreak{}:\allowbreak{} true,\allowbreak{} "\allowbreak{}suggested\_\allowbreak{}expansions"\allowbreak{}:\allowbreak{} 0,\allowbreak{} "\allowbreak{}requested\_\allowbreak{}routes"\allowbreak{}:\allowbreak{} 0,\allowbreak{} "\allowbreak{}guidance"\allowbreak{}:\allowbreak{} "\allowbreak{}Stage 1 is complete.\allowbreak{} The calling agent may answer now or request one or more compatible expansion routes only when a load-\allowbreak{}bearing weak,\allowbreak{} missing,\allowbreak{} or conflicting link could change the conclusion.\allowbreak{}"\allowbreak{}\}\allowbreak{},\allowbreak{} "\allowbreak{}report"\allowbreak{}:\allowbreak{} \{\allowbreak{}"\allowbreak{}summary"\allowbreak{}:\allowbreak{} "\allowbreak{}No complete connection path was reconstructed for PD-\allowbreak{}L1.\allowbreak{} Provider results and unresolved gaps are retained for diagnosis.\allowbreak{}"\allowbreak{},\allowbreak{} "\allowbreak{}support\_\allowbreak{}grade"\allowbreak{}:\allowbreak{} "\allowbreak{}incomplete"\allowbreak{},\allowbreak{} "\allowbreak{}conflicts"\allowbreak{}:\allowbreak{} [\allowbreak{}]\allowbreak{},\allowbreak{} "\allowbreak{}conflict\_\allowbreak{}count"\allowbreak{}:\allowbreak{} 0,\allowbreak{} "\allowbreak{}unresolved\_\allowbreak{}gaps"\allowbreak{}:\allowbreak{} [\allowbreak{}]\allowbreak{},\allowbreak{} "\allowbreak{}unresolved\_\allowbreak{}gap\_\allowbreak{}count"\allowbreak{}:\allowbreak{} 0,\allowbreak{} "\allowbreak{}optional\_\allowbreak{}expansions"\allowbreak{}:\allowbreak{} \{\allowbreak{}"\allowbreak{}policy"\allowbreak{}:\allowbreak{} "\allowbreak{}suggest"\allowbreak{},\allowbreak{} "\allowbreak{}suggested"\allowbreak{}:\allowbreak{} 0,\allowbreak{} "\allowbreak{}requested"\allowbreak{}:\allowbreak{} 0\}\allowbreak{},\allowbreak{} "\allowbreak{}diagnostics"\allowbreak{}:\allowbreak{} \{\allowbreak{}"\allowbreak{}non\_\allowbreak{}load\_\allowbreak{}bearing\_\allowbreak{}conflict\_\allowbreak{}count"\allowbreak{}:\allowbreak{} 226\}\allowbreak{},\allowbreak{} "\allowbreak{}interpretation"\allowbreak{}:\allowbreak{} "\allowbreak{}Provider-\allowbreak{}native scores are shown separately and were not averaged.\allowbreak{} A support grade describes evidence structure,\allowbreak{} not treatment efficacy.\allowbreak{}"\allowbreak{}\}\allowbreak{},\allowbreak{} "\allowbreak{}graph\_\allowbreak{}warnings"\allowbreak{}:\allowbreak{} [\allowbreak{}]\allowbreak{}\}\allowbreak{}\}\allowbreak{}\par
\addvspace{3pt}
\noindent \{\allowbreak{} "\allowbreak{}stderr"\allowbreak{}:\allowbreak{} "\allowbreak{}"\allowbreak{} \}\allowbreak{}\par}
\end{stepT}
\begin{stepR}{REASONING \textbullet\ STEP 17 \textbullet\ Classify the hint as class-adjacent, not direct evidence}
\stepsub{REASONING SUMMARY}\noindent{\footnotesize \textbf{Evaluating detail\_\allowbreak{}command usage limits} \quad \textbf{Scoring dostarlimab relevance for breast cancer}}\par\stepsub{MODEL NARRATION}\noindent{\footnotesize\itshape I have enough bounded evidence to make a calibrated call:\allowbreak{} strong checkpoint-\allowbreak{}mechanism plausibility,\allowbreak{} but no resolved stage-\allowbreak{}II breast-\allowbreak{}cancer anchor or direct dostarlimab-\allowbreak{}breast-\allowbreak{}cancer path in the sanctioned KG summaries.\allowbreak{} Submitting now.\allowbreak{}}\par
\end{stepR}
\begin{stepS}{SUBMISSION \textbullet\ STEP 18 \textbullet\ Submit}
\stepsub{FINAL ANSWER}\noindent{\footnotesize \textbf{promisingness} 0.3 \quad \textbf{confidence} 0.48\par\smallskip\noindent Dostarlimab is a validated anti-\allowbreak{}PD-\allowbreak{}1/\allowbreak{}PD-\allowbreak{}L1 checkpoint antibody with tissue-\allowbreak{}agnostic and multiple-\allowbreak{}cancer activity,\allowbreak{} including dMMR solid tumors,\allowbreak{} so the mechanism is biologically plausible in immunogenic breast-\allowbreak{}cancer subsets.\allowbreak{} However,\allowbreak{} the mandated tool\_\allowbreak{}A searches did not resolve Stage II Breast Cancer as a grounded disease anchor and reconstructed no candidate paths,\allowbreak{} connections,\allowbreak{} relationship pages,\allowbreak{} or load-\allowbreak{}bearing evidence for Dostarlimab to Stage II Breast Cancer even at the max tier.\allowbreak{} A PD-\allowbreak{}L1 related-\allowbreak{}page query supports checkpoint relevance in oncology and surfaced triple-\allowbreak{}negative breast cancer evidence for another PD-\allowbreak{}L1 inhibitor,\allowbreak{} but that is class-\allowbreak{}adjacent rather than dostarlimab-\allowbreak{}specific or stage-\allowbreak{}II-\allowbreak{}specific.\allowbreak{} I therefore score the pair below Phase-\allowbreak{}2-\allowbreak{}success territory but above purely preclinical,\allowbreak{} reflecting plausible repurposing with no sanctioned direct clinical support for this exact pair.\allowbreak{}}\par\stepsub{KEY CLAIMS (7)}\begin{itemize}[leftmargin=1.2em,itemsep=3pt,topsep=1pt]
\item \footnotesize Dostarlimab was resolved as a drug anchor,\allowbreak{} but Stage II Breast Cancer remained unresolved or ambiguous in the Stage-\allowbreak{}1 search.\allowbreak{}\newline{\scriptsize\ttfamily source: tool:\allowbreak{}tool\_\allowbreak{}A connect Dostarlimab + Stage II Breast Cancer -\allowbreak{}-\allowbreak{}tier extra}\newline{\scriptsize support: resolved\_\allowbreak{}entities included DOSTARLIMAB as type drug,\allowbreak{} while anchor\_\allowbreak{}quality listed unresolved <\allowbreak{}local\_\allowbreak{}id>\allowbreak{} with reason/\allowbreak{}source surface\_\allowbreak{}form and ambiguous true.\allowbreak{}}
\item \footnotesize No complete Dostarlimab to Stage II Breast Cancer connection was reconstructed in the bounded Stage-\allowbreak{}1 search.\allowbreak{}\newline{\scriptsize\ttfamily source: tool:\allowbreak{}tool\_\allowbreak{}A connect Dostarlimab + Stage II Breast Cancer -\allowbreak{}-\allowbreak{}tier extra}\newline{\scriptsize support: report.\allowbreak{}summary:\allowbreak{} No complete connection path was reconstructed for Dostarlimab $\leftrightarrow$ Stage II Breast Cancer;\allowbreak{} candidate\_\allowbreak{}paths,\allowbreak{} connections,\allowbreak{} load\_\allowbreak{}bearing\_\allowbreak{}evidence,\allowbreak{} and relationship\_\allowbreak{}pages were all empty.\allowbreak{}}
\item \footnotesize The max-\allowbreak{}tier rerun still found no direct path or relationship evidence for the exact pair,\allowbreak{} leaving the disease anchor unresolved.\allowbreak{}\newline{\scriptsize\ttfamily source: tool:\allowbreak{}tool\_\allowbreak{}A connect Dostarlimab + Stage II Breast Cancer -\allowbreak{}-\allowbreak{}tier max}\newline{\scriptsize support: report.\allowbreak{}summary:\allowbreak{} No complete connection path was reconstructed for Dostarlimab $\leftrightarrow$ Stage II Breast Cancer;\allowbreak{} unresolved\_\allowbreak{}gaps reason:\allowbreak{} unresolved\_\allowbreak{}or\_\allowbreak{}local\_\allowbreak{}query\_\allowbreak{}anchor;\allowbreak{} candidate\_\allowbreak{}paths shown 0 total 0 and connections shown 0 total 0.\allowbreak{}}
\item \footnotesize Dostarlimab is an FDA-\allowbreak{}approved anti-\allowbreak{}PD-\allowbreak{}1/\allowbreak{}PD-\allowbreak{}L1 antibody with cancer-\allowbreak{}treatment use and tumor-\allowbreak{}agnostic relevance.\allowbreak{}\newline{\scriptsize\ttfamily source: PMID:\allowbreak{}39282229;\allowbreak{} PMID:\allowbreak{}41904276;\allowbreak{} PMID:\allowbreak{}0 via <\allowbreak{}record\_\allowbreak{}1>\allowbreak{} page}\newline{\scriptsize support: Overview:\allowbreak{} Dostarlimab is an FDA-\allowbreak{}approved anti-\allowbreak{}PD-\allowbreak{}1/\allowbreak{}PD-\allowbreak{}L1 antibody widely used in cancer treatment [\allowbreak{}ref-\allowbreak{}9]\allowbreak{},\allowbreak{} [\allowbreak{}ref-\allowbreak{}11]\allowbreak{}.\allowbreak{}.\allowbreak{}.\allowbreak{} As a tumor-\allowbreak{}agnostic therapy based on genetic changes,\allowbreak{} dostarlimab is highly relevant for drug repurposing across multiple histologies [\allowbreak{}ref-\allowbreak{}10]\allowbreak{}.\allowbreak{}}
\item \footnotesize Dostarlimab has strong support in dMMR solid tumors and several non-\allowbreak{}breast cancers,\allowbreak{} including endometrial and rectal/\allowbreak{}colorectal settings,\allowbreak{} but the retrieved drug page did not list breast cancer as a supported direction.\allowbreak{}\newline{\scriptsize\ttfamily source: <\allowbreak{}record\_\allowbreak{}1>\allowbreak{} page inspected by tool\_\allowbreak{}A inspect <\allowbreak{}artifact\_\allowbreak{}1>\allowbreak{} /\allowbreak{}coarse/\allowbreak{}provider\_\allowbreak{}results/\allowbreak{}provider\_\allowbreak{}B/\allowbreak{}pages/\allowbreak{}0/\allowbreak{}text}\newline{\scriptsize support: Strongly supported directions listed cancer,\allowbreak{} advanced solid tumors,\allowbreak{} dMMR solid tumors,\allowbreak{} endometrial malignancies,\allowbreak{} rectal cancer,\allowbreak{} locally advanced rectal cancer,\allowbreak{} colorectal cancer,\allowbreak{} and NSCLC;\allowbreak{} no breast cancer direction appeared in that section.\allowbreak{}}
\item \footnotesize The mechanism is checkpoint blockade:\allowbreak{} dostarlimab binds PD-\allowbreak{}1 and suppresses PD-\allowbreak{}1 interactions with PD-\allowbreak{}L1/\allowbreak{}PD-\allowbreak{}L2 to restore T-\allowbreak{}cell activity.\allowbreak{}\newline{\scriptsize\ttfamily source: PMID:\allowbreak{}36762991;\allowbreak{} PMID:\allowbreak{}38365286;\allowbreak{} PMID:\allowbreak{}38605944;\allowbreak{} PMID:\allowbreak{}36005013 via <\allowbreak{}record\_\allowbreak{}1>\allowbreak{} page}\newline{\scriptsize support: Dostarlimab acts primarily by binding with high affinity (\allowbreak{}KD 300 pM)\allowbreak{} to human and cynomolgus monkey pd-\allowbreak{}1.\allowbreak{}.\allowbreak{}.\allowbreak{} As a functional antagonist,\allowbreak{} it suppresses the interaction between PD-\allowbreak{}1 and pd-\allowbreak{}l1 to support T cell activation.\allowbreak{}.\allowbreak{}.\allowbreak{} inhibits.\allowbreak{}.\allowbreak{}.\allowbreak{} PD-\allowbreak{}L2 and PD-\allowbreak{}1.\allowbreak{}.\allowbreak{}.\allowbreak{} blockade restores T cell activity and enhances immune infiltration.\allowbreak{}}
\item \footnotesize PD-\allowbreak{}L1 is a broad oncology checkpoint target and the related-\allowbreak{}page hints surfaced class-\allowbreak{}adjacent triple-\allowbreak{}negative breast cancer evidence for atezolizumab,\allowbreak{} supporting plausibility but not direct dostarlimab evidence.\allowbreak{}\newline{\scriptsize\ttfamily source: tool:\allowbreak{}tool\_\allowbreak{}A connect PD-\allowbreak{}L1 -\allowbreak{}-\allowbreak{}a-\allowbreak{}id <\allowbreak{}record\_\allowbreak{}2>\allowbreak{}}\newline{\scriptsize support: PD-\allowbreak{}L1 Overview:\allowbreak{} PD-\allowbreak{}L1.\allowbreak{}.\allowbreak{}.\allowbreak{} is an immune checkpoint protein and a critical target in oncology.\allowbreak{}.\allowbreak{}.\allowbreak{} related\_\allowbreak{}page\_\allowbreak{}hints included title:\allowbreak{} atezolizumab $\rightarrow$ triple-\allowbreak{}negative breast cancer,\allowbreak{} n\_\allowbreak{}papers:\allowbreak{} 221.\allowbreak{}}
\end{itemize}

\end{stepS}

\subsection{Process score}

{\footnotesize
\setlength{\LTcapwidth}{\textwidth}
\begin{xltabular}{\textwidth}{@{}c c >{\raggedright\arraybackslash}p{0.155\textwidth} X@{}}
\caption{\textsc{traces} process score with all sub-rubrics. A capability is the weighted mean of
its sub-rubric bands ($0$/$1$/$2$) rescaled to $[0,4]$. A sub-rubric marked \textsc{critical} caps its
capability when it falls below the top band. \textsc{nr} marks a sub-rubric the trajectory gave no
opportunity to exercise; it is excluded from the mean rather than scored zero.}
\label{tab:case-traces}\\
\toprule
\endfirsthead
\multicolumn{4}{@{}l}{\footnotesize\itshape Table~\thetable\ (continued)}\\
\toprule
\endhead
\endfoot
\bottomrule
\endlastfoot
\midrule
\multicolumn{4}{@{}l@{}}{%
\capbadge{C}\; \textbf{Long-horizon State Coherence}\hfill \textbf{\large 3.00}\,/\,4}\\[2pt]
\multicolumn{4}{@{}p{\dimexpr\textwidth-2\tabcolsep}@{}}{%
\footnotesize\itshape The assigned pair is held from task intake through both searches to submission; the final answer is built from what was retrieved rather than reasoning around it. Two bands are lost: the supplied identifiers were available but were not used to anchor the search, and the reported confidence is not tied to a named retrieved quantity.}\\[3pt]
& \bandbox{2} & \footnotesize Cross-step integration & \footnotesize The final judgement traceably incorporates the main tool\_\allowbreak{}A returns:\allowbreak{} the max-\allowbreak{}tier query'\allowbreak{}s lack of a reconstructed connection and zero candidate paths are explicitly used to lower the promisingness,\allowbreak{} while the PD-\allowbreak{}L1 hint is used only as class-\allowbreak{}adjacent plausibility rather than direct evidence.\allowbreak{} The submission also lists key claims tying each conclusion to the extra/\allowbreak{}max queries,\allowbreak{} the inspected drug page,\allowbreak{} and the PD-\allowbreak{}L1 related-\allowbreak{}page query,\allowbreak{} so retrieved evidence entries are folded into the final score rather than ignored.\allowbreak{} \\
\addlinespace[1.5pt]
& \bandbox{2} & \footnotesize State \& constraint retention\;\critbadge & \footnotesize The trajectory consistently centers the supplied pair from task inspection through the main tool\_\allowbreak{}A calls,\allowbreak{} using sanctioned offline the domain toolkit tooling for the exact Dostarlimab + Stage II Breast Cancer query.\allowbreak{} The final status claim that no direct connection was found is directly retrieved from the max-\allowbreak{}tier sanctioned result.\allowbreak{} Although the agent queries PD-\allowbreak{}L1 and notes atezolizumab/\allowbreak{}triple-\allowbreak{}negative breast cancer as plausibility context,\allowbreak{} it explicitly does not present that as direct evidence for the asked pair,\allowbreak{} calling it class-\allowbreak{}adjacent rather than dostarlimab-\allowbreak{} or stage-\allowbreak{}II-\allowbreak{}specific.\allowbreak{} \\
\addlinespace[1.5pt]
& \bandbox{1} & \footnotesize Task decomposition & \footnotesize The trajectory had a real dependency-\allowbreak{}ordered workflow:\allowbreak{} it inspected the task,\allowbreak{} made sanctioned tool\_\allowbreak{}A calls,\allowbreak{} inspected/\allowbreak{}expanded evidence,\allowbreak{} and only then submitted.\allowbreak{} But it does not reach the score-\allowbreak{}2 anchor requirement because the supplied <\allowbreak{}drug\_\allowbreak{}id\_\allowbreak{}ns>\allowbreak{}/\allowbreak{}<\allowbreak{}disease\_\allowbreak{}id\_\allowbreak{}ns>\allowbreak{} IDs were available yet the connect call used surface names,\allowbreak{} and the disease anchor remained LOCAL/\allowbreak{}unresolved;\allowbreak{} the final relies on recognizing that unresolved state rather than first anchoring the pair correctly by drug\_\allowbreak{}id/\allowbreak{}disease\_\allowbreak{}id.\allowbreak{} This fits score 1 rather than 0 or 2.\allowbreak{} \\
\addlinespace[1.5pt]
& \bandbox{1} & \footnotesize Terminal coherence & \footnotesize I agree with the reviewer that t=\allowbreak{}18 contains a strong qualitative contrast:\allowbreak{} biologically plausible mechanism/\allowbreak{}class evidence versus no sanctioned direct support for the exact pair,\allowbreak{} and that this is a spelled-\allowbreak{}out evidence-\allowbreak{}category-\allowbreak{}to-\allowbreak{}band mapping for the 0.\allowbreak{}3 promisingness.\allowbreak{} I also agree the key\_\allowbreak{}claims trace the rationale to concrete retrieved entries,\allowbreak{} including no-\allowbreak{}path/\allowbreak{}unresolved-\allowbreak{}anchor findings and class-\allowbreak{}adjacent PD-\allowbreak{}L1 breast-\allowbreak{}cancer evidence.\allowbreak{} However,\allowbreak{} the score-\allowbreak{}2 anchor specifically requires promisingness,\allowbreak{} confidence,\allowbreak{} and rationale each trace back with the evidence-\allowbreak{}strength-\allowbreak{}to-\allowbreak{}value mapping spelled out.\allowbreak{} A qualitative evidence-\allowbreak{}category-\allowbreak{}to-\allowbreak{}band justification can be sufficient;\allowbreak{} exact numeric calibration to 0.\allowbreak{}48 is not inherently required.\allowbreak{} But here the final never explicitly connects those categories to confidence as opposed to promisingness/\allowbreak{}rationale.\allowbreak{} The confidence value is present,\allowbreak{} but the text does not say that confidence is intermediate because evidence is tool-\allowbreak{}grounded yet indirect/\allowbreak{}incomplete.\allowbreak{} Thus the concession that the mapping to 0.\allowbreak{}48 was not spelled out remains material and is priced into a score of 1.\allowbreak{} \\
\addlinespace[1.5pt]
\midrule
\multicolumn{4}{@{}l@{}}{%
\capbadge{E}\; \textbf{Evidence Fidelity}\hfill \textbf{\large 3.50}\,/\,4}\\[2pt]
\multicolumn{4}{@{}p{\dimexpr\textwidth-2\tabcolsep}@{}}{%
\footnotesize\itshape Strongest dimension. Every key claim names a resolvable retrieved record and its support is traceable to that same record; retrieved fact, own inference, and acknowledged unknown are kept apart in the prose --- the mechanism is called \emph{plausible}, not established, and the class-level hint is explicitly labelled class-adjacent. One band is lost because load-bearing evidence is not accompanied by an explicit recency statement.}\\[3pt]
& \bandbox{2} & \footnotesize Claim grounding & \footnotesize The submitted key\_\allowbreak{}claims consistently identify resolvable tool commands or inspected pages as sources,\allowbreak{} and their supports are traceable to those same records.\allowbreak{} For example,\allowbreak{} the unresolved anchor claim in the final is supported by the same Stage-\allowbreak{}1 output showing the local unresolved breast-\allowbreak{}cancer anchor,\allowbreak{} and the no-\allowbreak{}connection claim matches the Stage-\allowbreak{}1 report text.\allowbreak{} The mechanistic and tumor-\allowbreak{}agnostic claims are directly supported by the inspected dostarlimab page text.\allowbreak{} Even the PD-\allowbreak{}L1/\allowbreak{}breast-\allowbreak{}cancer support is careful:\allowbreak{} it cites the PD-\allowbreak{}L1 query and characterizes the hint as class-\allowbreak{}adjacent rather than direct evidence,\allowbreak{} avoiding overstating the record.\allowbreak{} \\
\addlinespace[1.5pt]
& \bandbox{2} & \footnotesize Fact / inference / unknown separation & \footnotesize The final answer cleanly distinguishes source-\allowbreak{}backed findings from inference and unknowns.\allowbreak{} It labels the mechanistic extension as biologically plausible,\allowbreak{} not established;\allowbreak{} explicitly states the tool\_\allowbreak{}A limitation that the disease anchor was not resolved;\allowbreak{} calls PD-\allowbreak{}L1/\allowbreak{}TNBC evidence class-\allowbreak{}adjacent rather than direct;\allowbreak{} and frames the score as plausible repurposing with no sanctioned direct clinical support for the exact pair.\allowbreak{} I did not find a load-\allowbreak{}bearing conjecture presented as a settled tool\_\allowbreak{}A fact.\allowbreak{} \\
\addlinespace[1.5pt]
& \bandbox{2} & \footnotesize Source integrity\;\critbadge & \footnotesize The submitted answer'\allowbreak{}s load-\allowbreak{}bearing negative claims point to returned named fields from the tool\_\allowbreak{}A output,\allowbreak{} including zero candidate paths/\allowbreak{}connections/\allowbreak{}load-\allowbreak{}bearing evidence in step 14.\allowbreak{} Positive mechanism/\allowbreak{}context claims cite locatable returned records:\allowbreak{} PMIDs and the drug page uid/\allowbreak{}artifact path from returned/\allowbreak{}inspected pages.\allowbreak{} The PD-\allowbreak{}L1 class-\allowbreak{}adjacent claim cites the actual PD-\allowbreak{}L1 connect call and returned hints.\allowbreak{} I found no load-\allowbreak{}bearing item cited merely as vague '\allowbreak{}literature'\allowbreak{} or '\allowbreak{}KG'\allowbreak{} support.\allowbreak{} \\
\addlinespace[1.5pt]
& \bandbox{1} & \footnotesize Temporal-leakage discipline & \footnotesize The trajectory stayed within the sanctioned offline environment:\allowbreak{} bash was used to read local toolkit/\allowbreak{}access files and the main evidence gathering used tool\_\allowbreak{}A,\allowbreak{} with the access file noting the the sandbox has no network.\allowbreak{} However,\allowbreak{} load-\allowbreak{}bearing claims such as the final FDA/\allowbreak{}checkpoint-\allowbreak{}antibody rationale were stated as current without any explicit recency handling.\allowbreak{} The retrieved drug page carried an updated stamp (\allowbreak{}2026-\allowbreak{}07-\allowbreak{}06)\allowbreak{},\allowbreak{} but the submission did not state that status or caveat possible staleness for updated/\allowbreak{}undated synthesized pages.\allowbreak{} Thus there is no score-\allowbreak{}0 leakage breach,\allowbreak{} but the required recency discipline for score 2 is missing.\allowbreak{} \\
\addlinespace[1.5pt]
\midrule
\multicolumn{4}{@{}l@{}}{%
\capbadge{A}\; \textbf{Hypothesis Management}\hfill \textbf{\large 3.00}\,/\,4}\\[2pt]
\multicolumn{4}{@{}p{\dimexpr\textwidth-2\tabcolsep}@{}}{%
\footnotesize\itshape A concrete testable mechanism is stated with named targets and a cellular effect, and the negative retrieval is retained in the final answer and used to lower the score rather than being discarded. Two bands are lost: only one biological mechanism is advanced where the sub-rubric asks for a competing no-benefit hypothesis to be set beside it, and no explicit before/after revision of the estimate is shown as the negative evidence arrived.}\\[3pt]
& \bandbox{1} & \footnotesize Alternative paths considered & \footnotesize On revision,\allowbreak{} the reviewer'\allowbreak{}s point is persuasive:\allowbreak{} the agent mainly advances one biological mechanism,\allowbreak{} PD-\allowbreak{}1/\allowbreak{}PD-\allowbreak{}L1 checkpoint blockade,\allowbreak{} and the statement that the tool found no candidate paths or load-\allowbreak{}bearing evidence is an evidence-\allowbreak{}gap statement,\allowbreak{} not an explicitly formulated competing biological '\allowbreak{}no benefit'\allowbreak{} or '\allowbreak{}symptomatic-\allowbreak{}only'\allowbreak{} hypothesis.\allowbreak{} The final does qualify the single mechanism by noting class-\allowbreak{}adjacent and non-\allowbreak{}specific evidence,\allowbreak{} so it acknowledges weakness/\allowbreak{}other possibilities,\allowbreak{} but it never places two mechanistic paths side by side nor explicitly argues that the drug would have no benefit beyond symptomatic-\allowbreak{}only care.\allowbreak{} Therefore this fits score 1 rather than 2.\allowbreak{} \\
\addlinespace[1.5pt]
& \bandbox{2} & \footnotesize Contradiction retention & \footnotesize The trace offered a clear null/\allowbreak{}contradictory retrieval:\allowbreak{} max-\allowbreak{}tier tool\_\allowbreak{}A found no complete path and labeled support incomplete.\allowbreak{} The submission retained this in the rationale,\allowbreak{} noting the unresolved disease anchor and no direct evidence,\allowbreak{} and explicitly used it to lower the deliverable promisingness/\allowbreak{}confidence by scoring below Phase-\allowbreak{}2-\allowbreak{}success territory while still acknowledging mechanistic plausibility/\allowbreak{}class-\allowbreak{}adjacent support.\allowbreak{} \\
\addlinespace[1.5pt]
& \bandbox{2} & \footnotesize Hypotheses made explicit & \footnotesize The final deliverable explicitly states a testable checkpoint-\allowbreak{}blockade mechanism with named targets (\allowbreak{}PD-\allowbreak{}1/\allowbreak{}PD-\allowbreak{}L1,\allowbreak{} plus PD-\allowbreak{}L2 in key claims)\allowbreak{},\allowbreak{} a cellular process (\allowbreak{}restored T-\allowbreak{}cell activity)\allowbreak{},\allowbreak{} and relates it to the disease context through immunogenic breast-\allowbreak{}cancer subsets.\allowbreak{} Although it caveats that evidence for exact Stage II Breast Cancer is indirect,\allowbreak{} the mechanism itself is concrete and disease-\allowbreak{}relevant,\allowbreak{} meeting the high band.\allowbreak{} \\
\addlinespace[1.5pt]
& \bandbox{1} & \footnotesize Hypothesis update on evidence & \footnotesize The trajectory maps retrieved records into the final score:\allowbreak{} the max-\allowbreak{}tier search found no complete path (\allowbreak{}t14)\allowbreak{},\allowbreak{} while the final rationale explicitly uses that negative evidence plus the PD-\allowbreak{}L1 related-\allowbreak{}page evidence to justify a below-\allowbreak{}Phase-\allowbreak{}2 but above-\allowbreak{}preclinical score (\allowbreak{}t18)\allowbreak{}.\allowbreak{} However,\allowbreak{} I found no visible prior,\allowbreak{} baseline,\allowbreak{} interim score,\allowbreak{} or explicit upward/\allowbreak{}downward update caused by a named retrieved record;\allowbreak{} the score is a static synthesis rather than a demonstrated ranking movement.\allowbreak{} This matches score 1 rather than 2.\allowbreak{} \\
\addlinespace[1.5pt]
\midrule
\multicolumn{4}{@{}l@{}}{%
\capbadge{S}\; \textbf{Boundary \& Failure Reasoning}\hfill \textbf{\large 2.50}\,/\,4}\\[2pt]
\multicolumn{4}{@{}p{\dimexpr\textwidth-2\tabcolsep}@{}}{%
\footnotesize\itshape Weakest dimension, and the honest reading of it. Research-use framing is maintained throughout and the evidence gap for this exact pair is stated plainly. But the counter-case is argued almost entirely as an \emph{absence of evidence}: the submission does not name a concrete biological or clinical reason this repurposing could fail in this setting, does not raise an indication-specific safety or feasibility constraint, and never explains why the stated confidence value is the appropriate one.}\\[3pt]
& \bandbox{2} & \footnotesize Boundary \& scope\;\critbadge & \footnotesize The submission stays within research-\allowbreak{}use framing for the exact drug-\allowbreak{}disease pair:\allowbreak{} it describes mechanistic plausibility and explicitly qualifies the evidence as lacking direct clinical support for Stage II Breast Cancer.\allowbreak{} Its key claims state that no complete connection was reconstructed and that PD-\allowbreak{}L1/\allowbreak{}TNBC evidence is only class-\allowbreak{}adjacent,\allowbreak{} matching the retrieved incomplete support structure.\allowbreak{} I found no dosing,\allowbreak{} administration,\allowbreak{} monitoring,\allowbreak{} or recommendation to start/\allowbreak{}switch/\allowbreak{}treat patients with dostarlimab for this population.\allowbreak{} \\
\addlinespace[1.5pt]
& \bandbox{1} & \footnotesize Failure modes & \footnotesize On revision,\allowbreak{} I agree the cited counter-\allowbreak{}evidence is primarily an evidence-\allowbreak{}gap argument,\allowbreak{} not a concrete biological/\allowbreak{}clinical failure mechanism.\allowbreak{} The final submission says the mechanism is plausible in immunogenic breast-\allowbreak{}cancer subsets,\allowbreak{} but it does not state that Stage II tumors may lack immunogenicity,\allowbreak{} that dostarlimab may have insufficient target engagement,\allowbreak{} inadequate PK/\allowbreak{}exposure,\allowbreak{} limited tissue delivery,\allowbreak{} or a wrong endpoint/\allowbreak{}population mechanism causing failure.\allowbreak{} Its concerns that the evidence is class-\allowbreak{}adjacent,\allowbreak{} not dostarlimab-\allowbreak{} or stage-\allowbreak{}II-\allowbreak{}specific,\allowbreak{} and that there is no direct clinical support amount to limited-\allowbreak{}evidence hedging.\allowbreak{} Therefore this fits score 1 rather than score 2.\allowbreak{} \\
\addlinespace[1.5pt]
& \bandbox{1} & \footnotesize Safety \& feasibility gating & \footnotesize The trajectory had some generic feasibility/\allowbreak{}safety-\allowbreak{}adjacent material in the inspected drug page,\allowbreak{} including pharmacokinetics and no known drug-\allowbreak{}drug interactions,\allowbreak{} but it did not name an indication-\allowbreak{}specific safety or feasibility constraint for Stage II Breast Cancer.\allowbreak{} The submitted rationale and score were driven by lack of direct evidence and class-\allowbreak{}adjacent plausibility,\allowbreak{} not tolerability,\allowbreak{} contraindications,\allowbreak{} target-\allowbreak{}organ toxicity,\allowbreak{} administration,\allowbreak{} or monitoring burden.\allowbreak{} \\
\addlinespace[1.5pt]
& \bandbox{1} & \footnotesize Uncertainty calibration & \footnotesize The submission names uncertainty sources (\allowbreak{}unresolved disease anchor,\allowbreak{} no paths/\allowbreak{}load-\allowbreak{}bearing evidence,\allowbreak{} class-\allowbreak{}adjacent PD-\allowbreak{}L1 evidence)\allowbreak{} satisfying the score-\allowbreak{}1 prerequisite.\allowbreak{} However,\allowbreak{} the rationale never mentions '\allowbreak{}confidence'\allowbreak{} and never explains why confidence\_\allowbreak{}score=\allowbreak{}0.\allowbreak{}48 is appropriate.\allowbreak{} The only '\allowbreak{}therefore score'\allowbreak{} sentence references promisingness tier anchors (\allowbreak{}'\allowbreak{}below Phase-\allowbreak{}2-\allowbreak{}success territory but above purely preclinical'\allowbreak{})\allowbreak{},\allowbreak{} which maps to the 0.\allowbreak{}3 promisingness value,\allowbreak{} not the 0.\allowbreak{}48 confidence value.\allowbreak{} The task instructs '\allowbreak{}Report low confidence when evidence is thin'\allowbreak{} yet 0.\allowbreak{}48 is moderate,\allowbreak{} not low,\allowbreak{} and this choice receives no justification.\allowbreak{} Score 2 requires explaining a confidence\_\allowbreak{}score consistent with described evidence strength;\allowbreak{} merely naming uncertainty without connecting it to the numerical confidence value is score 1.\allowbreak{} \\
\addlinespace[1.5pt]
\midrule
\multicolumn{4}{@{}l@{}}{%
\capbadge{T}\; \textbf{Tool Use \& Execution State Management}\hfill \textbf{\large 3.50}\,/\,4}\\[2pt]
\multicolumn{4}{@{}p{\dimexpr\textwidth-2\tabcolsep}@{}}{%
\footnotesize\itshape Manuals were read before any retrieval, the mandated bounded search was issued in the documented form, returned artefacts were opened and their content reused in the final answer, and search depth was escalated once and then stopped with an explicit sufficiency judgement. One band is lost on call correctness, for the avoidable invalid-parameter attempt at step~12 --- the same event that makes Repair gradeable.}\\[3pt]
& \bandbox{1} & \footnotesize Call correctness & \footnotesize The agent did read the workflow/\allowbreak{}help first and its initial required Stage-\allowbreak{}1 call used the documented command form correctly.\allowbreak{} However,\allowbreak{} it later made an avoidable argument error by inventing/\allowbreak{}using an unsupported tier '\allowbreak{}deep'\allowbreak{},\allowbreak{} which the tool rejected,\allowbreak{} then patched by retrying with the valid '\allowbreak{}max'\allowbreak{} tier.\allowbreak{} Also the pair was passed by surface names rather than the task'\allowbreak{}s supplied IDs,\allowbreak{} and the disease resolved as local/\allowbreak{}unknown,\allowbreak{} so this does not meet the high-\allowbreak{}band exact anchor/\allowbreak{}argument standard.\allowbreak{} The calls were ultimately usable,\allowbreak{} fitting score 1 rather than 0 or 2.\allowbreak{} \\
\addlinespace[1.5pt]
& \bandbox{2} & \footnotesize Cost \& stopping discipline & \footnotesize The agent did not stop after a single weak return:\allowbreak{} it escalated from an extra-\allowbreak{}tier pair query to max tier,\allowbreak{} inspected the truncated drug page,\allowbreak{} and made one relevant PD-\allowbreak{}L1 hint query.\allowbreak{} It then submitted a low/\allowbreak{}moderate confidence conclusion with an explicit sufficiency judgment:\allowbreak{} the exact pair still had no reconstructed path or load-\allowbreak{}bearing evidence even at max tier,\allowbreak{} while the PD-\allowbreak{}L1 breast-\allowbreak{}cancer evidence was only class-\allowbreak{}adjacent,\allowbreak{} hence the stated below-\allowbreak{}Phase-\allowbreak{}2 score.\allowbreak{} There is a minor non-\allowbreak{}incremental invalid deep-\allowbreak{}tier attempt,\allowbreak{} but it immediately self-\allowbreak{}corrected to max and converged rather than looping.\allowbreak{} \\
\addlinespace[1.5pt]
& \bandbox{2} & \footnotesize Result absorption & \footnotesize The agent opened and read returned artifacts,\allowbreak{} including the full drug page and concrete summary fields,\allowbreak{} then made those details resurface in key\_\allowbreak{}claims/\allowbreak{}source/\allowbreak{}support.\allowbreak{} Because the search returned no paths or edges,\allowbreak{} there were no edge directions/\allowbreak{}confidences to absorb;\allowbreak{} nevertheless it used specific returned items such as empty candidate\_\allowbreak{}paths/\allowbreak{}connections/\allowbreak{}load\_\allowbreak{}bearing\_\allowbreak{}evidence,\allowbreak{} unresolved/\allowbreak{}no complete path reporting,\allowbreak{} drug-\allowbreak{}page mechanism/\allowbreak{}approval text,\allowbreak{} and PD-\allowbreak{}L1 related-\allowbreak{}page hint details to support the final judgment.\allowbreak{} This exceeds count/\allowbreak{}headline-\allowbreak{}only absorption.\allowbreak{} \\
\addlinespace[1.5pt]
& \bandbox{2} & \footnotesize Tool selection\;\critbadge & \footnotesize The agent used retrieval rather than recall and chose tool\_\allowbreak{}A commands aligned with the open questions.\allowbreak{} It began with the required pairwise connect for Dostarlimab and Stage II Breast Cancer,\allowbreak{} then the returned summary showed no suggested expansions,\allowbreak{} making skipped expand acceptable.\allowbreak{} It later followed a decision-\allowbreak{}relevant related-\allowbreak{}page hint for PD-\allowbreak{}L1 using the correct target id and stage2 suggest form.\allowbreak{} The invalid deep-\allowbreak{}tier attempt is a parameter/\allowbreak{}call-\allowbreak{}correctness problem,\allowbreak{} not evidence that the subcommand choice was unrelated to the pair.\allowbreak{} \\
\addlinespace[1.5pt]
\midrule
\multicolumn{4}{@{}l@{}}{%
\capbadge{R}\; \textbf{Self-correction under Verification}\hfill \textbf{\large 4.00}\,/\,4}\\[2pt]
\multicolumn{4}{@{}p{\dimexpr\textwidth-2\tabcolsep}@{}}{%
\footnotesize\itshape \textbf{This is why the case was selected.} A real failure occurred at step~12 and was repaired at step~14, so this capability is genuinely \emph{rated} rather than skipped. Four of its five sub-rubrics score at the top band: the rejection was read as a self-made bad argument rather than a tool or data fault; the fix changed \emph{only} the offending parameter and resumed the same evidence line; the repaired call's substantive content was then actually used, not merely checked for success; and the same error class never recurred. The fifth is \textsc{nr}: with the error corrected on the very next action, the trace never exhibited a \emph{sustained} localisation problem to grade separately.}\\[3pt]
& \textsc{nr} & \footnotesize Error localisation & \footnotesize The rubric'\allowbreak{}s trigger condition requires '\allowbreak{}a non-\allowbreak{}ok/\allowbreak{}rejected tool status,\allowbreak{} an explicit error,\allowbreak{} or an empty/\allowbreak{}negative result the agent had relied on'\allowbreak{} and explicitly excludes '\allowbreak{}A call that succeeded (\allowbreak{}exit 0)\allowbreak{}'\allowbreak{}.\allowbreak{} The only incident was t12'\allowbreak{}s invalid -\allowbreak{}-\allowbreak{}tier deep flag,\allowbreak{} which the agent immediately corrected to -\allowbreak{}-\allowbreak{}tier max at t14.\allowbreak{} The connection calls at t8 and t14 both succeeded with exit\_\allowbreak{}code 0 and returned structured partial results with budget\_\allowbreak{}exhausted:\allowbreak{} true,\allowbreak{} error:\allowbreak{} null,\allowbreak{} timed\_\allowbreak{}out:\allowbreak{} false.\allowbreak{} The task brief explicitly frames such partial results as '\allowbreak{}usable bounded coverage-\allowbreak{}-\allowbreak{}-\allowbreak{}not a provider failure or negative evidence.\allowbreak{}'\allowbreak{} The unresolved disease anchor appeared in anchor\_\allowbreak{}quality.\allowbreak{}unresolved with documented reason '\allowbreak{}unresolved\_\allowbreak{}or\_\allowbreak{}local\_\allowbreak{}query\_\allowbreak{}anchor'\allowbreak{},\allowbreak{} but this was an operational limitation in a successful API response,\allowbreak{} not a repairable tool-\allowbreak{}call error.\allowbreak{} The documented workflow in <\allowbreak{}manual\_\allowbreak{}B>\allowbreak{} instructs entity names with no mention of <\allowbreak{}disease\_\allowbreak{}id\_\allowbreak{}ns>\allowbreak{} IDs or -\allowbreak{}-\allowbreak{}b-\allowbreak{}id usage for disease resolution,\allowbreak{} and the KG provider uses internal hash IDs (\allowbreak{}disease:\allowbreak{}<\allowbreak{}record>\allowbreak{} format)\allowbreak{} not external ontology identifiers.\allowbreak{} There is no evidence that -\allowbreak{}-\allowbreak{}b-\allowbreak{}id <\allowbreak{}disease\_\allowbreak{}id>\allowbreak{} would have resolved the anchor.\allowbreak{} Because the connection calls succeeded and returned documented partial coverage per the API design,\allowbreak{} no qualifying repairable incident occurred under the rubric'\allowbreak{}s definition.\allowbreak{} \\
\addlinespace[1.5pt]
& \bandbox{2} & \footnotesize Failure comprehension & \footnotesize The trajectory contains a clear repairable incident:\allowbreak{} the agent tried an unsupported tier,\allowbreak{} and the tool explained that '\allowbreak{}deep'\allowbreak{} was an invalid choice with valid options including max.\allowbreak{} The next action selected the valid max tier,\allowbreak{} showing the agent read the failure as a bad argument rather than tool/\allowbreak{}data failure.\allowbreak{} The empty/\allowbreak{}no-\allowbreak{}path results were also represented in the deliverable as unresolved coverage with zero candidate paths and connections,\allowbreak{} not overstated as definitive negative evidence.\allowbreak{} \\
\addlinespace[1.5pt]
& \bandbox{2} & \footnotesize No repeated error & \footnotesize There was a clear repairable incident:\allowbreak{} the agent attempted an unsupported tier,\allowbreak{} and the tool reported that '\allowbreak{}deep'\allowbreak{} was invalid while listing allowed tiers.\allowbreak{} The next analogous call changed the argument to '\allowbreak{}-\allowbreak{}-\allowbreak{}tier max'\allowbreak{} and succeeded with exit\_\allowbreak{}code 0.\allowbreak{} I found no recurrence of the same invalid tier error afterward,\allowbreak{} so the error class never returned.\allowbreak{} \\
\addlinespace[1.5pt]
& \bandbox{2} & \footnotesize Re-verification & \footnotesize The trajectory had a qualifying incident:\allowbreak{} the attempted deep tier produced an explicit invalid-\allowbreak{}choice error.\allowbreak{} The agent substituted the valid max tier for the same Dostarlimab/\allowbreak{}Stage II Breast Cancer connection query,\allowbreak{} and the returned record was not merely checked for success;\allowbreak{} its substantive result (\allowbreak{}no reconstructed connection,\allowbreak{} unresolved disease anchor,\allowbreak{} zero paths/\allowbreak{}connections)\allowbreak{} was used in the final rationale/\allowbreak{}key claims.\allowbreak{} The failed deep-\allowbreak{}tier line was not carried forward as evidence.\allowbreak{} \\
\addlinespace[1.5pt]
& \bandbox{2} & \footnotesize Targeted repair & \footnotesize The trace contains a clear repairable incident:\allowbreak{} the agent attempted the same tool\_\allowbreak{}A connect line with an unsupported -\allowbreak{}-\allowbreak{}tier deep,\allowbreak{} and the tool reported the valid choices.\allowbreak{} The next relevant action made the minimal root-\allowbreak{}cause fix by changing only the invalid tier to the supported max tier while keeping the same Dostarlimab/\allowbreak{}Stage II Breast Cancer evidence line,\allowbreak{} and that repaired command succeeded.\allowbreak{} \\
\addlinespace[1.5pt]
\end{xltabular}}

\end{document}